%% file: main.tex
\documentclass[nohyperref]{article}

\usepackage{microtype}
\usepackage{graphicx}
\usepackage{subfigure}
\usepackage{booktabs} 
\usepackage{multirow}
\usepackage[dvipsnames]{xcolor}

\usepackage{hyperref}

\usepackage[accepted]{icml_style_file/icml2022}

\usepackage{amsmath}
\usepackage{amssymb}
\usepackage{mathtools}
\usepackage{amsthm}
\usepackage{bbm}
\usepackage[capitalize,noabbrev]{cleveref}

\theoremstyle{plain}
\newtheorem{theorem}{Theorem}[section]
\newtheorem{proposition}[theorem]{Proposition}

\theoremstyle{definition}

\newtheorem{assumption}[theorem]{Assumption}
\theoremstyle{remark}

\newcommand{\R}{\mathbb{R}}
\newcommand{\cN}{\mathcal{N}}
\newcommand{\sT}{\mathsf{T}}
\newcommand{\bzero}{{\boldsymbol 0}}
\newcommand{\bx}{{\boldsymbol x}}
\newcommand{\bX}{{\boldsymbol X}}
\newcommand{\by}{{\boldsymbol y}}
\newcommand{\bH}{{\boldsymbol H}}

\newcommand{\byt}{\widetilde{\boldsymbol y}}
\newcommand{\bxt}{\widetilde{ \boldsymbol x}}
\newcommand{\bXt}{\widetilde{ \boldsymbol X}}
\newcommand{\bSigma}{{\boldsymbol \Sigma}}
\newcommand{\bSigmat}{\widetilde{ \boldsymbol \Sigma}}
\newcommand{\bP}{{\boldsymbol P}}
\newcommand{\bPt}{\widetilde{ \boldsymbol P}}
\newcommand{\bI}{{\boldsymbol I}}
\newcommand{\yt}{\widetilde{y}}
\newcommand{\ytp}{\widetilde{y}^{\,p}}
\newcommand{\bu}{{\boldsymbol u}}
\newcommand{\bv}{{\boldsymbol v}}
\newcommand{\btheta}{{\boldsymbol \theta}}
\newcommand{\bthetah}{\smash{\widehat{\boldsymbol\theta}}}
\newcommand{\bthetat}{\smash{\widetilde{\boldsymbol\theta}}}
\newcommand{\Dtrain}{\mathcal{D}_{\text{train}}}
\newcommand{\Dtest}{\mathcal{D}_{\text{test}}}
\newcommand{\err}{\textsf{TestError}}
\newcommand{\tl}{\textsf{TestLoss}}
\newcommand{\pj}{\textsf{ProjNorm}}
\newcommand{\pjl}{\textsf{ProjNormLinear}}
\newcommand{\grad}{\nabla}
\newcommand{\<}{\langle}
\renewcommand{\>}{\rangle}
\newcommand{\flin}{f^{\textsf{Lin}}}

\definecolor{pink}{RGB}{255,20,147}
\definecolor{PINK}{RGB}{255,20,147}

\definecolor{Gray}{gray}{0.85}

\usepackage[textsize=tiny]{todonotes}

\begin{document}

\twocolumn[
\icmltitle{Predicting Out-of-Distribution Error with the Projection Norm}


\icmlsetsymbol{equal}{*}

\begin{icmlauthorlist}
\icmlauthor{Yaodong Yu}{berkeley,equal}
\icmlauthor{Zitong Yang}{berkeley,equal}
\icmlauthor{Alexander Wei}{berkeley}
\icmlauthor{Yi Ma}{berkeley}
\icmlauthor{Jacob Steinhardt}{berkeley}
\end{icmlauthorlist}
\icmlaffiliation{berkeley}{University of California, Berkeley}
\icmlcorrespondingauthor{Yaodong Yu}{yyu@eecs.berkeley.edu}
\icmlcorrespondingauthor{Zitong Yang}{zitong@berkeley.edu}
\icmlcorrespondingauthor{Alexander Wei}{awei@berkeley.edu}
\icmlcorrespondingauthor{Yi Ma}{yima@eecs.berkeley.edu}
\icmlcorrespondingauthor{Jacob Steinhardt}{jsteinhardt@berkeley.edu}

\vskip 0.3in
]



\printAffiliationsAndNotice{\icmlEqualContribution} 

\begin{abstract}
We propose a metric---\emph{Projection Norm}---to predict a model's performance on out-of-distribution (OOD) data without  access to ground truth labels. 
Projection Norm first uses model predictions to pseudo-label test samples and then trains a new model on the pseudo-labels. The more the new model's parameters differ from an in-distribution model, the greater the predicted OOD error. 
Empirically, our approach outperforms existing methods on both image and text classification tasks and across different network architectures. Theoretically, we connect our approach to a bound on the test error for overparameterized linear models. Furthermore, we find  that Projection Norm is the only approach that achieves non-trivial detection performance on adversarial examples. Our code is available at \url{https://github.com/yaodongyu/ProjNorm}. 
\end{abstract}

\input{sec_intro}

\input{sec_method}

\input{sec_exp}

\input{sec_linearmodel}

\input{sec_ntkemp}

\input{sec_stree}

\input{sec_relatedwork}

\input{sec_discussion}

\subsection*{Acknowledgements}
We would like to thank Aditi Raghunathan, Yu Sun, and Chong You for their valuable feedback and comments. We would also like to thank Keyang Xu and Xiang Zhou for helpful discussions on the NLI experiments.

\bibliography{reference}
\bibliographystyle{icml_style_file/icml2022}

\appendix

\input{sec_appendix}

\end{document}

%% file: sec_intro.tex
\section{Introduction}\label{sec:intro}
\begin{figure*}[ht!]
    \centering
    \subfigure{\includegraphics[width=.65\textwidth]{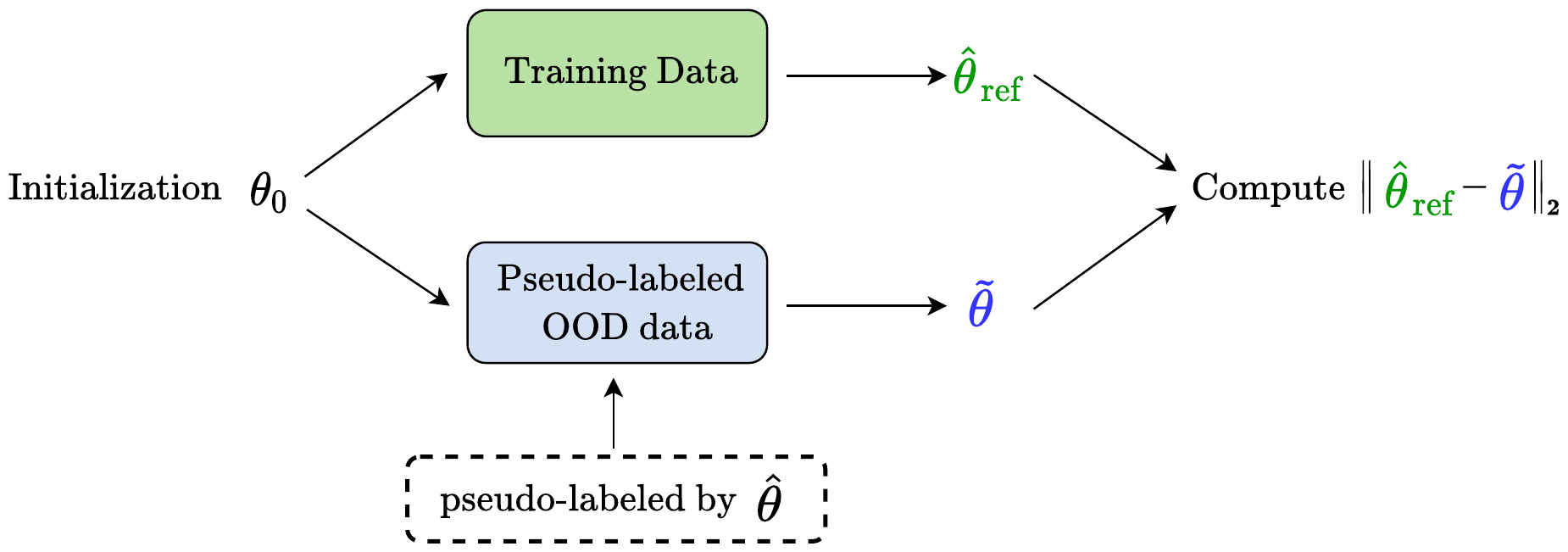}}
    \subfigure{\includegraphics[width=.275\textwidth]{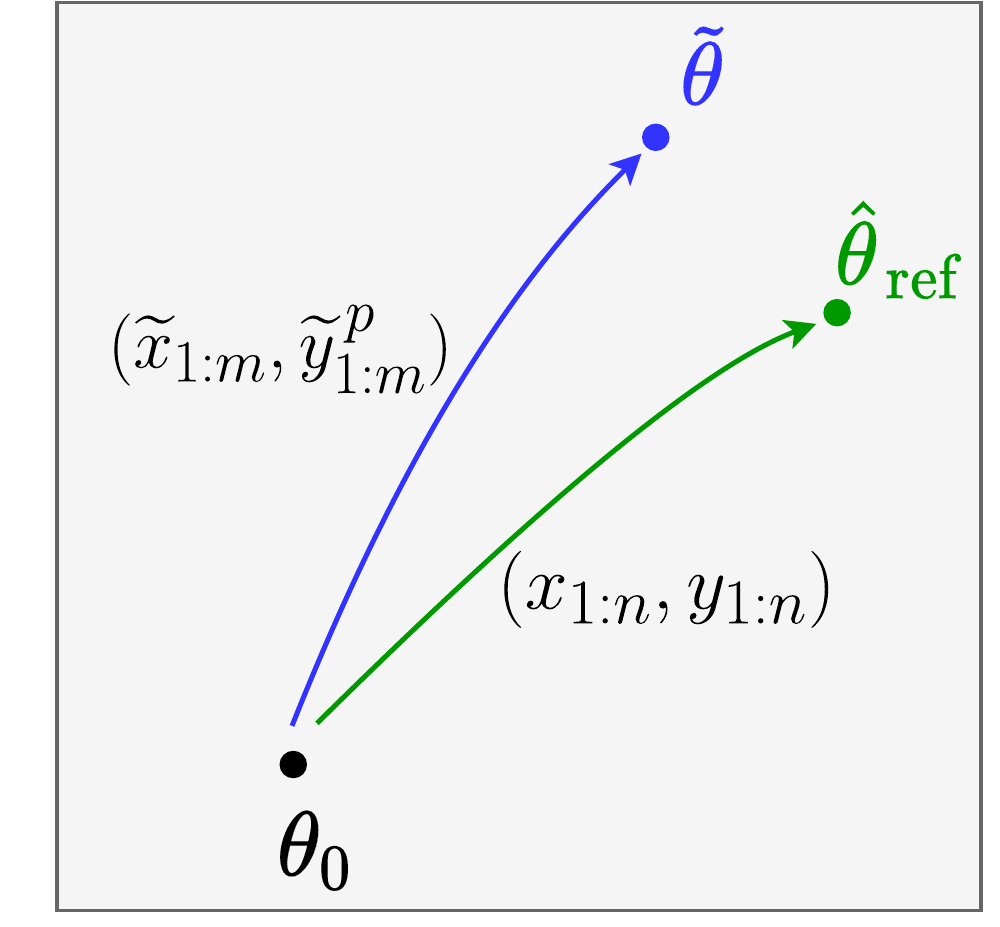}}
    \vspace{-0.1in}
    \caption{\textbf{How to compute Projection Norm on unlabeled OOD data}. (\textbf{Left}) Given a classifier $\bthetah$, we first pseudo-label the OOD data $\bxt_{1:m}$ using the predictions of $\bthetah$. Next, we obtain a new network $\bthetat$ that is initialized with $\btheta_{0}$ and trained on the pseudo-labeled OOD data. Finally, we train a reference network $\bthetah_{\text{ref}}$ on the training data (with the same initialization $\btheta_{0}$) and output the $\pj(\Dtrain, \bxt_{1:m}) = \|\bthetah_{\text{ref}} - \bthetat\|_2$. (\textbf{Right}) Schematic of $\bthetat$ and $\bthetah_{\text{ref}}$ are obtained. See Algorithm~\ref{alg:projnorm} for details on computing \pj. }
    \label{fig:projnorm-compute}
    \vspace{-0.15in}
\end{figure*}

To reliably deploy machine learning models in practice, we must understand the model's performance on unseen test samples.
Conventional machine learning wisdom suggests using a held-out validation set to estimate the model's test-time performance~\citep{esl}. However, this fails to account for distribution shift.
For deep neural networks, even simple distribution shifts can lead to large drops in performance~\citep{quinonero2008dataset, koh2021wilds}.
Thus, it is crucial to understand, especially in safety-critical applications, how a model might perform on out-of-distribution (OOD) data.
Finally, understanding OOD performance helps shed light on the structure of natural covariate shifts, which remain poorly understood from a conceptual standpoint~\citep{hendrycks2021many}.

To this end, we propose \emph{Projection Norm}, which uses unlabeled test samples to help predict the OOD test error. 
Let $\bthetah$ be the model whose test error we aim to predict. 
At a high level, the Projection Norm algorithm pseudo-labels the test samples using $\bthetah$ and then uses these pseudo-labels to train a new model $\bthetat$.
Finally, it compares the distance between $\bthetah$ and $\bthetat$, with a larger distance corresponding to higher test error.
We formally present this algorithm in Section \ref{sec:method}.

Empirically, we demonstrate that Projection Norm predicts test error more accurately than existing methods~\citep{deng2021does, guillory2021predicting, garg2021leveraging}, across several vision and language benchmarks and for different neural network architectures~(Section~\ref{subsec:exp-main-results}).
Moreover, while the errors of existing methods are highly correlated with each other, the errors of Projection Norm are nearly uncorrelated with those of existing methods~(Section~\ref{subsec:stat}), so combining Projection Norm with these methods results in even better prediction performance. 
Finally, we stress test our method against adversarial examples, an extreme type of distribution shift, and we find that
Projection Norm is the only method that achieves non-trivial performance (Section~\ref{sec:stress}).

Projection Norm also has a natural theoretical motivation.
We show for overparameterized linear models that Projection Norm measures the projection (hence the name) of a ``ground truth model'' onto the overlap of the training and test data (Section \ref{sec:linearmodel}).
In this linear setting, many common methods focus only on the logits and thus cannot capture information that is orthogonal to the training manifold. In contrast, Projection Norm can, which explains why it provides information complementary to that of other methods. 
We also connect Projection Norm to a mathematical bound on the test loss, based on assumptions backed by empirical studies on vision data (Section \ref{sec:ntkemp}).

In summary, we propose a new metric for predicting OOD error that provides a more accurate and orthogonal signal in comparison to existing approaches.
Our method is easy to implement and is applicable to a wide range of prediction tasks.
In addition, our method connects naturally to the theory of high-dimensional linear models and attains non-trivial performance even for adversarial examples.

%% file: sec_method.tex
\section{Our Method: Projection Norm}\label{sec:method}
In this section, we formulate the problem of predicting OOD performance at test time and then present the Projection Norm algorithm.

\textbf{Problem formulation.} Consider solving a $K$-class classification task using a neural network parameterized by $\btheta$. 
Let $f_1,\dots, f_K$ be functions representing the last layer of the neural network and $C(\bx; {\btheta}) = \arg\max_i f_i(\bx; \btheta)$ be the corresponding classifier.
Given a training set $\Dtrain = \{(\bx_i, y_i)\}_{i=1,\dots, n}$, we use a pre-trained network, denoted by $\btheta_0$, for initialization and fine-tune the network on $\Dtrain$ by approximately minimizing the training loss (e.g.~via SGD). 
We denote the parameters of the fine-tuned network by $\bthetah$.

At test time, the fine-tuned classifier $C(\cdot; \bthetah)$ is then tested on $m$ (out-of-distribution) test samples $\bxt_{1:m}$ with corresponding unobserved labels $\yt_{1:m}$. 
The test error on OOD data $\bxt_{1:m}$ is defined as
\vspace{-0.075in}
\begin{equation*}
    \err(\bxt_{1:m}, \yt_{1:m}, \bthetah) = \frac{1}{m} \sum_{j=1}^m \mathbf{1}\big\{C(\bxt_j; \bthetah)\neq \yt_j\big\}.
\vspace{-0.075in}
\end{equation*}
Our goal is to propose a quantity, {\em without access to the test labels $\yt_{1:m}$},  that correlates well with the test error across different distribution shifts. 
\vspace{-0.1in}
\subsection{Projection norm}\label{sec:pjmethod}
To this end, we introduce the Projection Norm metric, denoted by $\pj(\Dtrain, \bxt_{1:m})$, which empirically correlates well with the test error.
At a high level, our method consists of three steps (illustrated in Figure~\ref{fig:projnorm-compute}):
\vspace{-0.1in}

\begin{itemize}
\item \textbf{Step 1: Pseudo-label the test set.} Given a classifier $C(\cdot; \bthetah)$
and test samples $\bxt_{1:m}$, compute ``pseudo-labels'' $\ytp_j = C(\bxt_j; \bthetah)$.
\item \textbf{Step 2: Fine-tune on the pseudo-labels.}  Initialize a fresh network with pre-trained parameters $\btheta_{0}$, then fine-tune on the $m$ pseudo-labeled OOD data points to obtain a model $\bthetat$. 
\item \textbf{Step 3: Compute the distance to a reference model.} Finally, we define the \textit{Projection Norm} as the Euclidean distance to a reference model $\bthetah_{\text{ref}}$: 
\vspace{-0.01in}
\begin{equation}\label{eqn:pjnn}
\pj(\Dtrain, \bxt_{1:m}) = \|\bthetah_{\text{ref}} - \bthetat\|_2.
\end{equation}
\end{itemize}
\vspace{-0.075in}
We can take $\bthetah_{\text{ref}} = \bthetah$; however, $\bthetah$ may be trained on many more samples than $\bthetat$, 
so an intuitive choice for $\bthetah_{\text{ref}}$ is to instead fine-tune  $\btheta_0$ 
on $m$ samples from the training set, using the same fine-tuning procedure as \textbf{Step\,2}. We find that both choices yield similar performance (Section~\ref{subsec:exp-analysis}), and use the latter for our mainline experiments. 
Fine-tuning $\bthetah_{\text{ref}}$ and $\bthetat$ requires $m$ to be reasonably large to achieve meaningful results (see Section~\ref{subsec:exp-analysis}).

We will see in Section \ref{sec:linearmodel} that \textbf{Steps\,1} and \textbf{2} essentially perform a ``nonlinear projection'' of $\bthetah$ onto the span of OOD samples $\bxt_{1:m}$, which is where the name Projection Norm came from.
Intuitively, $\bthetat$ has a subset of the information in $\bthetah$ (since it is trained on the latter model's pseudo-labels). The smaller the overlap between train and test, the less this information will be retained and the further 
$\bthetat$ will be from the reference model.

As we will show in Section~\ref{sec:linearmodel}, an advantage of our method is that it captures information orthogonal to the training manifold (in contrast to other methods) 
and can be connected to a bound on the test error. Before diving into theoretical analysis, we first study the empirical performance of $\pj$ to demonstrate its effectiveness.

%% file: sec_exp.tex
\begin{table*}[ht]
\centering
\caption{\textbf{Summary of prediction performance on CIFAR10, CIFAR100, and MNLI.} We compute coefficients of determination~($R^{2}$) and rank correlations~($\rho$) for existing methods and \pj{} to compare prediction performance (higher is better). The highest $R^2$ and $\rho$ quantities in each row are in \textbf{bold}. 
}
\vspace{0.05in}
\small
\label{table:main-table}
\begin{tabular}{@{}cccccccccccccc@{}}
\toprule
\multirow{2}{*}{Dataset} 
& \multirow{2}{*}{Network}
& \multicolumn{2}{c}{Rotation}
& \multicolumn{2}{c}{ConfScore}
& \multicolumn{2}{c}{Entropy}
& \multicolumn{2}{c}{AgreeScore}
& \multicolumn{2}{c}{ATC} 
& \multicolumn{2}{c}{ProjNorm}
\\ \cmidrule(l){3-14}
& & $R^2$ & $\rho$ & $R^2$ & $\rho$ & $R^2$ & $\rho$ & $R^2$ & $\rho$ & $R^2$ & $\rho$ & $R^2$ & $\rho$
\\ \midrule
\multirow{5}{*}{CIFAR10} 
& ResNet18 & 0.839 & 0.953 & 0.847 & 0.981 & 0.872 & 0.983 & 0.556 & 0.871 & 0.860 & 0.983 & \textbf{0.962} & \textbf{0.992} \\
& ResNet50 & 0.784 & 0.950 & 0.935 & 0.993 & 0.946 & \textbf{0.994} & 0.739 & 0.961 & 0.949  & \textbf{0.994} & \textbf{0.951} & 0.991 \\ 
& VGG11 & 0.826 & 0.876 & 0.929 & 0.988 & 0.927 & 0.989 & 0.907 & 0.989 & \textbf{0.931} & 0.989 & 0.891 & \textbf{0.991} \\ 
\cmidrule(l){2-14} 
&{\color{Blue}{Average}} & {\color{Blue}{0.816}} & {\color{Blue}{0.926}} & {\color{Blue}{0.904}} & {\color{Blue}{0.987}}  & {\color{Blue}{0.915}}  & {\color{Blue}{0.989}} & {\color{Blue}{0.734}} & {\color{Blue}{0.940}} & {\color{Blue}{0.913}} & {\color{Blue}{0.989}} & {\color{Blue}{\textbf{0.935}}} & {\color{Blue}{\textbf{0.991}}} \\ 
\midrule
\multirow{5}{*}{CIFAR100} 
& ResNet18 & 0.903 & 0.955  & 0.917 & 0.958 & 0.879 & 0.938 & 0.939 & 0.969 & 0.934 & 0.966 & \textbf{0.978} & \textbf{0.989} \\
& ResNet50   & 0.916 & 0.963  & 0.932 & 0.986 & 0.905 & 0.980 & 0.927 & 0.985 & 0.947 & 0.989 & \textbf{0.984} & \textbf{0.993} \\
& VGG11  & 0.780 & 0.945 & 0.899 & 0.981 & 0.880  & 0.979 & 0.919 & 0.988 & 0.935 & 0.986 & \textbf{0.953} & \textbf{0.993} \\ 
\cmidrule(l){2-14} 
& {\color{Blue}{Average}} &  {\color{Blue}{0.866}} &   {\color{Blue}{0.954}} &   {\color{Blue}{0.916}} &   {\color{Blue}{0.975}}  &   {\color{Blue}{0.888}}  &   {\color{Blue}{0.966}} &   {\color{Blue}{0.928}} &   {\color{Blue}{0.981}} &   {\color{Blue}{0.939}} &   {\color{Blue}{{0.980}}} &   {\color{Blue}{\textbf{0.972}}} &   {\color{Blue}{\textbf{0.992}}} \\ 
\midrule
\multirow{4}{*}{MNLI} 
& BERT & - & - & 0.516 & 0.671 & 0.533  & \textbf{0.734} & 0.318 & 0.524 & 0.524 & 0.699 & \textbf{0.585} & 0.664 \\
& RoBERTa & - & - & 0.493 & 0.727 & 0.498 & 0.734 & 0.499 & 0.762 & 0.519 & 0.734 & \textbf{0.621} & \textbf{0.790} \\
\cmidrule(l){2-14} 
& {\color{Blue}{Average}}& 
{\color{Blue}-}&
{\color{Blue}-} & 
{\color{Blue}{0.505}}&  {\color{Blue}{0.699}}& {\color{Blue}{0.516}}& {\color{Blue}\textbf{{0.734}}}& {\color{Blue}{0.409}}& {\color{Blue}{0.643}}& {\color{Blue}{0.522}}& {\color{Blue}{0.717}}& {\color{Blue}\textbf{{0.603}}} & {\color{Blue}{0.727}}  \\ 
\bottomrule
\end{tabular}
\vspace{-0.15in}
\end{table*}

\begin{figure*}[t!]
    \centering
    \subfigure[ConfScore]{\includegraphics[width=.28\textwidth]{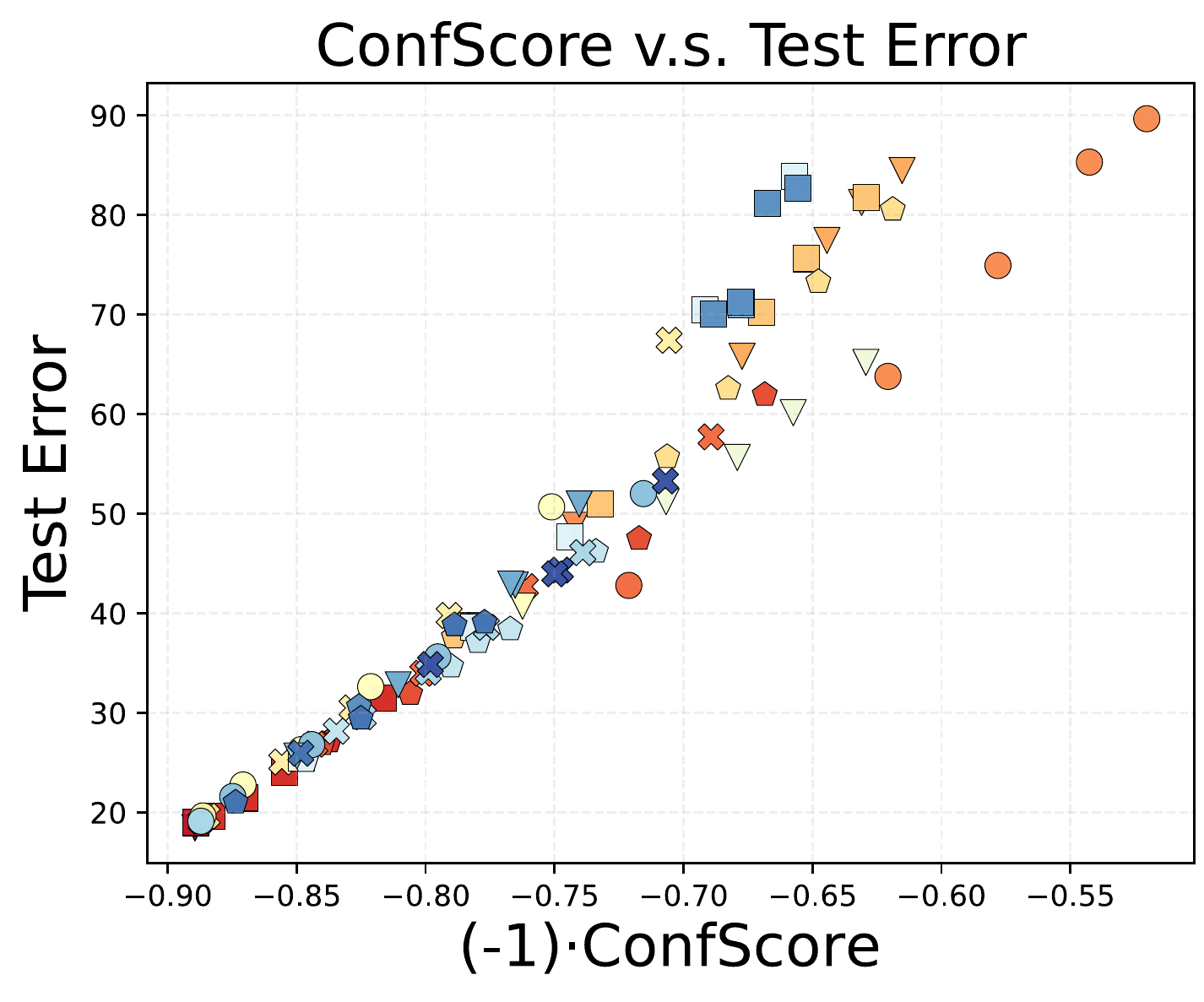}}
    \subfigure[ATC]{\includegraphics[width=.28\textwidth]{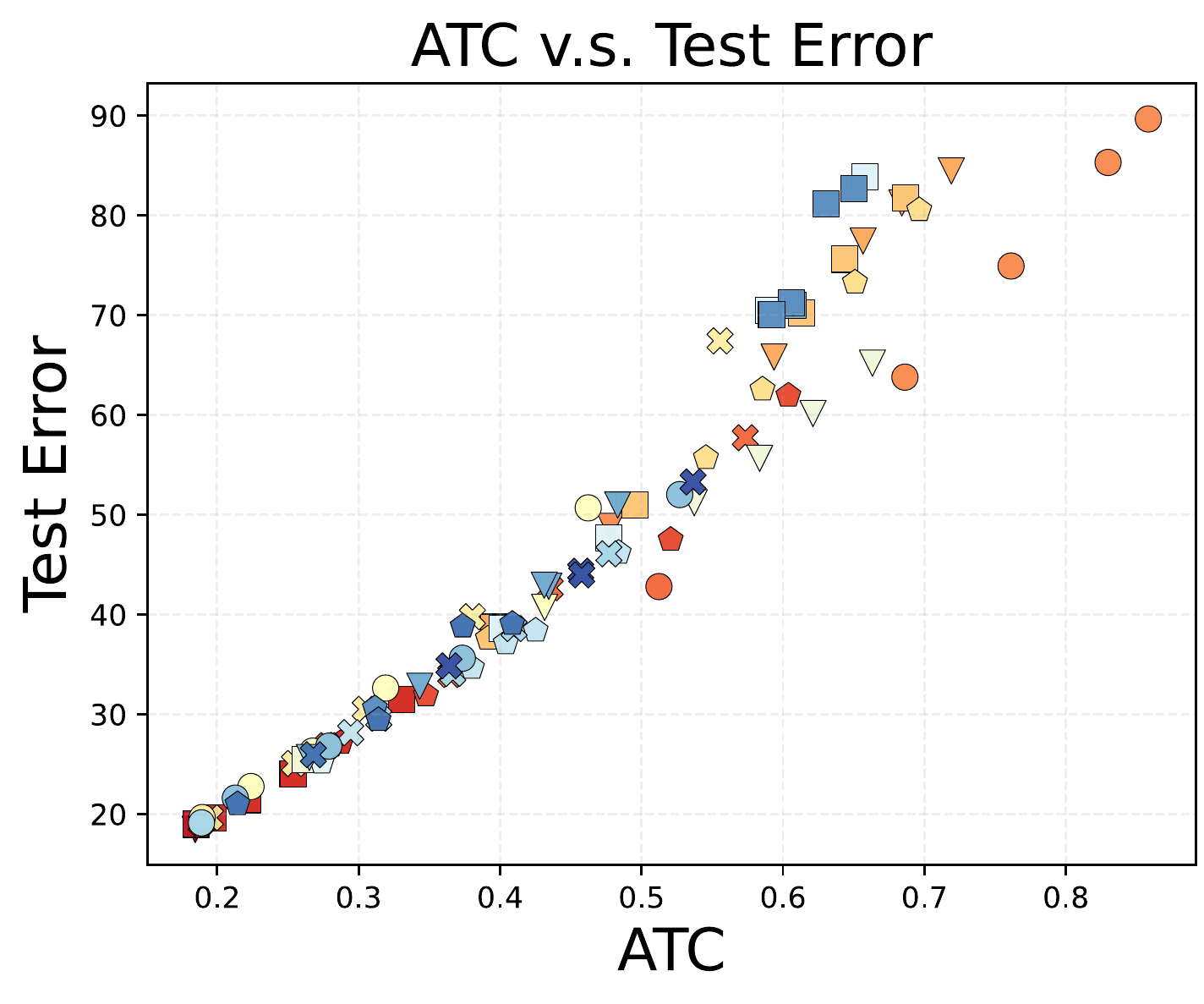}}
    \subfigure[ProjNorm]{\includegraphics[width=.373\textwidth]{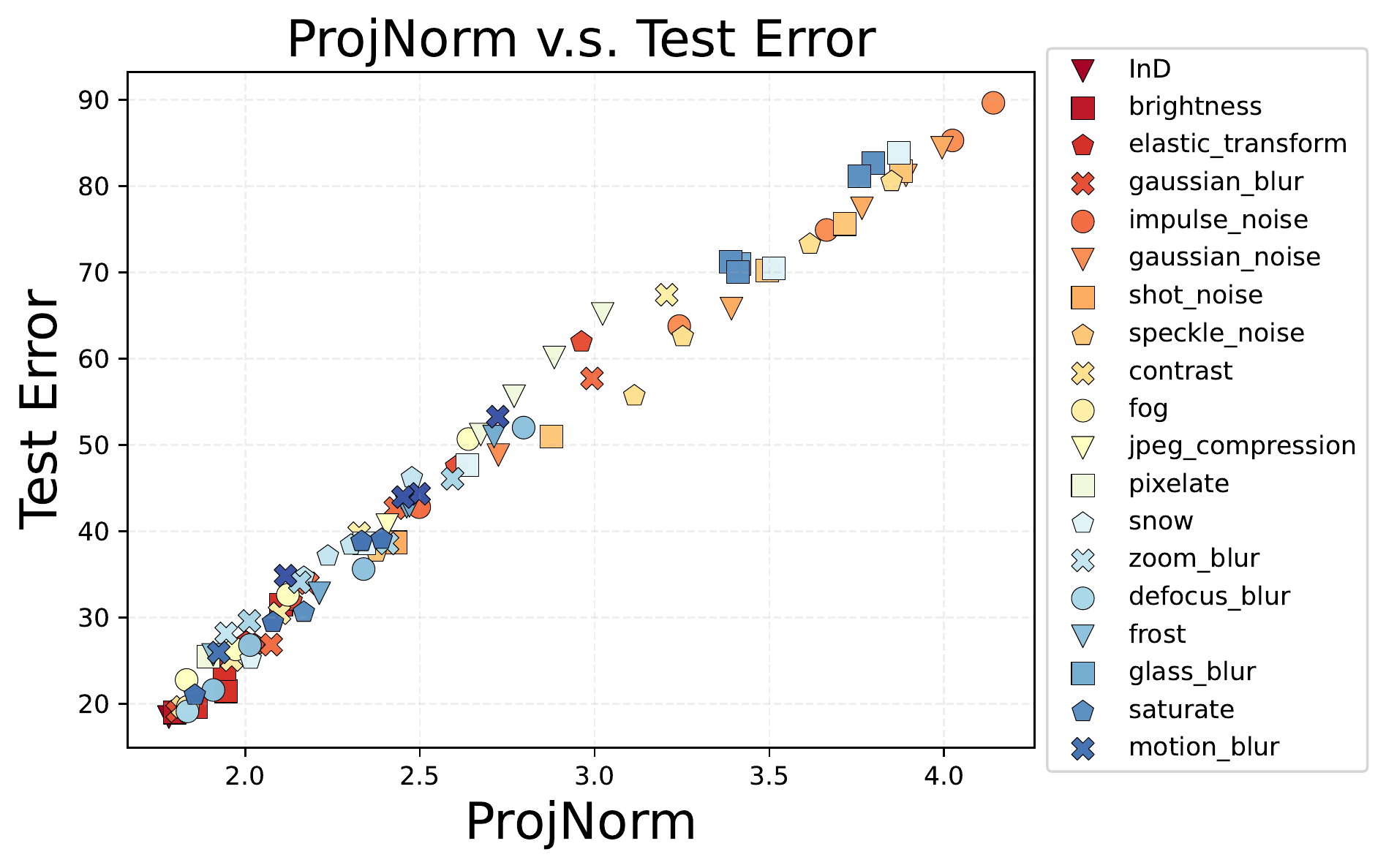}}
    \vspace{-0.15in}
    \caption{\textbf{Generalization prediction versus test error on CIFAR100 with ResNet50.} Compare out-of-distribution  prediction performance of  ConfScore (left), ATC (middle), and \pj{} (right) on CIFAR100. We plot the actual test error and the method prediction on each OOD dataset.
    Each point represents one InD/OOD dataset, and points with the same color and marker shape are the same corruption but with different severity levels.
    }
    \label{fig:compare-main}
    \vspace{-0.15in}
\end{figure*}

\section{Experimental Results}\label{sec:exp}
We evaluate the $\pj$ algorithm on several out-of-distribution datasets in the vision and language domains. We first compare our method with existing methods and demonstrate its effectiveness (Section~\ref{subsec:exp-main-results}). Next, we study the sensitivity of $\pj$ to hyperparameters and data set size~(Section~\ref{subsec:exp-analysis}). Finally, we show that the errors of $\pj$ are nearly uncorrelated with those of existing methods (Section~\ref{subsec:stat}), and use this to construct an ensemble method that is even more accurate than $\pj$ alone.

\vspace{-0.05in}
\textbf{Datasets.} We evaluate each method we consider on the image classification tasks CIFAR10, CIFAR100~\citep{krizhevsky2009learning} and the natural language inference task MNLI~\citep{williams2017broad}. To generate out-of-distribution data, for the CIFAR datasets we use the ``common corruptions'' of \citet{hendrycks2019benchmarking}, CIFAR10-C and CIFAR100-C, spanning 18 types of corruption with 5 severity levels. 
For MNLI, we use BREAK-NLI~\citep{glockner2018breaking}, EQUATE~\citep{ravichander2019equate}, HANS~\citep{mccoy2019right}, MNLI-M, MNLI-MM, SICK~\citep{marelli2014sick}, SNLI~\citep{bowman2015large}, STRESS-TEST~\citep{naik2018stress}, and SICK~\citep{marelli2014sick} as out-of-distribution datasets, with STRESS-TEST containing 5 sub-datasets. These OOD datasets include shifts such as swapping words, word overlap, length mismatch, etc. (More comprehensive descriptions of these datasets can be found in \citet{zhou2020curse}.)

\textbf{Methods.}  We consider five existing methods for predicting OOD error: \textit{Rotation Prediction}~(Rotation)~\citep{deng2021does}, \textit{Averaged Confidence}~(ConfScore)~\citep{hendrycks2016baseline}, \textit{Entropy}~\citep{guillory2021predicting}, \textit{Agreement Score}~(AgreeScore)~\citep{madani2004co, nakkiran2020distributional, jiang2021assessing}, and \textit{Averaged Threshold Confidence}~(ATC)~\citep{garg2021leveraging}. 
Rotation evaluates rotation prediction accuracy on test samples to predict test error. AgreeScore measures agreement rate  between two independently trained classifiers on unlabeled test data. ConfScore, Entropy, and ATC predict test error on OOD data based on softmax outputs of the model. 
See Appendix~\ref{sec:methods-detail-appendix} for more details of these existing methods.

\begin{figure*}[t!]
    \centering
    \subfigure[ResNet18.]{\includegraphics[width=.295\textwidth]{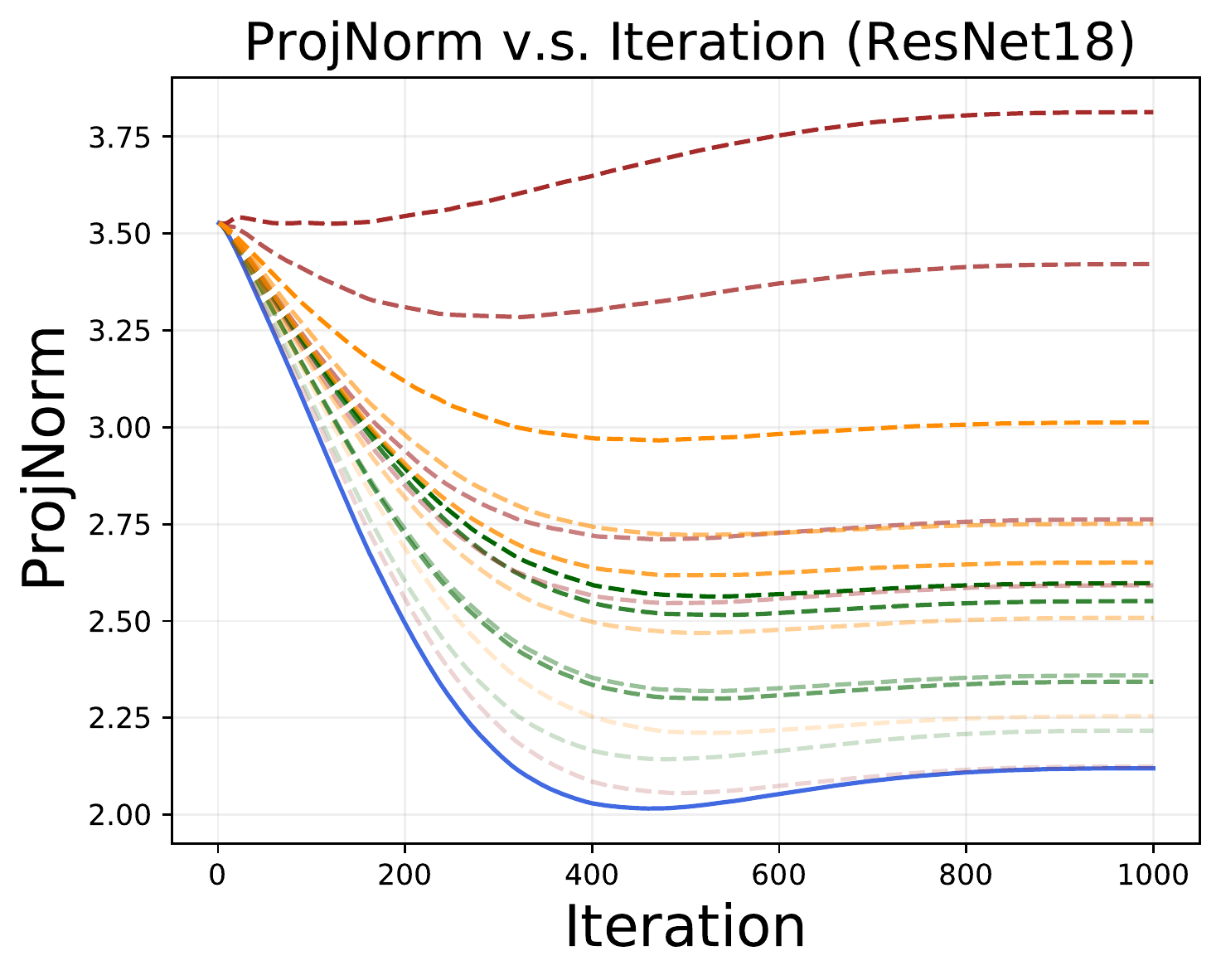}}
    \subfigure[ResNet50.]{\includegraphics[width=.29\textwidth]{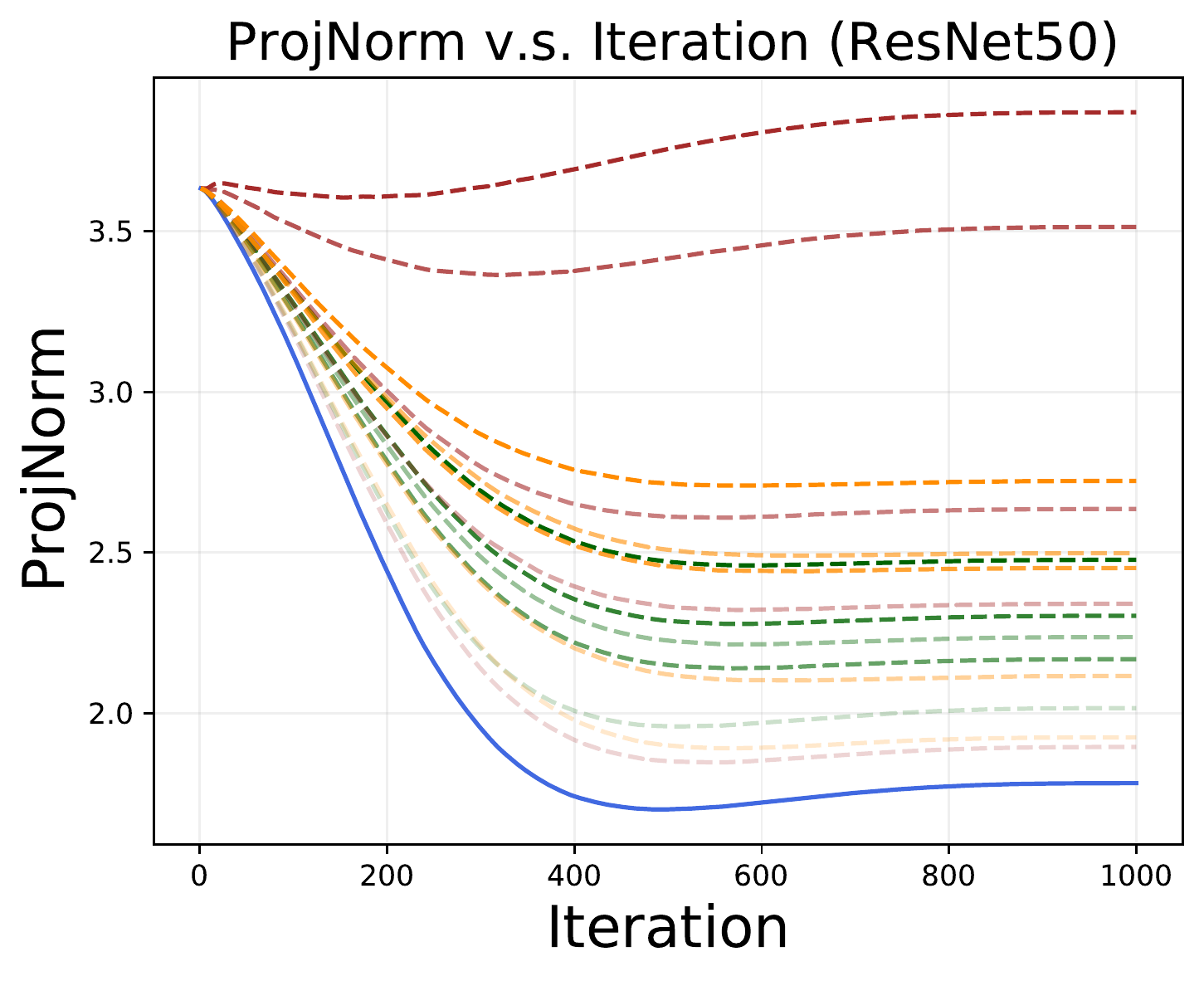}}
    \subfigure[VGG11.]{\includegraphics[width=.385\textwidth]{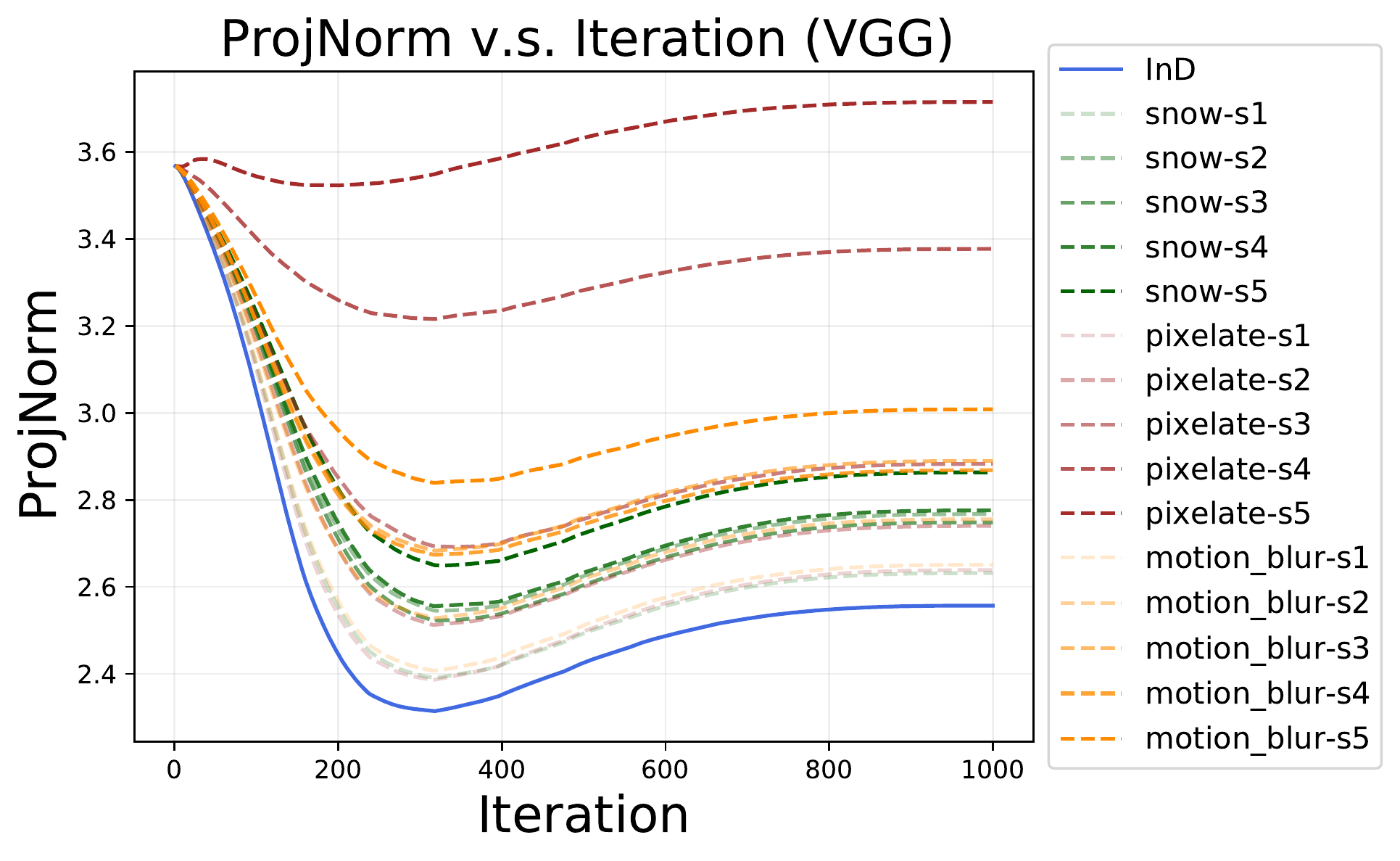}}
    \vspace{-0.15in}
    \caption{\textbf{Evaluation of {\normalfont\pj{}} as training progresses on CIFAR100.} We visualize how the \pj{} changes as the number of training iterations increases for (a) ResNet18, (b) ResNet50, and (c) VGG11 on CIFAR100. We show results on three corruptions (snow, pixelate, and motion blur) as well as the in-distribution dataset (InD). For complete results, see Appendix~\ref{sec:additional-exp-results}.}
    \label{fig:projnorm-analysis-main}
    \vspace{-0.2in}
\end{figure*}

\begin{table*}[ht]
\centering
\caption{\textbf{Hyperparameter sensitivity of {\normalfont\pj}}. We vary the number of ``pseudo-label projection'' training iterations ($T$) and the number of test samples ($m$) of \pj, and evaluate the $R^{2}$ statistic. The performance of \pj{} is relatively stable, but decreases for sample sizes below $1000$.}
\vspace{0.05in}
\label{table:table-sensitivity}
\begin{tabular}{@{}ccccc|ccccc@{}}
\toprule
\multirow{2}{*}{Dataset} 
& \multicolumn{3}{c}{Training iterations ($m$=1000)}
& \multicolumn{1}{c}{}
& \multicolumn{5}{c}{Test samples (set $T$=$m/10$)}
\\ \cmidrule(l){2-10}
& $T$=1000 & $T$=500 & $T$=200 & & $m$=5000 & $m$=2000 & $m$=1000 & $m$=500 & $m$=100
\\ \midrule
\multirow{1}{*}{CIFAR10} 
 &  0.962 &  0.985 & 0.983 & &  0.973 & 0.977 &  0.980 & 0.946 & 0.784 \\
\midrule
\multirow{1}{*}{CIFAR100} 
 &  0.978 &  0.980 &  0.959 & & 0.972 &  0.942 & 0.942 & 0.903 & 0.466 \\
\bottomrule
\end{tabular}
\vspace{-0.15in}
\end{table*}

\vspace{-0.05in}
\textbf{Pre-trained models and training setup.} We use pre-trained models and fine-tune on the in-distribution training dataset.  For image classification, we use ResNet18, ResNet50~\citep{he2016deep}, and VGG11~\citep{simonyan2014very}, all pre-traineded on ImageNet~\citep{deng2009imagenet}. We consider BERT~\citep{devlin2018bert} and RoBERTa~\citep{liu2019roberta}
for the natural language inference task, fine-tuned on the MNLI training set. For the CIFAR datasets, we fine-tune using SGD with learning rate $10^{-3}$, momentum $0.9$, and cosine learning rate decay~\citep{loshchilov2016sgdr}. For MNLI, we use AdamW~\citep{loshchilov2017decoupled} with learning rate $2\cdot 10^{-5}$ and linear learning rate decay. For computing \pj, we apply the same optimizer as fine-tuning on each dataset. The default number of training iterations for \pj{} is $1000$. For further details, see Appendix~\ref{sec:appendix-exp-detail}.

\textbf{Metrics.} To evaluate performance, we compute the correlation between the predictions and the actual test accuracies across the OOD test datasets, using $R^{2}$ and rank correlation (Spearman's $\rho$). We also present scatter plots to compare different methods qualitatively.

\vspace{-0.05in}
\subsection{Main results: comparison of all methods}\label{subsec:exp-main-results}
We summarize results for all methods and datasets in Table~\ref{table:main-table}. We find that \pj{} achieves better performance than existing methods in most settings. On CIFAR100, \pj{} achieves an averaged $R^{2}$ of $0.972$, while the second-best method (ATC) only obtains $0.939$. The prediction performance of \pj{} is also more stable than other methods. For Spearman's $\rho$ on CIFAR10/100, ATC varies from $0.966$ to $0.994$ and AgreeScore varies from $0.871$ to $0.989$. In contrast, \pj{} achieves $\rho>0.989$ in all settings.

\vspace{-0.05in}
We also provide scatter plots on CIFAR100 in Figure~\ref{fig:compare-main}. \pj's better performance primarily comes from better predicting harder OOD datasets. While all methods do well when the test error is below $40\%$, ConfScore and ATC often underpredict the larger test errors. In contrast, \pj{} does well even for errors of $90\%$. In Section~\ref{sec:linearmodel}, we argue that this is because \pj{} better captures directions ``orthogonal'' to the training set. 
Scatter plots on other methods/datasets can be found in Appendix~\ref{sec:additional-exp-results}.

\subsection{Sensitivity analysis and ablations}\label{subsec:exp-analysis}
We investigate the following four questions for \pj: (1)~To improve computational efficiency, can we use fewer training iterations to compute \pj{} while still achieving similar prediction performance? (2)~How many test samples $m$ are needed for \pj{} to perform well? 
(3)~How important is the choice of reference model?  
(4)~What role do  the pseudo-labels play in \pj's performance?

\vspace{-0.05in}
\textbf{Training iterations.} We first visualize how \pj{} changes with respect to the number of training iterations. 
We evaluate \pj{} at training steps from $1$ to $1000$ and display results for snow, pixelate, and motion blur corruptions in Figure~\ref{fig:projnorm-analysis-main} (see Figure~\ref{fig:projnorm-analysis-appendix} for results on all corruptions). For most corruptions in CIFAR10-C and CIFAR100-C, we find that \pj{} initially decreases with more training iterations, then slowly increases and before converging.  
Importantly, from Figure~\ref{fig:projnorm-analysis-main} we see that the iteration count usually does not affect the ranking of different distribution shifts.  
Table~\ref{table:table-sensitivity} displays $R^2$ values for different iteration counts $T$, and shows that \pj{} still achieves good performance with as few as $200$ training iterations. 

\vspace{-0.05in}
\textbf{Sample size.} We next consider the effect of the number of test samples $m$, varying $m$ from $5000$ to $100$ (from a default size of $10000$).  
Results are shown in Table~\ref{table:table-sensitivity}, where we observe that \pj{} achieves reasonable performance down to around $1000$ to $2000$ samples, but performs poorly below that. 
In general, we conjecture that \pj\,performs well once the number of samples is large enough for fine-tuning 
to generalize well.

\textbf{Reference model.} We consider directly using $\bthetah$ as the reference model, rather than fine-tuning a new one. 
As shown in Table~\ref{table:table-sensitivity-thetahat-appendix} and Figure~\ref{fig:compare-appendix-cifar10-thetahat} in the appendix, using $\bthetah_{\text{ref}}=\bthetah$ achieves similar performance compared to the default version of \pj{} on CIFAR10. 

\textbf{Pseudo-labels.} Finally, we investigate the role of pseudo-labels in our method. We modify \textbf{Step 2} of \pj{} by training $\bthetat$ using the \textit{ground truth labels} of the OOD data.  From Table~\ref{table:table-sensitivity-truelabel-appendix} and  Figure~\ref{fig:compare-appendix-cifar10-truelabel}, we find that \pj{} with pseudo-label performs much better than \pj{} with ground truth label, which suggests that pseudo-labeling is an essential component in \pj.

\subsection{Correlation analysis}
\label{subsec:stat}
In this section, we provide a short statistical analysis of using different measurements to predict test error.
We focus on the CIFAR100 dataset and Resnet18 architecture. 
We show that \pj{} captures signal  that existing methods fail to detect, so that ensembling with the existing approaches leads to even better performance. 

For each method, we first compute residuals when predicting the test error by performing simple linear regression.
Then we compute the correlation between the residual errors for each pair of methods.

\begin{table}[t!]
\vspace{-0.1in}
    \caption{Correlation of residuals of regressing test error against different  measurements  with CIFAR100 and ResNet18.}
    \vspace{0.075in}
    \centering
    {\renewcommand{\arraystretch}{1.2}
    \begin{tabular}{l|c|c|c|c|c}
    \hline
    \toprule
    ~ & Ent. & ConfS. & ATC & Rota. &Proj.\\ \hline
    Agree.S. & ~0.85 & ~0.87 & ~0.84 & ~0.80 & ~0.05\\ \hline
    Ent. & - & ~0.98 & ~0.93 & ~0.67 & -0.07\\ \hline
    ConfS. & - & - & ~0.98 & ~0.67 & -0.14\\ \hline
    ATC & - & - & - & ~0.65 & -0.19\\ \hline
    Rota. & - & - & - & - & ~0.03 \\
    \bottomrule
    \hline
    \end{tabular}
    }
    \vspace{-0.15in}
    \label{table:cor}
\end{table}

\noindent
We see from Table \ref{table:cor} that the correlation among all existing methods is high: strictly larger than $0.6$.
The correlations among ConfScore, ATC and Entropy are especially high ($>\!0.9$) suggesting they are almost equivalent approaches.
This high correlation is unsurprising since these methods are all different ways of manipulating the logits.

In contrast, the correlation between \pj{} and existing methods is always less than $0.05$, and often negative. Intriguingly, while the correlations among existing methods are positive, \pj{} sometimes has negative correlation with existing methods.
This means \pj{} underestimates the test error when other methods overestimate it.

The low correlation implies that \pj{} provides very different signal compared to existing methods and suggests a natural ensembling approach for improving performance further.
Indeed, if we average \pj{} and ATC (the second best method), normalized by standard deviation, we further improve $R^2$ from 0.978 (using \pj{} only) to 0.982 (averaging \pj{} and ATC).

%% file: sec_linearmodel.tex
\section{Insights from an Overparameterized Linear Model}\label{sec:linearmodel}
In this section, we provide some insights for Projection Norm by studying its behavior on  high-dimensional linear models.
We demonstrate an extreme example where Projection Norm has a qualitative advantage over other methods such as Confidence Score.
We also show that Projection Norm is tied to an upper bound on the test loss under certain assumptions, which we empirically validate on the CIFAR10 dataset.

We consider a linear model with covariates $\bx\in\R^d$ and response $y\in\R$. Let $\bX\in\R^{n\times d}$ and $\by\in\R^n$ denote the training set $\Dtrain = \{(\bx_i, y_i)\}_{i=1,\dots, n}$. 
We focus on the $d \gg n$ regime and take $\bthetah$ to be the minimum-norm interpolating solution,
\begin{equation}
  \bthetah = \min_{\bX\btheta=\by} \|\btheta\|_2 = \bX^\sT (\bX\bX^\sT)^{-1}\by.
\end{equation}
Let $\bXt\in\R^{m\times d}$ denote the out-of-distribution test covariates and $\byt\in\R^m$ the corresponding ground truth response vector.
Our goal is to estimate the test loss
\begin{equation}
 \label{eqn:testloss}
  \tl=\frac{1}{m}\|\bXt\bthetah-\byt\|_2^2
\end{equation}
using only $\bX$, $\by$, and $\bXt$---that is, without having access to the ground truth response $\byt$. 

Note that most existing methods in Section~\ref{sec:exp} (such as the Confidence Score) only look at the outputs of the model $\bthetah$. 
In this linear setting, this corresponds to the vector $\bXt\bthetah$. 
We show (Section~\ref{sec:bstar}) that any method with this property has severe limitations, while the linear version of Projection Norm overcomes these. 
Then we present results connecting this linear version of Projection Norm to the test loss (Section~\ref{sec:spectral}).

\subsection{Motivating Projection Norm}\label{sec:bstar}
To analyze the linear setting, we assume that the responses $\by$ and $\byt$ are noiseless and differ only due to covariate shift:
\begin{assumption}[Covariate shift]\label{ass:btstar}
There exists a ground truth $\btheta_\star\in\R^d$ relating the covariates and responses such that $\bX\btheta_\star=\by$ and $\bXt\btheta_\star=\byt$, i.e., for both the in-distribution training data and out-of-distribution  test data.
\end{assumption}

\vspace{-0.075in}
By Assumption~\ref{ass:btstar}, the minimum-norm solution reduces to
\vspace{-0.05in}
\begin{equation}
  \bthetah = \bX^\sT (\bX\bX^\sT)^{-1}\bX\btheta_\star = \bP\btheta_\star,
\end{equation}
where $\bP$ is defined as the orthogonal projection matrix onto the row space of $\bX$, i.e.,  $\bP=\bX^\sT (\bX\bX^\sT)^{-1}\bX$.
Similarly, let $\bPt$ be the projection matrix for the row space of $\bXt$. 
Using the fact that $\bthetah = \bP\btheta^*$ and $\byt = \bXt\btheta^*$, the test loss in \eqref{eqn:testloss} can be written as
\vspace{-0.05in}
\begin{equation}\label{eqn:testloss2}
  \tl = \frac{1}{m}\|\bXt(\bI - \bP)\btheta_\star\|_2^2.
\end{equation}
From Eq.~\eqref{eqn:testloss2}, we see that the test loss depends on the portion of $\btheta_\star$ that is orthogonal to $\bX$---i.e., in the span of $\bXt$ but not $\bX$.
Now consider any method that depends only on the model output $\bXt\bthetah=\bXt\bP\btheta_\star$---such a method is not sensitive to this orthogonal component at all! 
We can see this concretely through the following setting with Gaussian covariates:
\begin{align}
    \bx_i&\overset{i.i.d.}{\sim}\cN\left(\bzero, \left[ \begin{array}{cc} \bI_{d_1} & \bzero \\ \bzero & \bzero \end{array} \right] \right),\;i=1,\ldots,n,\label{eqn:toyx}\\
    \bxt_j&\overset{i.i.d.}{\sim}\cN\left(\bzero, \left[ \begin{array}{cc} \bI_{d_1} & \bzero \\ \bzero & \sigma^2 \bI_{d_2} \end{array} \right] \right),\;j=1,\ldots,m.\label{eqn:toyxt}
\end{align}
Here we decompose the $d$-dimensional covariate space into two orthogonal components $d=d_1+d_2$, where the last $d_2$ components appear only at test time. We display empirical results for this distribution in Figure~\ref{fig:toy} (see Appendix~\ref{sec:appendix-toy} for full experimental details).
Methods that depend only on the model outputs---such as the confidence score---are totally insensitive to the parameter $\sigma$.

\vspace{-0.05in}
\textbf{Advantage of projection norm.} We next define a linear version of the Projection Norm:
\vspace{-0.05in}
\begin{equation}\label{eqn:pjlinear}
  \pjl = \|\bthetah - \bPt\bthetah\|_2.
\vspace{-0.05in}
\end{equation}
This computes the difference between the reference model $\bthetah$ and a projected model $\bPt\bthetah$.
Before justifying \pjl~ as an adaptation of \pj, we first examine its performance on the example introduced above.

In particular, we show that \pjl~ has the right dependence on $\sigma$ whereas ConfScore does not.
In \pjl, the less overlap $\bXt$ has with $\bX$, the smaller $\bPt\bthetah$ will be, so 
the quantity in Eq.~\eqref{eqn:pjlinear} does track the orthogonal component of $\bXt$. Results for $\pjl$, also shown in Figure~\ref{fig:toy}, confirm this. In contrast to the confidence score, \pjl\,does vary with $\sigma$, better tracking the test error. 

We next explain why $\pjl$ is the linear version of $\pj$ as defined in Eq.~\eqref{eqn:pjnn} (Section~\ref{sec:method}).
To draw the connection, first note that the projection step $\bPt\bthetah$ is equivalent to finding the minimum $\ell_2$-norm solution of 
\vspace{-0.03in}
\begin{equation}\label{eqn:proplin}
  \min_{\btheta} \|\bXt\btheta - \bXt\bthetah\|_2^2.
\vspace{-0.05in}
\end{equation}
In the linear setting, the minimum-norm solution can be obtained by initializing at $\btheta_{0}=\mathbf{0}$ and performing gradient descent to convergence~\cite{wilson2017marginal, hastie2020surprises}. 
If we write $\flin(\bx; \btheta) = \<\bx, \btheta\>$, then Eq.~\eqref{eqn:proplin} can be equivalently written as
\vspace{-0.04in}
\begin{equation}\label{eqn:propnn}
  \min_{\btheta}\, \sum_{j=1}^m \Big(\flin(\bxt_j; \btheta) - \flin(\bxt_j; \bthetah\,)\Big)^2.
\vspace{-0.04in}
\end{equation}
In other words, it minimizes the squared loss relative to the pseudo-labels $\flin(\bxt_j, \btheta)$.
The metric \pjl\,is thus the squared difference between the original ($\bthetah$) and pseudo-labeled model ($\bthetat=\bPt\bthetah$) in parameter space, akin to the \pj{} in the non-linear setting. 
Note that in this linear setting, there is no distinction between $\bthetah_\text{ref}$ and $ \bthetah$ as in Section \ref{sec:method}.

\begin{figure}[t]
    \centering
    \includegraphics[width=0.42\textwidth]{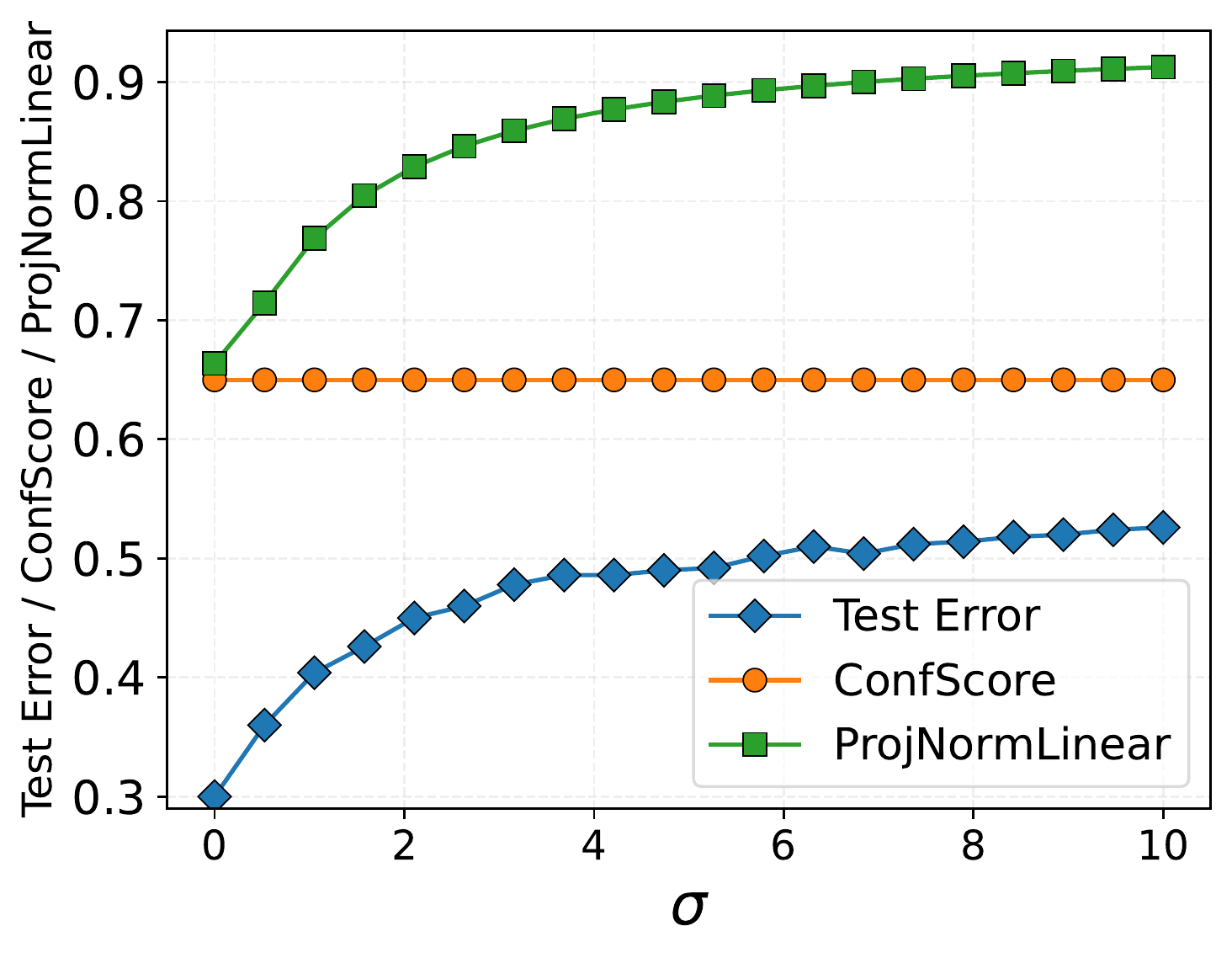}
    \vspace{-0.2in}
    \caption{A synthetic binary classification experiment with data distributions defined in Eq.~\eqref{eqn:toyx} and Eq.~\eqref{eqn:toyxt}. The test error increases with $\sigma$, and the linearized version of ProjNorm tracks this, but the confidence score does not.}
    \label{fig:toy}
\end{figure}

\subsection{Analyzing projection norm}\label{sec:spectral}
To further explain why \pjl\,performs well, we connect it to an upper bound on the test loss, under assumptions 
that we will empirically investigate in Section~\ref{sec:ntkemp}.  
Our first assumption states that $\btheta_\star$ has the same complexity when projected onto the train and OOD test distributions.
\begin{assumption}[Projected norm]\label{ass:norm}
We assume that $\|\bP\btheta_\star\|_2 = \|\bPt\btheta_\star\|_2$.
\end{assumption}
\vspace{-0.05in}
Our second assumption is on the spectral properties of the covariance matrices:
\begin{assumption}[Spectral properties]\label{ass:spectral}
Write the eigendecomposition of the empirical training and test covariance as 
\vspace{-0.03in}
\begin{equation}
    \bSigma = \frac{1}{n}\bX^\sT\bX = \frac{1}{n}\sum_{i=1}^n \mu_i \bu_i\bu_i^\sT,
\end{equation}
\vspace{-0.03in}
\begin{equation}
    \bSigmat = \frac{1}{m}\bXt^\sT\bXt = \frac{1}{m}\sum_{i=1}^m \lambda_j \bv_j\bv_j^\sT,
\vspace{-0.05in}
\end{equation}
where $\mu_1\geq\dots\geq\mu_n$ and $\lambda_1\geq\dots\geq\lambda_m$. We assume there exists some constant $0<k<\min(m, n)$ such that
\begin{equation}
    \text{Span}\{\bu_1, \dots, \bu_k\} = \text{Span}\{\bv_1, \dots, \bv_k\}
\vspace{-0.075in}
\end{equation}
\vspace{-0.02in}
and 
\begin{equation}
\text{Span}\{\bu_{k+1}, \dots, \bu_{n}\} \,\cap\, \text{Span}\{\bv_{k+1}, \dots, \bv_{m}\} = \mathbf{0}.
\end{equation}
\end{assumption}
\vspace{-0.05in}
In other words, we assume the large eigenvectors of the train and OOD test covariates span a common subspace, while the small eigenvectors are orthogonal. 
Under these assumptions, we show that the $\tl$ is bounded by a (constant) multiple of $\pjl$.
\begin{proposition}\label{proposition:ntk-eigen}
Under Assumptions \ref{ass:btstar}, \ref{ass:norm}, and \ref{ass:spectral},
\begin{equation*}
    \frac{\lambda_{m}}{m} \leq \frac{\tl}{\pjl^{~2}} \leq \frac{\lambda_{k+1}}{m},
\end{equation*}
where $\lambda_m, \lambda_{k+1}$
are the $m$-th and $(k+1)$-th eigenvalue of the covariance matrix $\bXt^\sT\bXt/m$.
\vspace{-0.05in}
\end{proposition}
This offers mathematical intuition for the effectiveness of Projection Norm that we observed in Section~\ref{sec:exp}.

%% file: sec_ntkemp.tex
\vspace{-0.075in}
\subsection{Checking assumptions on linearized representations}\label{sec:ntkemp}
\begin{figure*}[h]
    \centering
    \subfigure[Eigenvector alignment $\bH$.\label{fig:ntk-heatmap-eigenvec}]{\includegraphics[width=.27\textwidth]{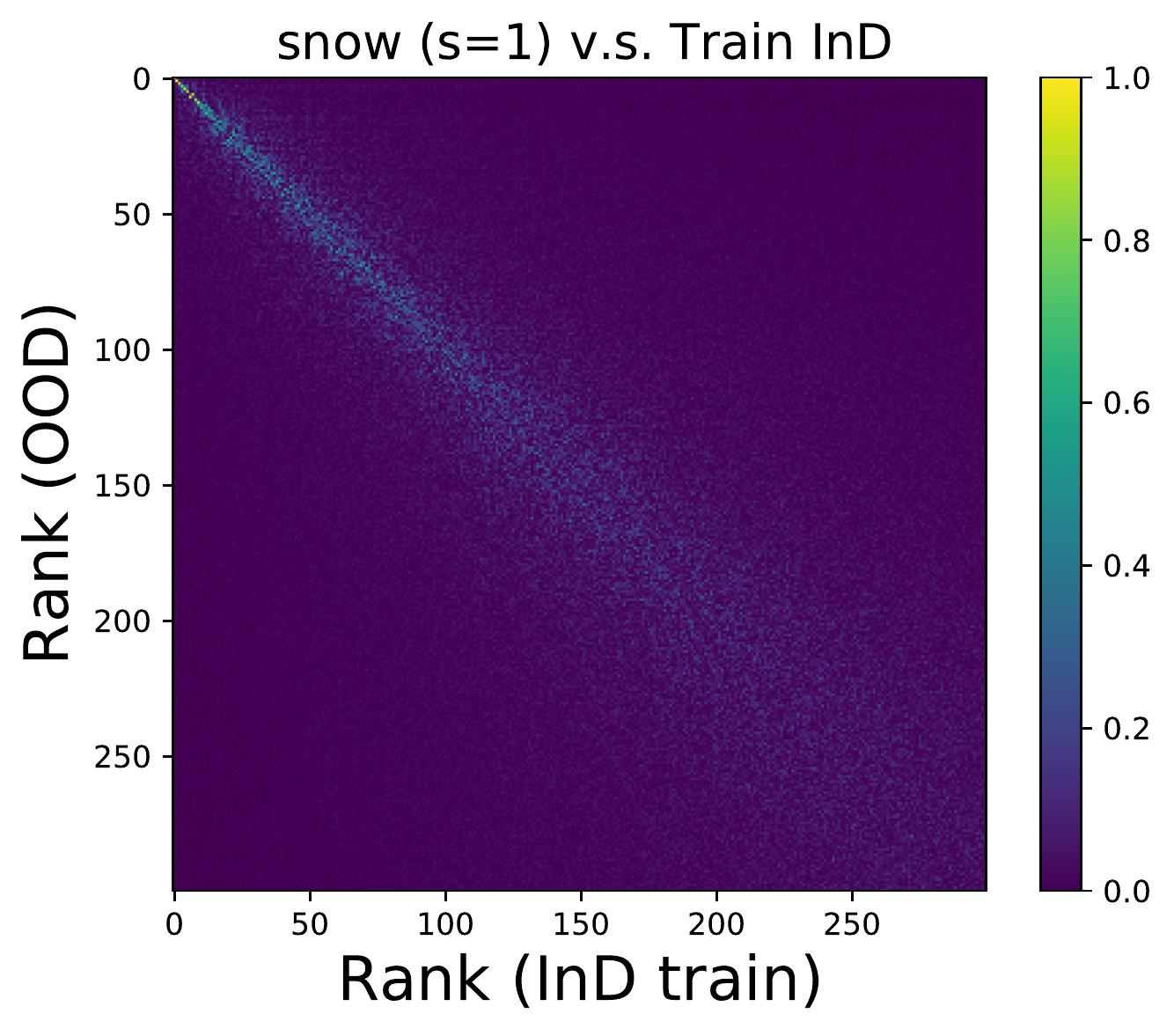}}
    \subfigure[Eigenvalue decay.\label{fig:ntk-heatmap-eigenvalue}]{\includegraphics[width=.3\textwidth]{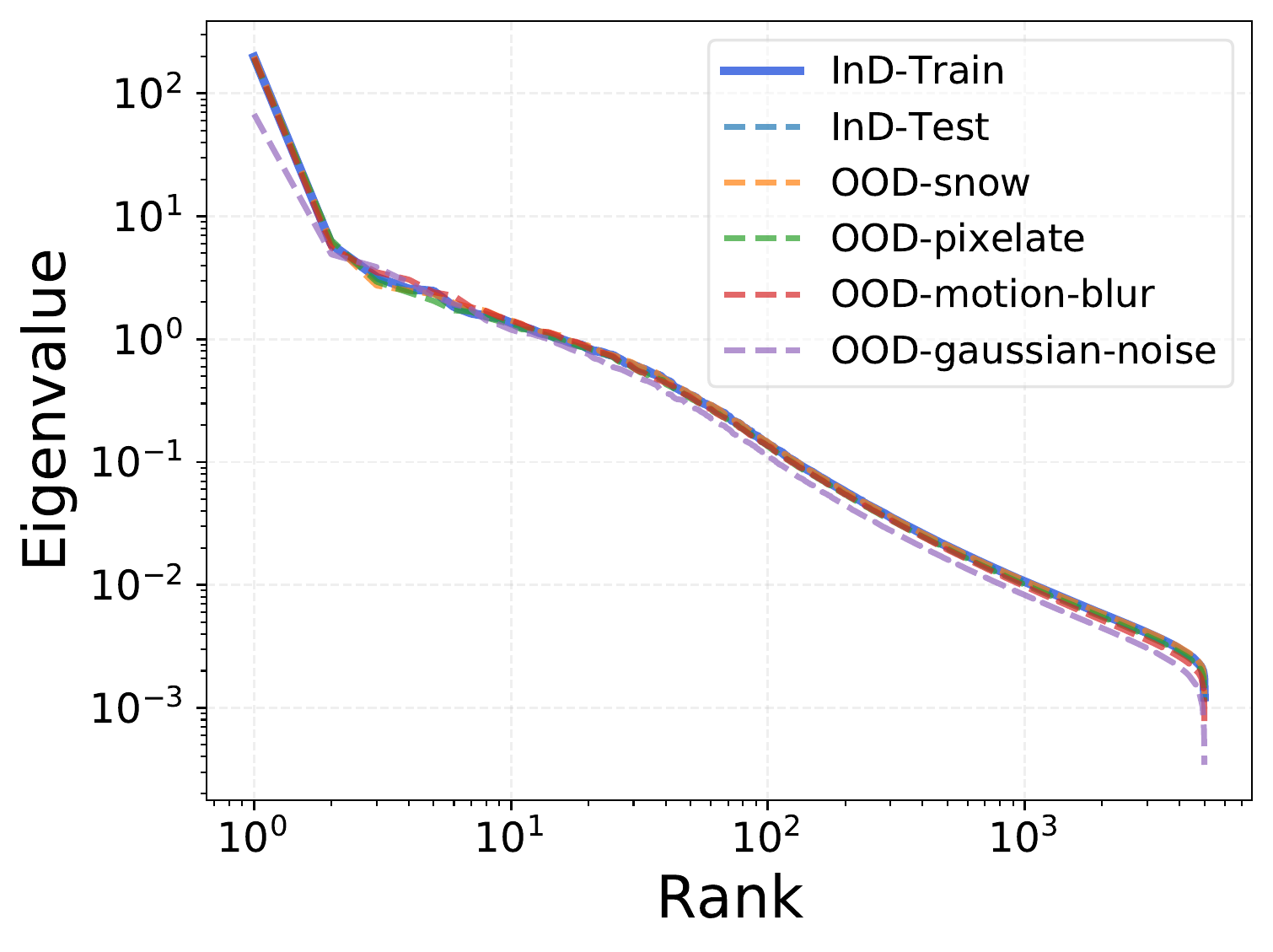}}
    \subfigure[\pjl-NTK predicts OOD error.\label{fig:ntk-heatmap-predict-error}]{\includegraphics[width=.38\textwidth]{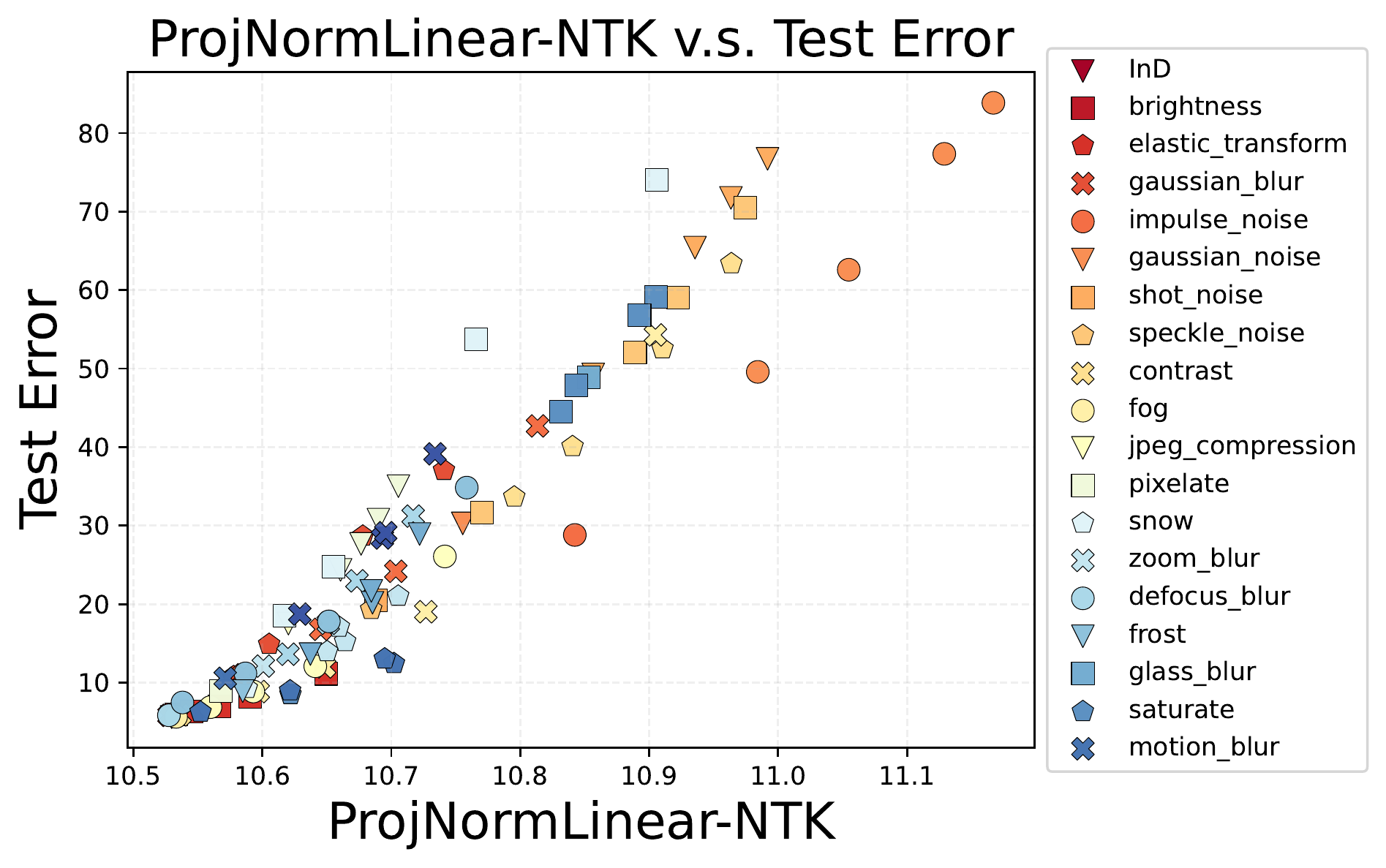}}
    \vspace{-0.15in}
    \caption{Experimental results on NTKs. (\textbf{a}) We visualize the alignment matrix $\bH\in\R^{300\times300}$ between top-300 eigenvectors of covariance matrices for in-distribution ($\{\bu_{1},\cdots,\bu_{300}\}$) and OOD (snow) ($\{\bv_{1},\cdots,\bv_{300}\}$) datasets, where $\bH_{ij}=|\langle \bv_{i}, \bu_{j}\rangle|$ for $i,j\in[n]$. (\textbf{b}) Eigenvalue decay of kernels on a $\log$-$\log$ scale, including in-distribution train, in-distribution test, snow, pixelate, motion blur, and Gaussian noise with severity 1 from CIFAR10-C. (\textbf{c}) Scatter plot of \pjl{} \textit{computed on NTK representations} versus true test error of \textit{model fine-tuned with SGD} on CIFAR10-C ($R^{2}=0.914$; $\rho=0.960$). See Appendix~\ref{sec:appendix-ntk} for more results on different corruptions/severities on CIFAR10-C.}
    \label{fig:ntk-heatmap}
    \vspace{-0.15in}
\end{figure*}

\begin{figure*}[t!]
    \centering
    \subfigure[ConfScore.]{\includegraphics[width=.32\textwidth]{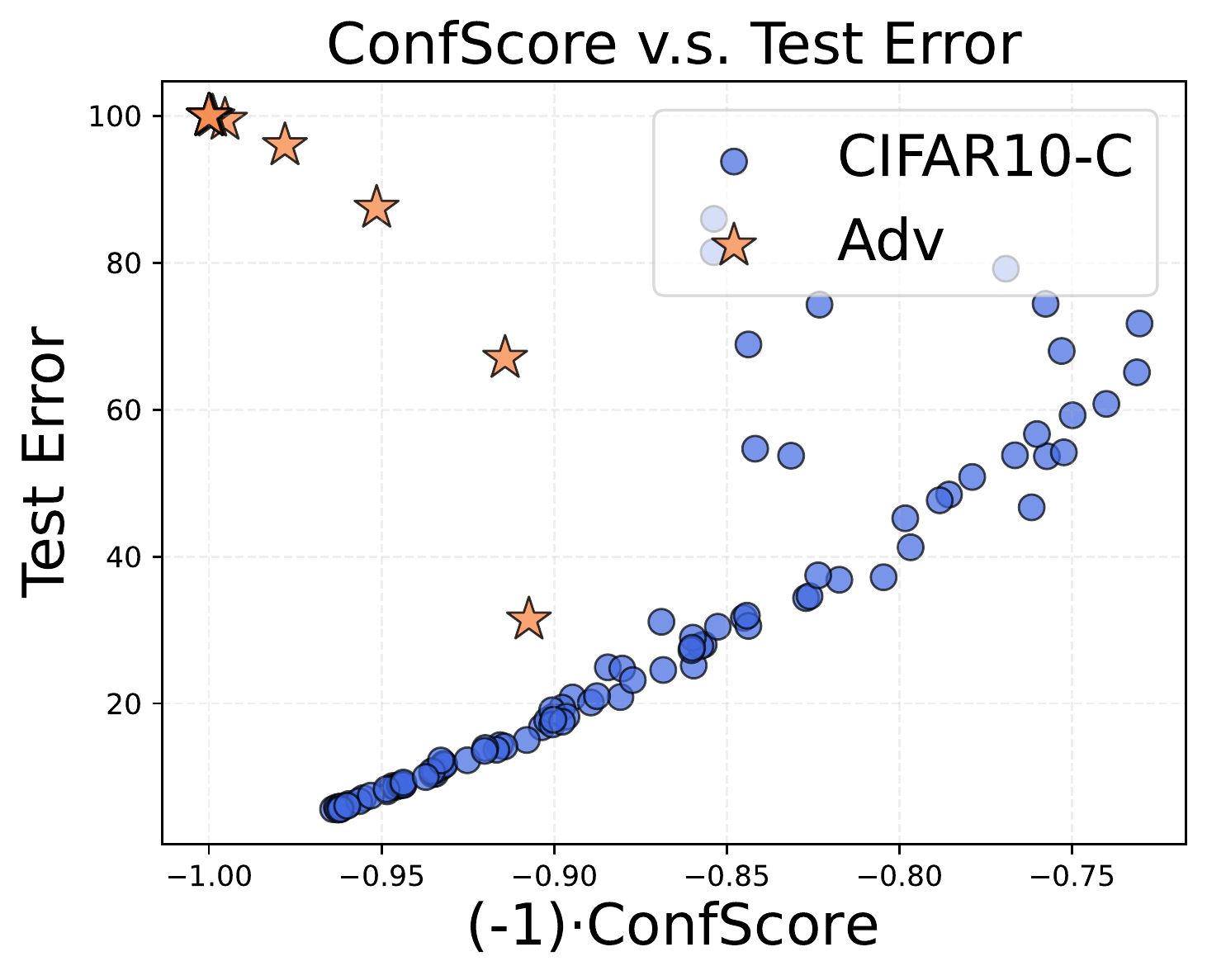}}
    \subfigure[ATC.]{\includegraphics[width=.32\textwidth]{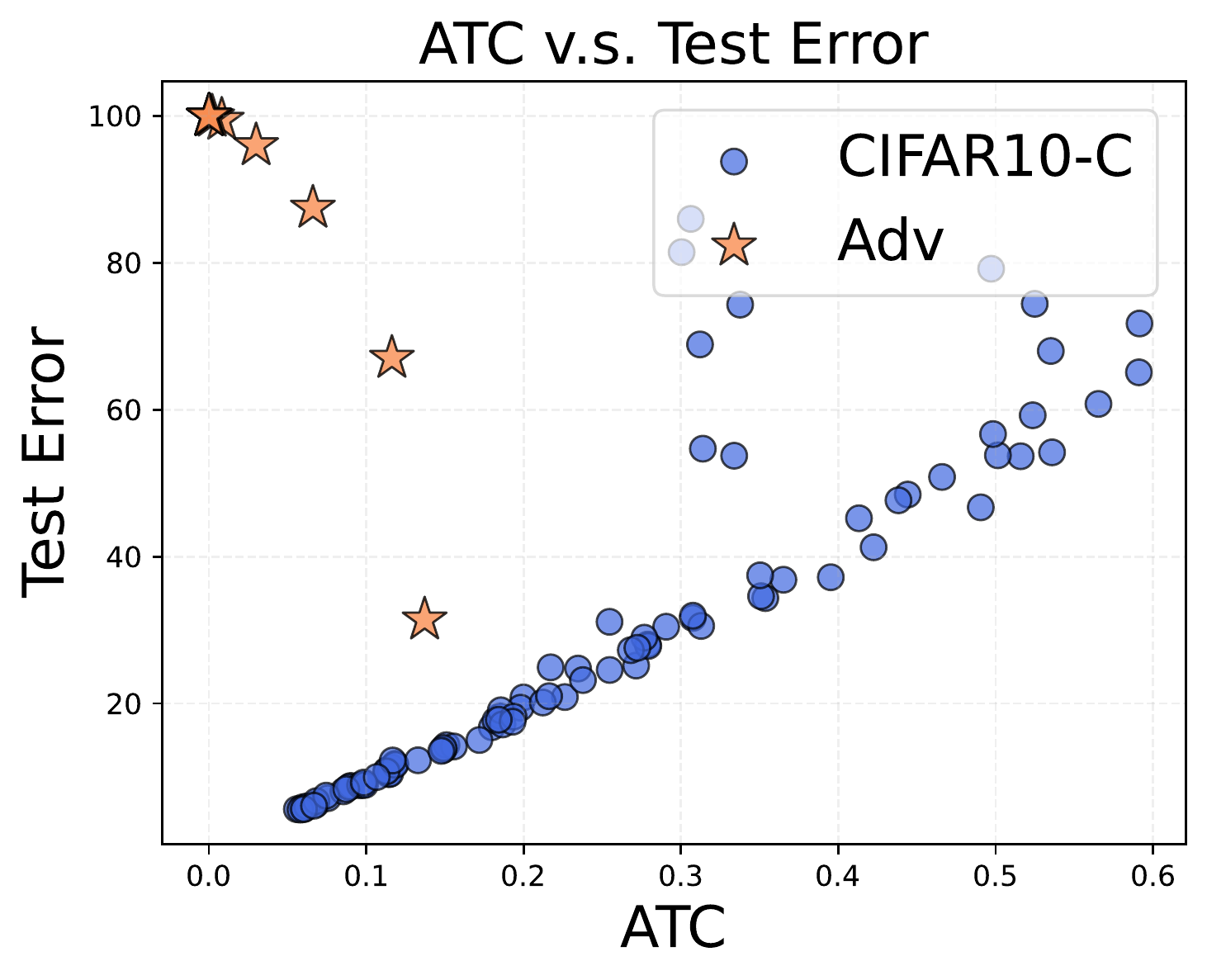}}
    \subfigure[ProjNorm.]{\includegraphics[width=.32\textwidth]{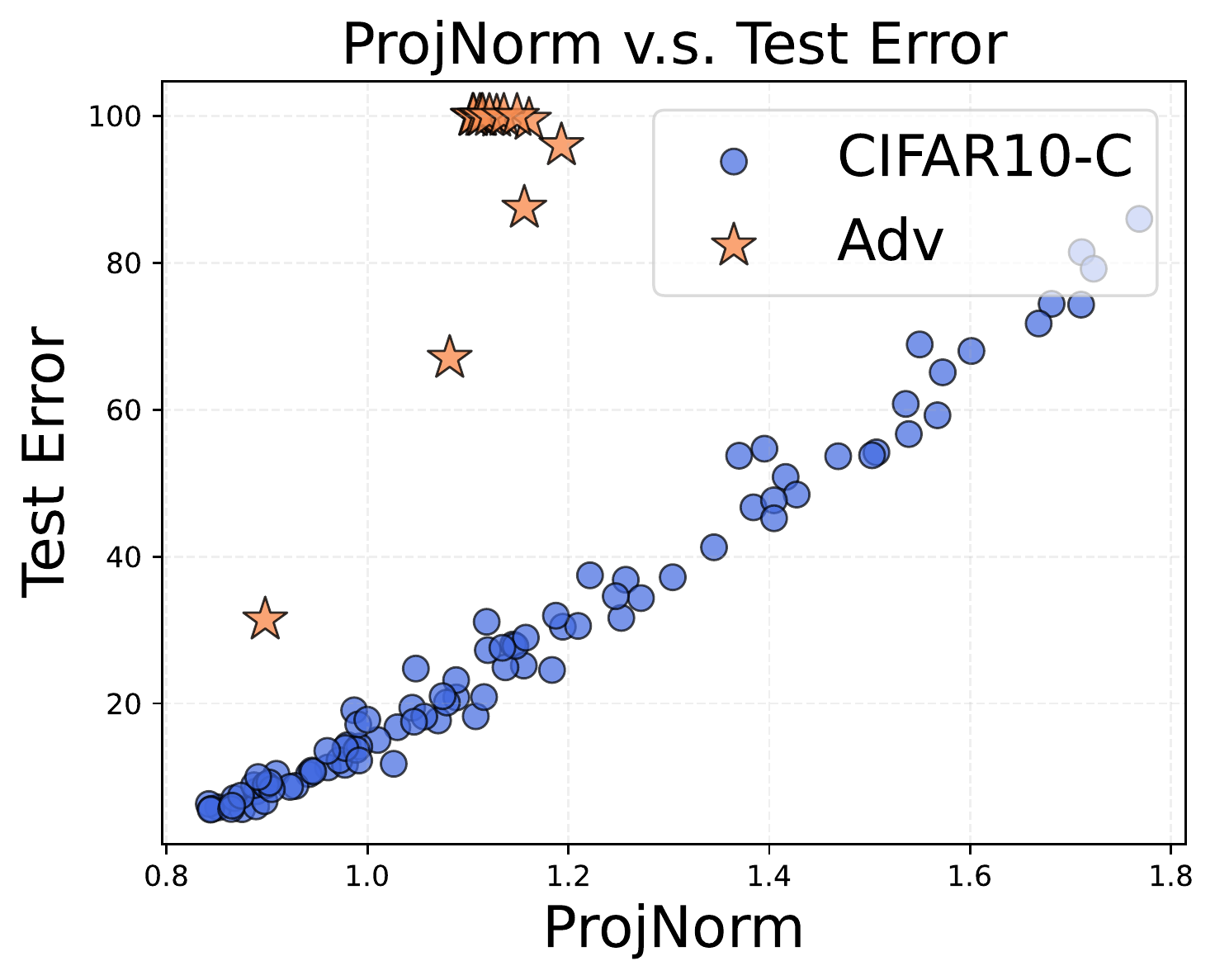}}
    \vspace{-0.15in}
    \caption{Evaluation of ConfScore, ATC, and {\normalfont\pj{}} on predicting OOD error under adversarial attack. Blue circles are results evaluated on CIFAR10-C~(each point corresponds to one corrupted test dataset), and orange stars are results evaluated on adversarial examples~(each point corresponds to one perturbation radius $\varepsilon$). }
    \label{fig:adv-main}
    \vspace{-0.15in}
\end{figure*}
In this subsection, we check Assumptions~\ref{ass:norm} and \ref{ass:spectral} on linear representations derived from the CIFAR datasets. To construct the linear representation, consider an image input $\bx_{\text{img}}$ and a neural network $f(\,\cdot\,; \btheta)$. The behavior of the network can be locally approximated by its linearized counterpart~\cite{JacotNEURIPS2018, lee2019wide}, i.e.,
\begin{equation*}
    f(\bx_{\text{img}}; \btheta) \approx f(\bx_{\text{img}}; \btheta_{\text{0}}) + \<\grad_\btheta f(\bx_{\text{img}}; \btheta_{\text{0}}), \btheta-\btheta_{\text{0}}\>.
\end{equation*}
Under this approximation, we can replace the neural network training on the raw data $\bx_{\text{img}}$ by linear regression on its Neural Tangent Kernel (NTK) representation $\bx_{\text{ntk}}$:
\vspace{-0.05in}
\begin{equation}
    \bx_{\text{ntk}} = \grad_\btheta f(\bx_{\text{img}}; \btheta_{0}) \in \R^d.
\vspace{-0.01in}
\end{equation}
We therefore test the assumptions from Section~\ref{sec:spectral} on these NTK representations.

In the most of our experiments, we derive NTK representations from a pretrained ResNet18, which has dimension $d=500,000$ (we randomly subsample 500,000 parameters from a total of 11,177,025 parameters). See Appendix~\ref{sec:appendix-ntk} for more details.

\textbf{Justification of Assumption~\ref{ass:norm} and \ref{ass:spectral}.} We first compute the NTK representations of the training data and OOD data on CIFAR10 with sample size $n=m=5,000$.  
Then we evaluate $\|\bPt\btheta_\star\|_2$ on each OOD datasets in CIFAR10-C and compare with $\|\bP\btheta_\star\|_2$. As shown in Figure~\ref{fig:thetastar-appendix}, $\|\bPt\btheta_\star\|_2$ and $\|\bP\btheta_\star\|_2$ are within a multiplicative factor of $2$ on most of the OOD datasets. 

Next, we compute the eigenvalues and top-$K$ ($K=300$) eigenvectors of $\bSigma_{\text{ntk}} = \bX_{\text{ntk}}^\sT\bX_{\text{ntk}}/n$ and $\bSigmat_{\text{ntk}} = \bXt_{\text{ntk}}^\sT\bXt_{\text{ntk}}/m$. 
As shown in Figure~\ref{fig:ntk-heatmap-eigenvec}, the top-$k$ ($k\leq200$) eigenvectors of in-distribution and OOD covariance matrices align well with each other. When $k$ is large, the in-distribution and OOD eigenvectors become more orthogonal to each other. 
This suggests that our assumptions on covariance matrices (i.e., Assumption~\ref{ass:spectral}) approximately align with real data.

We also visualize the eigenvalues of $\bSigma_{\text{ntk}}$ and $\bSigmat_{\text{ntk}}$ in Figure~\ref{fig:ntk-heatmap-eigenvalue}. We find that the eigenvalues of both the in-distribution and OOD covariance matrices approximately follow power-law scaling relations with respect to the index of the eigenvalue.

\textbf{Linear representations predict nonlinear OOD error.}
To check that our linear analysis actually captures nonlinear neural network behavior, we use $\pjl$ on the NTK representation to 
predict the error of the original, nonlinear neural network (i.e.~fine-tuned Resnet18 on CIFAR10). 
We display the results in Figure~\ref{fig:ntk-heatmap-predict-error}. 
We find that \pjl{} computed on NTK representations predicts the OOD error of its nonlinear counterpart trained by SGD ($R^{2}=0.914$). 
Compared to results in the first row of Table~\ref{table:main-table}, \pjl{} is less accurate than \pj~($R^{2}=0.962$), but still more accurate than all existing methods in terms of $R^2$.

%% file: sec_stree.tex
\vspace{-0.075in}
\section{Stress Test: Adversarial Examples}\label{sec:stress}
Finally, we construct a ``stress test'' to explore the limits of our method. We test our method against adversarial examples, optimized to fool the network into misclassifying, but not specifically optimized to evade detection. 

In more detail, we consider white-box $\ell_{\infty}$ attacks on the CIFAR10 dataset, with adversarial perturbation budget $\varepsilon$ ranging from $0.25$ to $8.0$. We generate attacks using $20$ steps of projected gradient descent (PGD), using the untargeted attack of~\citet{KurakinAdv2017}. The adversarial OOD test distribution is obtained by computing an adversarial example from each image in the CIFAR10 test set. 

We present  scatter plots of the performances of \pj{}, ATC, and ConfScore in Figure~\ref{fig:adv-main}.
For large adversarial perturbation budgets, ATC and ConfScore perform trivially (assigning a minimal score even though the test error is maximal). While \pj{} also struggles, underpredicting the test error significantly, it stands apart by making non-trivial predictions even for large budgets.

\vspace{-0.025in}
To quantify this numerically, we convert each method to an OOD error estimate by calibrating on CIFAR10-C (i.e.~running linear regression on the blue circles in Figure~\ref{fig:adv-main}). 
For $\varepsilon = 8$, \pj{} predicts an error of $28.1\%$ when the true error is $100.0\%$, whereas predictions of other methods are smaller than $0.0\%$. 
Full results for all methods are in Table~\ref{table:table-detail-adv-appendix}.

\vspace{-0.02in}
Such a stress test could be an interesting target for future work. While detecting adversarial examples is notoriously difficult~\citep{carlini2017adversarial}, this setting may be more tractable because an entire distribution of data points is observed, rather than a single point.

%% file: sec_relatedwork.tex
\section{Related Work}\label{sec:relatedwork}

\textbf{Predicting OOD generalization.}  
Predicting OOD error from test samples is also called unsupervised 
risk estimation~\citep{donmez2010unsupervised}. 
\citet{balasubramanian2011unsupervised} address this task using Gaussian mixture models, and \citet{steinhardt2016unsupervised} use conditional independence 
assumptions and the method of moments. 
In a different direction, \citet{chuang2020estimating} propose using domain-invariant representations~\citep{ben2007analysis} to estimate model generalization. \citet{deng2021labels} and \citet{deng2021does} apply rotation prediction to estimate classifier accuracy on vision tasks. Other works propose using the model's (softmax) predictions on the OOD data~\citep{guillory2021predicting, jiang2021assessing, garg2021leveraging}. \citet{chen2021mandoline} propose an importance weighting approach that leverages prior knowledge.

\textbf{Robustness.} Recent works develop benchmarks for evaluating model performance under various distribution shifts, including  vision and language benchmarks~\citep{geirhos2018imagenet, recht2019imagenet,  hendrycks2019benchmarking, shankar2021image,  hendrycks2021natural, santurkar2020breeds, hendrycks2021many, naik2018stress, mccoy2019right,  miller2020effect,  koh2021wilds}. Several recent works~\citep{taori2020measuring, allen2019learning} identify the ``accuracy on the line'' phenomenon---a linear trend between in-distribution accuracy and OOD accuracy. \citet{taori2020measuring} and \citet{hendrycks2021many} find that using larger models pre-trained on more (diverse) datasets are two effective techniques for improving robustness. \citet{sun2020test} propose a test-time-training method to improve robustness.

\textbf{OOD detection.} The goal of OOD detection is to identify whether a test sample comes from a different distribution than the training data, which is closely related to the task we study.  \citet{hendrycks2016baseline} and \citet{geifman2017selective} use model softmax outputs to detect OOD samples. \citet{lee2018simple} propose to use a generative classifier for OOD detection.  \citet{liang2018enhancing} find that temperature scaling~\citep{guo2017calibration} and adversarial perturbations~\citep{goodfellow2014explaining} can improve  detection performance. Other work utilizes pre-trained models to improve OOD detection performance~\citep{hendrycks2020pretrained, Xu2021}. Our method can potentially be extended to perform OOD detection.

\textbf{Domain adaptation.} A large body of work studies how to learn representations that transfer from a source domain to a target domain during training~\citep{ben2007analysis, ben2010theory, pan2010domain, long2015learning, ganin2016domain, tzeng2017adversarial, zhao2019learning}. The goal of domain adaptation is to improve model performance on a target (OOD) domain, whereas we focus on predicting performance of a fixed model on OOD data. An interesting direction for future work would be to explore the application of \pj{} in domain adaptation.

\textbf{NTK and overparameterized linear models.} A recent line of theoretical work tries to connect deep neural network training to neural tangent kernels (NTK)~\citep{JacotNEURIPS2018, lee2019wide, du2019gradient, allen2019learning, zou2019gradient}, showing that infinite-width networks converge to a limiting kernel. Several recent works study the  benign overfitting phenomenon in deep learning through overparameterized linear models~\citep{bartlett2020benign, tsigler2020benign, koehler2021uniform}.
\citet{tripuraneni2021covariate} computes the exact asymptotics of generalization error for random feature models under certain assumptions of distribution shift.

%% file: sec_discussion.tex
\section{Discussion}\label{sec:discussion} 
Thus far, we have focused on the advantages of Projection Norm in terms of empirical performance and theoretical interpretability. 
We now briefly discuss limitations of Projection Norm and future directions.
One limitation is that it needs sufficiently many samples (because of the fine-tuning step) to make accurate predictions on the OOD test dataset.
It would be useful to reduce the sample complexity of this method, with the ideal being a one-sample version of \pj.
Another issue is that \pj{} sometimes does poorly on ``easy'' shifts, as it looks for all differences between two distributions, including those that might make the problem easier. We illustrate this in Figure~\ref{fig:compare-appendix-labelshift} of the appendix, where \pj{} typically overpredicts the error under label shifts. 
A final limitation is \pj's performance on adversarial examples, which suggests an interesting avenue for future work.

Beyond predicting OOD error, \pj{} provides a general way to compute distances between distributions. 
For instance, it could be used to choose sample policies for active learning or exploration policies for reinforcement learning. We see \pj{} as a particularly promising approach for addressing ``novelty'' in high-dimensional settings.

%% file: sec_appendix.tex
\clearpage
\onecolumn
\section{Experimental Details}\label{sec:appendix-exp-detail}

\textbf{Details on {\normalfont\pj}.} Algorithm~\ref{alg:projnorm} provides a detailed description of the \pj{} algorithm.

\begin{algorithm}[H]
\caption{\pj}\label{alg:projnorm}
\begin{algorithmic}[1]
\vspace{0.03in}
\STATE \textbf{Input:} Classifier $C(\cdot; \bthetah)$ to be evaluated, initialization $\btheta_{0}$, training data $\mathcal{D}_{\text{train}}=\{(\bx_i, y_i)\}_{i=1}^{n}$, OOD unlabeled test data $\bxt_{1:m}=\{\bxt_j\}_{j=1}^{m}$.
\STATE \textbf{Parameters:} Number of training steps $T$, initial learning rate $\eta$.
\STATE \textbf{Step 1:} Pseudo-label OOD data with $C(\cdot; \bthetah)$, i.e., $\ytp_j = C(\bxt_j; \bthetah)$, $j\in[m]$.
\STATE \textbf{Step 2:} From initialization $\btheta_{0}$, train a new model $\bthetat$ on pseudo-labeled OOD data $\{(\bxt_j, \ytp_j)\}_{j=1}^{m}$ by performing $T$ steps of stochastic gradient descent updates with learning rate $\eta$.
\STATE \textbf{Step 2+:} From initialization $\btheta_{0}$, train a reference model $\bthetah_{\textsf{ref}}$ on training data $\{(\bx_i, y_i)\}_{i=1}^{n}$ by performing $T$ steps of stochastic gradient descent updates with learning rate $\eta$.
\STATE \textbf{Step 3:} Output  $\pj(\mathcal{D}_{\text{train}}, \bxt_{1:m}) := \|\bthetah_{\text{ref}} -\bthetat \|_{2}.$
\end{algorithmic}
\end{algorithm}

\vspace{-0.1in}
\textbf{Additional implementation details.} For the CIFAR  datasets, we fine-tune the pre-trained model on in-distribution training data for 20 and 50 epochs for CIFAR10 and CIFAR100, respectively. For MNLI, we fine-tune the pre-trained model for 4 epochs on in-distribution training data.

\subsection{Details of existing methods}\label{sec:methods-detail-appendix}
\textbf{Rotation.} The \textit{Rotation Prediction} (Rotation)~\citep{deng2021does} metric is defined as
\begin{equation}\label{eq:method-rotation}
    \text{Rotation} = \frac{1}{m}\sum_{j=1}^{m}\left\{ \frac{1}{4} \sum_{r \in \{0^{\circ}, 90^{\circ}, 180^{\circ}, 270^{\circ}\}} \mathbf{1}\left\{C^{r}(\bxt_j; \bthetah) \neq y_r \right\} \right\},
\end{equation}
where $y_r$ is the label for $r \in \{0^{\circ}, 90^{\circ}, 180^{\circ}, 270^{\circ}\}$, and $C^{r}(\bxt_j; \bthetah)$ predicts the rotation degree of an image $\bxt_j$.

\textbf{ConfScore.} The \textit{Averaged Confidence} (ConfScore) is defined as
\begin{equation}\label{eq:method-confscore}
    \text{ConfScore} = \frac{1}{m}\sum_{j=1}^{m} \max_{k} \text{Softmax}(f(\bxt_j; \bthetah))_k,
\end{equation}
where $\text{Softmax}(\cdot)$ is the softmax function.

\textbf{Entropy.} The \textit{Entropy} metric is defined as
\begin{equation}\label{eq:method-entropy}
    \text{Entropy} = \frac{1}{m}\sum_{j=1}^{m} \text{Ent}\left( \text{Softmax}(f(\bxt_j; \bthetah))\right),
\end{equation}
where $\text{Ent}(\boldsymbol{p}) = -\sum_{k=1}^{K}\boldsymbol{p}_k\cdot\log(\boldsymbol{p}_k)$.

\textbf{AgreeScore.} The \textit{Agreement
Score} (AgreeScore) is defined as
\begin{equation}\label{eq:method-agreescore}
    \text{AgreeScore} = \frac{1}{m}\sum_{j=1}^{m} \mathbf{1}\left\{C(\bxt_j; \btheta_1) \neq C(\bxt_j; \btheta_2)\right\},
\end{equation}
where $C(\bxt_j; \btheta_1)$ and $C(\bxt_j; \btheta_2)$ are two classifiers that are trained on in-distribution training data independently.

\textbf{ATC.} The \textit{Averaged Threshold
Confidence} (ATC)~\citep{garg2021leveraging} is defined as
\begin{equation}\label{eq:method-atc}
    \text{ATC} = \frac{1}{m}\sum_{j=1}^{m} \mathbf{1}\left\{ s(\text{Softmax}(f(\bxt_j; \bthetah))) < t \right\},
\end{equation}
where $s(\boldsymbol{p}) = \sum_{j=1}^{K}\boldsymbol{p}_k\log(\boldsymbol{p}_k)$, and $t$ is defined as the solution to the following equation,
\begin{equation}
    \frac{1}{m^{\text{val}}}\sum_{\ell=1}^{m^{\text{val}}} \mathbf{1}\left\{ s(\text{Softmax}(f({\bx}^{\text{val}}_{\ell}; \bthetah))) < t \right\} = \frac{1}{m^{\text{val}}}\sum_{\ell=1}^{m^{\text{val}}} \mathbf{1}\left\{ C({\bx}^{\text{val}}_{\ell}; \bthetah) \neq y^{\text{val}}_{\ell} \right\},
\end{equation}
where $({\bx}^{\text{val}}_{\ell}, y^{\text{val}}_{\ell})$, $\ell=1, \dots, m^{\text{val}}$, are in-distribution validation samples.

\section{Additional Experimental Results}\label{sec:additional-exp-results}

\begin{figure*}[ht]
    \centering
    \subfigure[Rotation.]{\includegraphics[width=.33\textwidth]{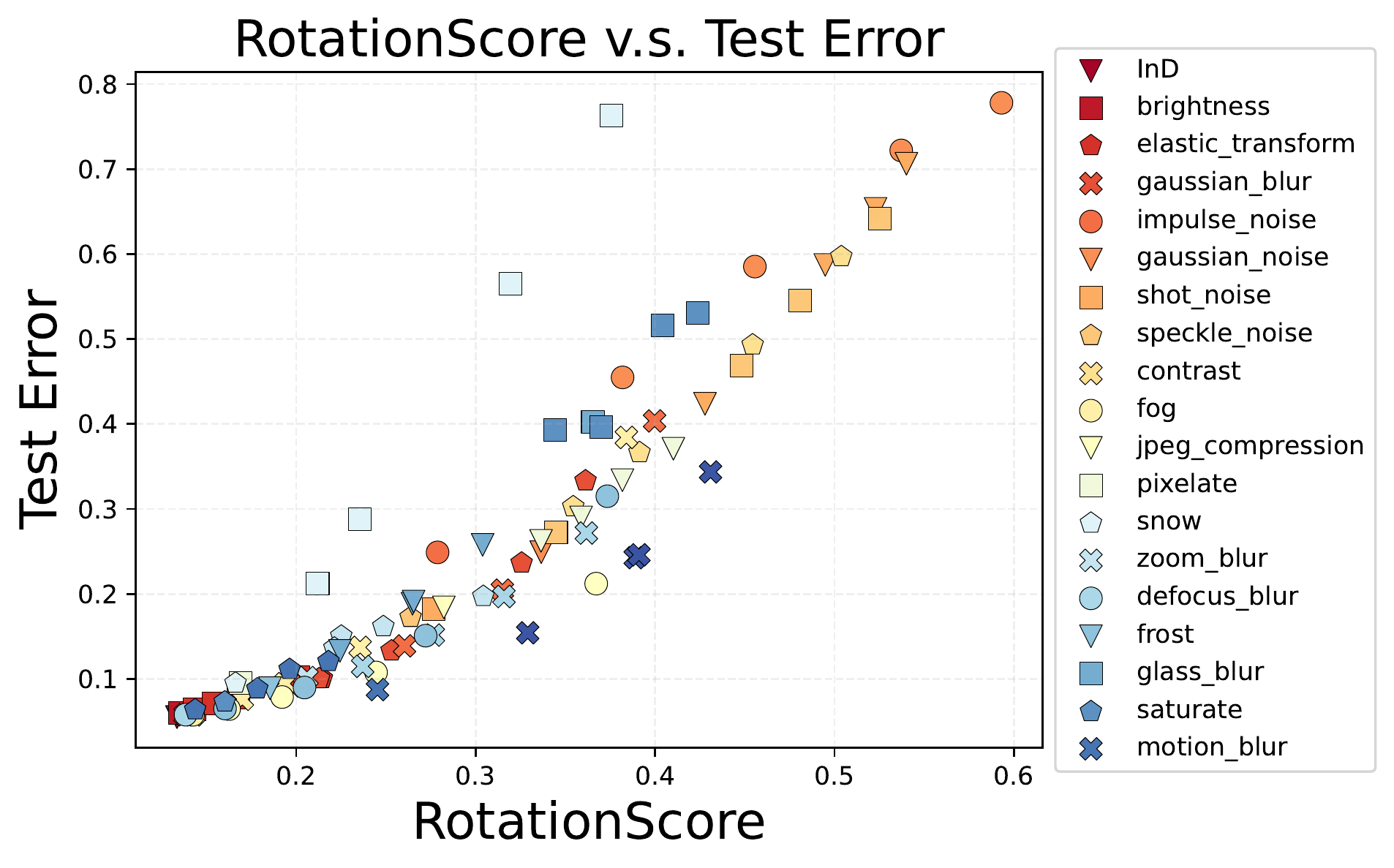}}
    \subfigure[ConfScore.]{\includegraphics[width=.33\textwidth]{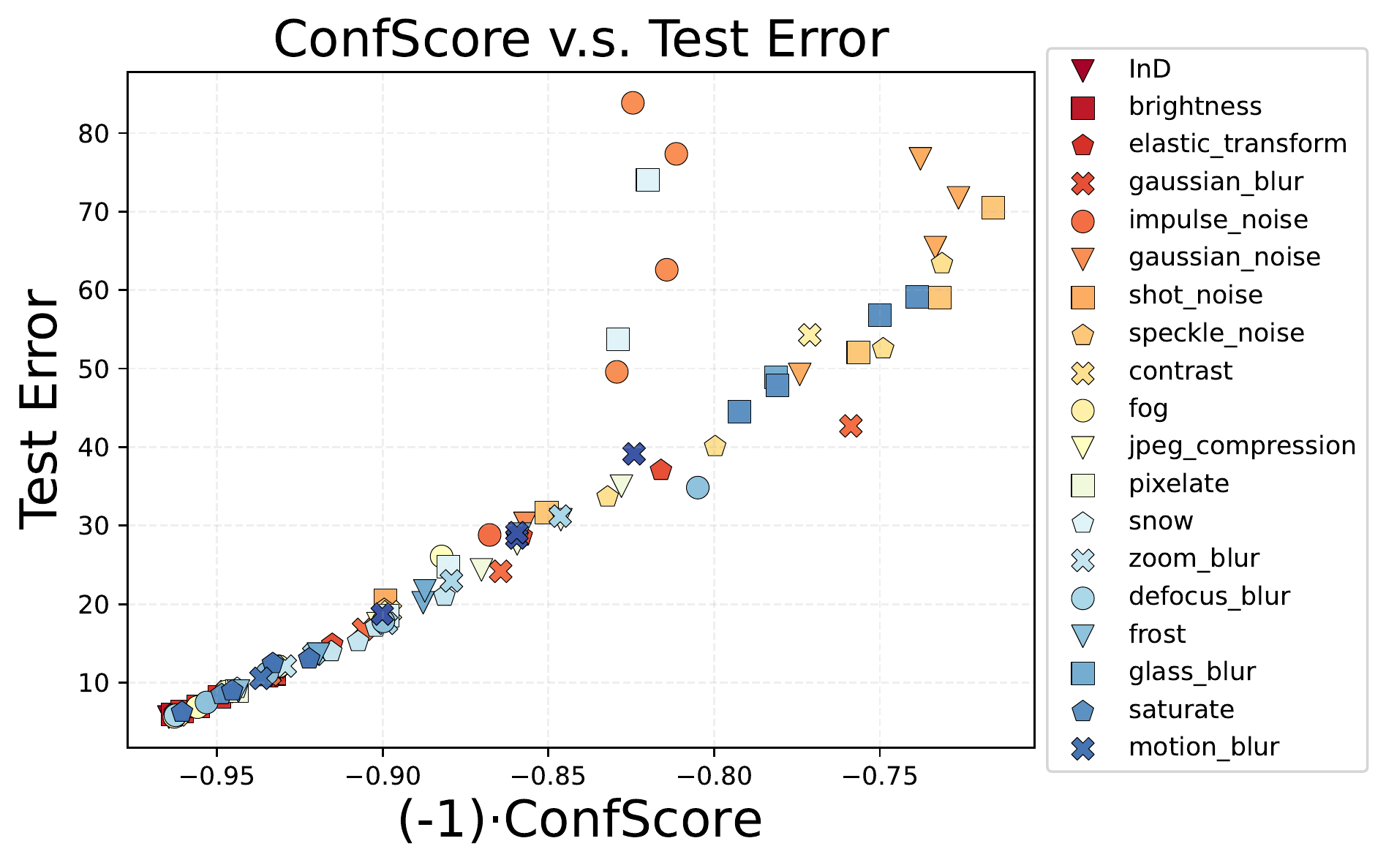}}
    \subfigure[Entropy.]{\includegraphics[width=.33\textwidth]{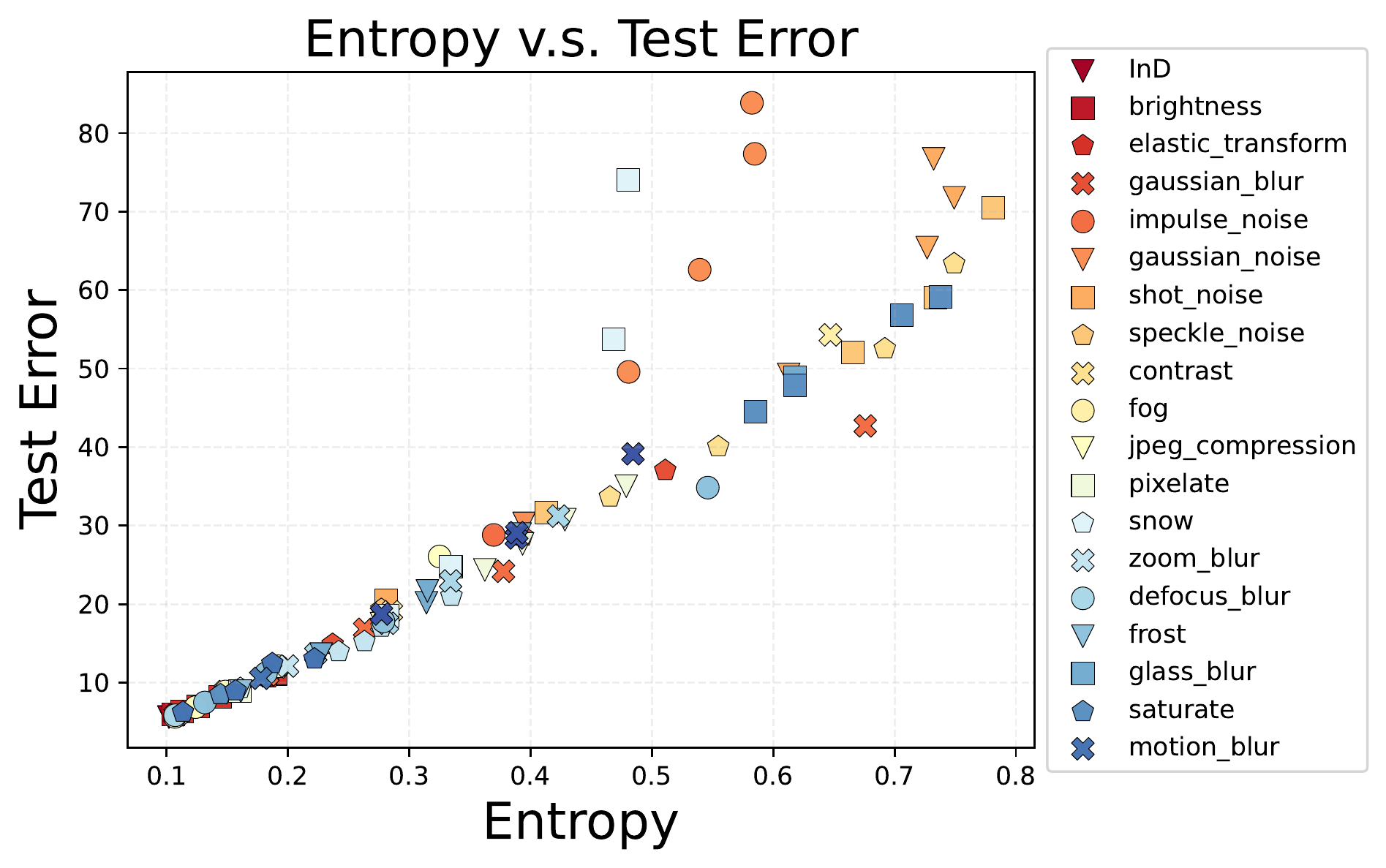}}
    \subfigure[AgreeScore.]{\includegraphics[width=.33\textwidth]{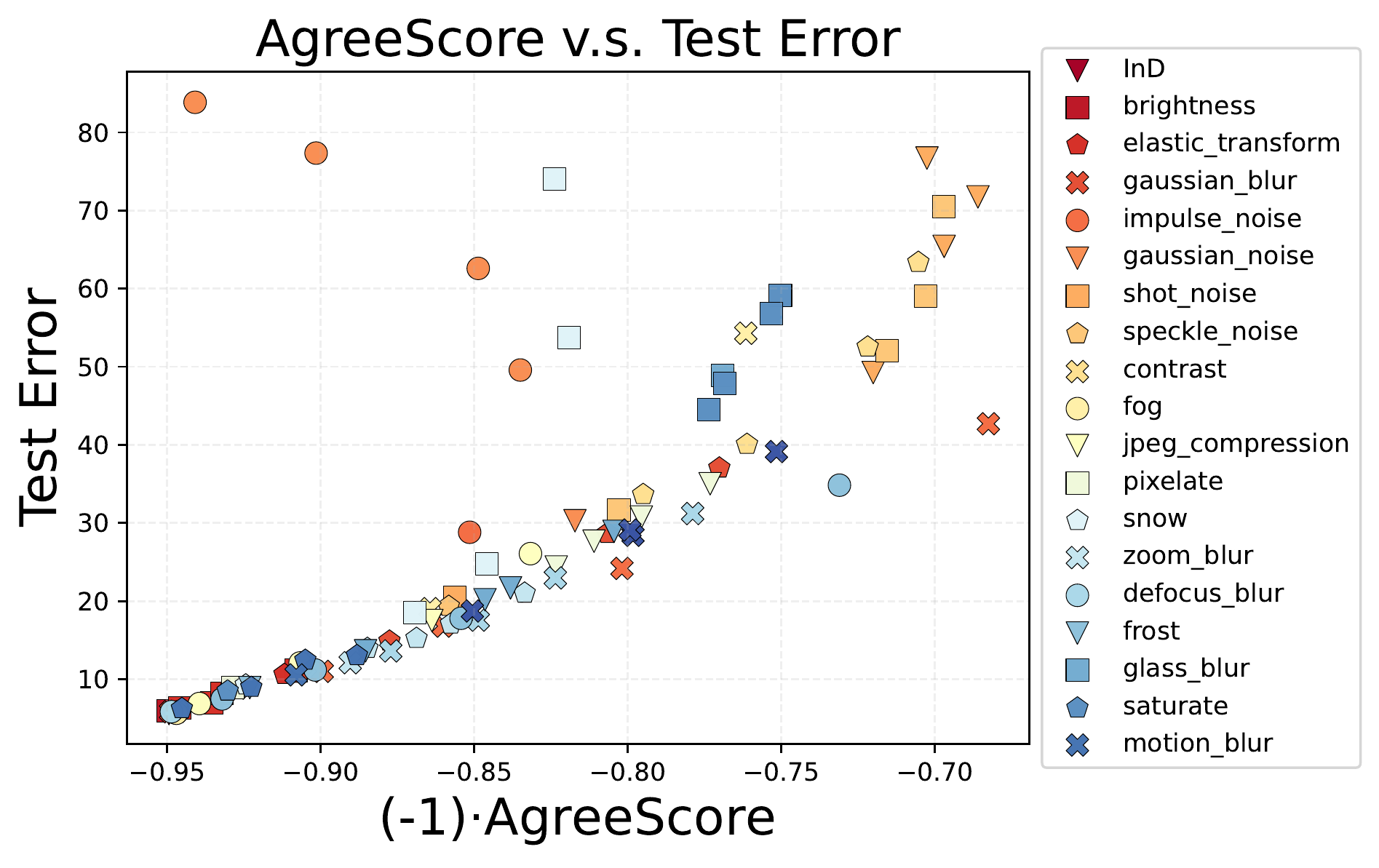}}
    \subfigure[ATC.]{\includegraphics[width=.33\textwidth]{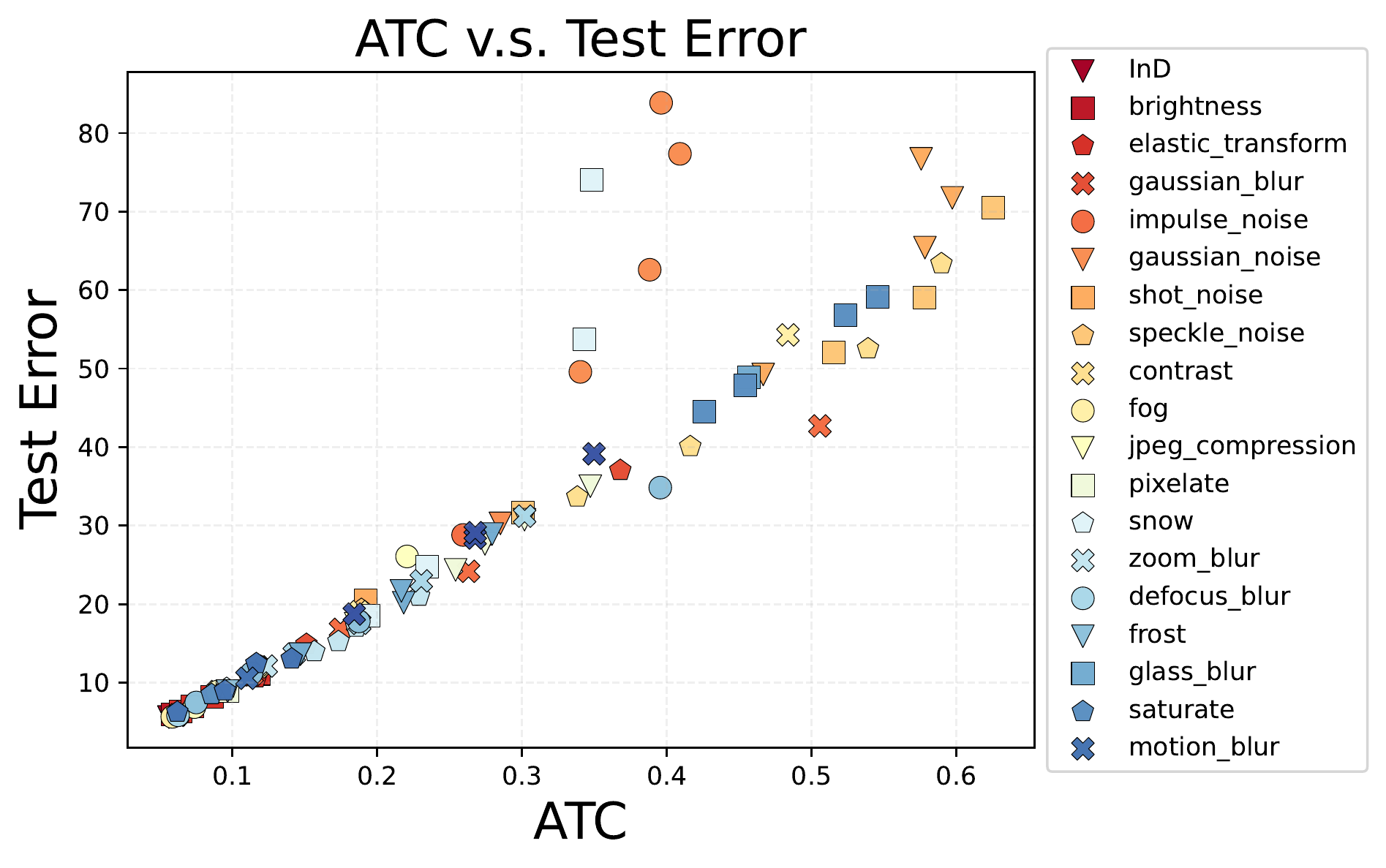}}
    \subfigure[ProjNorm.]{\includegraphics[width=.33\textwidth]{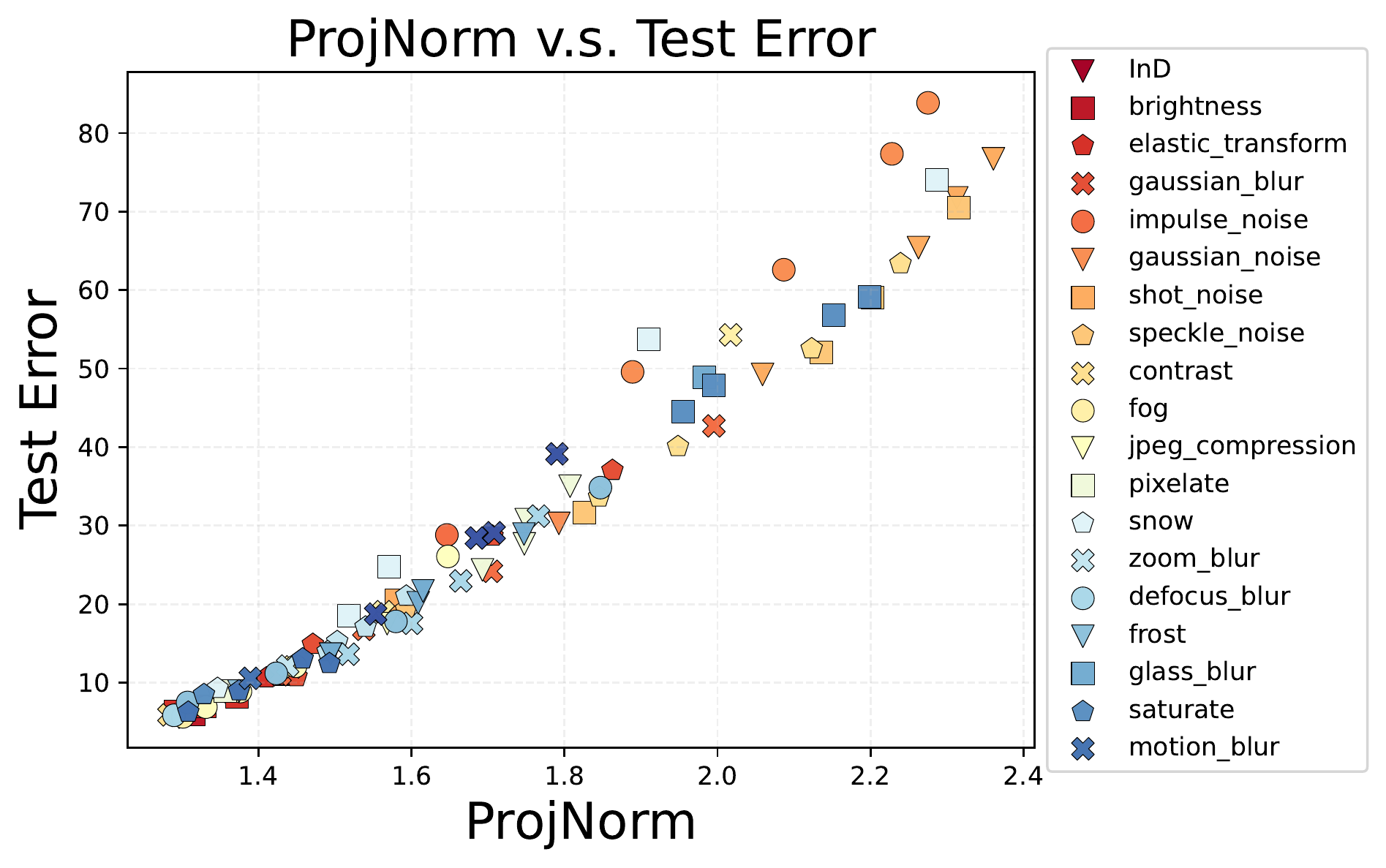}}
    \vspace{-0.1in}
    \caption{\textbf{Generalization prediction versus test error on CIFAR10 with ResNet18.} Compare out-of-distribution prediction performance of all methods. We plot the actual test error and the method prediction on each OOD dataset.
    Each point represents one InD/OOD dataset, and points with the same color and marker shape are the same corruption but with different severity levels.
    }
    \label{fig:compare-appendix-cifar10-resnet18}
    \vspace{-0.15in}
\end{figure*}

\begin{figure*}[ht]
    \centering
    \subfigure[Rotation.]{\includegraphics[width=.33\textwidth]{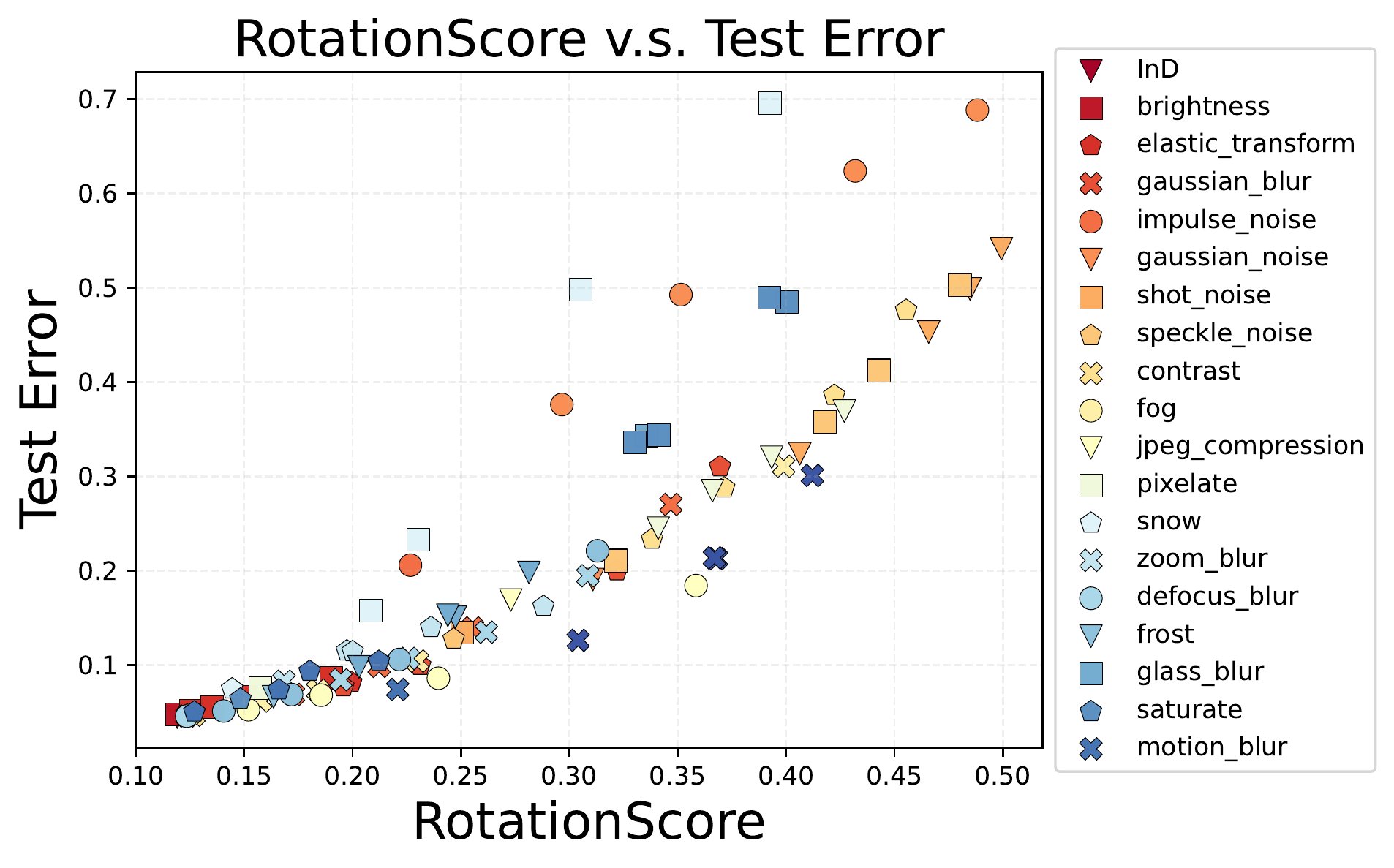}}
    \subfigure[ConfScore.]{\includegraphics[width=.33\textwidth]{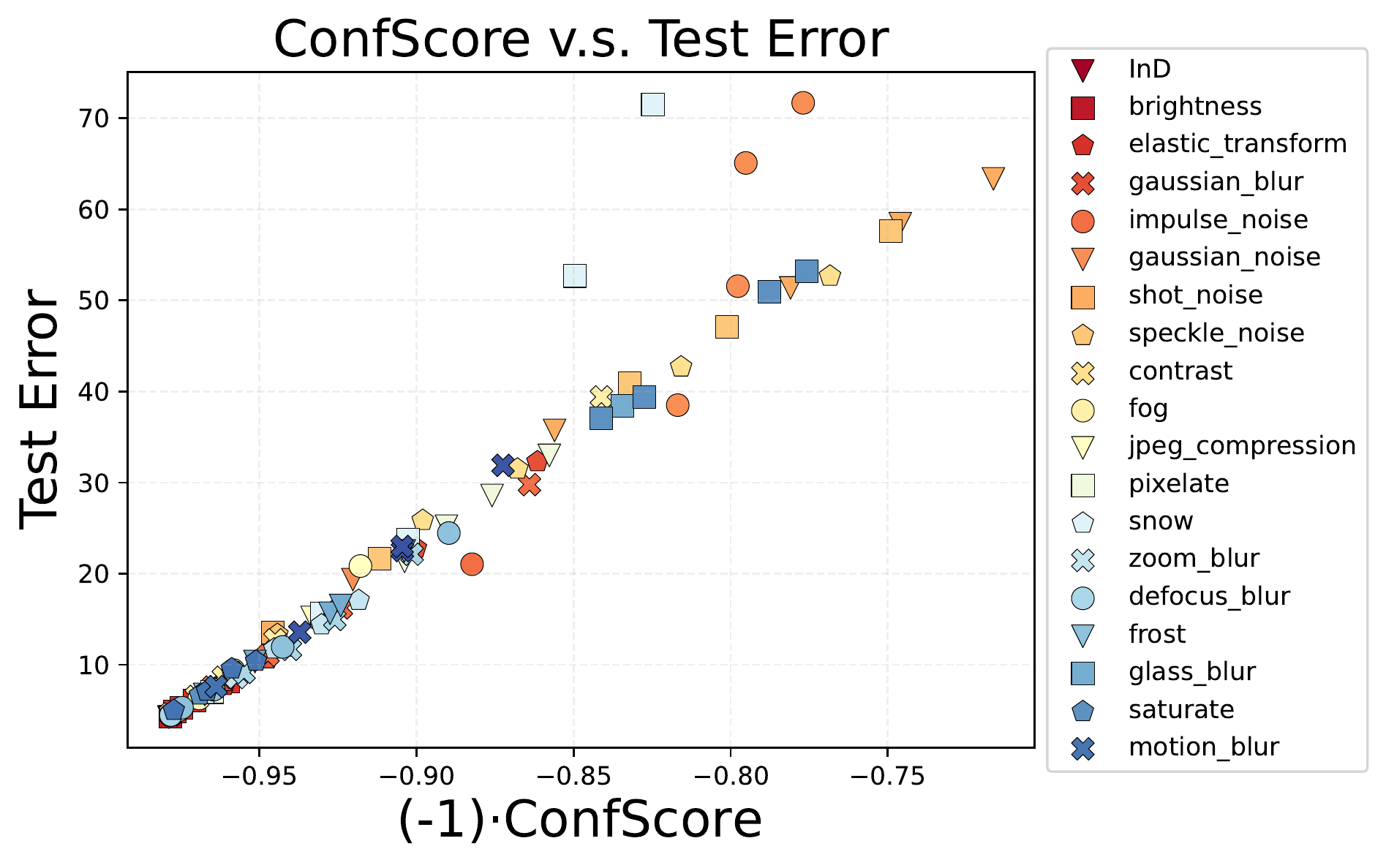}}
    \subfigure[Entropy.]{\includegraphics[width=.33\textwidth]{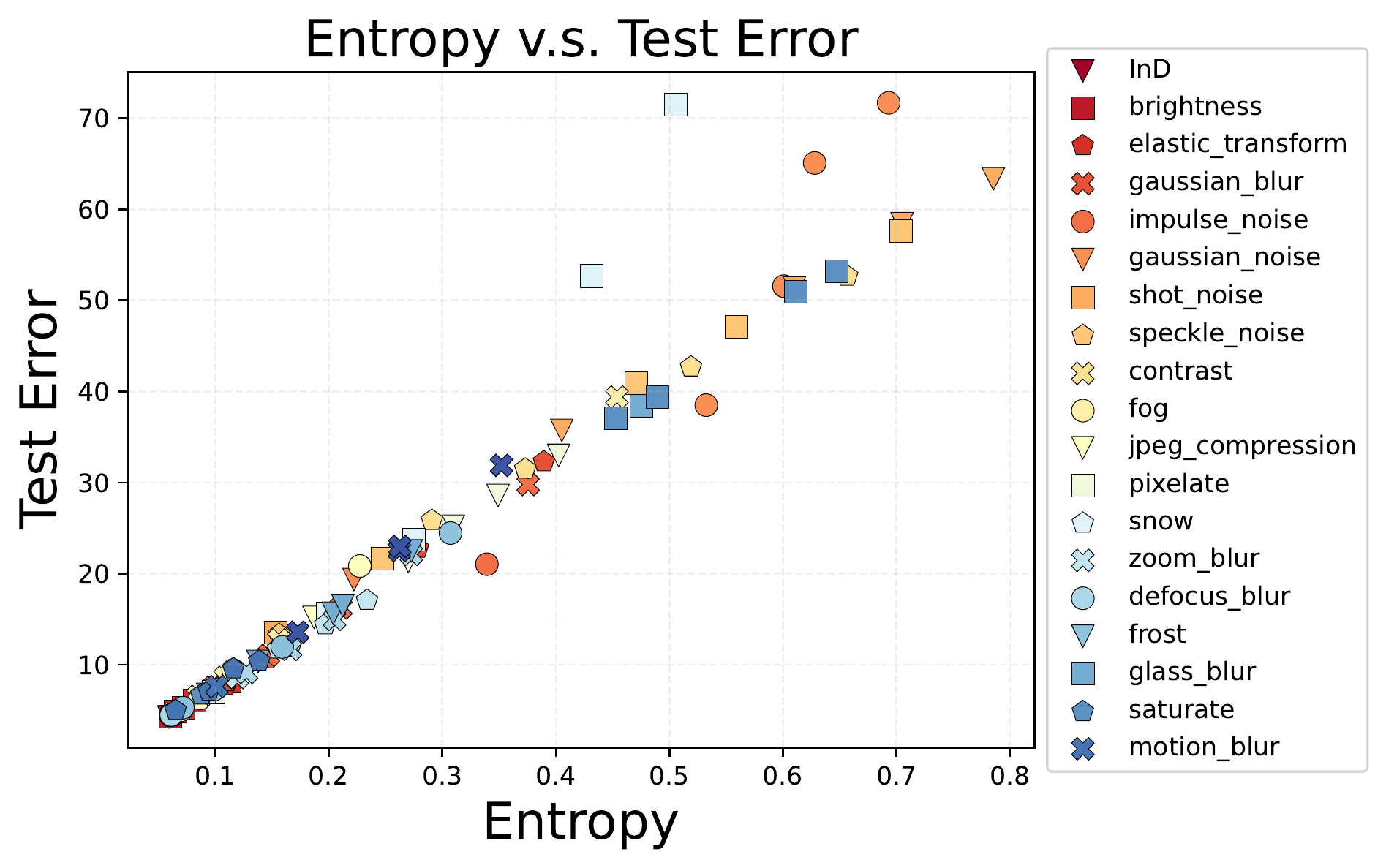}}
    \subfigure[AgreeScore.]{\includegraphics[width=.33\textwidth]{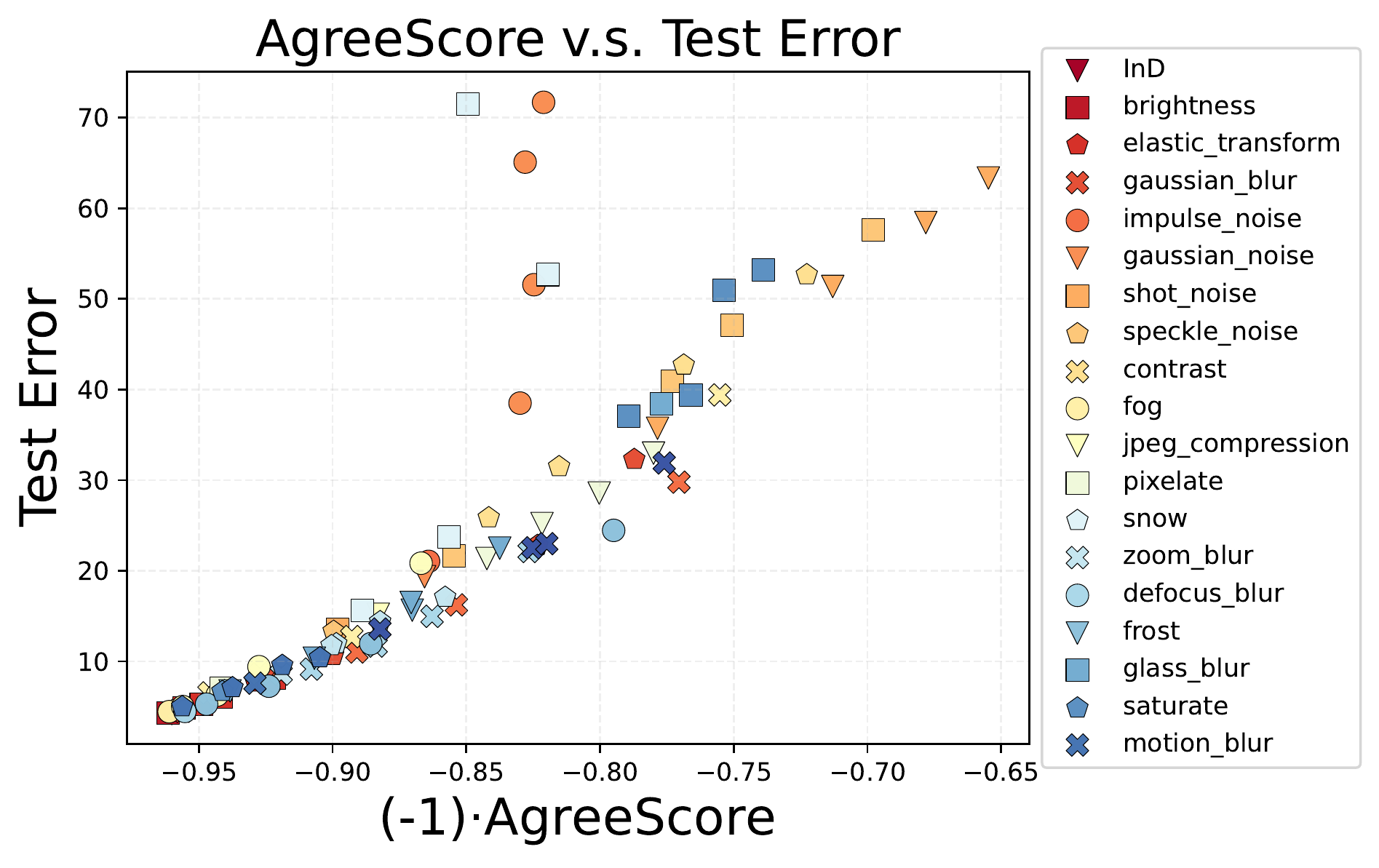}}
    \subfigure[ATC.]{\includegraphics[width=.33\textwidth]{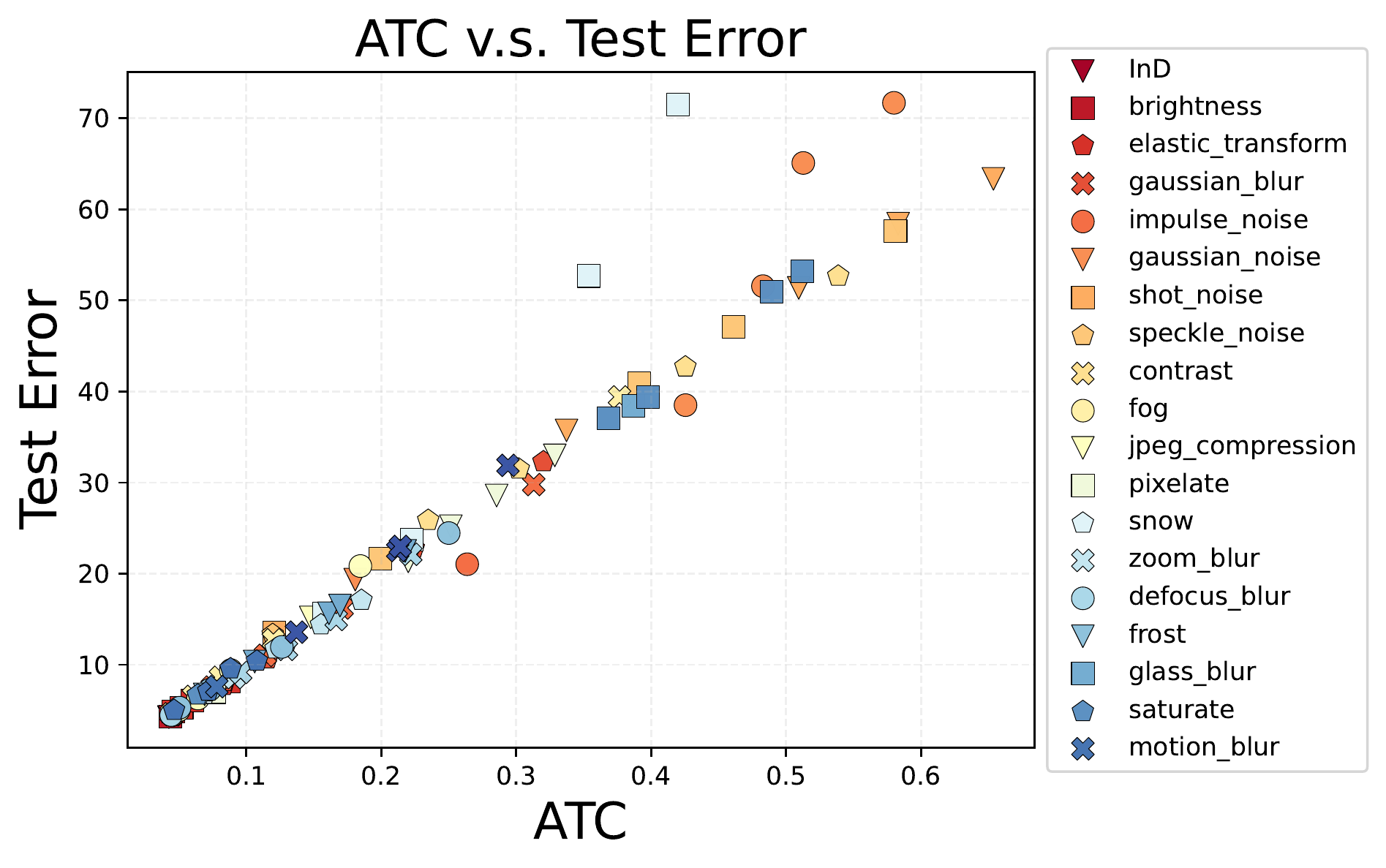}}
    \subfigure[ProjNorm.]{\includegraphics[width=.33\textwidth]{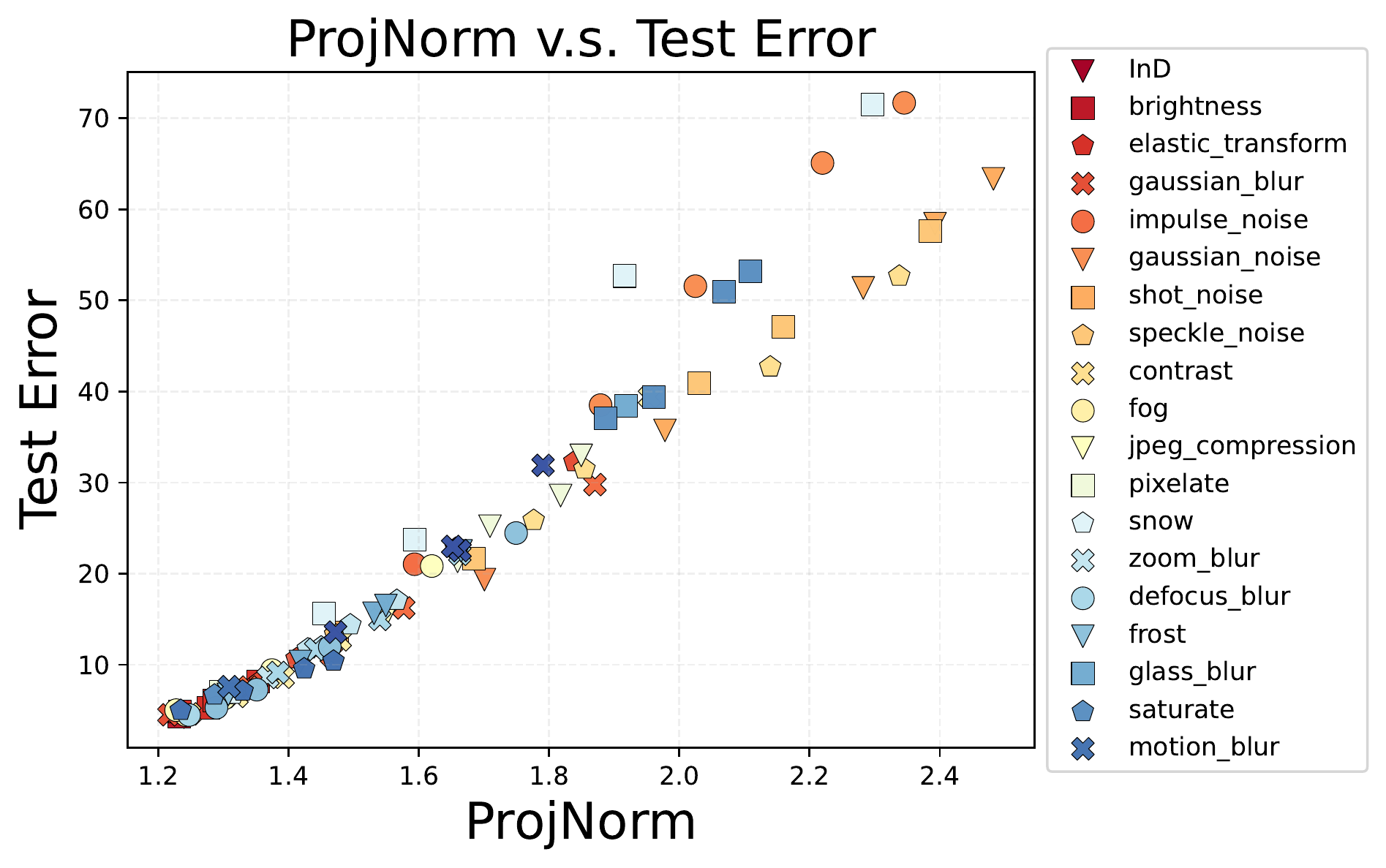}}
    \vspace{-0.1in}
    \caption{\textbf{Generalization prediction versus test error on CIFAR10 with ResNet50.} Compare out-of-distribution prediction performance of all methods. We plot the actual test error and the method prediction on each OOD dataset.
    Each point represents one InD/OOD dataset, and points with the same color and marker shape are the same corruption but with different severity levels.
    }
    \label{fig:compare-appendix-cifar10-resnet50}
    \vspace{-0.15in}
\end{figure*}

\begin{figure*}[ht]
    \centering
    \subfigure[Rotation.]{\includegraphics[width=.33\textwidth]{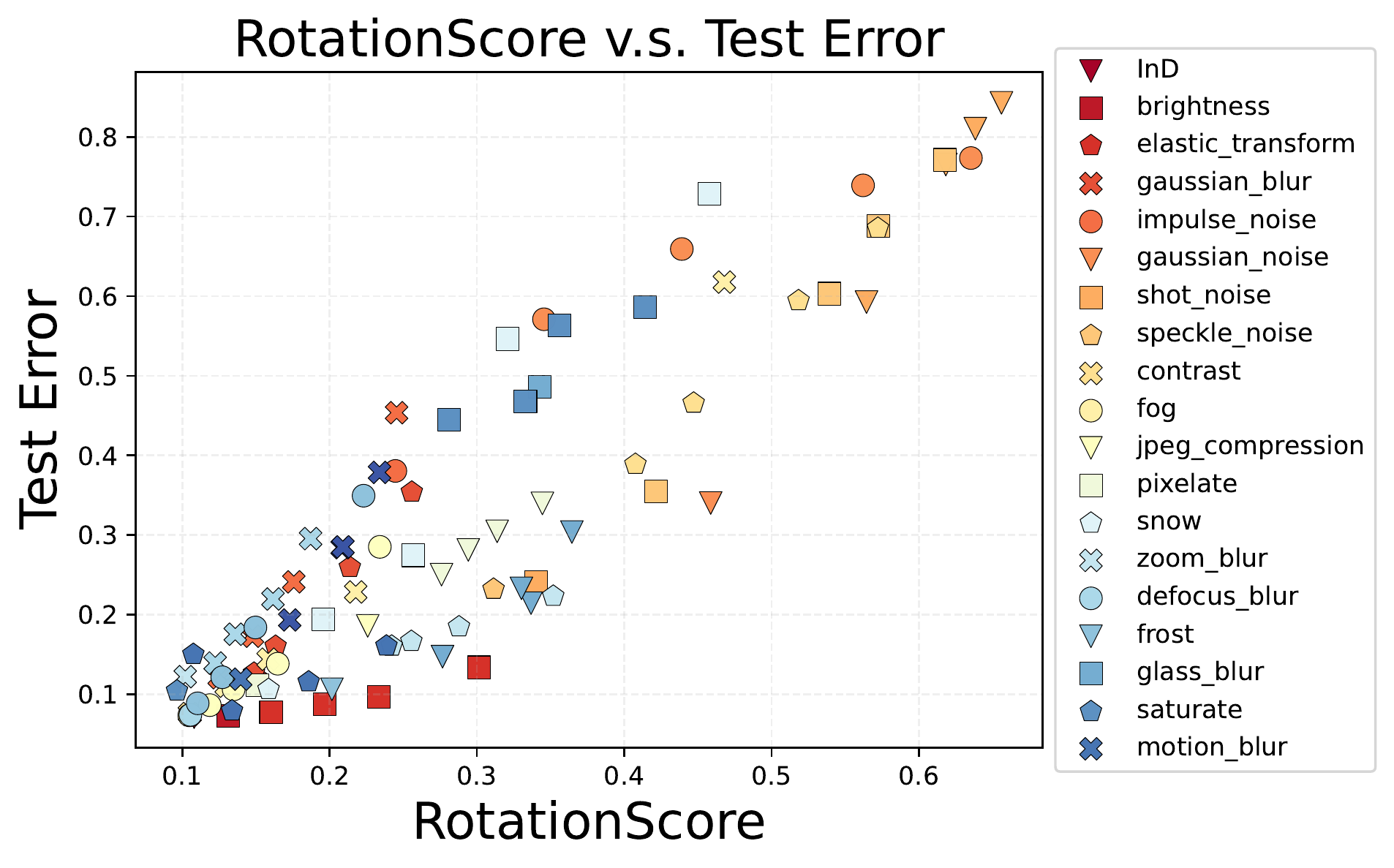}}
    \subfigure[ConfScore.]{\includegraphics[width=.33\textwidth]{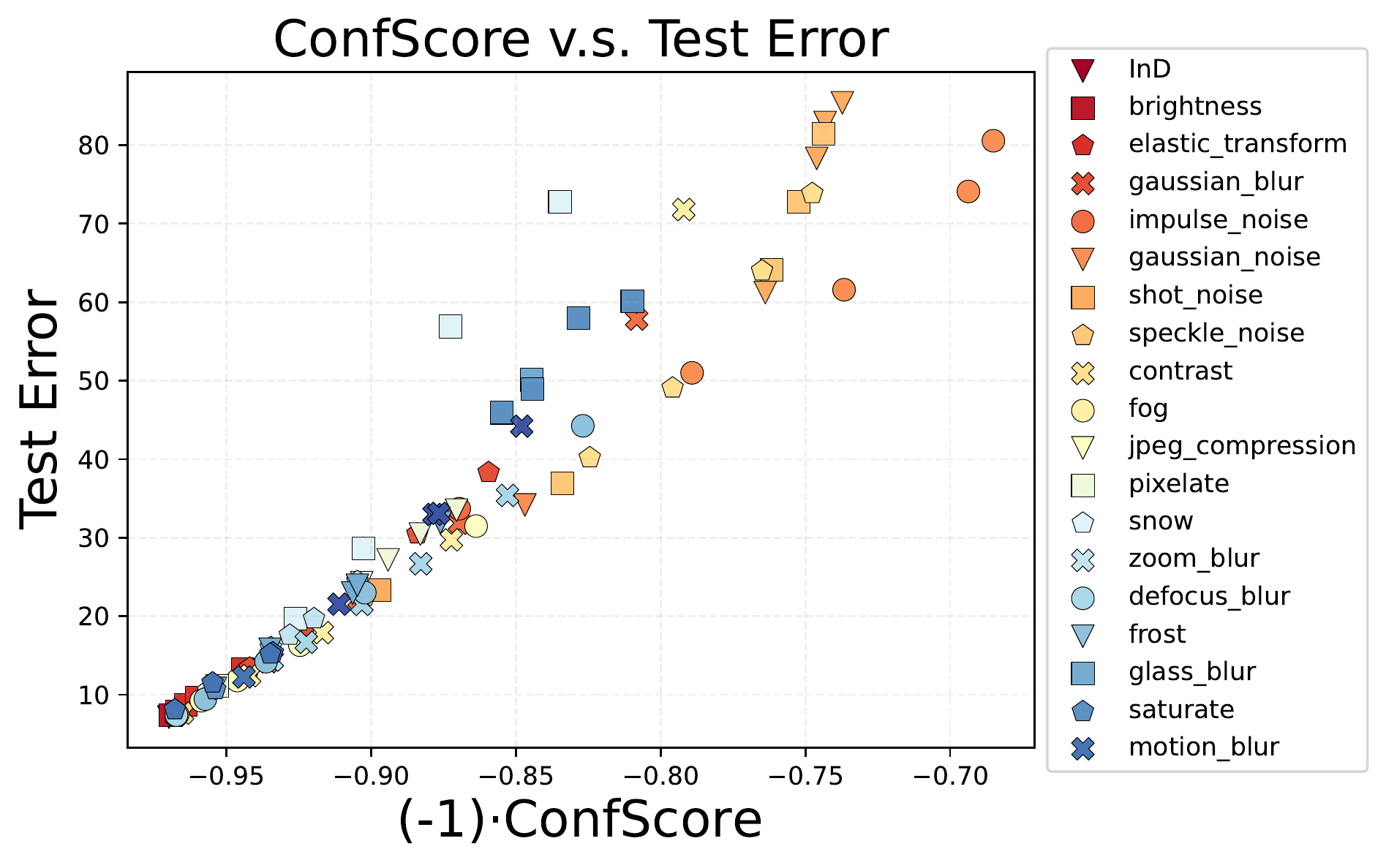}}
    \subfigure[Entropy.]{\includegraphics[width=.33\textwidth]{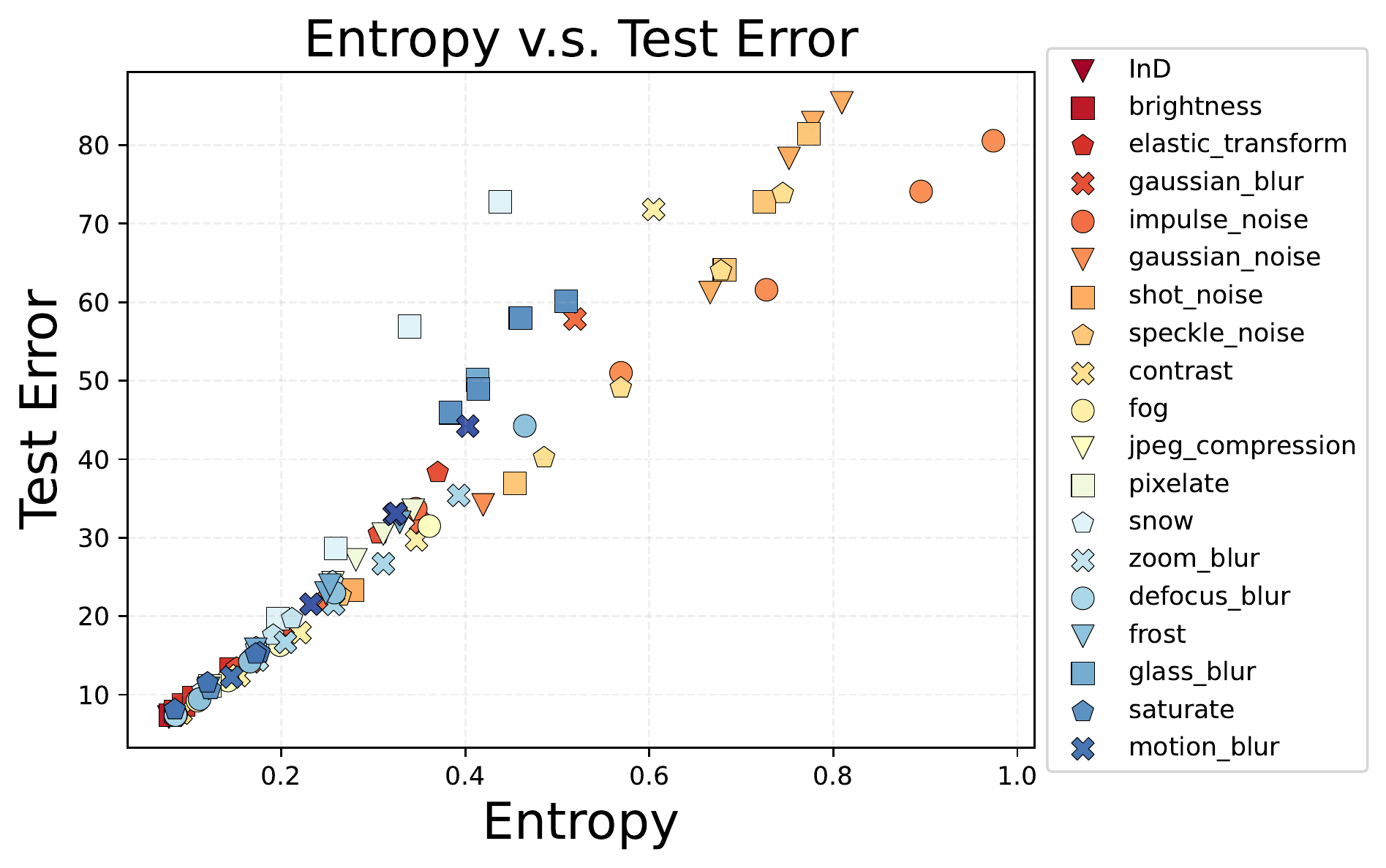}}
    \subfigure[AgreeScore.]{\includegraphics[width=.33\textwidth]{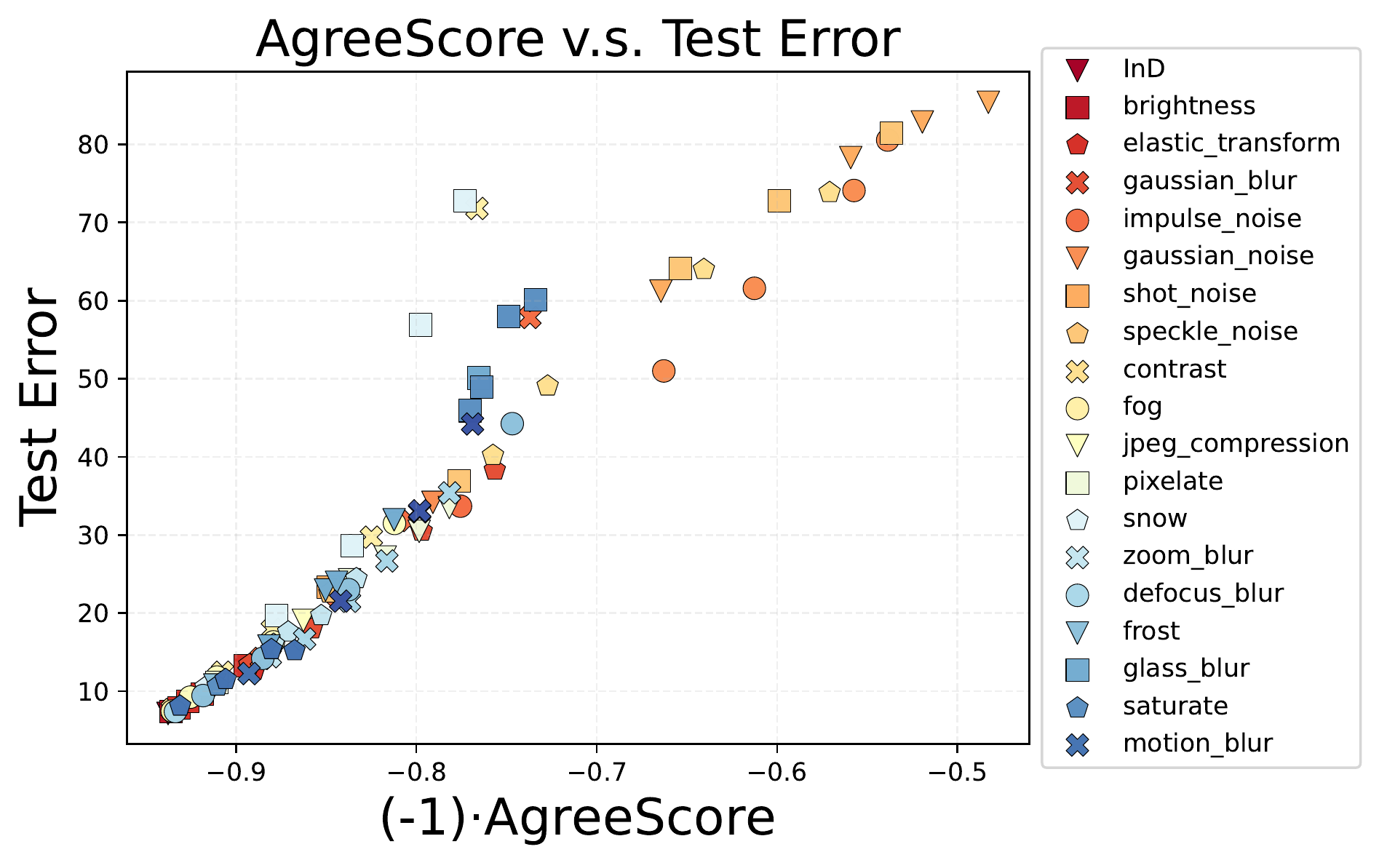}}
    \subfigure[ATC.]{\includegraphics[width=.33\textwidth]{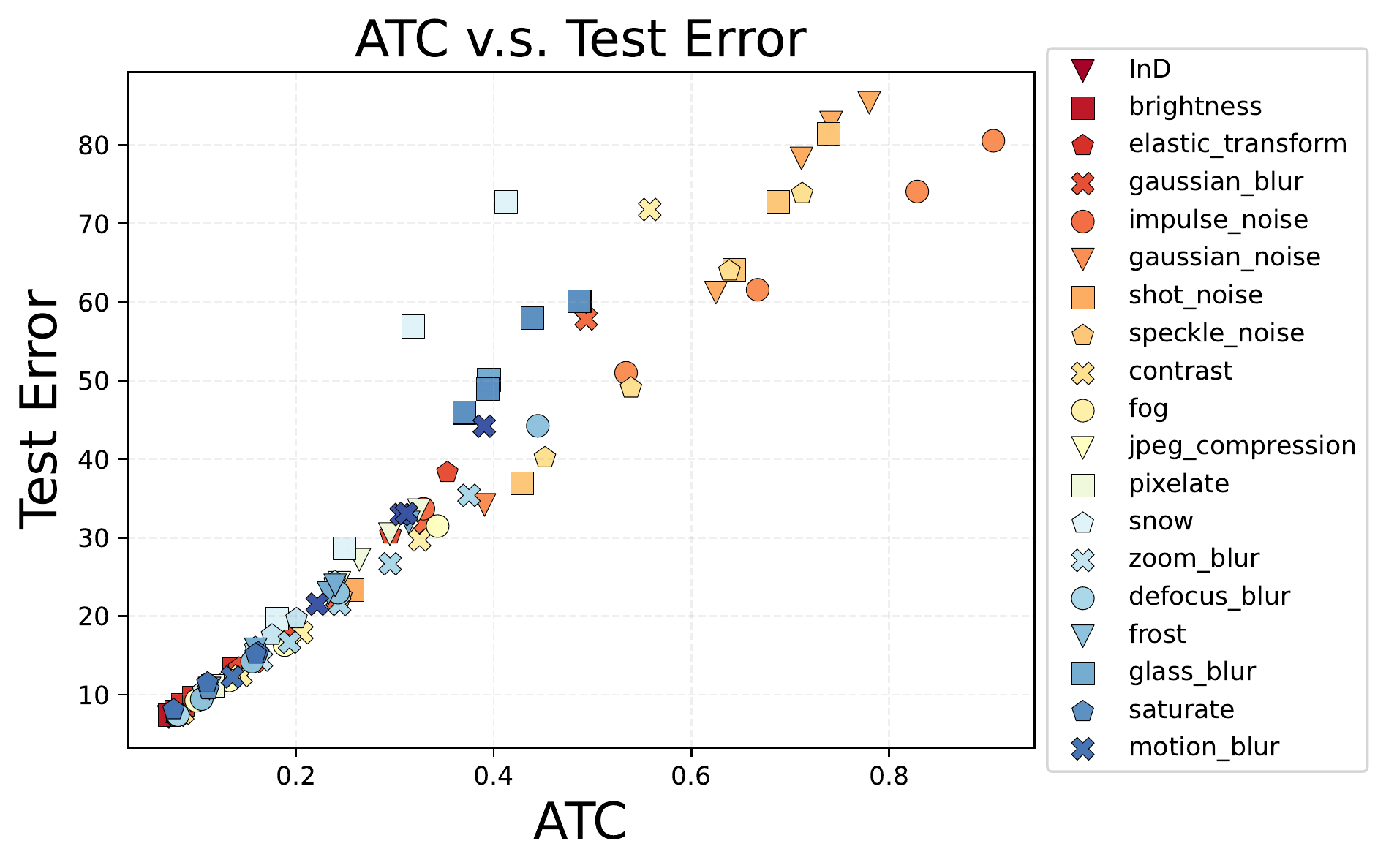}}
    \subfigure[ProjNorm.]{\includegraphics[width=.33\textwidth]{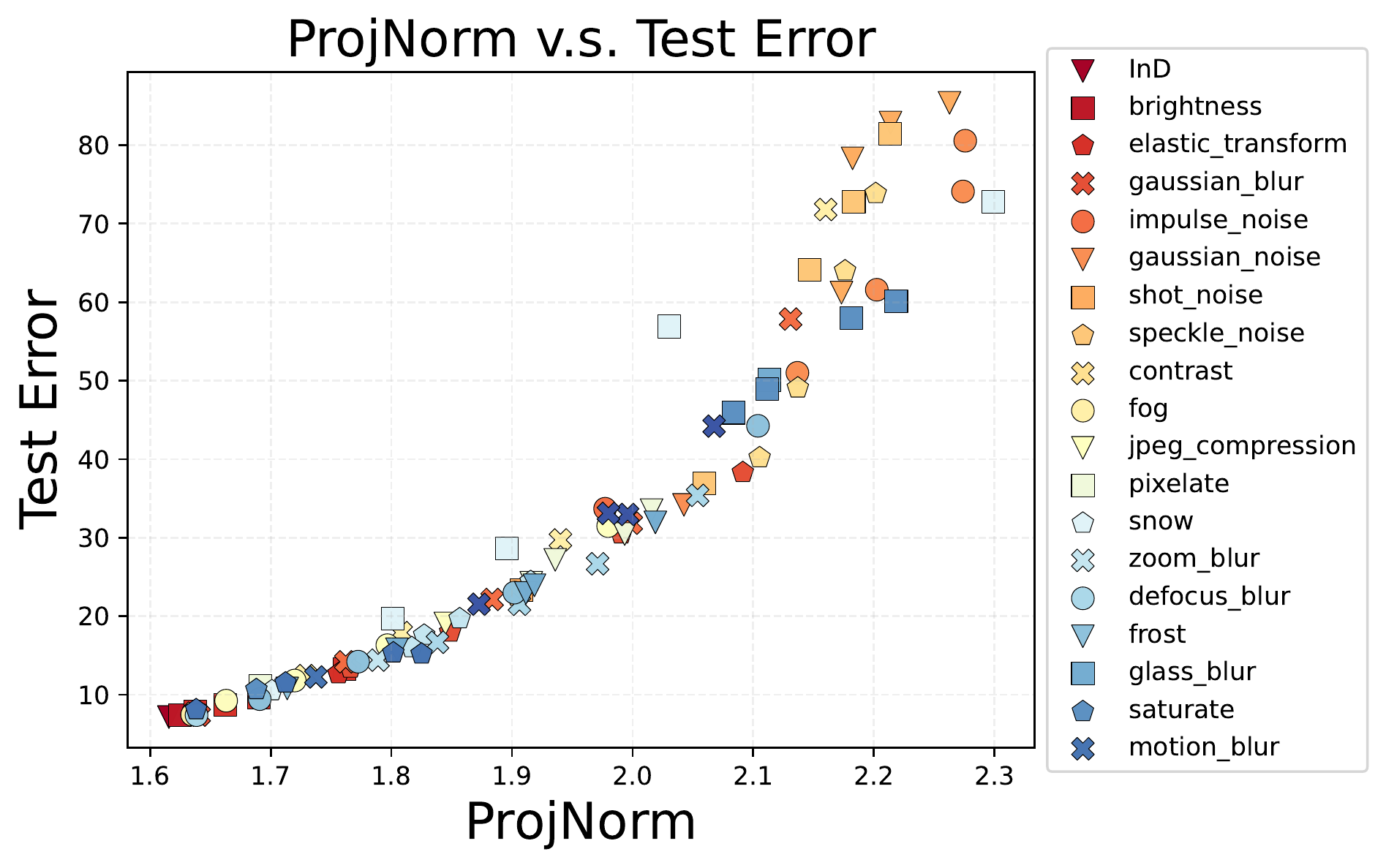}}
    \vspace{-0.1in}
    \caption{\textbf{Generalization prediction versus test error on CIFAR10 with VGG11.} Compare out-of-distribution prediction performance of all methods. We plot the actual test error and the method prediction on each OOD dataset.
    Each point represents one InD/OOD dataset, and points with the same color and marker shape are the same corruption but with different severity levels.
    }
    \label{fig:compare-appendix-cifar10-vgg}
    \vspace{-0.15in}
\end{figure*}

\begin{figure*}[ht]
    \centering
    \subfigure[Rotation.]{\includegraphics[width=.33\textwidth]{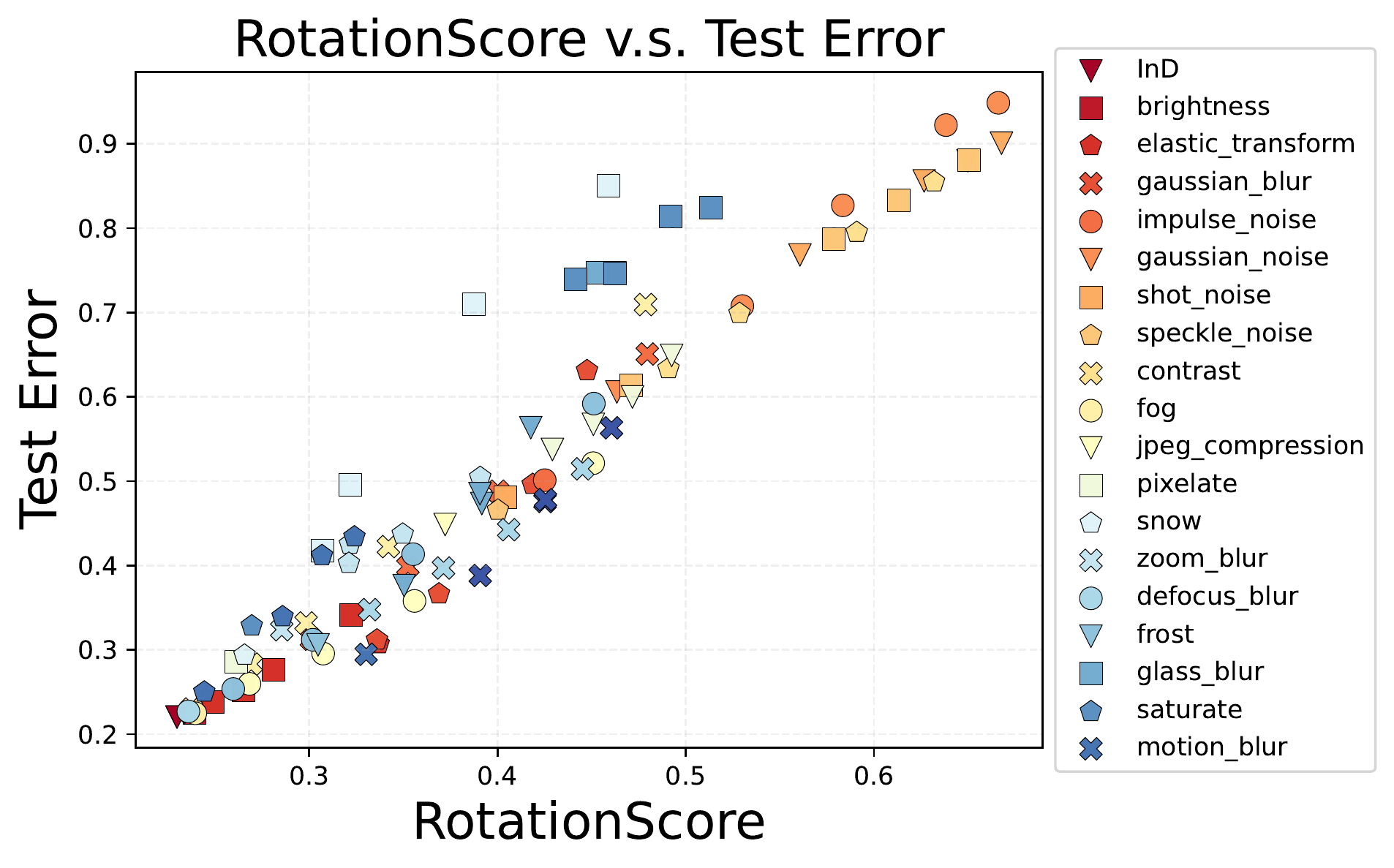}}
    \subfigure[ConfScore.]{\includegraphics[width=.33\textwidth]{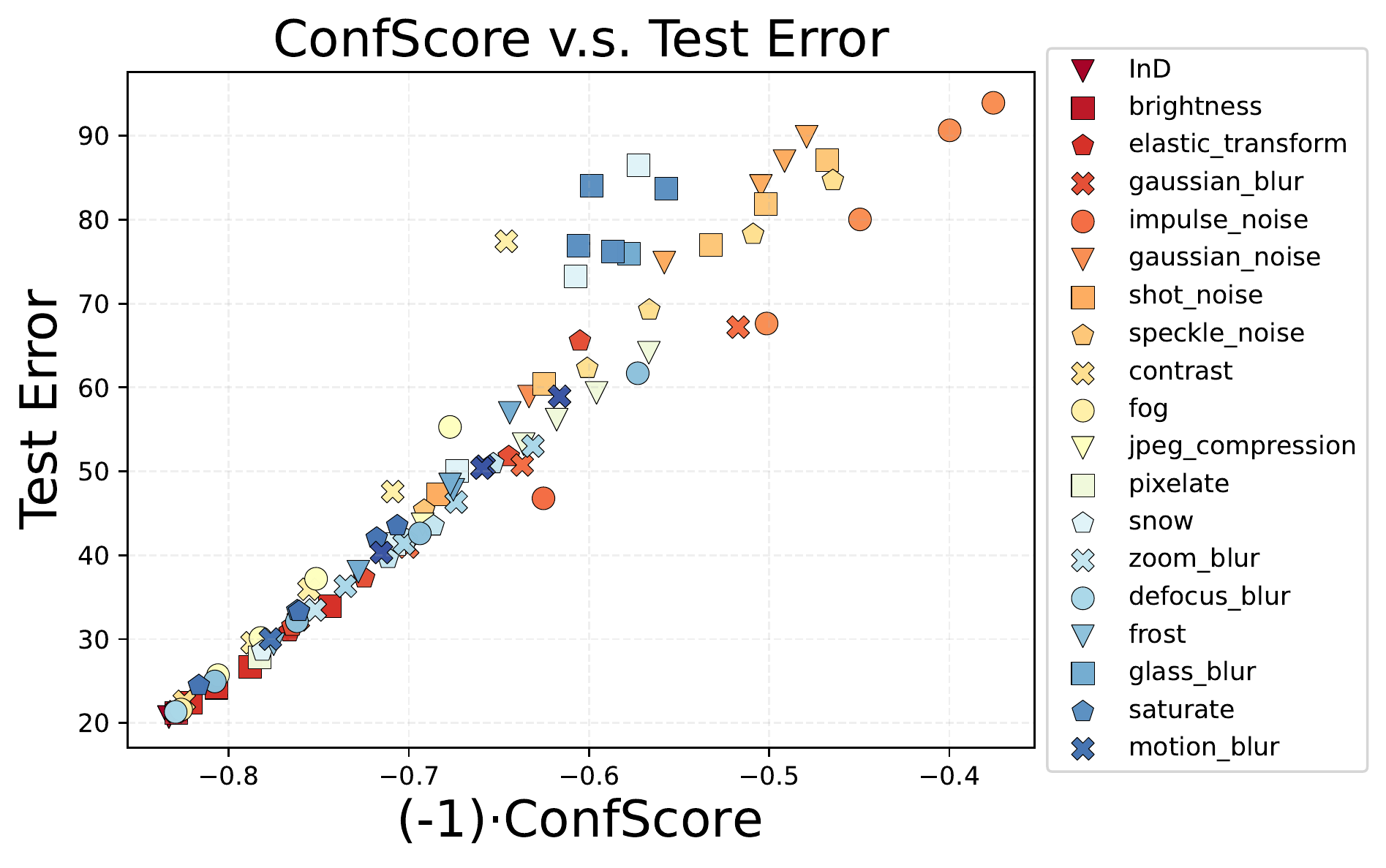}}
    \subfigure[Entropy.]{\includegraphics[width=.33\textwidth]{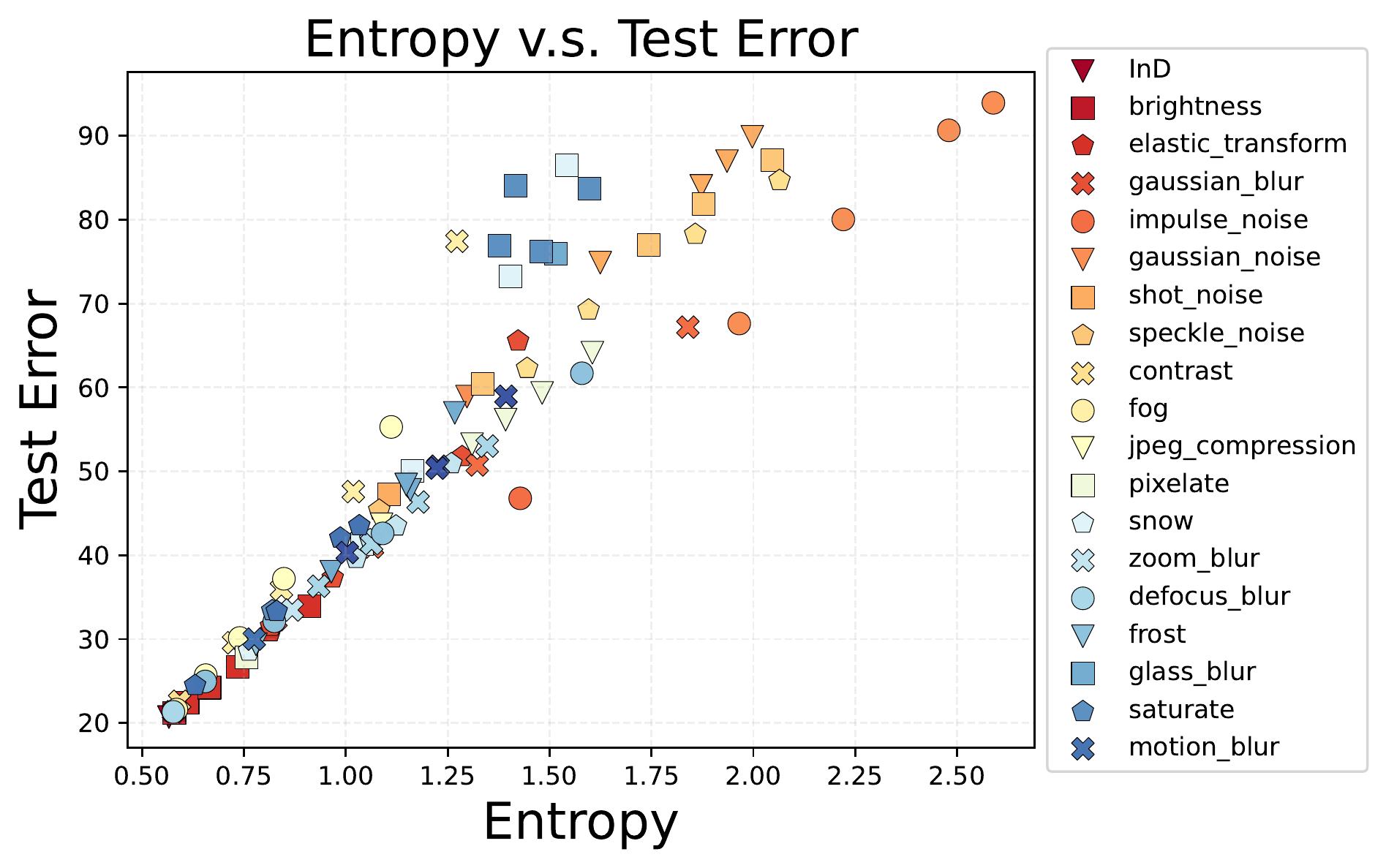}}
    \subfigure[AgreeScore.]{\includegraphics[width=.33\textwidth]{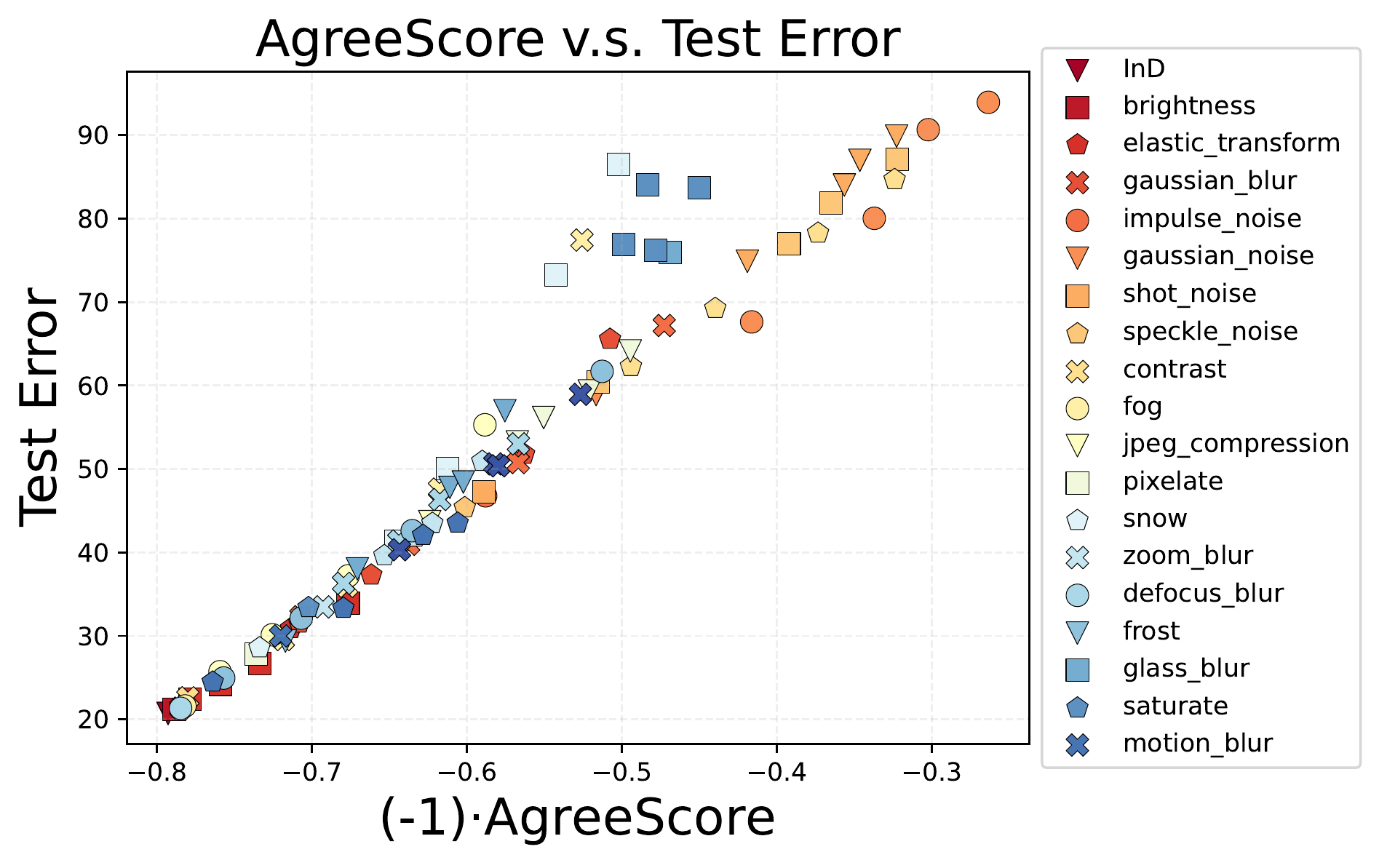}}
    \subfigure[ATC.]{\includegraphics[width=.33\textwidth]{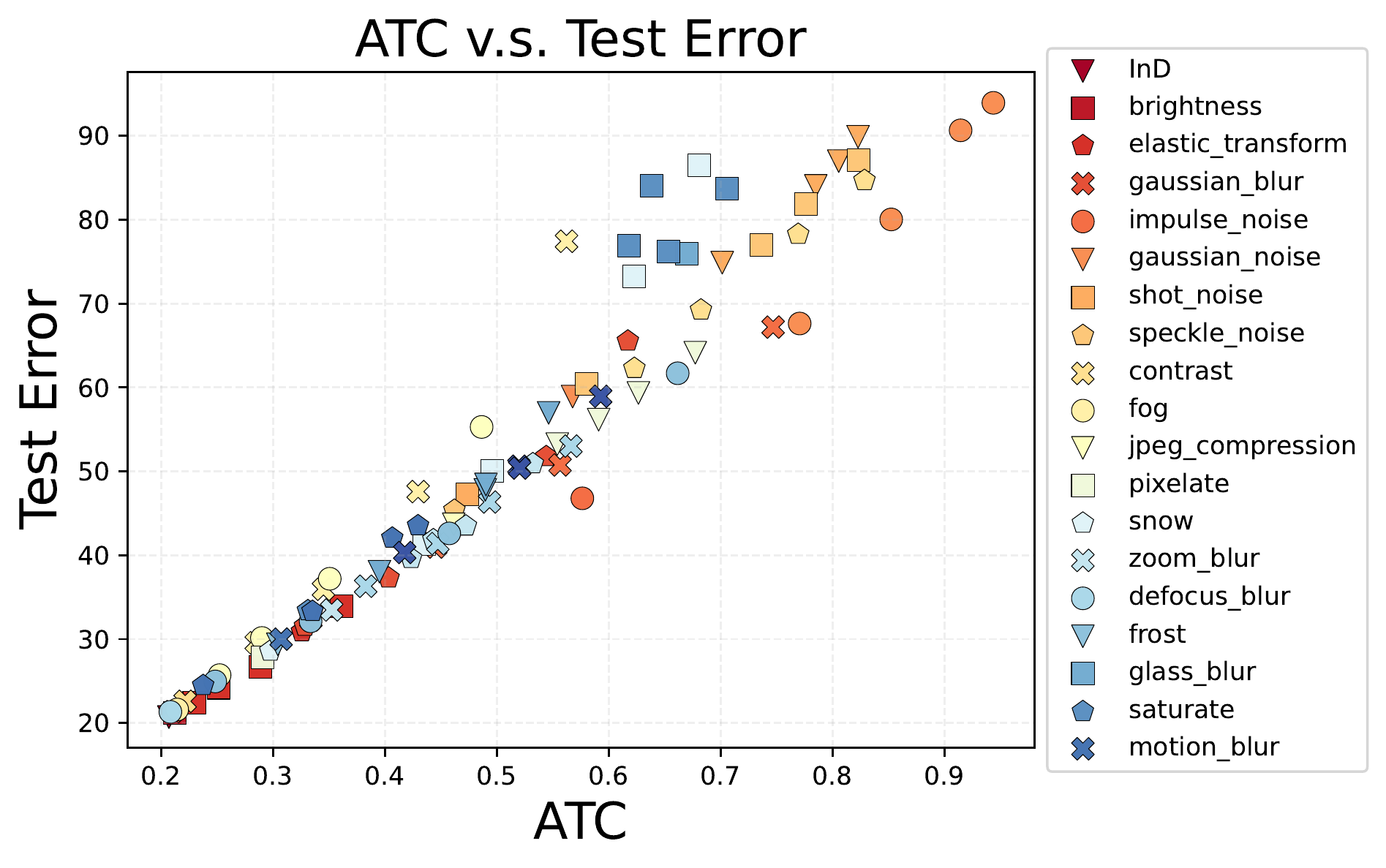}}
    \subfigure[ProjNorm.]{\includegraphics[width=.33\textwidth]{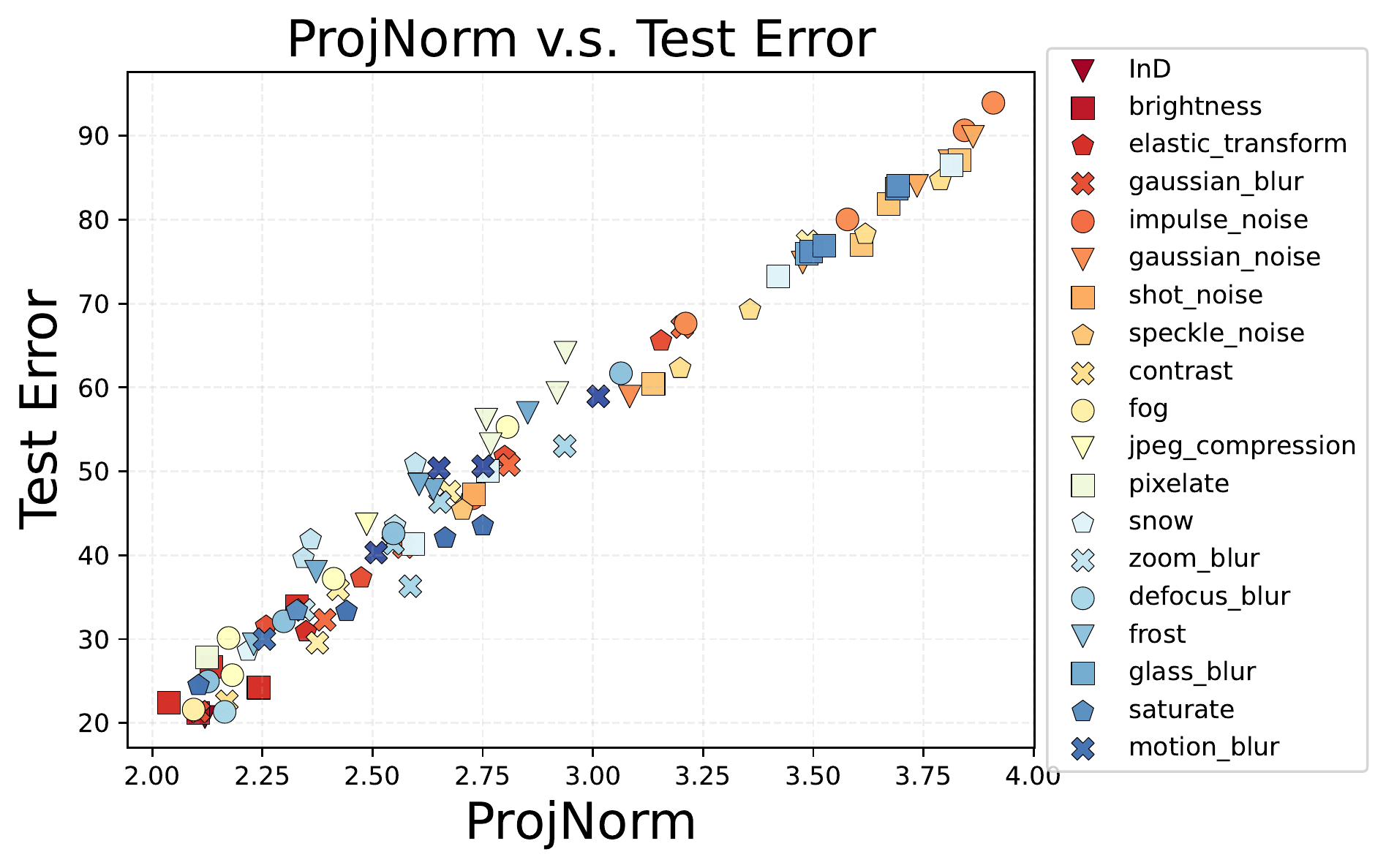}}
    \vspace{-0.1in}
    \caption{\textbf{Generalization prediction versus test error on CIFAR100 with ResNet18.} Compare out-of-distribution prediction performance of all methods. We plot the actual test error and the method prediction on each OOD dataset.
    Each point represents one InD/OOD dataset, and points with the same color and marker shape are the same corruption but with different severity levels.
    }
    \label{fig:compare-appendix-cifar100-resnet18}
    \vspace{-0.15in}
\end{figure*}

\begin{figure*}[ht]
    \centering
    \subfigure[Rotation.]{\includegraphics[width=.33\textwidth]{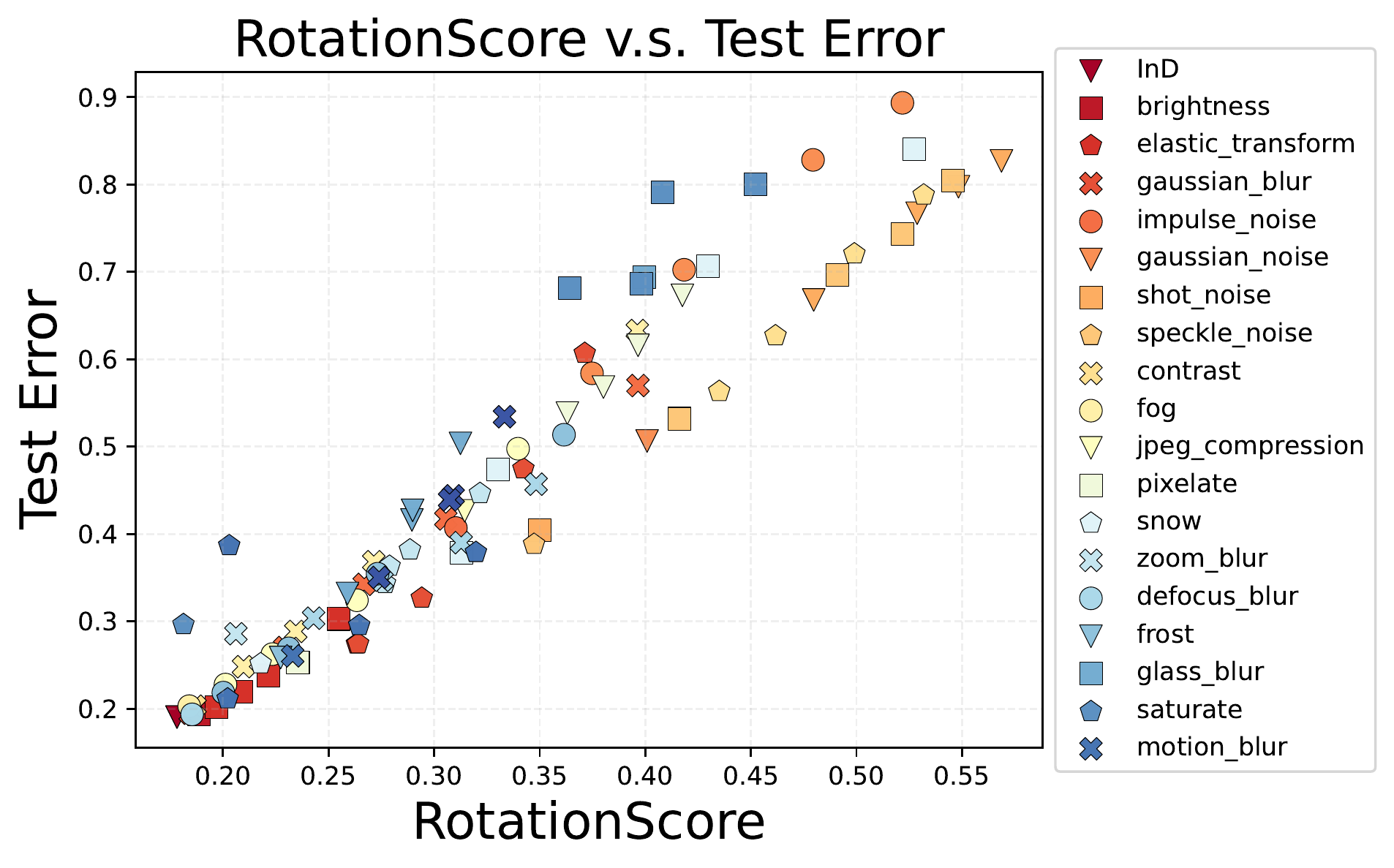}}
    \subfigure[ConfScore.]{\includegraphics[width=.33\textwidth]{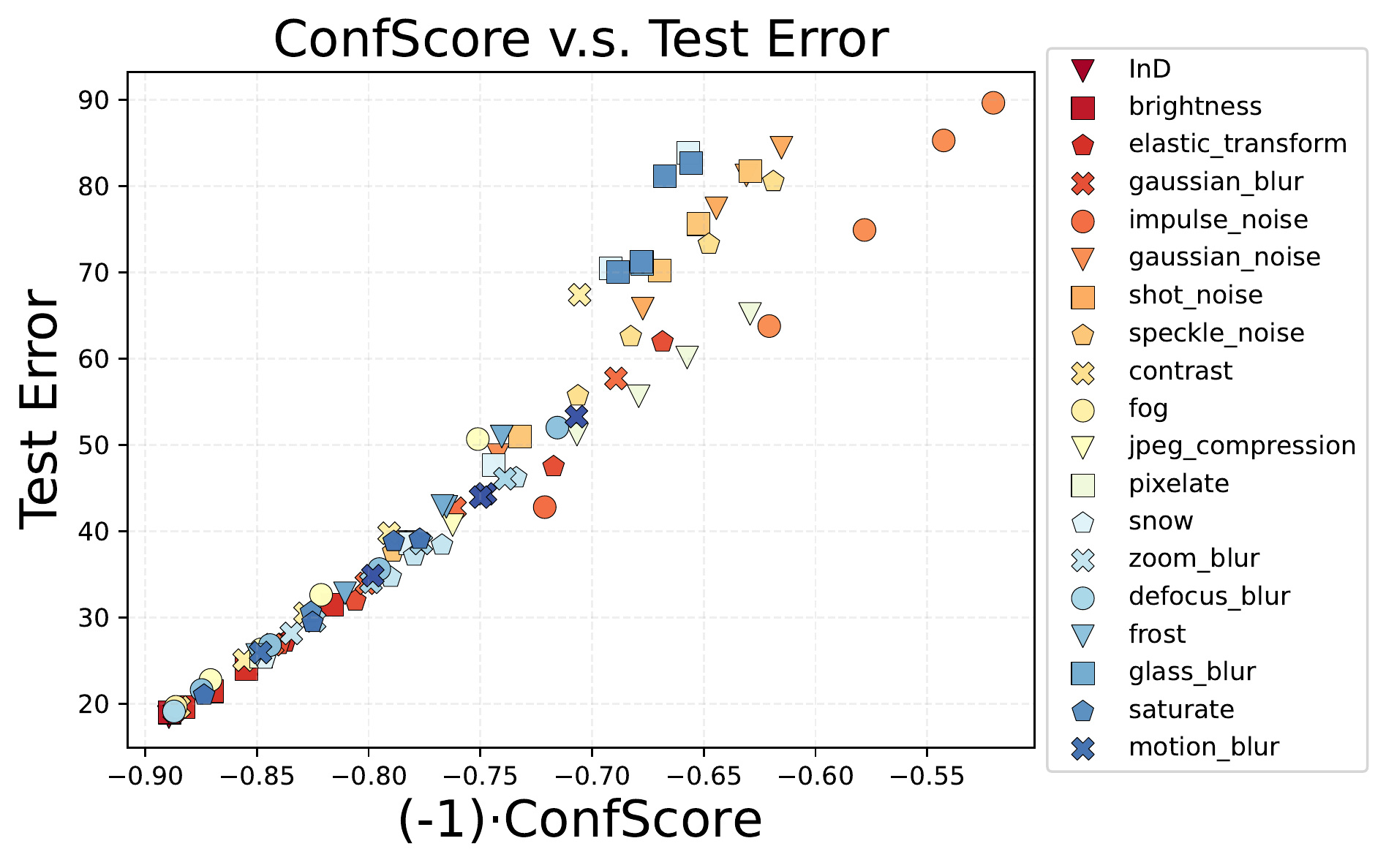}}
    \subfigure[Entropy.]{\includegraphics[width=.33\textwidth]{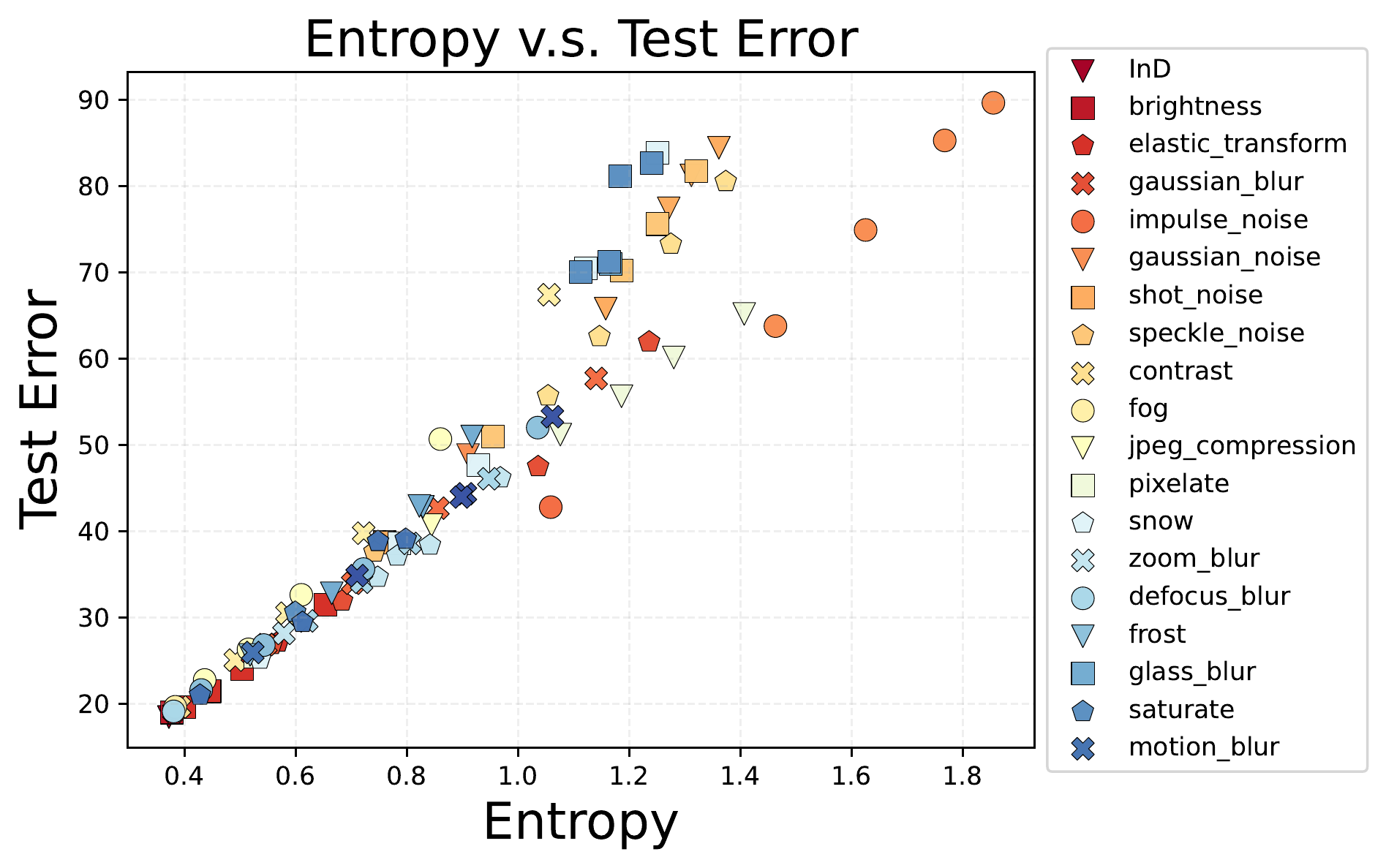}}
    \subfigure[AgreeScore.]{\includegraphics[width=.33\textwidth]{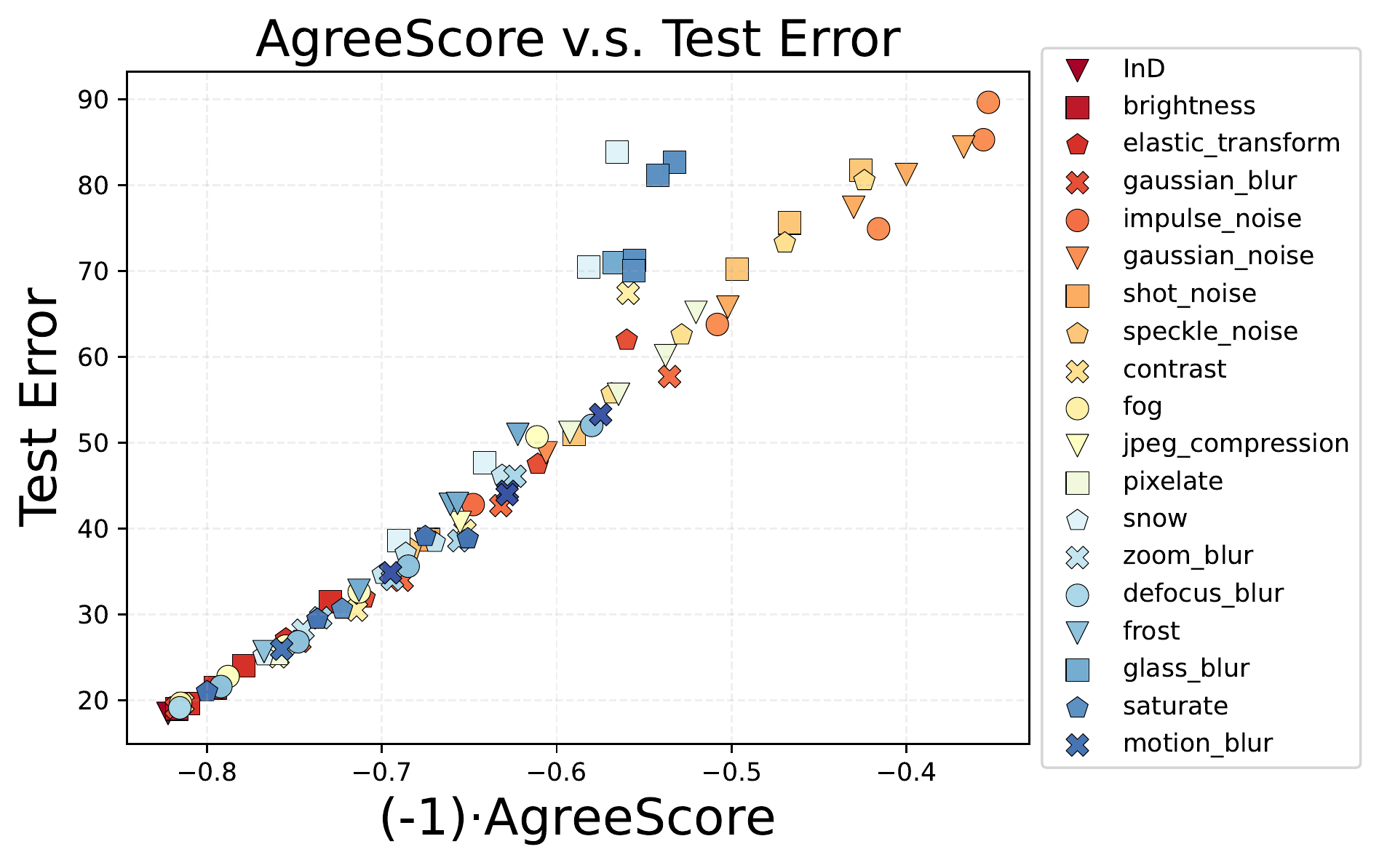}}
    \subfigure[ATC.]{\includegraphics[width=.33\textwidth]{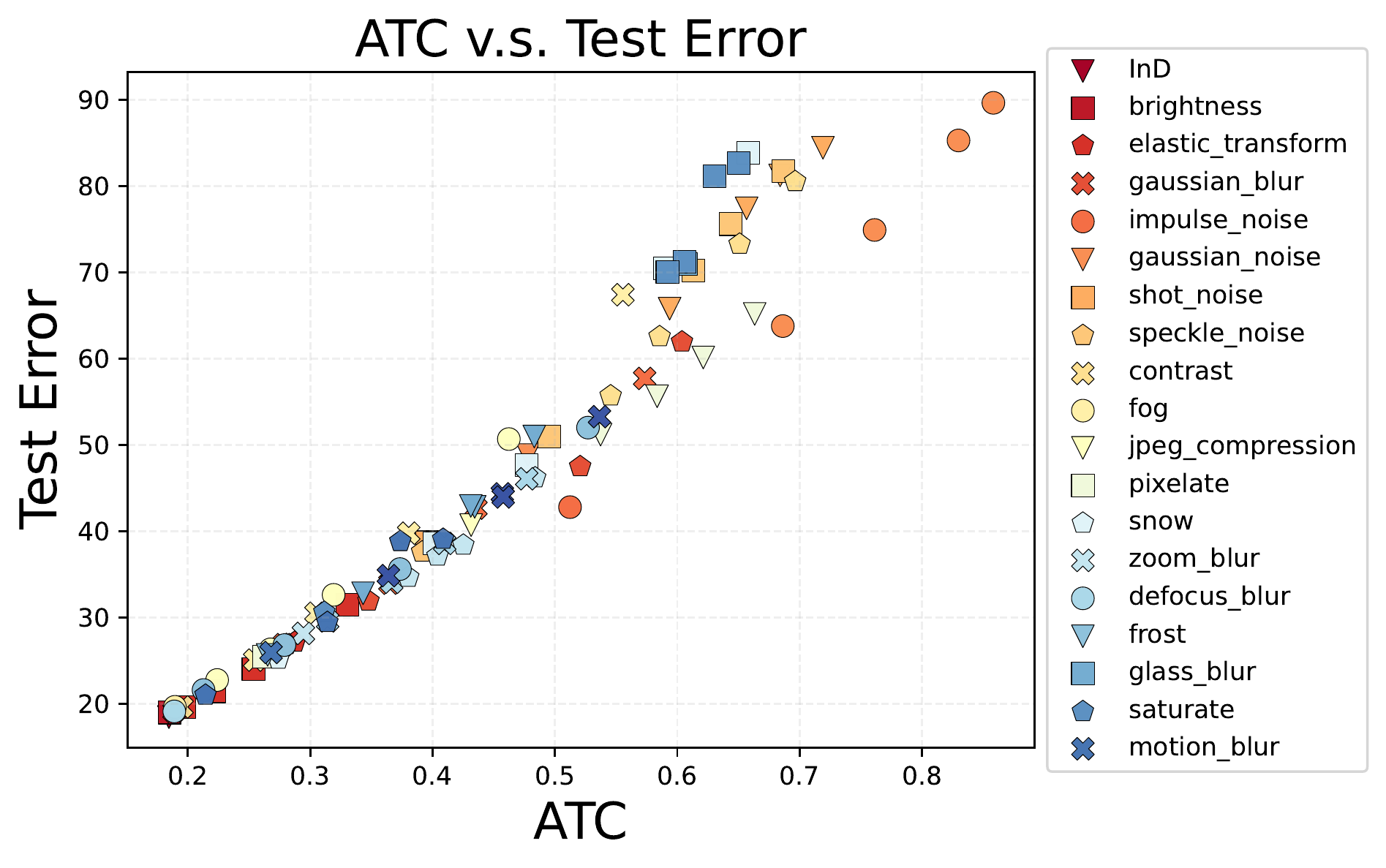}}
    \subfigure[ProjNorm.]{\includegraphics[width=.33\textwidth]{figs/cifar100/dnn_cifar100_resnet50_ProjNorm.pdf}}
    \vspace{-0.1in}
    \caption{\textbf{Generalization prediction versus test error on CIFAR100 with ResNet50.} Compare out-of-distribution prediction performance of all methods. We plot the actual test error and the method prediction on each OOD dataset.
    Each point represents one InD/OOD dataset, and points with the same color and marker shape are the same corruption but with different severity levels.
    }
    \label{fig:compare-appendix-cifar100-resnet50}
    \vspace{-0.15in}
\end{figure*}

\begin{figure*}[ht]
    \centering
    \subfigure[Rotation.]{\includegraphics[width=.33\textwidth]{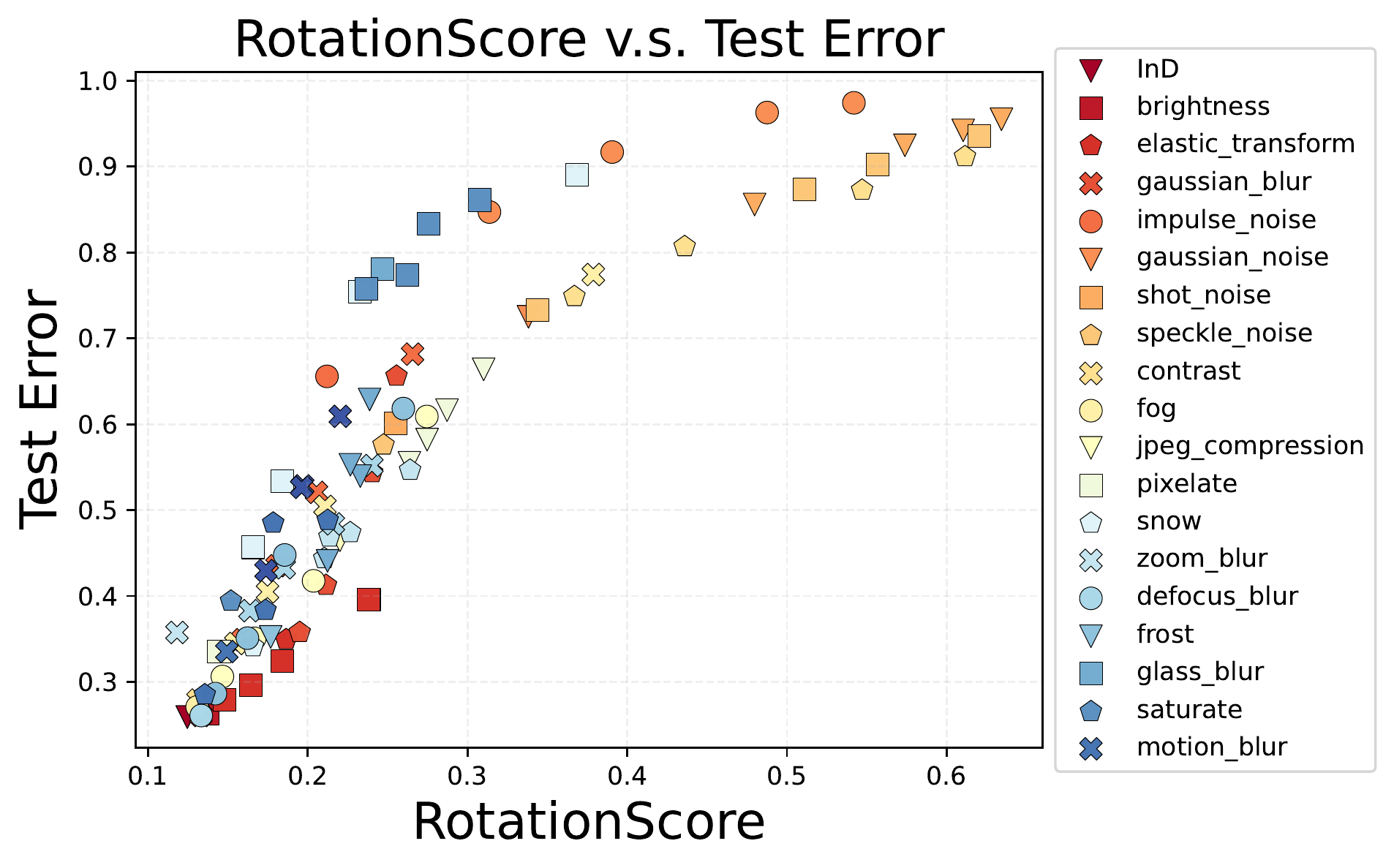}}
    \subfigure[ConfScore.]{\includegraphics[width=.33\textwidth]{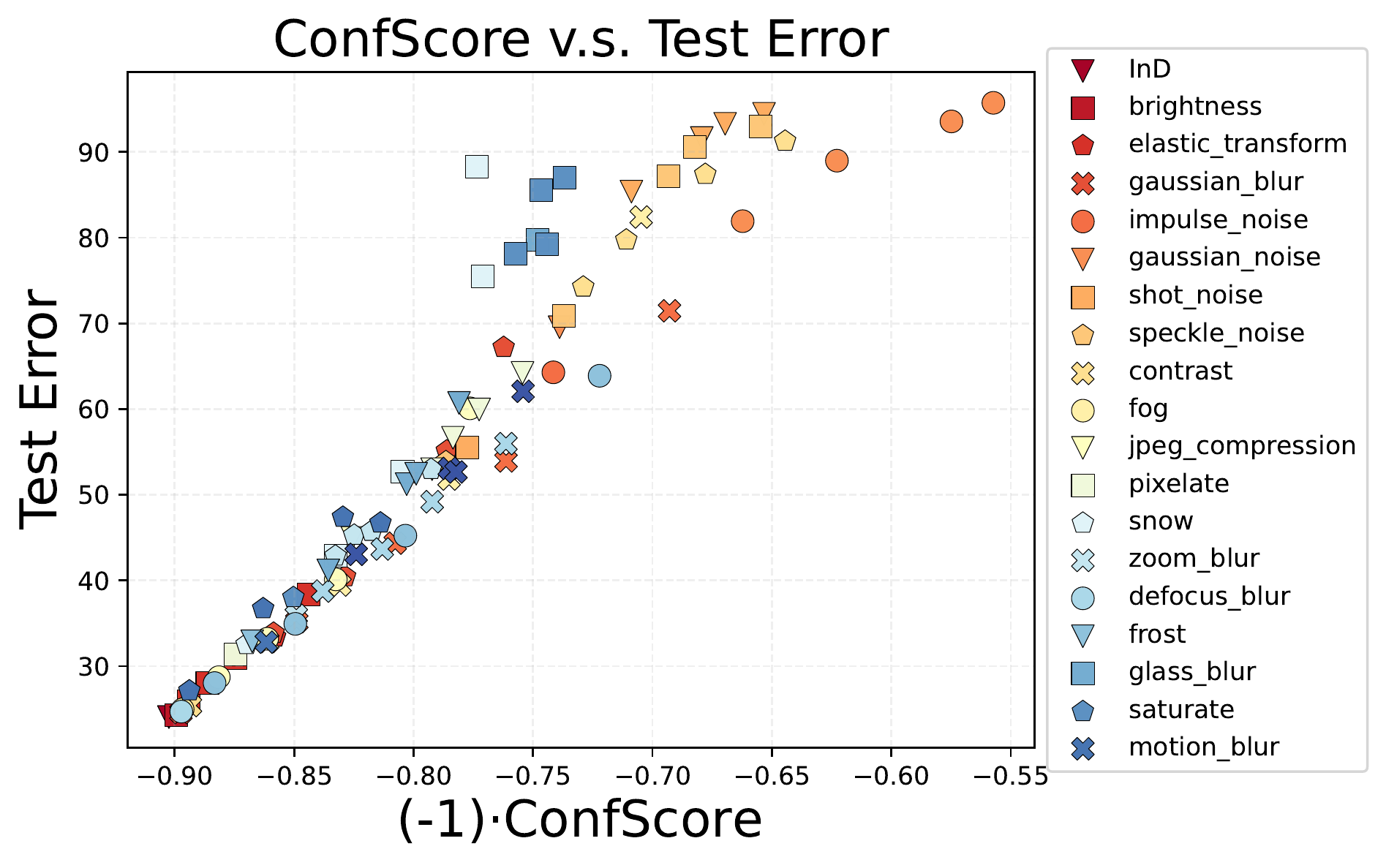}}
    \subfigure[Entropy.]{\includegraphics[width=.33\textwidth]{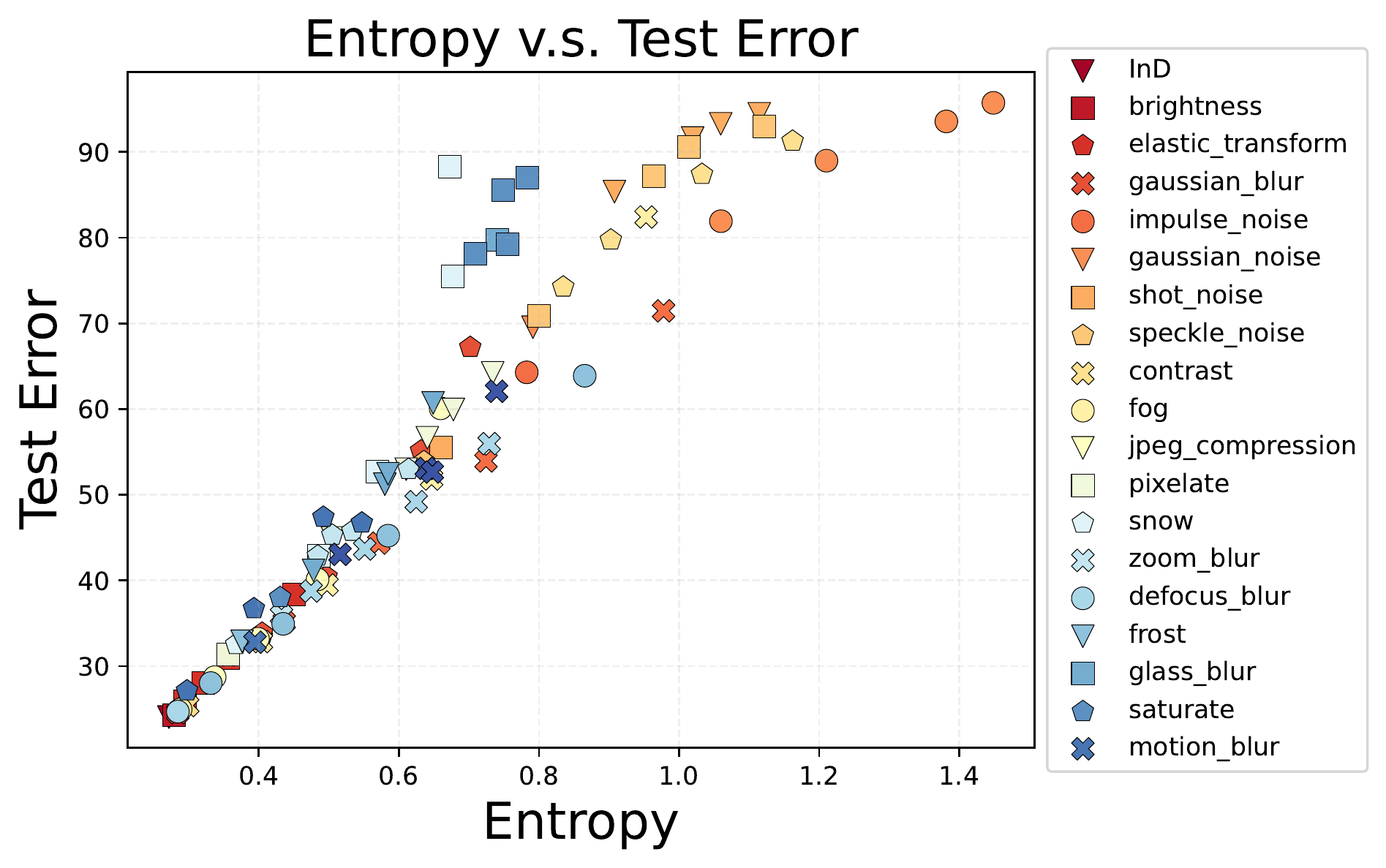}}
    \subfigure[AgreeScore.]{\includegraphics[width=.33\textwidth]{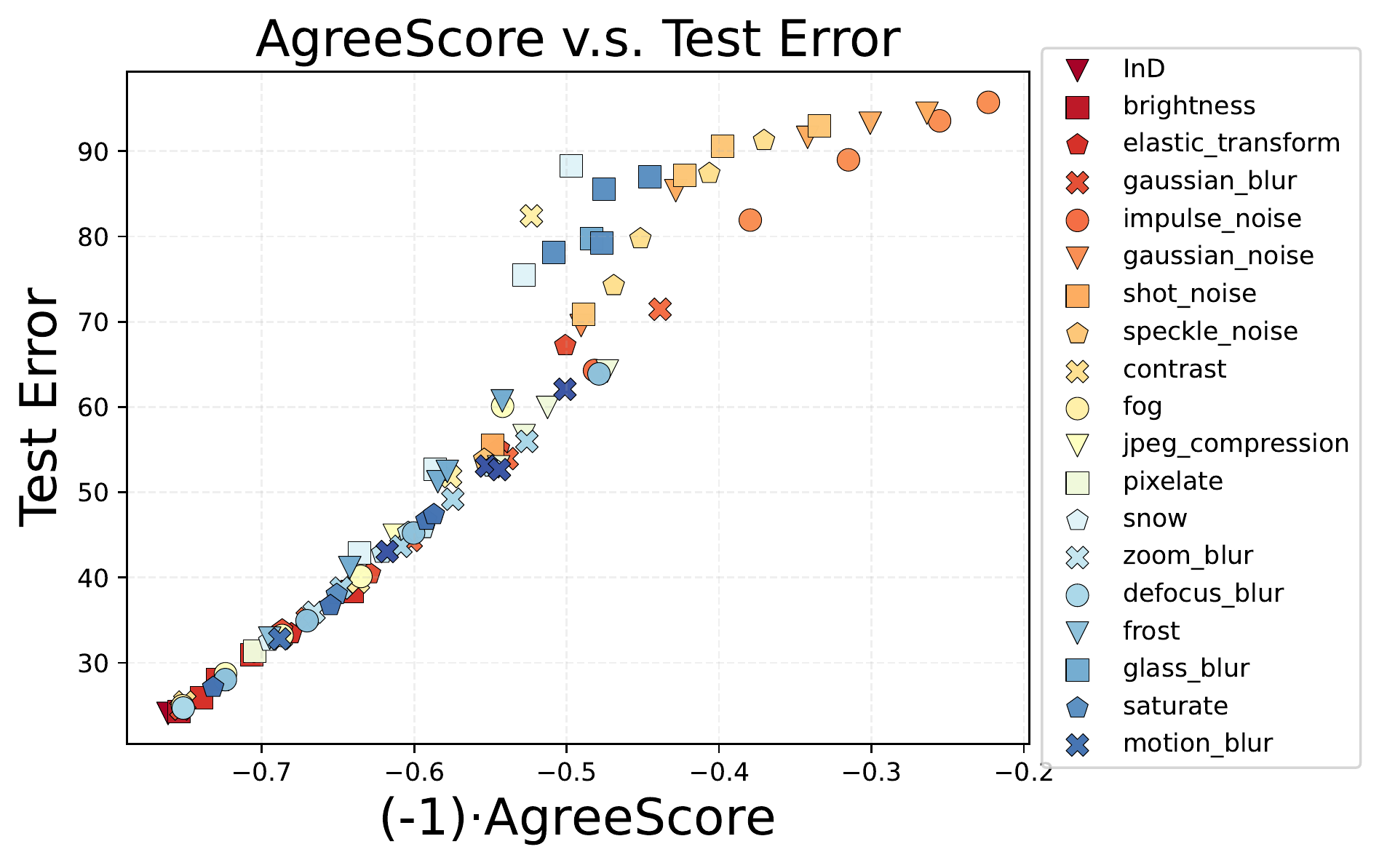}}
    \subfigure[ATC.]{\includegraphics[width=.33\textwidth]{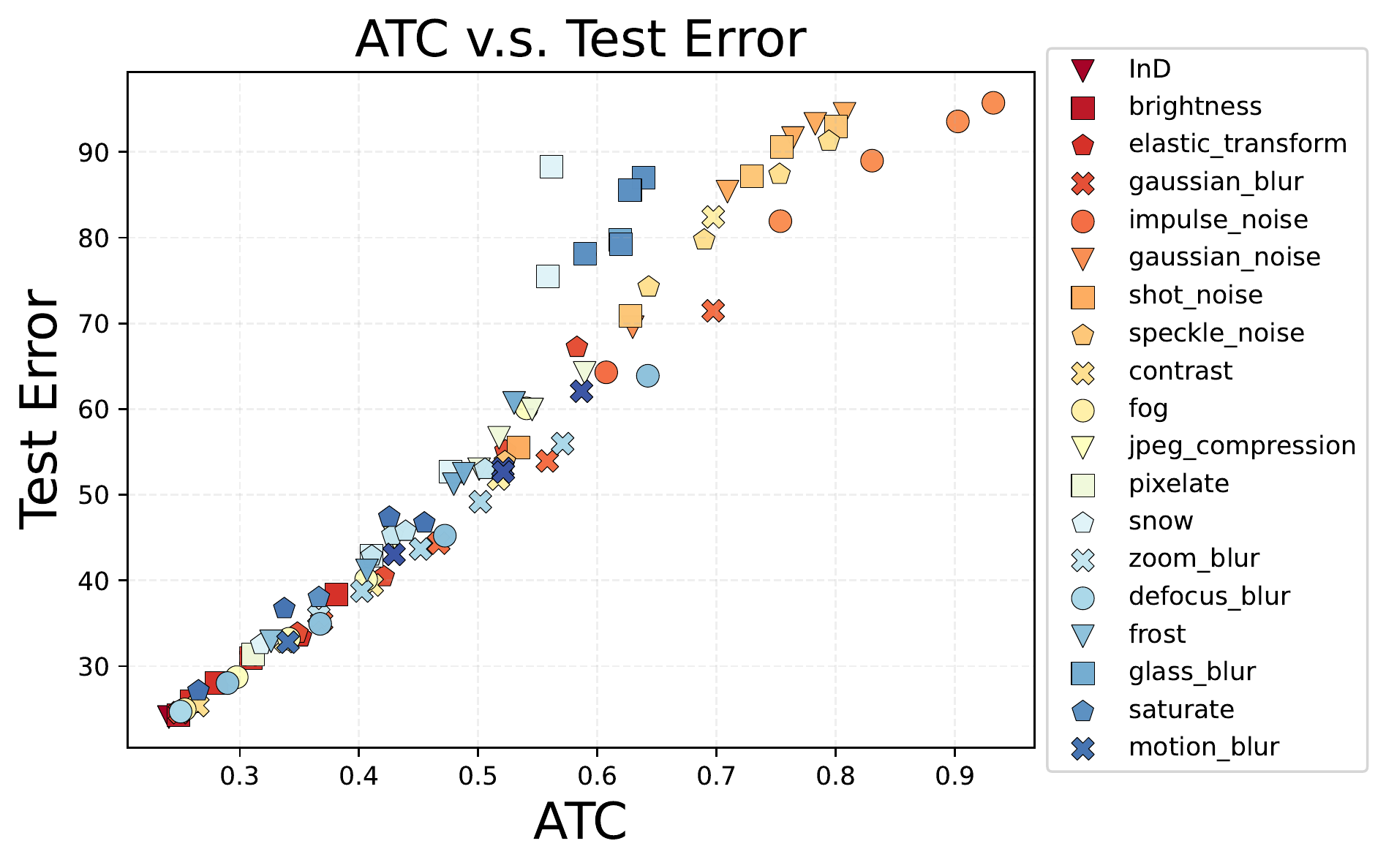}}
    \subfigure[ProjNorm.]{\includegraphics[width=.33\textwidth]{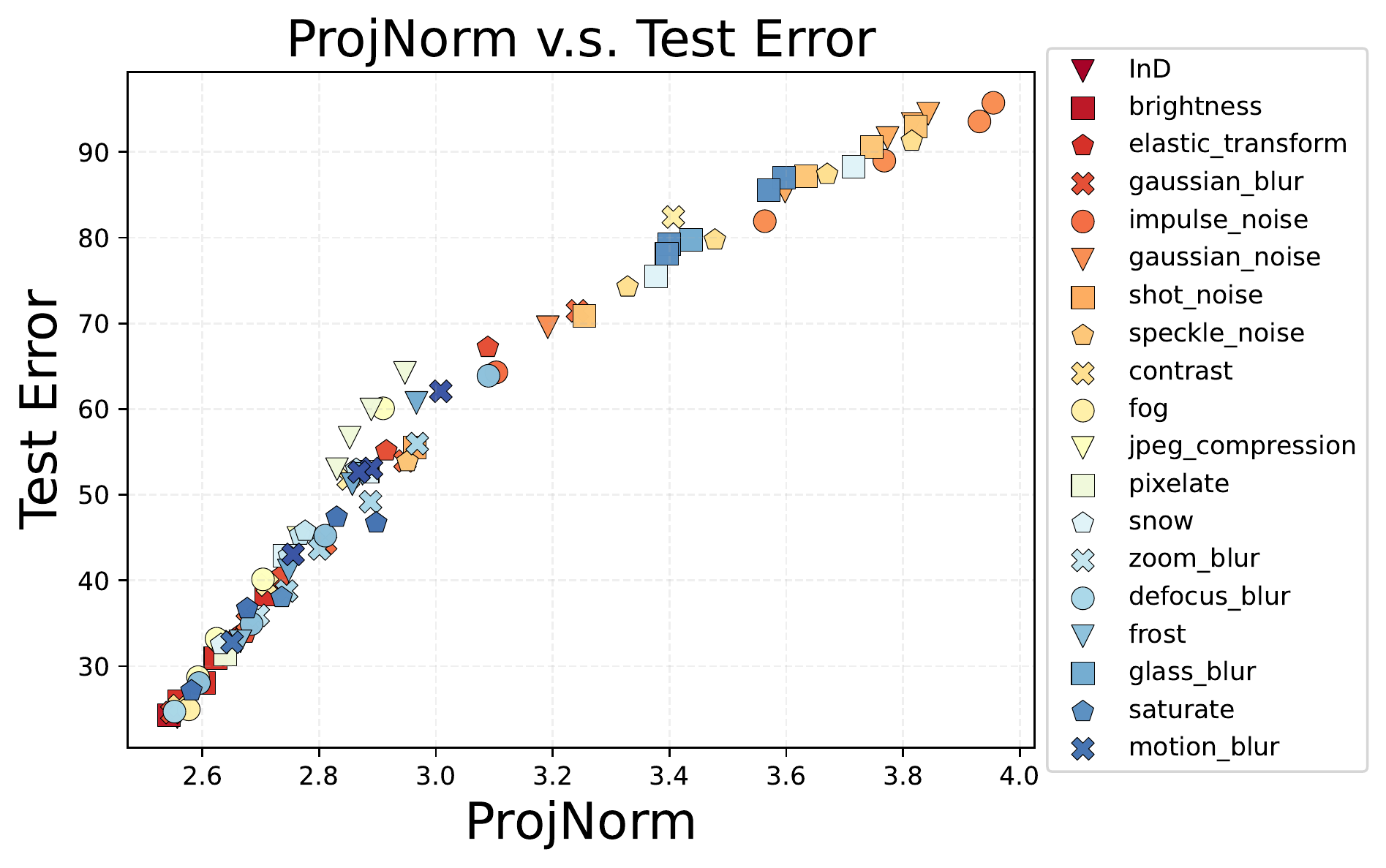}}
    \vspace{-0.1in}
    \caption{\textbf{Generalization prediction versus test error on CIFAR100 with VGG11.} Compare out-of-distribution prediction performance of all methods. We plot the actual test error and the method prediction on each OOD dataset.
    Each point represents one InD/OOD dataset, and points with the same color and marker shape are the same corruption but with different severity levels.
    }
    \label{fig:compare-appendix-cifar100-vgg}
    \vspace{-0.15in}
\end{figure*}

\begin{figure*}[ht]
    \centering
    \subfigure[ConfScore.]{\includegraphics[width=.33\textwidth]{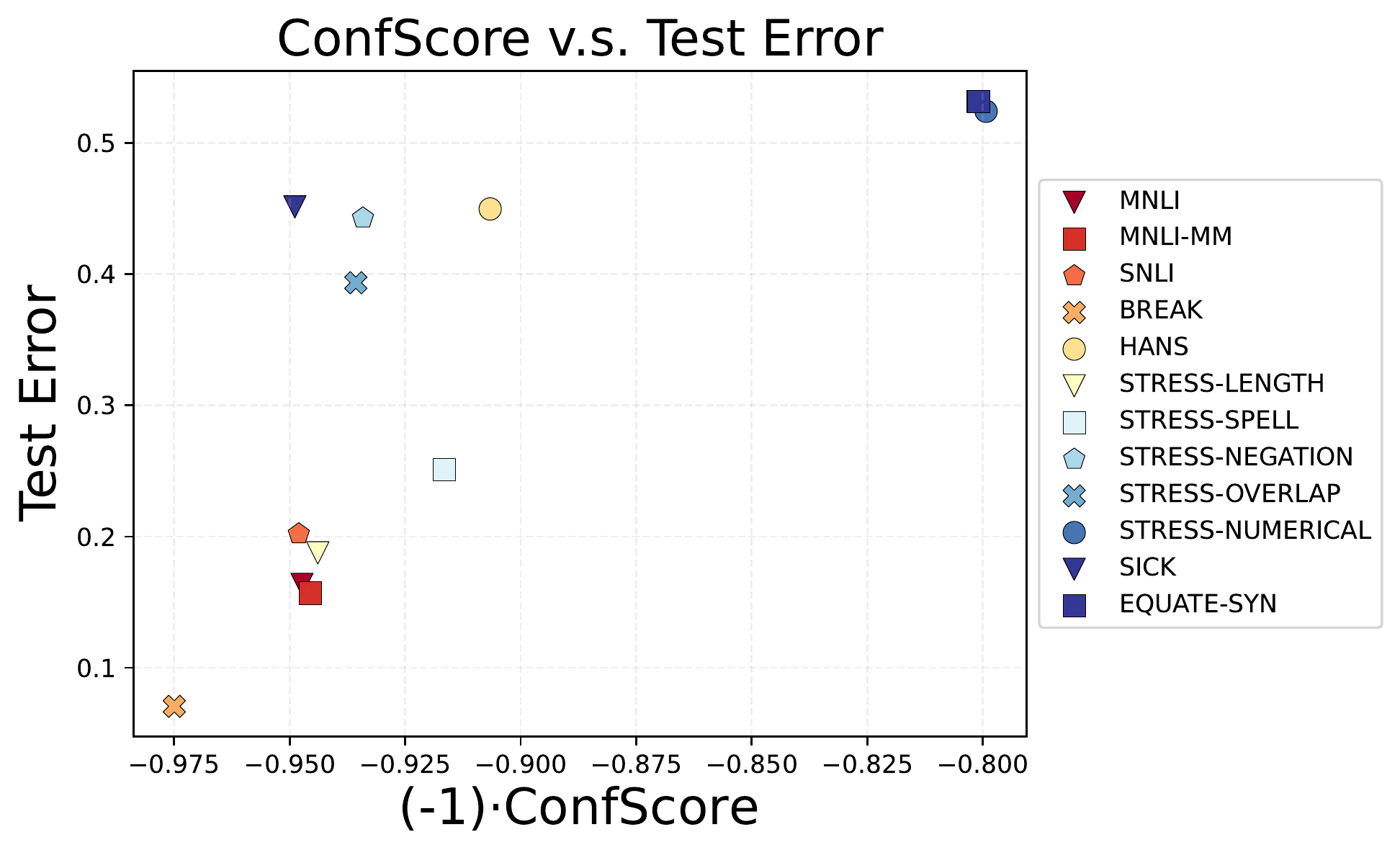}}
    \subfigure[Entropy.]{\includegraphics[width=.33\textwidth]{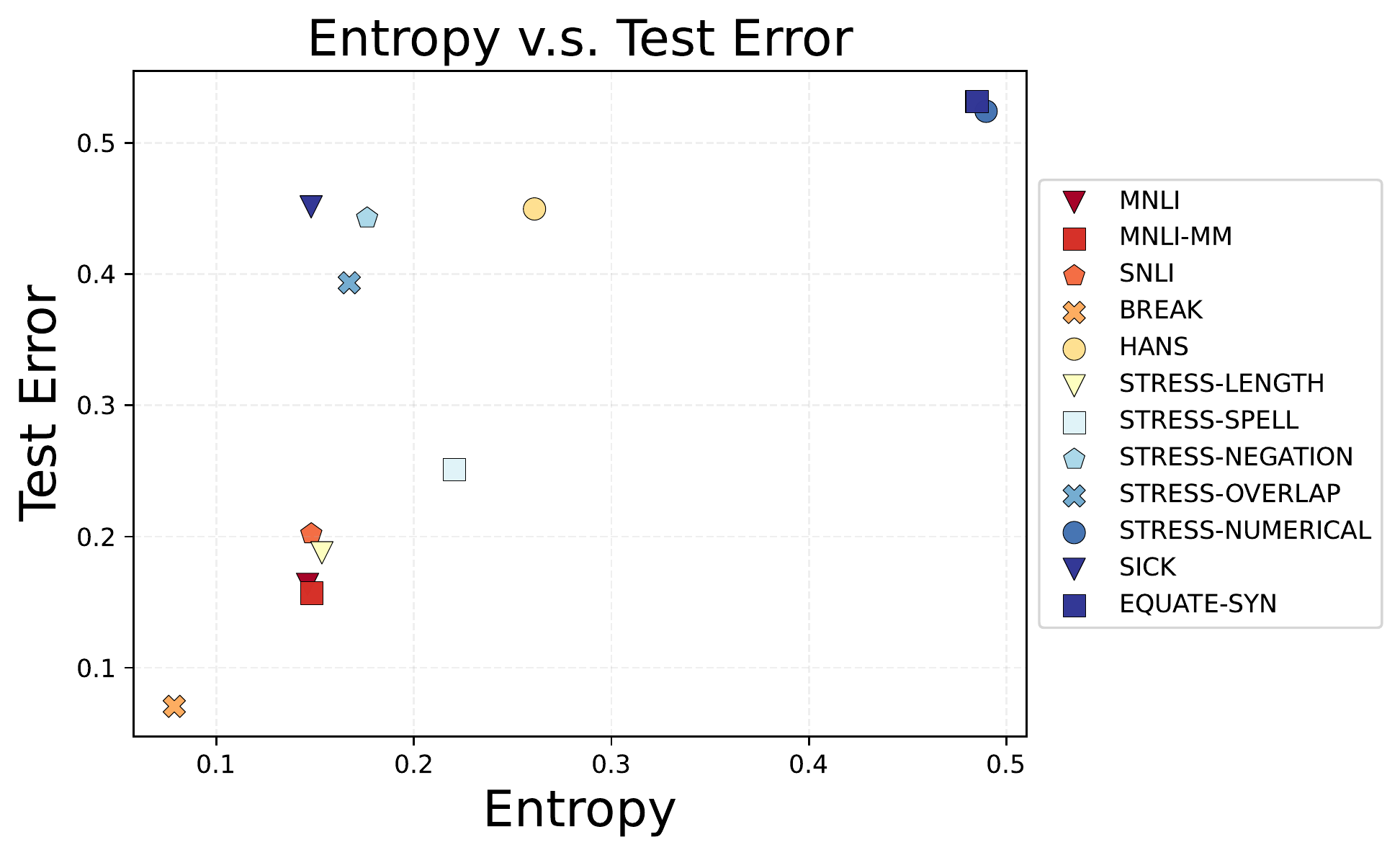}}
    \subfigure[AgreeScore.]{\includegraphics[width=.33\textwidth]{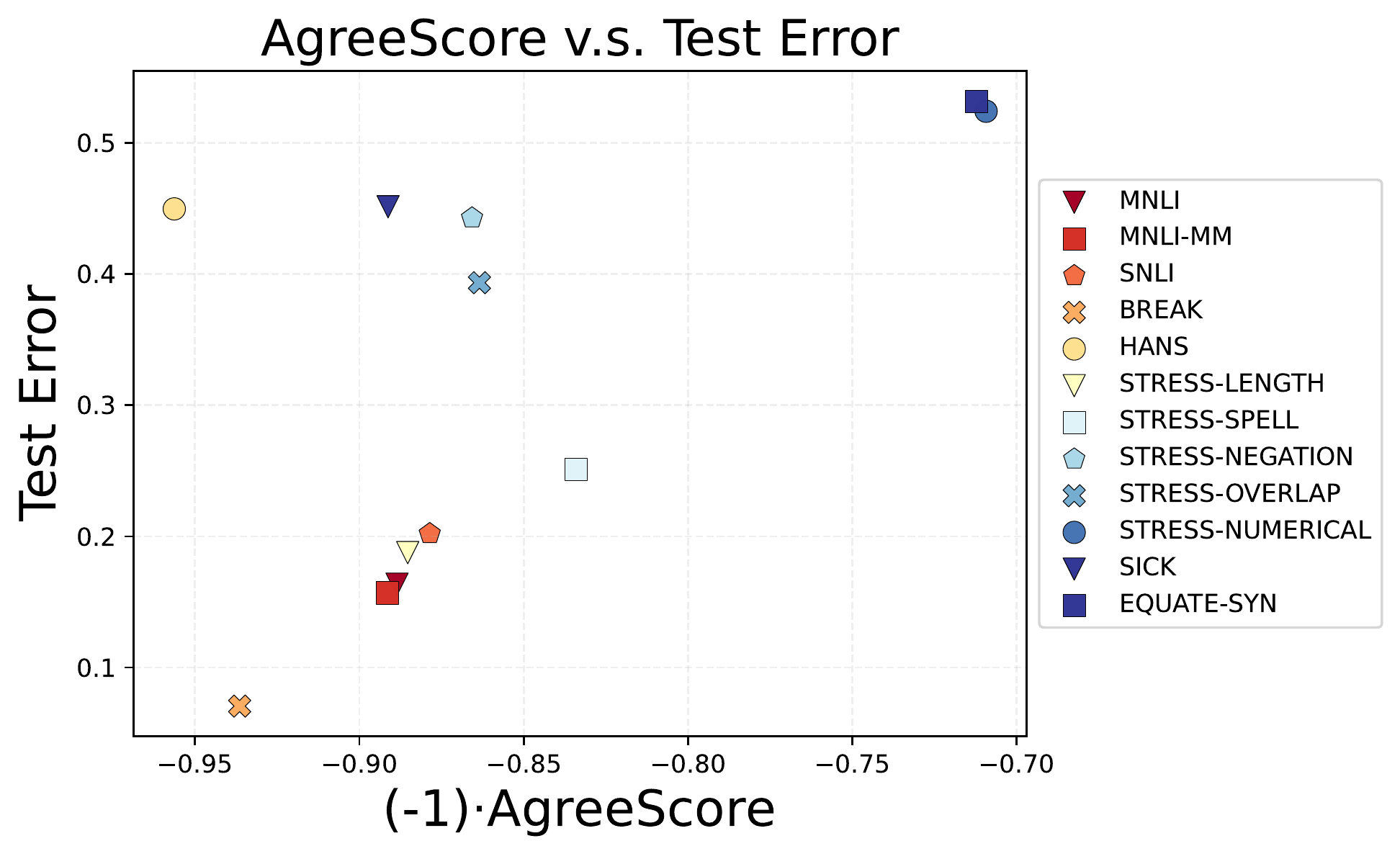}}
    \subfigure[ATC.]{\includegraphics[width=.33\textwidth]{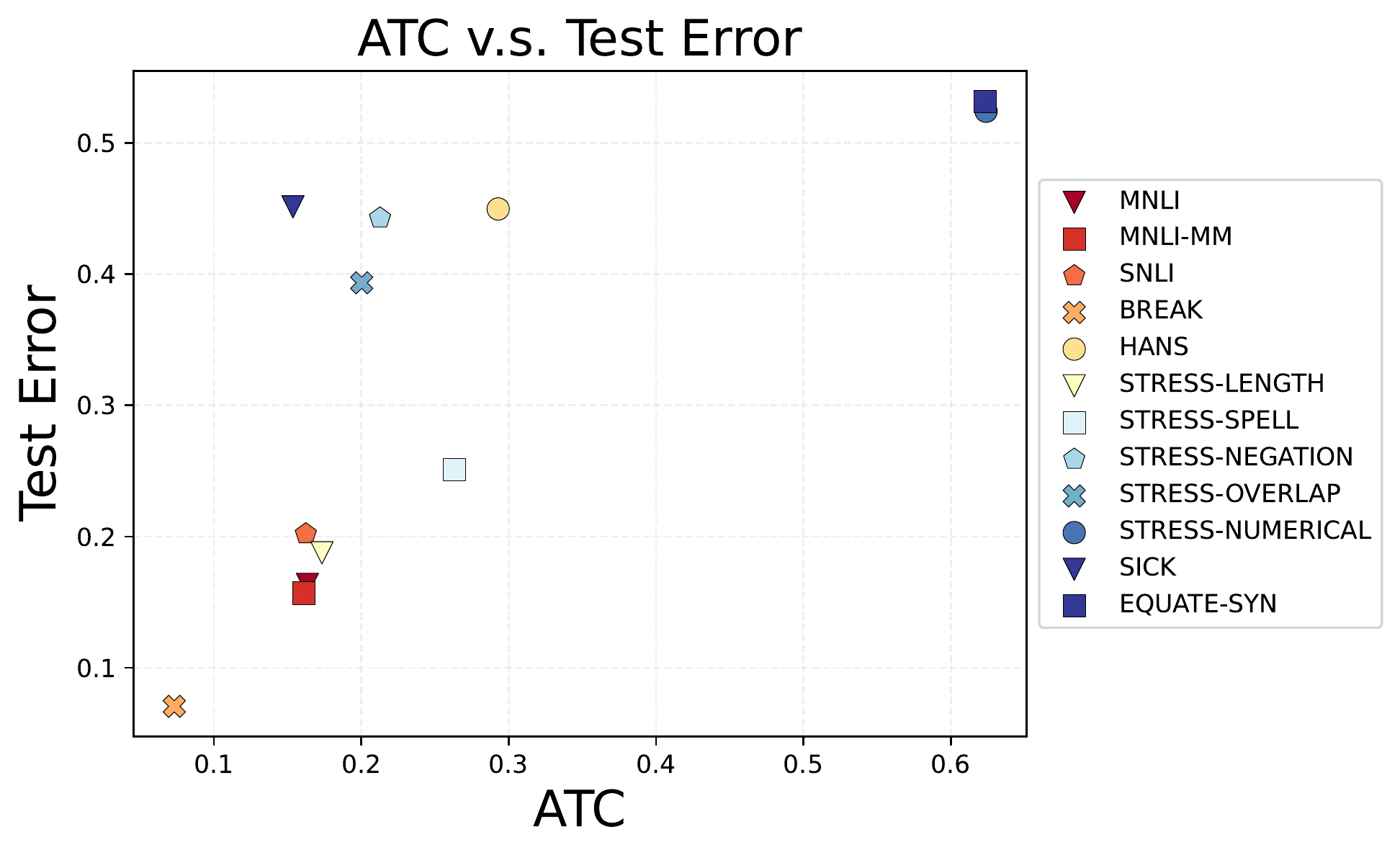}}
    \subfigure[ProjNorm.]{\includegraphics[width=.33\textwidth]{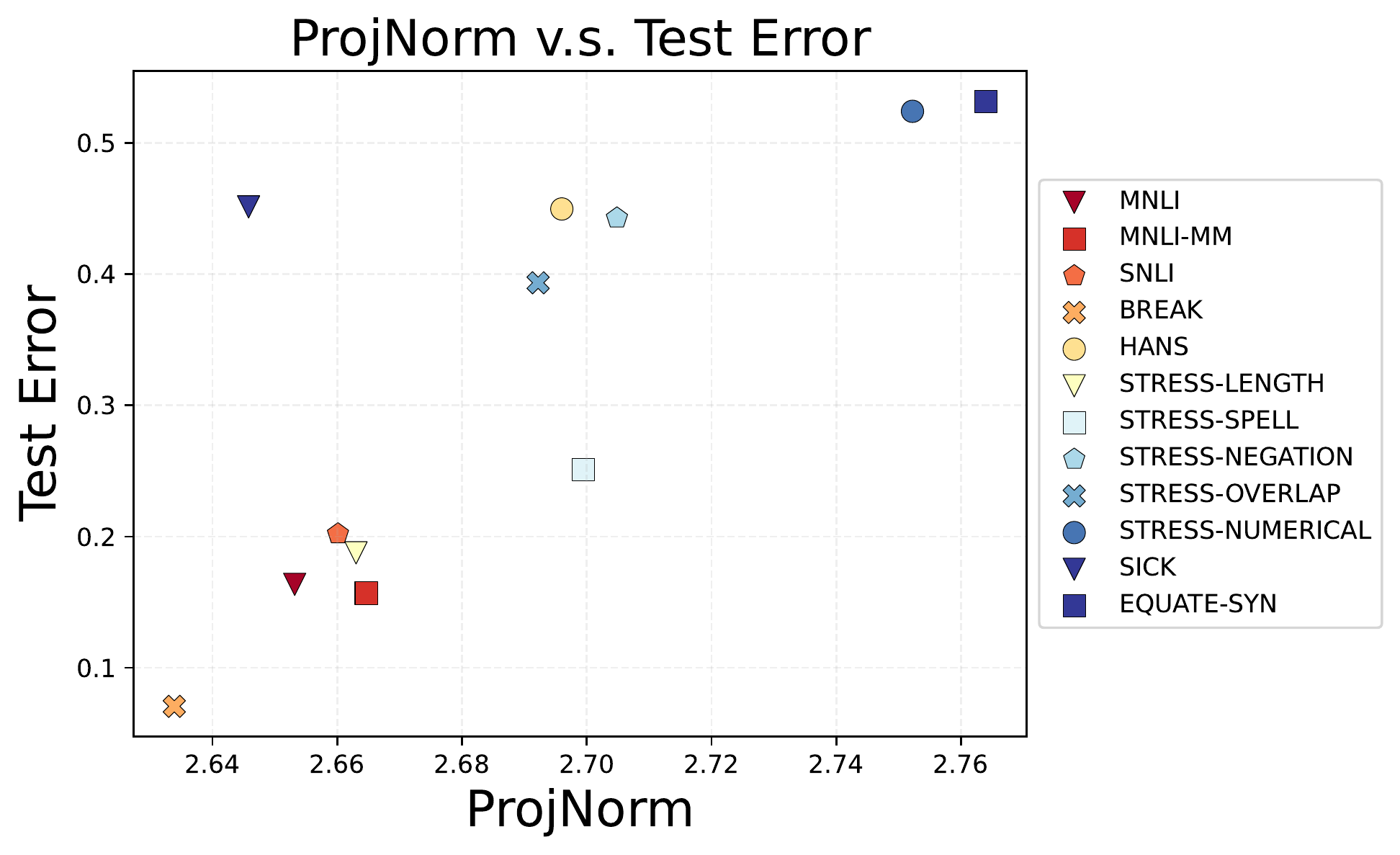}}
    \vspace{-0.1in}
    \caption{\textbf{Generalization prediction versus test error on MNLI with BERT.} Compare out-of-distribution prediction performance of all methods (except for Rotation). We plot the actual test error and the method prediction on each InD/OOD dataset.
    }
    \label{fig:compare-appendix-mnli-bert}
    \vspace{-0.15in}
\end{figure*}

\begin{figure*}[ht]
    \centering
    \subfigure[ConfScore.]{\includegraphics[width=.33\textwidth]{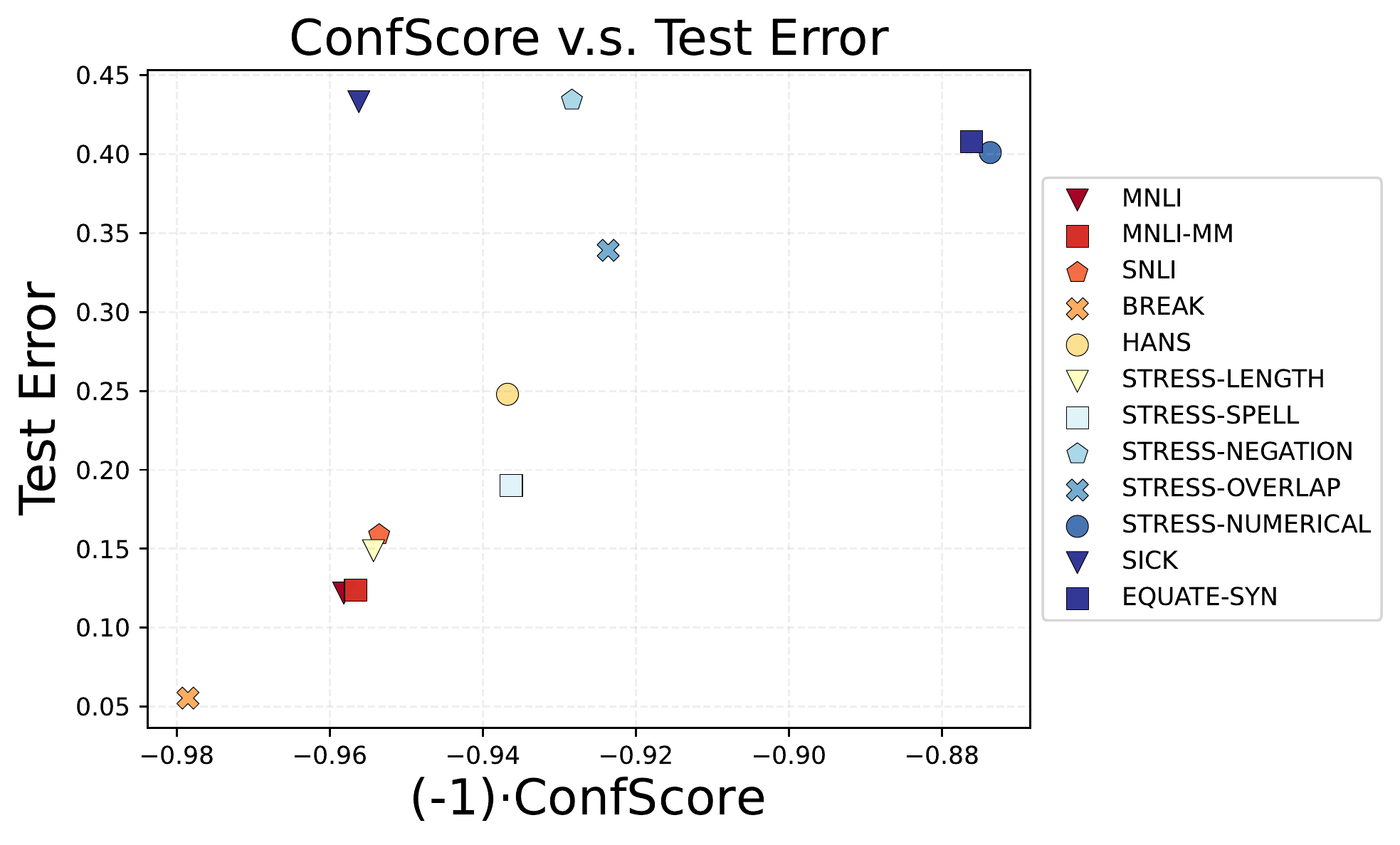}}
    \subfigure[Entropy.]{\includegraphics[width=.33\textwidth]{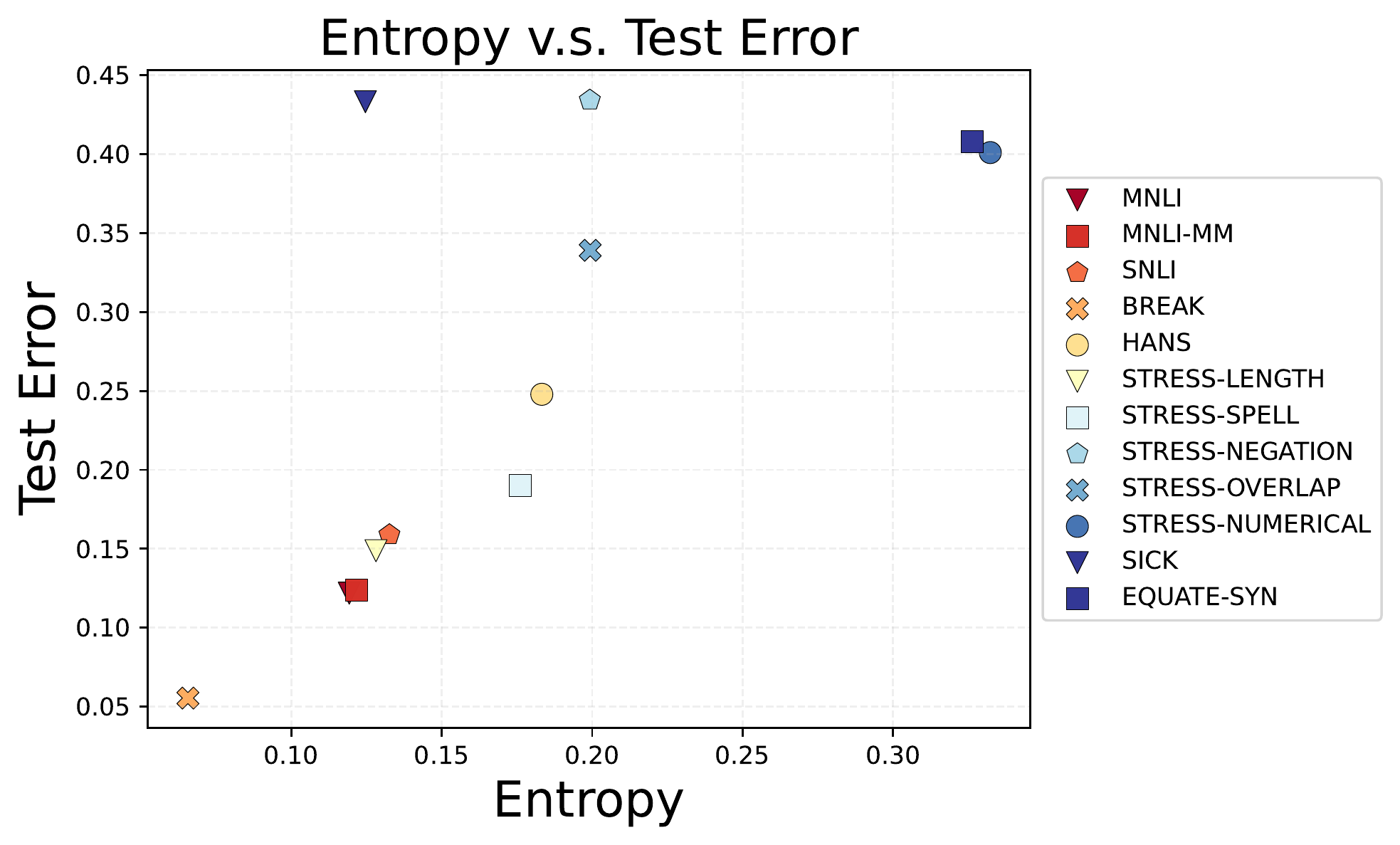}}
    \subfigure[AgreeScore.]{\includegraphics[width=.33\textwidth]{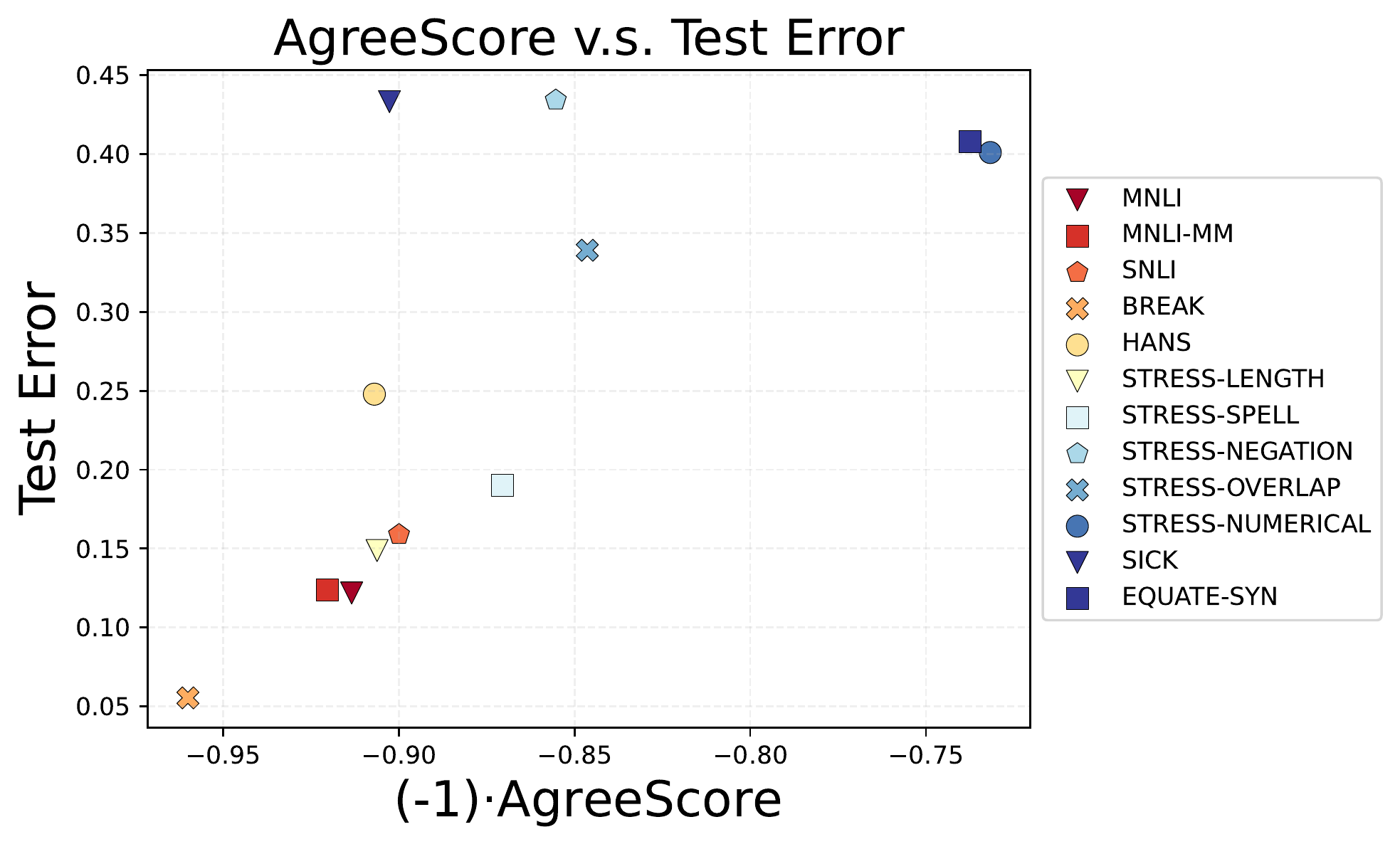}}
    \subfigure[ATC.]{\includegraphics[width=.33\textwidth]{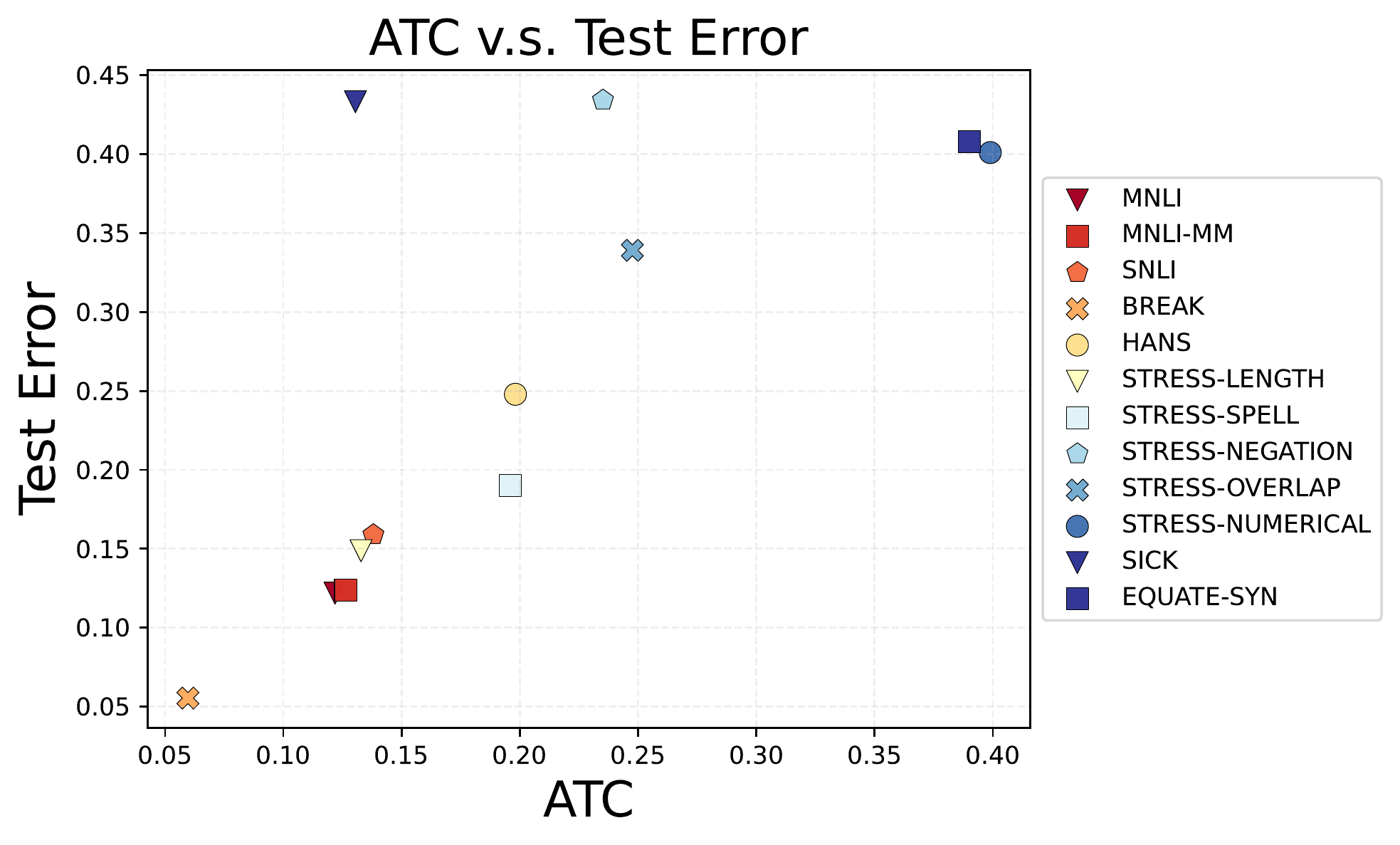}}
    \subfigure[ProjNorm.]{\includegraphics[width=.33\textwidth]{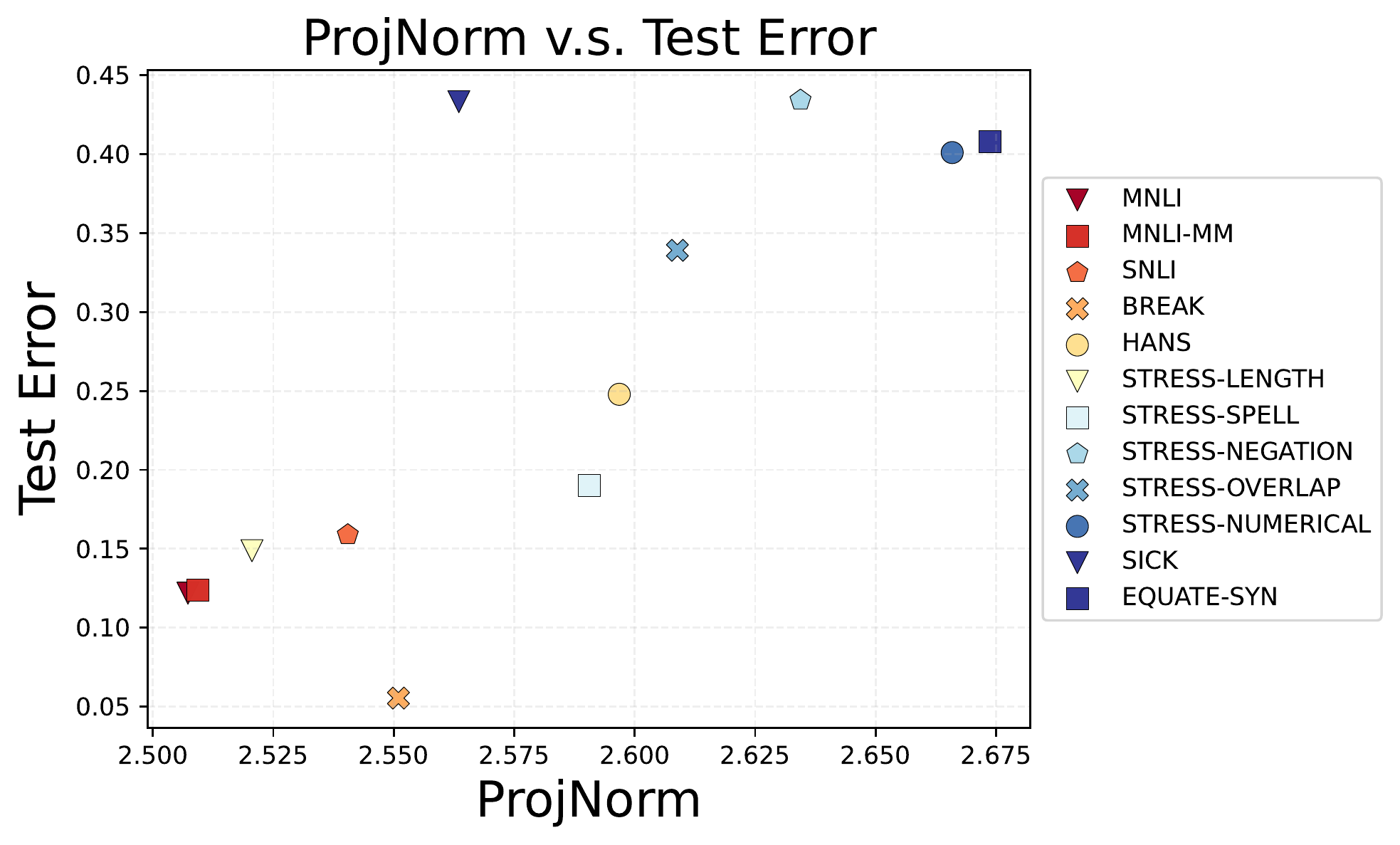}}
    \vspace{-0.1in}
    \caption{\textbf{Generalization prediction versus test error on MNLI with RoBERTa.} Compare out-of-distribution prediction performance of all methods (except for Rotation). We plot the actual test error and the method prediction on each InD/OOD dataset.
    }
    \label{fig:compare-appendix-mnli-roberta}
    \vspace{-0.15in}
\end{figure*}

\textbf{Scatter plots of generalization prediction versus test error.} We present the scatter plots for all methods (displayed in Table~\ref{table:main-table}) in Figures~\ref{fig:compare-appendix-cifar10-resnet18}--\ref{fig:compare-appendix-mnli-roberta}. More specifically, the figures plot results for the following models and datasets:
\vspace{-0.1in}
\begin{itemize}
    \item CIFAR10: ResNet18 (Figure~\ref{fig:compare-appendix-cifar10-resnet18}), ResNet50 (Figure~\ref{fig:compare-appendix-cifar10-resnet50}), and VGG11 (Figure~\ref{fig:compare-appendix-cifar10-vgg}).
    \item CIFAR100:   ResNet18 (Figure~\ref{fig:compare-appendix-cifar100-resnet18}),  ResNet50 (Figure~\ref{fig:compare-appendix-cifar100-resnet50}),  VGG11 (Figure~\ref{fig:compare-appendix-cifar100-vgg}).
    \item MNLI:  BERT (Figure~\ref{fig:compare-appendix-mnli-bert}),  RoBERTa (Figure~\ref{fig:compare-appendix-mnli-roberta}).
\end{itemize}

\textbf{Sensitivity analysis.} We present more results on the sensitivity analysis of \pj. We vary the number of iterations $T$~(Table~\ref{table:table-sensitivity-appendix-T}), the number of test samples $m$~(Table~\ref{table:table-sensitivity-appendix-m}), and the learning rate $\eta$~(Table~\ref{table:table-sensitivity-lr-appendix}).

\begin{figure*}[ht!]
    \centering
    \subfigure{\includegraphics[width=.23\textwidth]{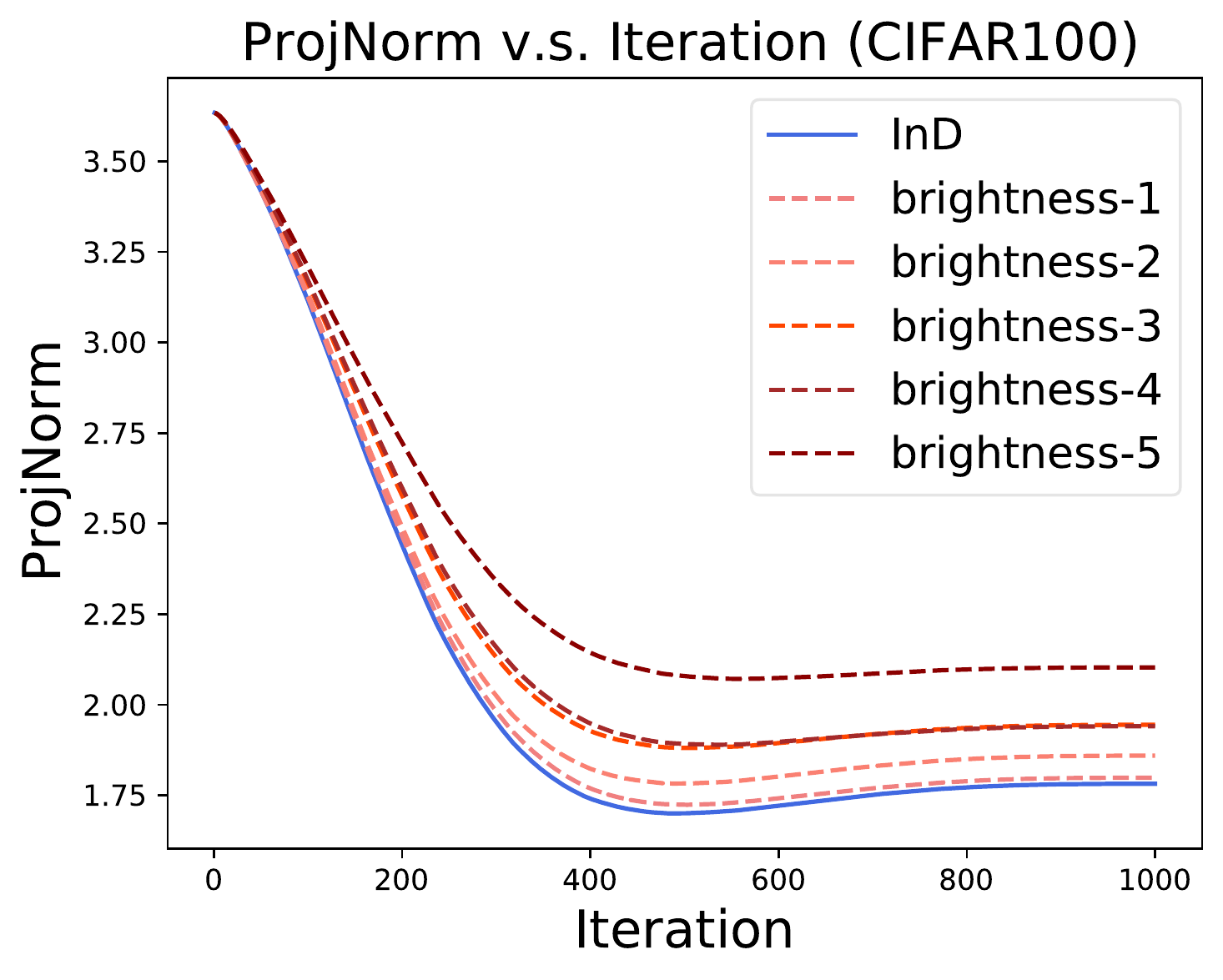}}
    \subfigure{\includegraphics[width=.23\textwidth]{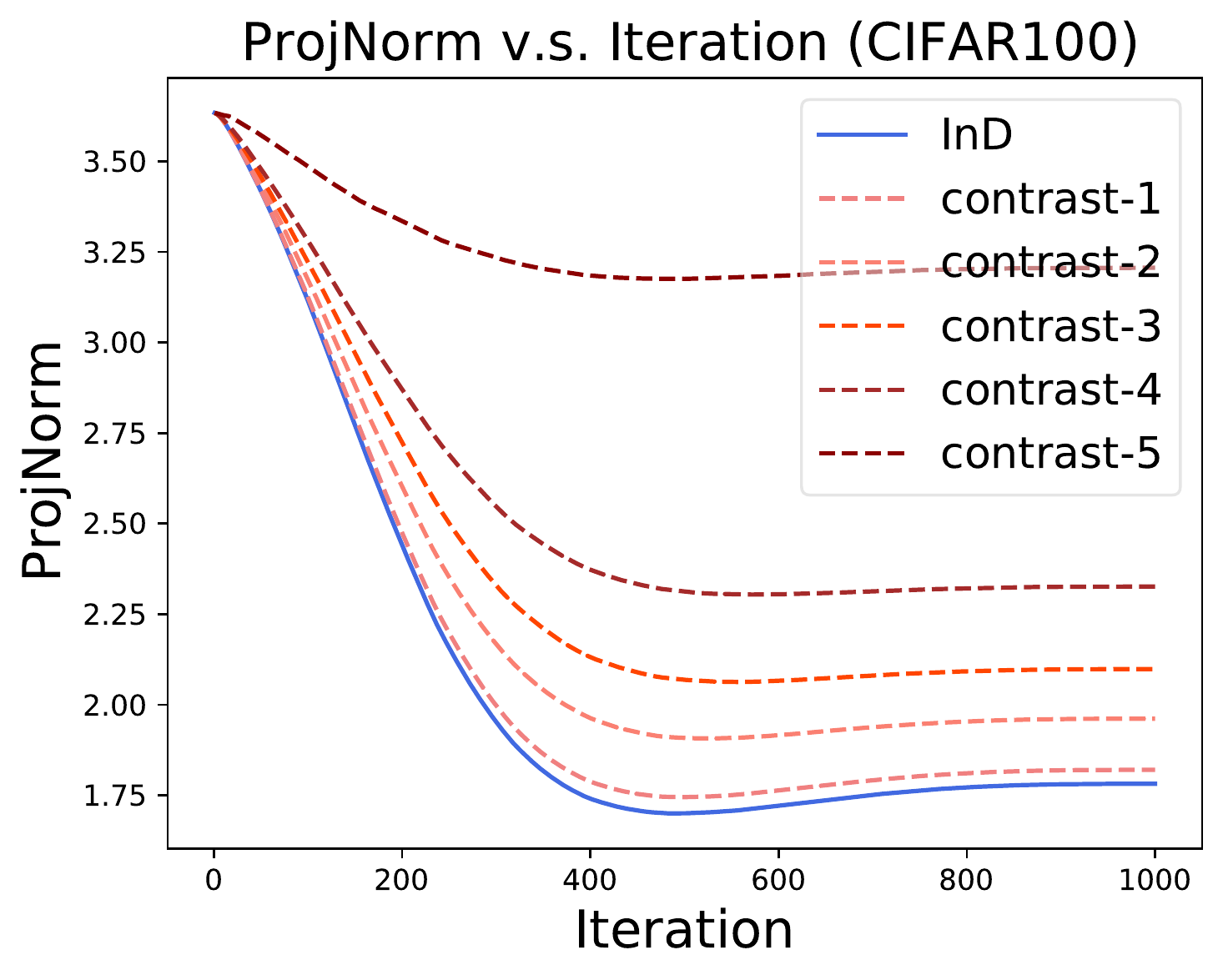}}
    \subfigure{\includegraphics[width=.23\textwidth]{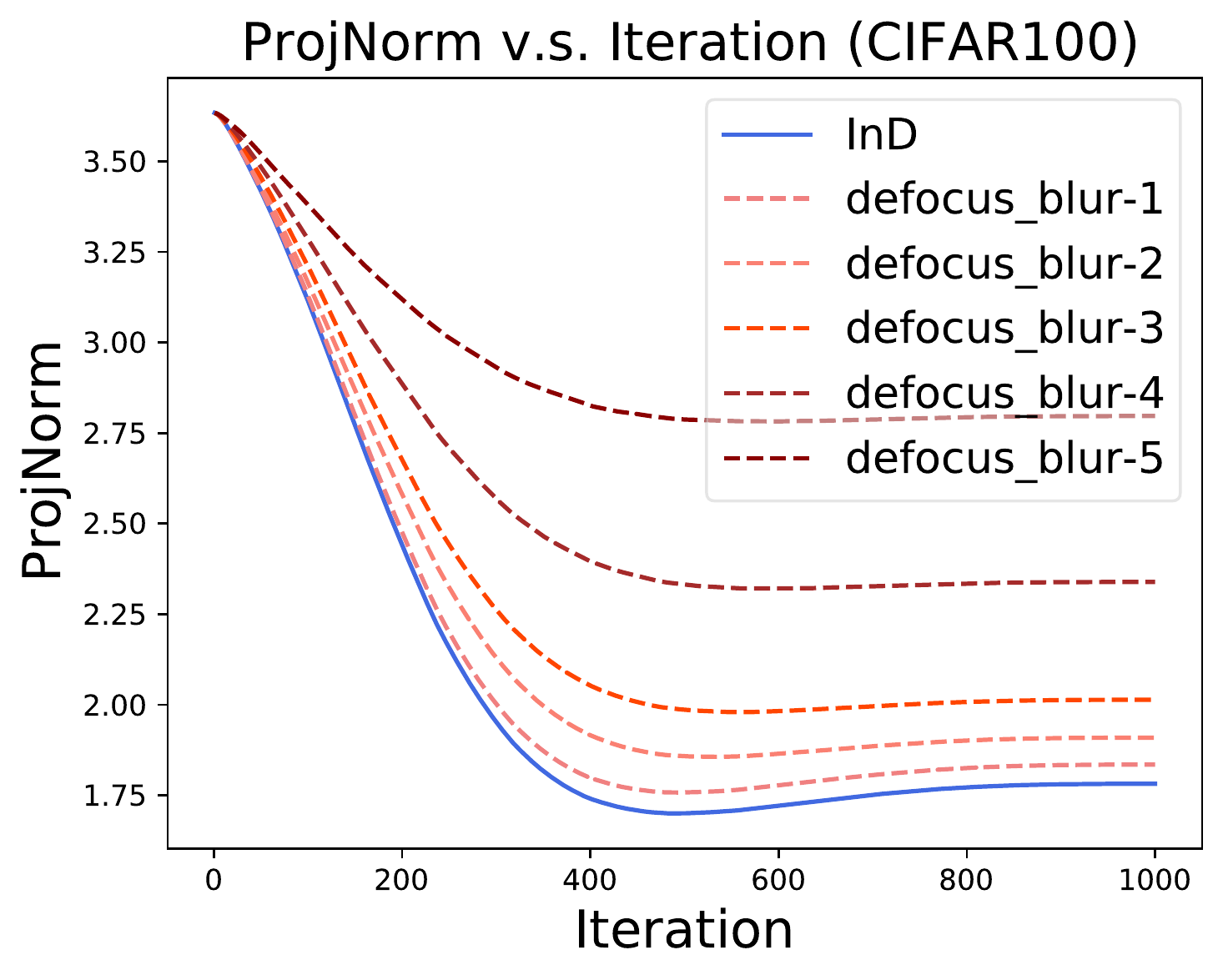}}
    \subfigure{\includegraphics[width=.23\textwidth]{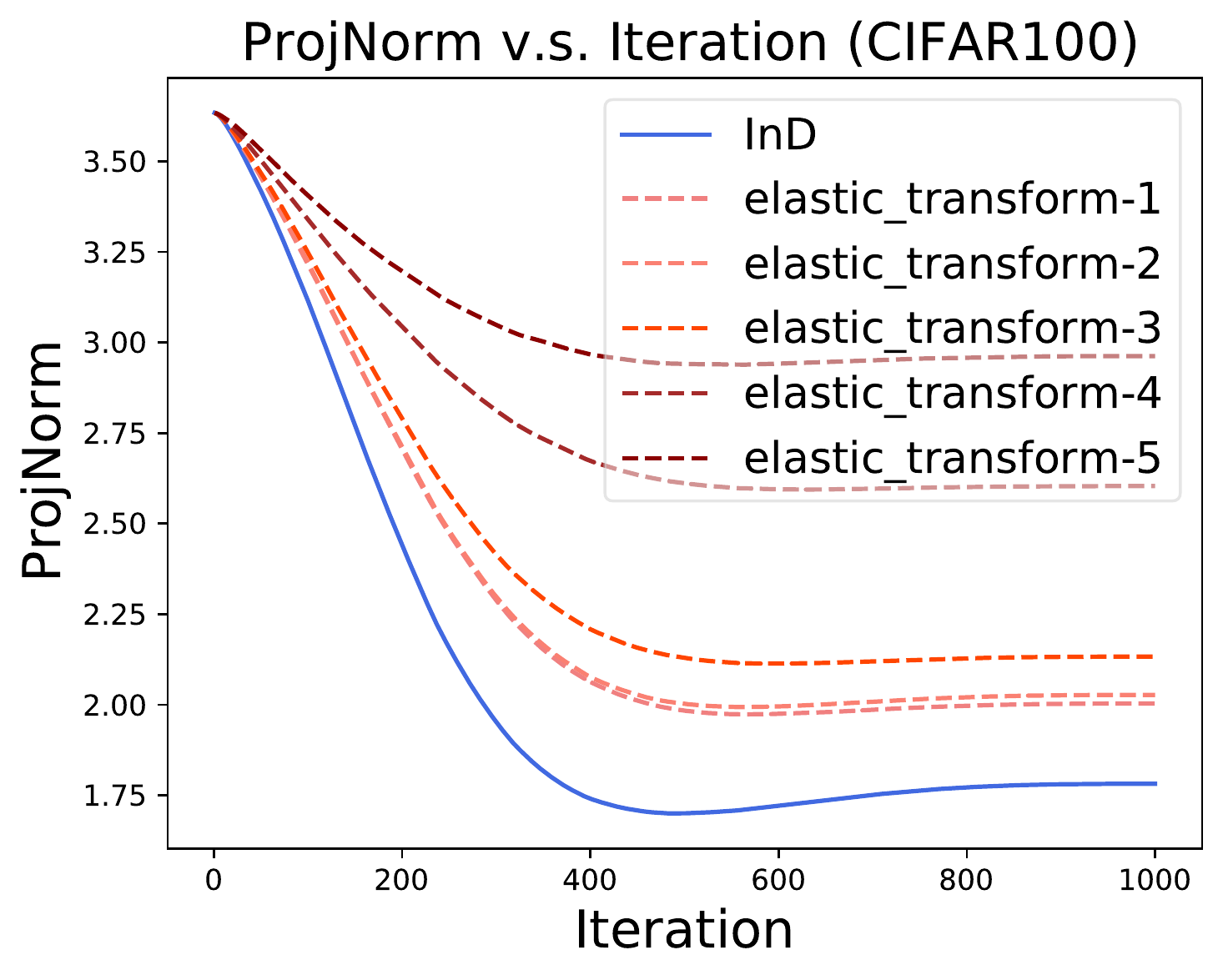}}
    \subfigure{\includegraphics[width=.23\textwidth]{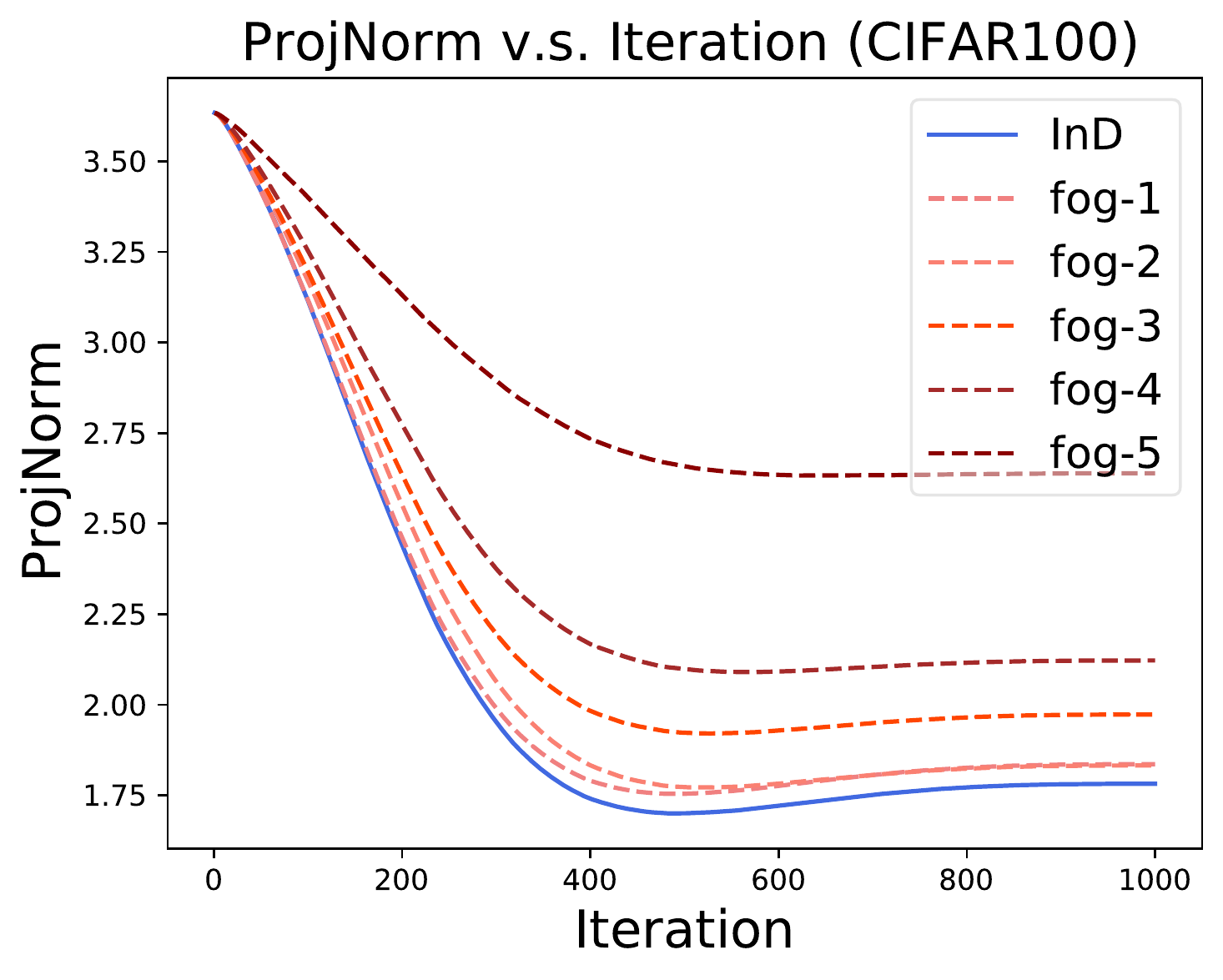}}
    \subfigure{\includegraphics[width=.23\textwidth]{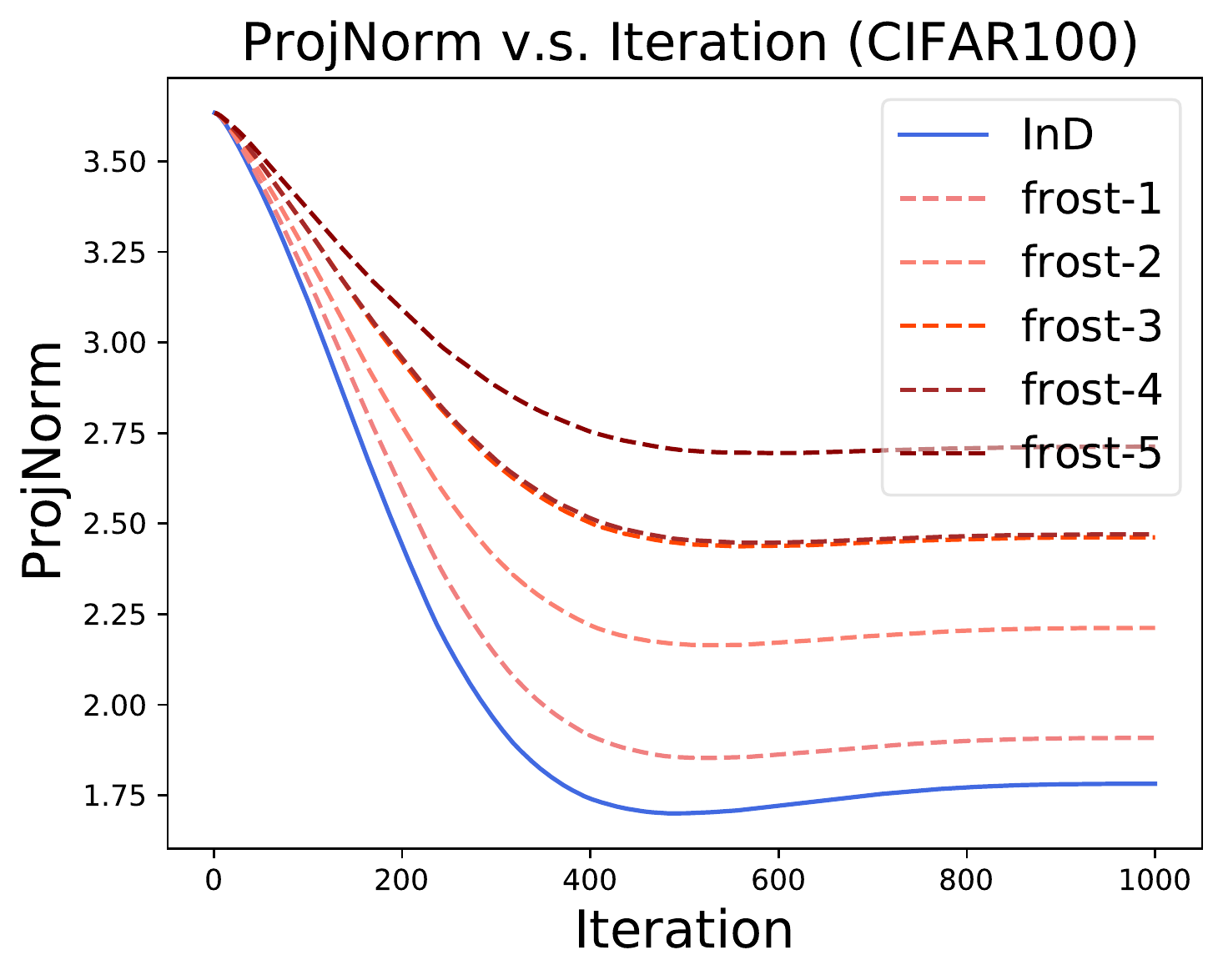}}
    \subfigure{\includegraphics[width=.23\textwidth]{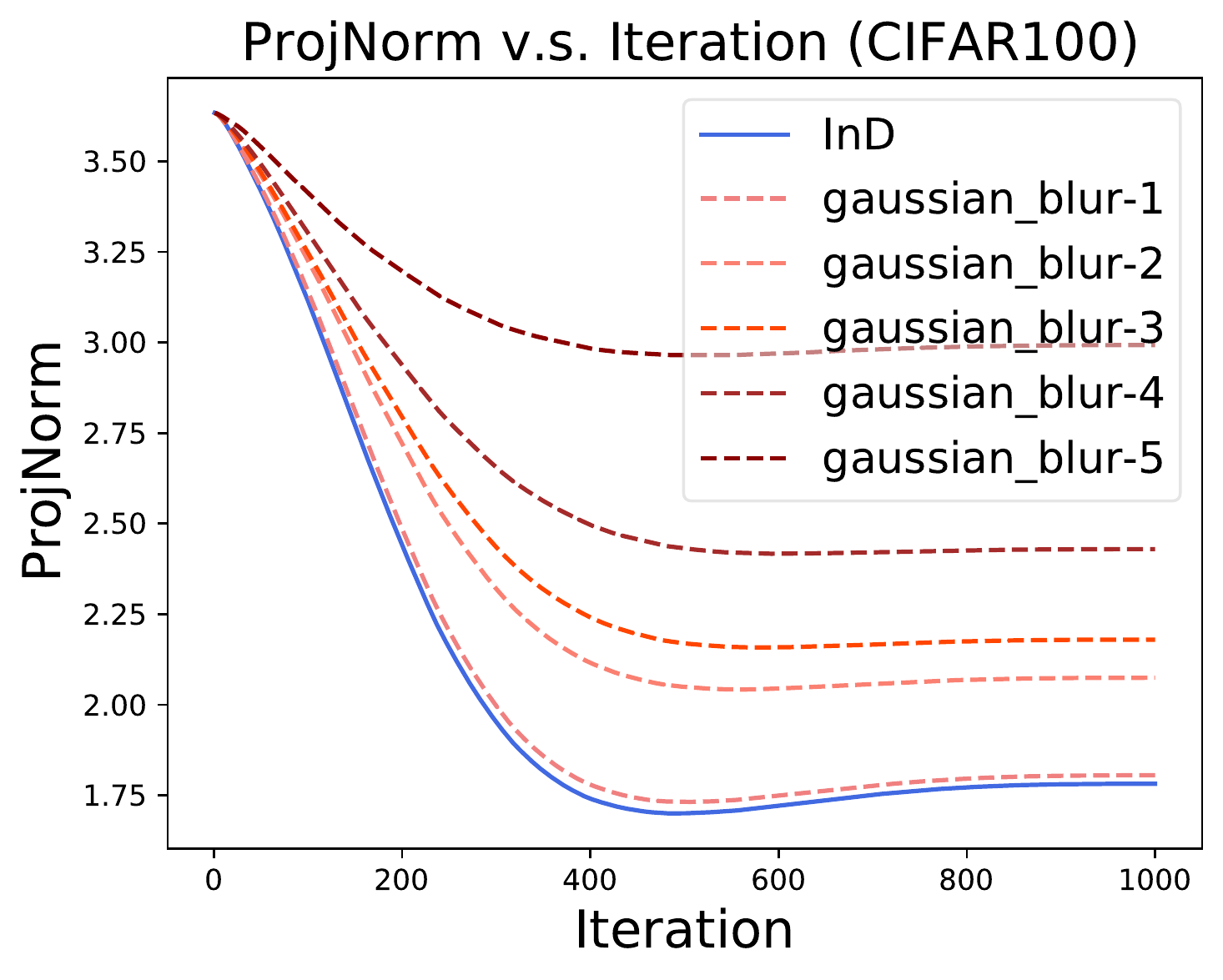}}
    \subfigure{\includegraphics[width=.23\textwidth]{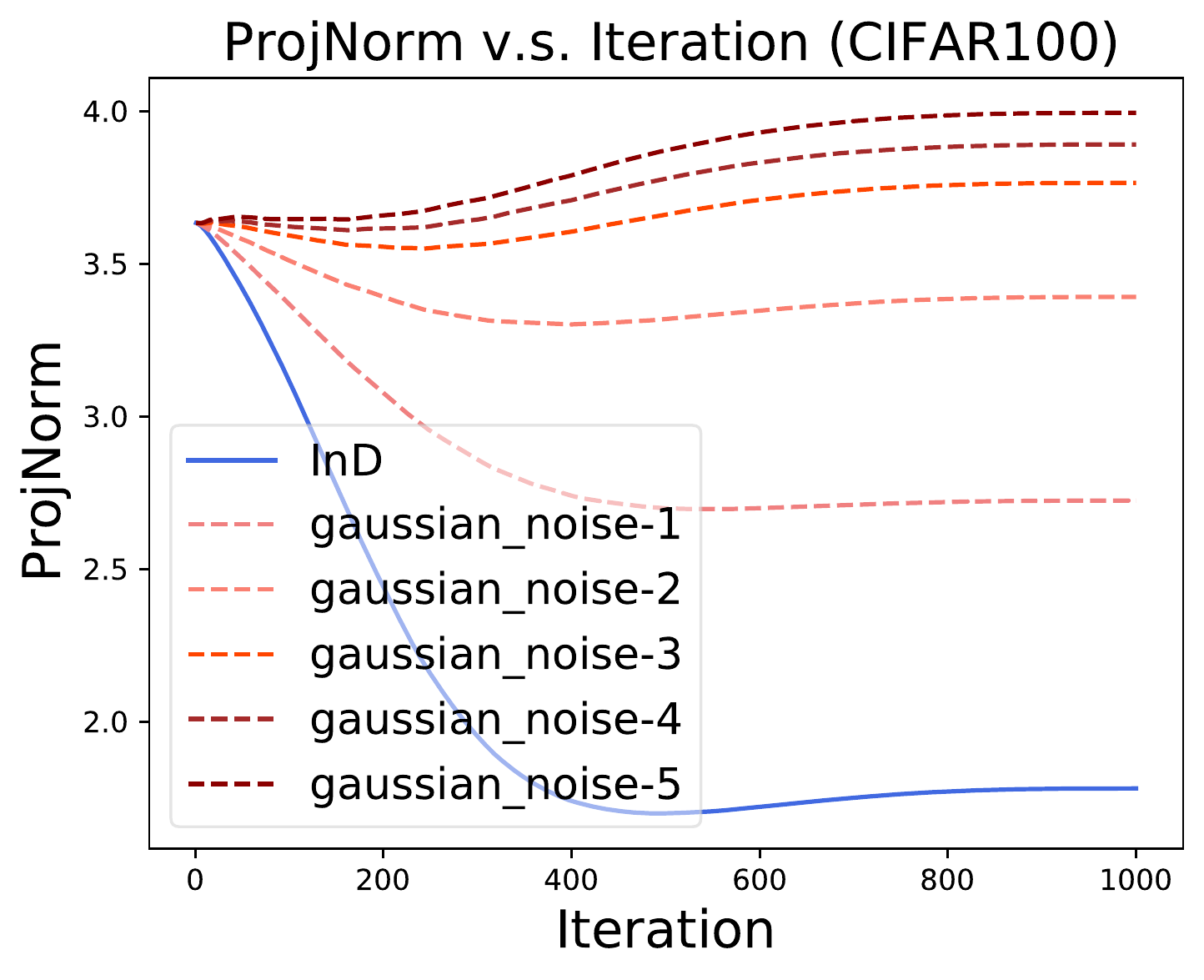}}
    \subfigure{\includegraphics[width=.23\textwidth]{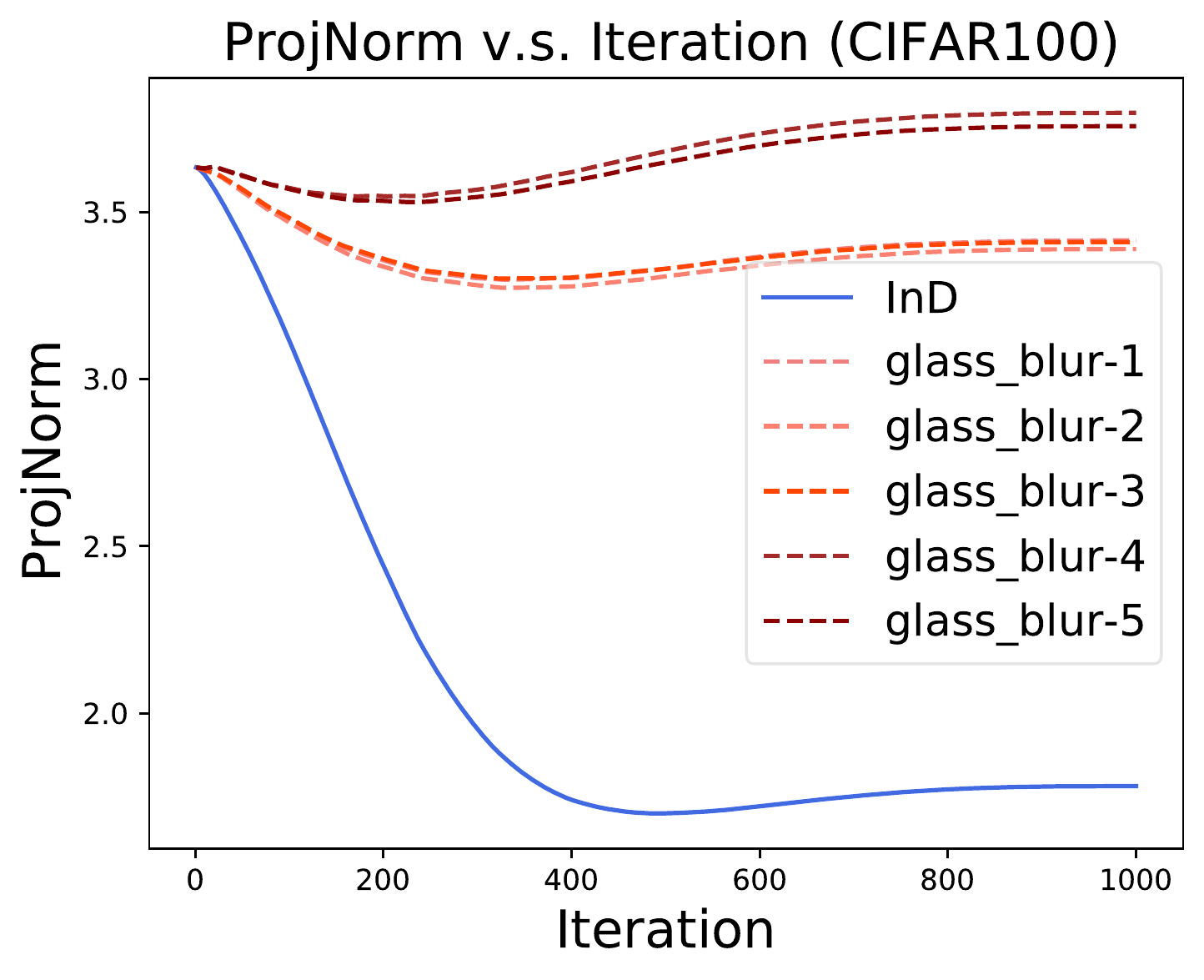}}
    \subfigure{\includegraphics[width=.23\textwidth]{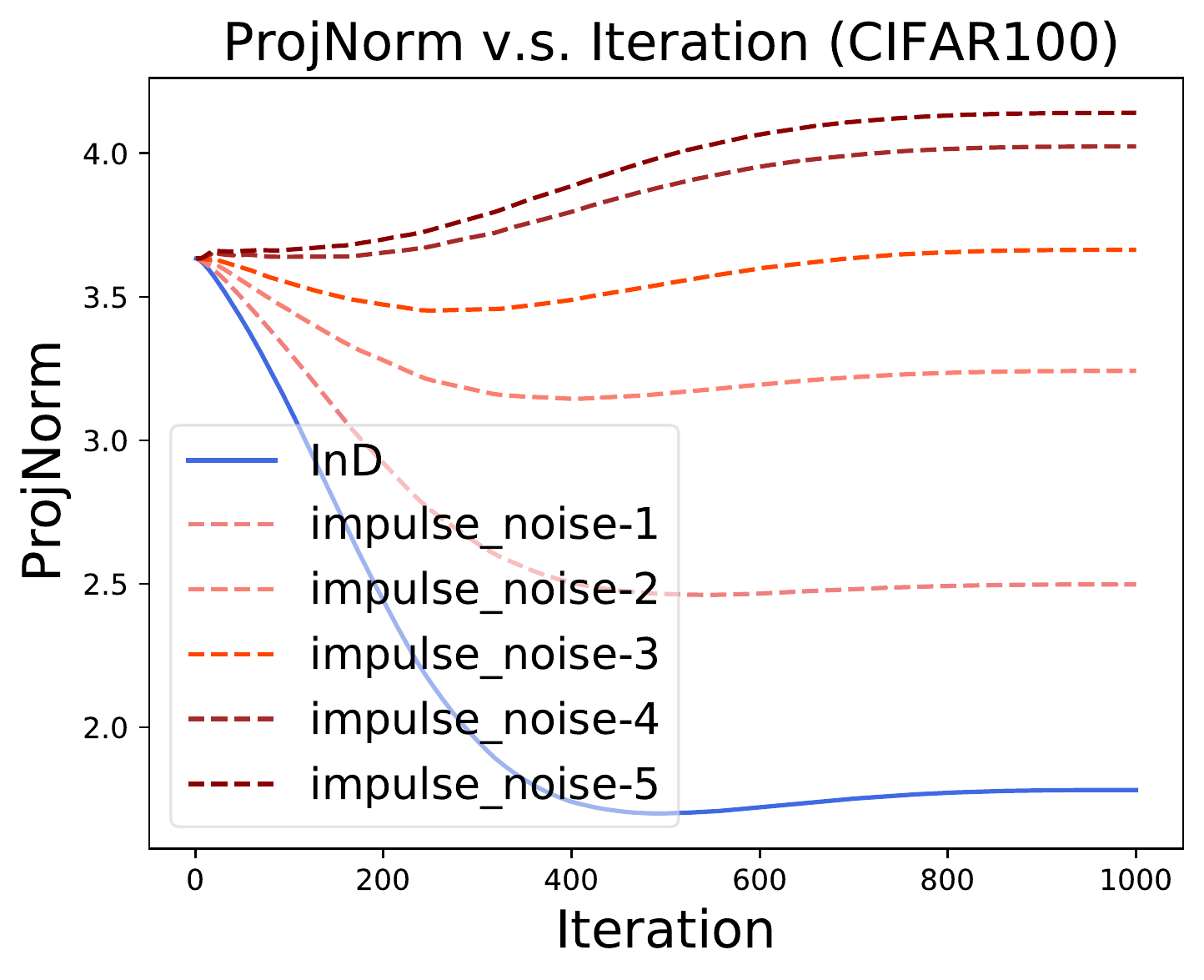}}
    \subfigure{\includegraphics[width=.23\textwidth]{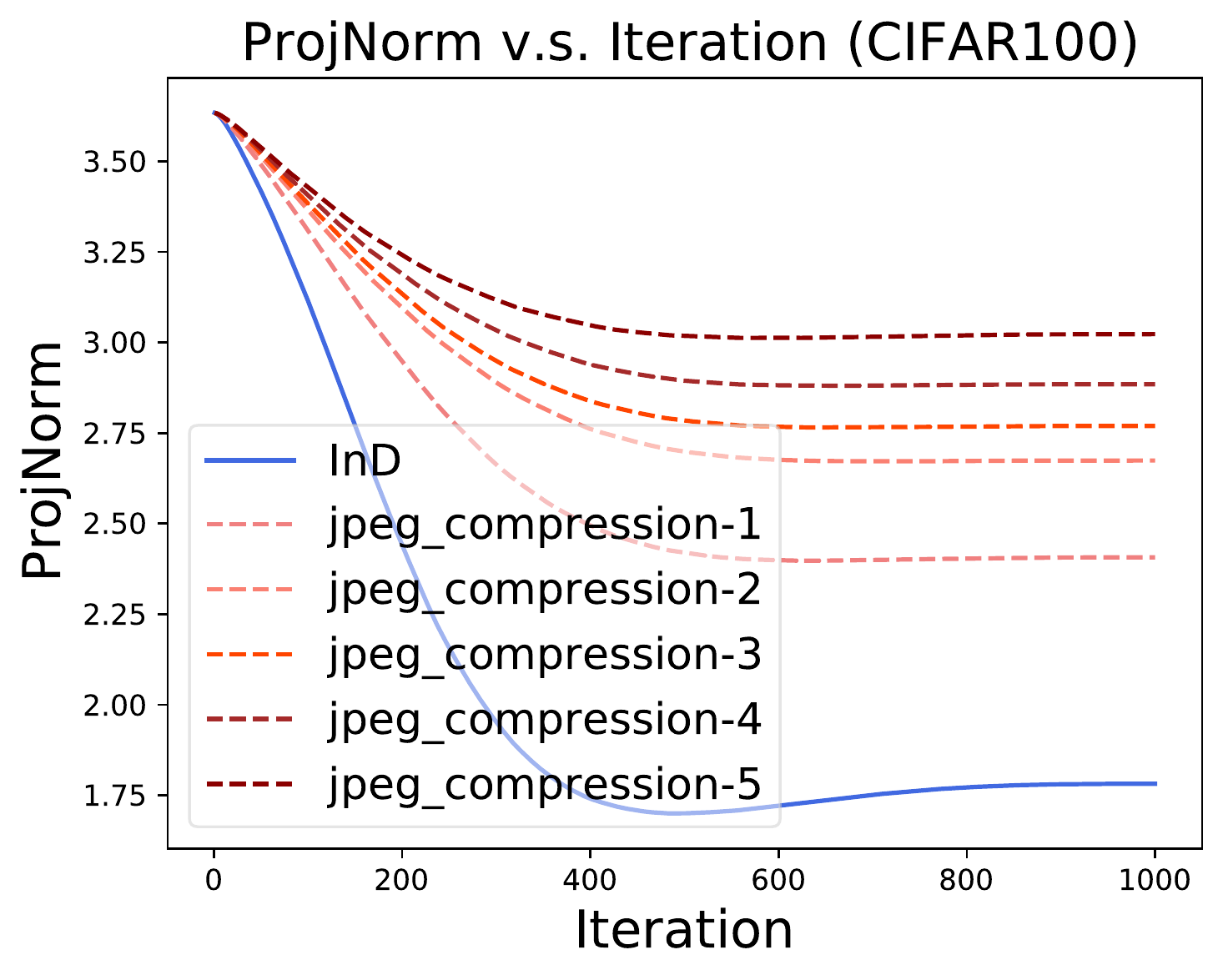}}
    \subfigure{\includegraphics[width=.23\textwidth]{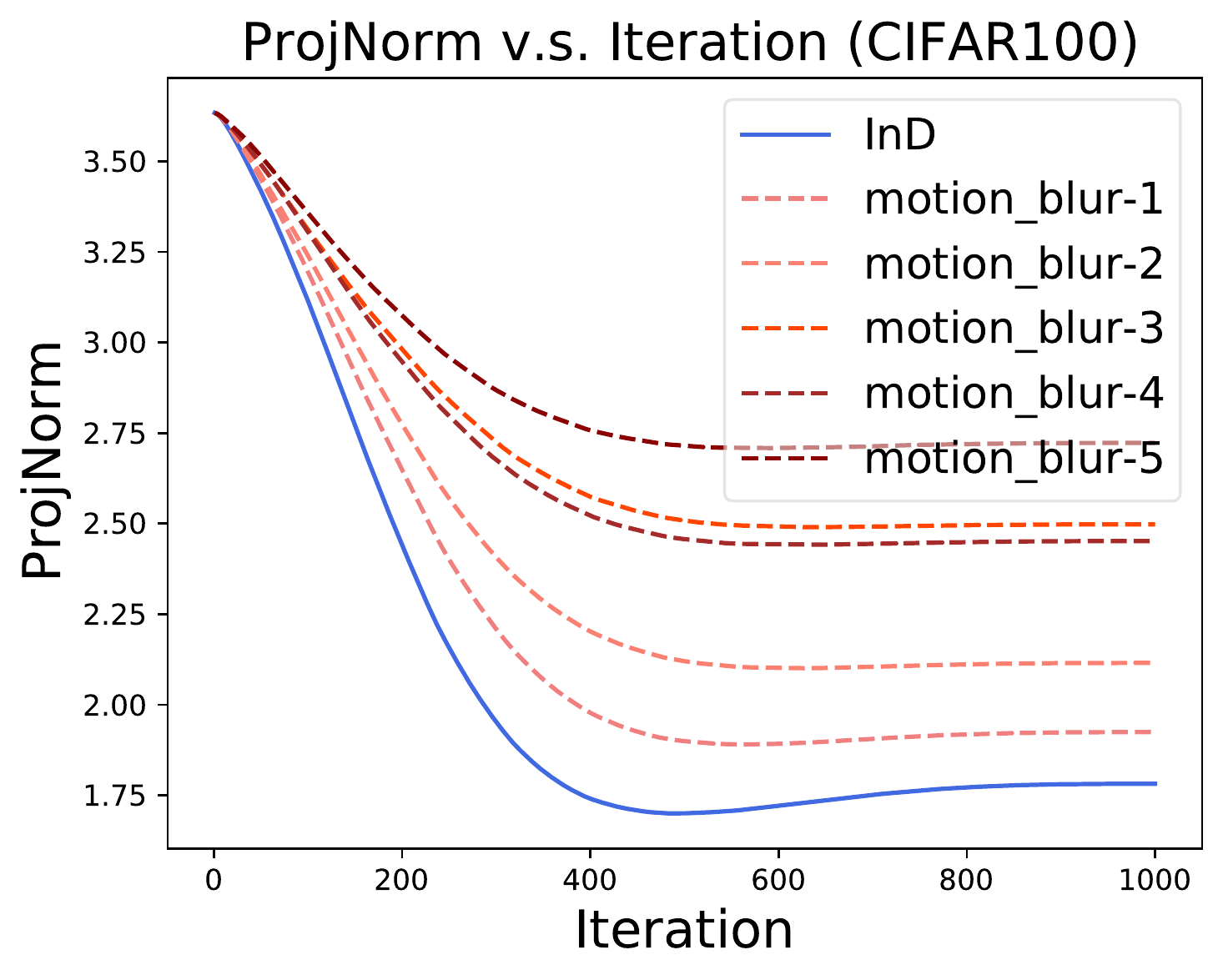}}
    \subfigure{\includegraphics[width=.23\textwidth]{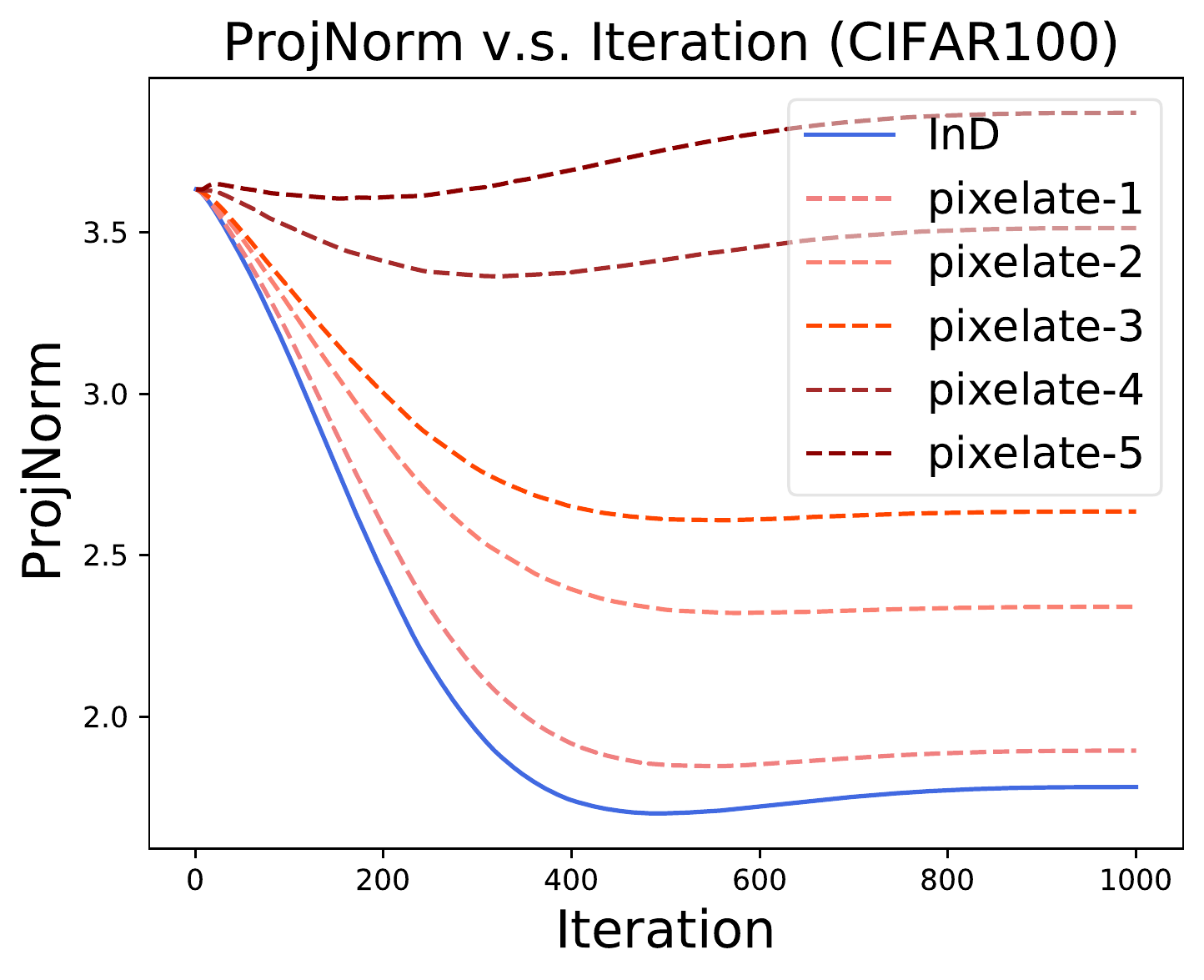}}
    \subfigure{\includegraphics[width=.23\textwidth]{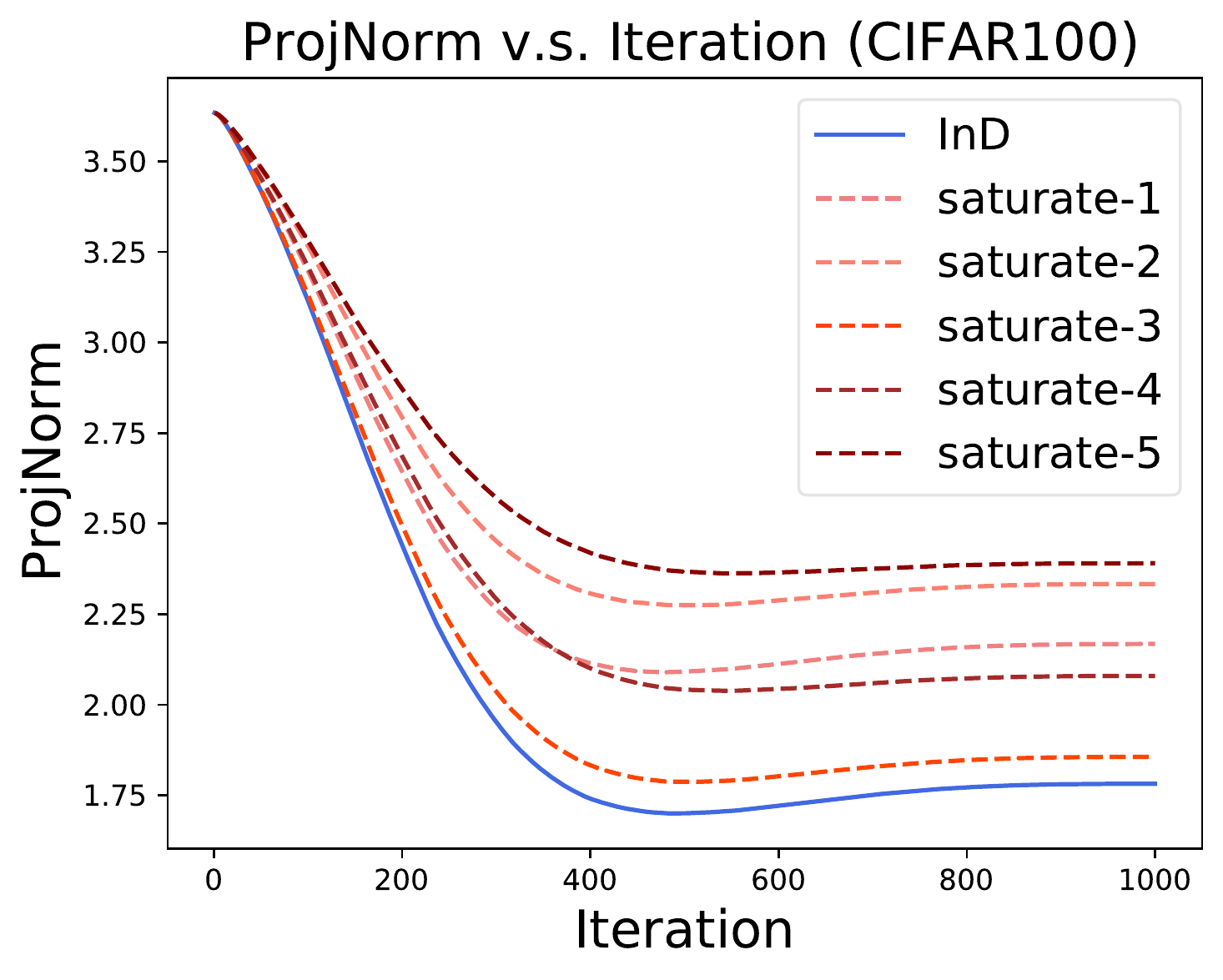}}
    \subfigure{\includegraphics[width=.23\textwidth]{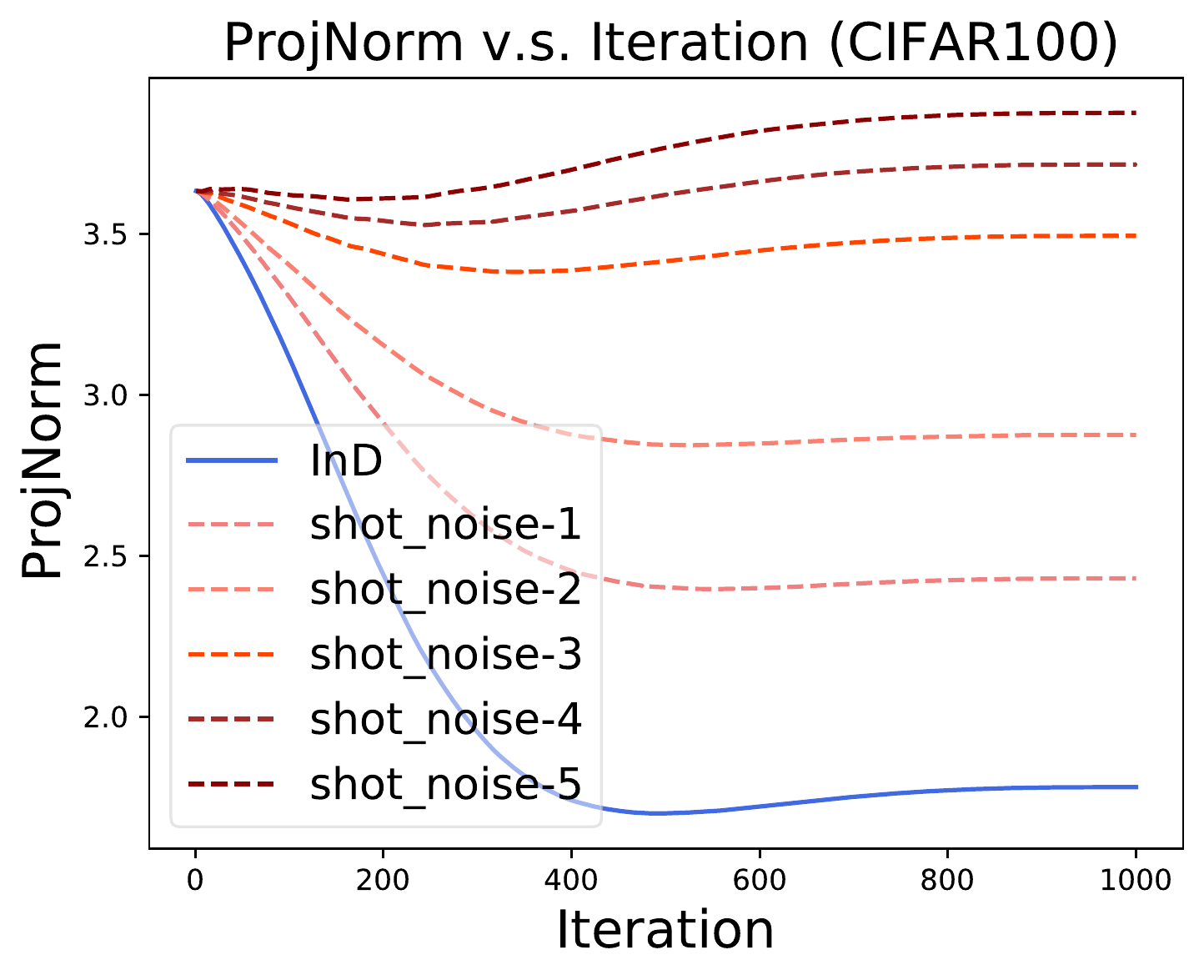}}
    \subfigure{\includegraphics[width=.23\textwidth]{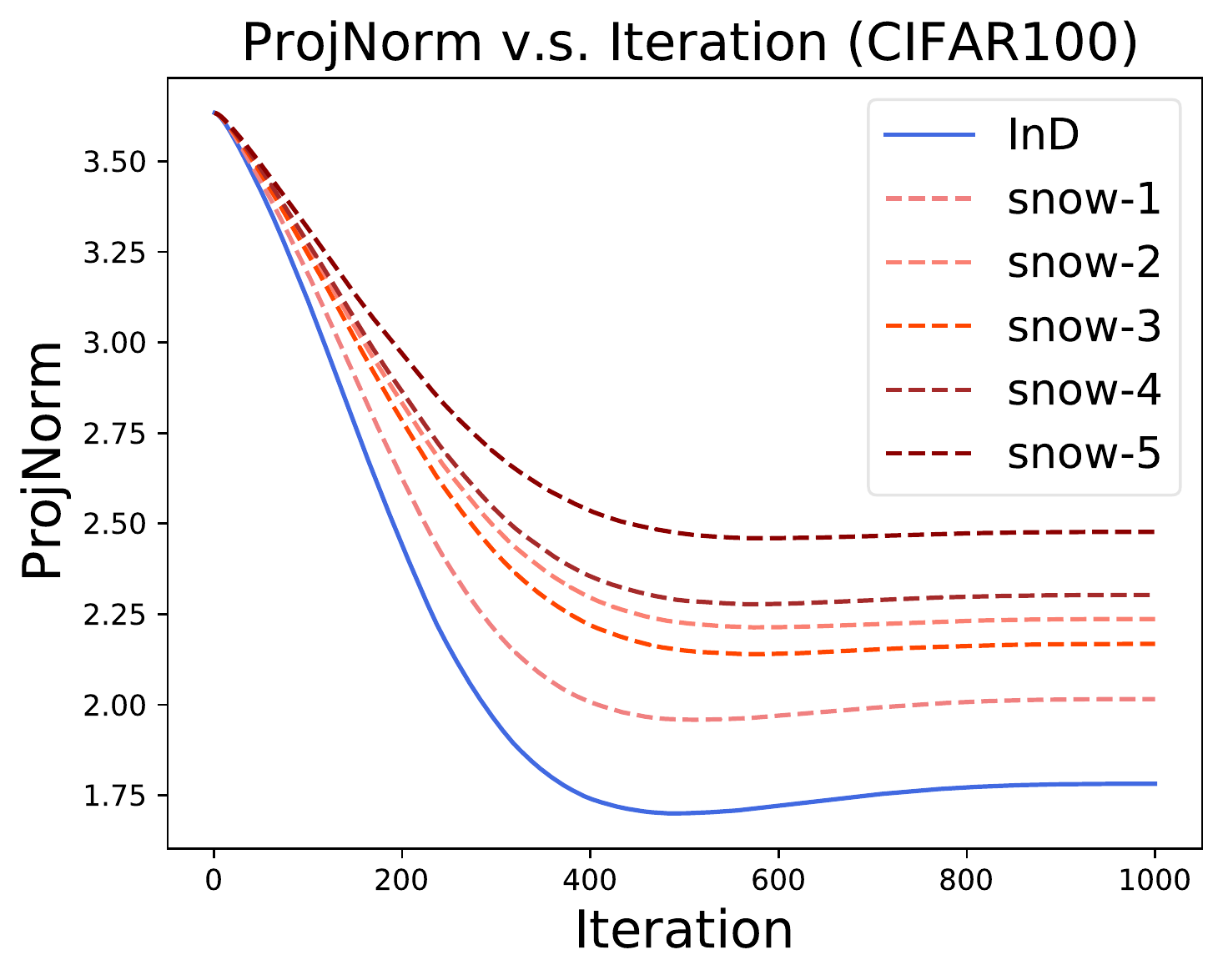}}
    \subfigure{\includegraphics[width=.23\textwidth]{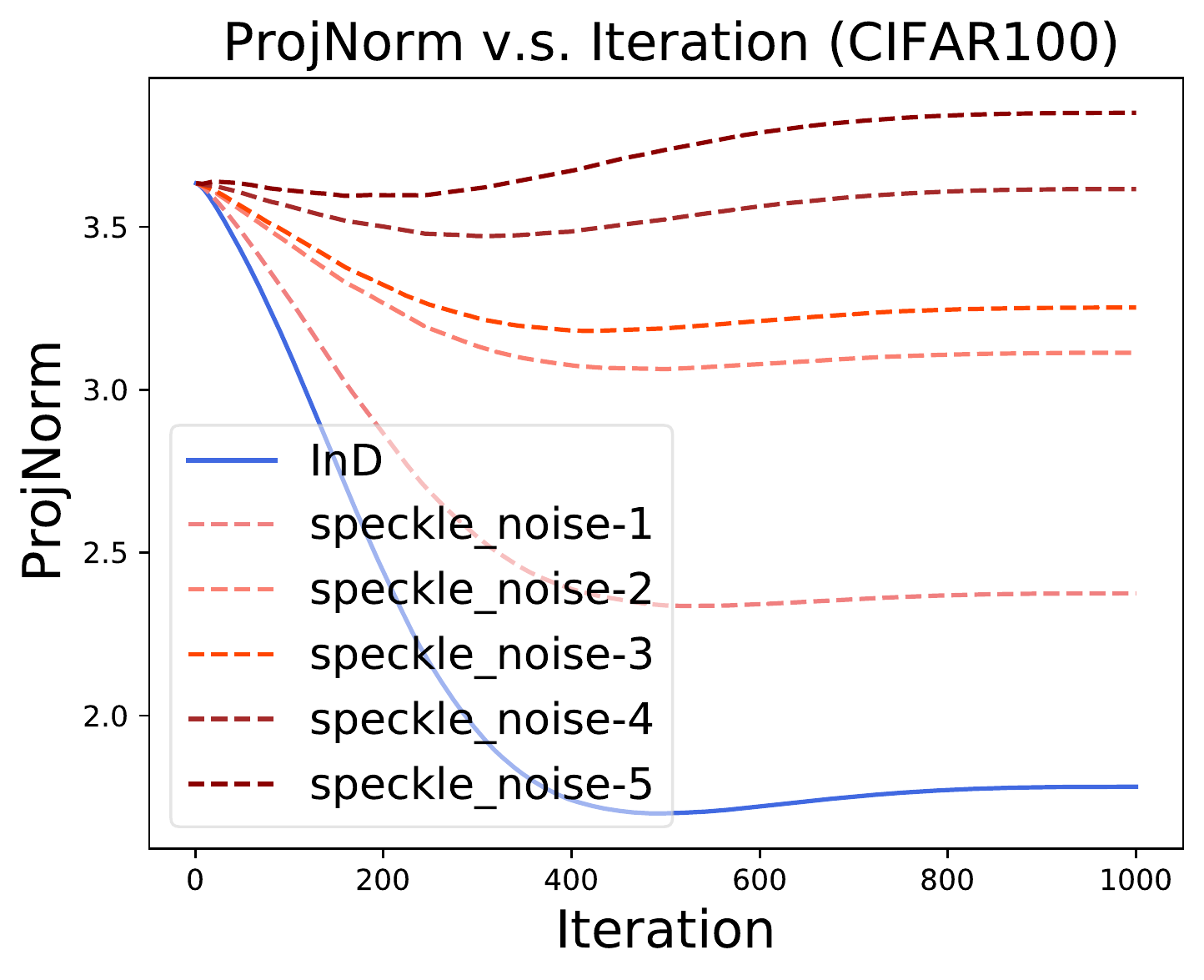}}
    \subfigure{\includegraphics[width=.23\textwidth]{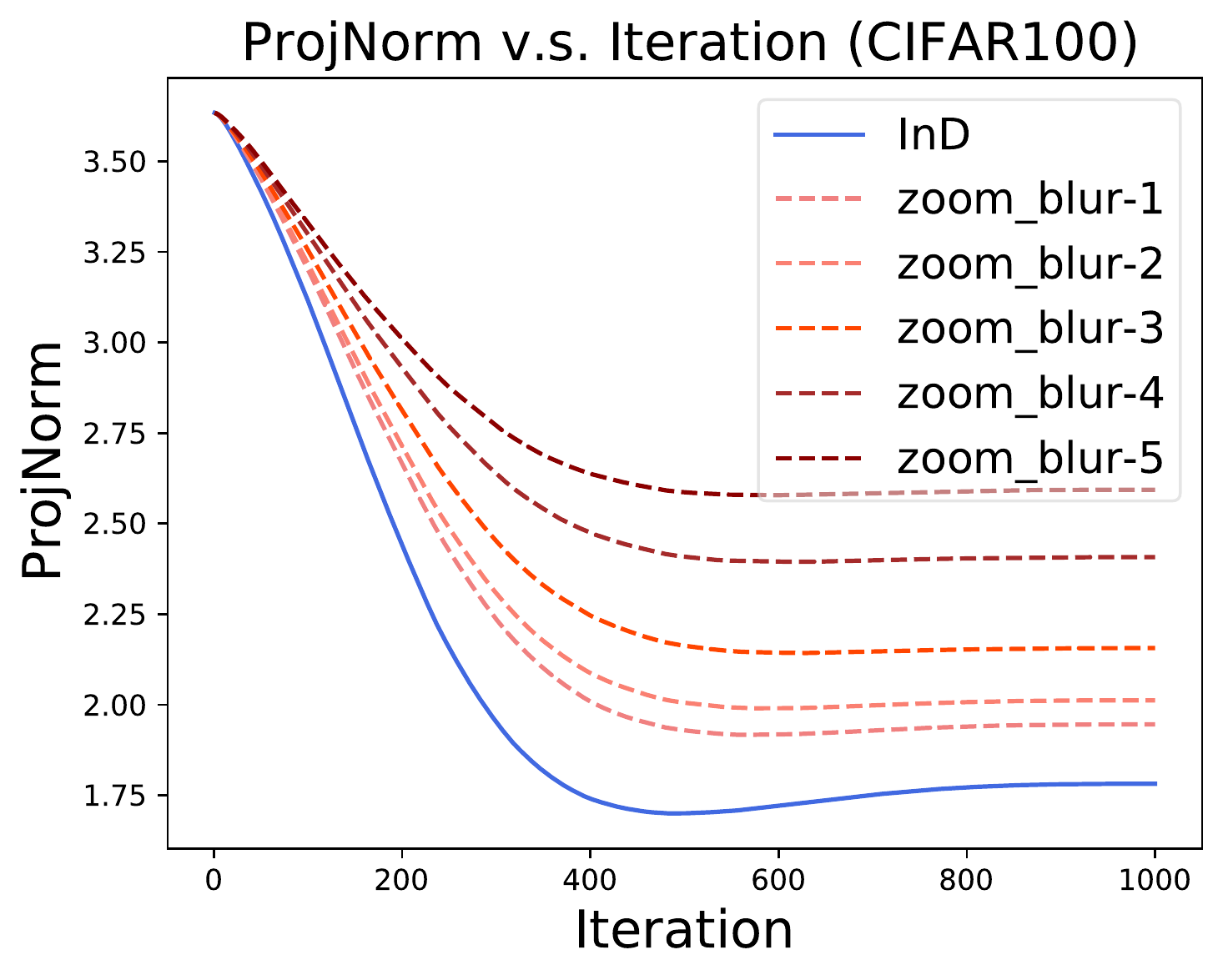}}
    \vspace{-0.1in}
    \caption{\textbf{Evaluation of {\normalfont\pj\,}as training progresses on all corruptions in CIFAR100-C.} We visualize how the \pj\,changes as the number of training iteration increases for ResNet50 on CIFAR100.}
    \label{fig:projnorm-analysis-appendix}
    \vspace{-0.2in}
\end{figure*}

\begin{table*}[ht]
\centering
\caption{\textbf{Hyperparameter sensitivity of {\normalfont\pj}} (w.r.t. sample size $T$). We vary the number of ``pseudo-label projection'' training iterations ($T$). We set the number of test samples $m=10,000$. The performance of \pj~is stable across different number of training iterations.}
\vspace{0.05in}
\label{table:table-sensitivity-appendix-T}
\begin{tabular}{@{}ccccccccccc@{}}
\toprule
\multirow{2}{*}{Dataset} 
& \multicolumn{2}{c}{$T$=1,000}
& \multicolumn{2}{c}{$T$=500}
& \multicolumn{2}{c}{$T$=200}
\\ \cmidrule(l){2-7}
& $R^2$ & $\rho$ & $R^2$ & $\rho$ & $R^2$ & $\rho$ 
\\ \midrule
\multirow{1}{*}{CIFAR10} 
 & 0.962  & \textbf{0.992}  & \textbf{0.985} & 0.987 & 0.983 &  0.986 \\
\midrule
\multirow{1}{*}{CIFAR100} 
 & 0.978 & \textbf{0.989}  & \textbf{0.980} & 0.986 &  0.959 & 0.968 \\
\bottomrule
\end{tabular}
\vspace{-0.15in}
\end{table*}

\begin{table*}[ht]
\centering
\caption{\textbf{Hyperparameter sensitivity of {\normalfont\pj}} (w.r.t. sample size $m$). We vary the number of test samples ($m$) of \pj. We set learning rate $\eta$=0.001 and number of training iterations $T$=$m/10$.}
\vspace{0.05in}
\label{table:table-sensitivity-appendix-m}
\begin{tabular}{@{}ccccccccccccc@{}}
\toprule
\multirow{2}{*}{Dataset} 
& \multicolumn{2}{c}{$m=10,000$}
& \multicolumn{2}{c}{$m=5,000$}
& \multicolumn{2}{c}{$m=2,000$}
& \multicolumn{2}{c}{$m=1,000$} 
& \multicolumn{2}{c}{$m=500$} 
& \multicolumn{2}{c}{$m=100$} 
\\ \cmidrule(l){2-13}
& $R^2$ & $\rho$ & $R^2$ & $\rho$ & $R^2$ & $\rho$ & $R^2$ & $\rho$ & $R^2$ & $\rho$ & $R^2$ & $\rho$ 
\\ \midrule
\multirow{1}{*}{CIFAR10} 
 & 0.962  & \textbf{0.992}  & 0.973 & 0.989 & 0.977 &  0.985 & \textbf{0.980} & 0.975 & 0.946 & 0.983 & 0.784 & 0.896\\
\midrule
\multirow{1}{*}{CIFAR100} 
 & \textbf{0.978} & \textbf{0.989}  & 0.972 & 0.983 & 0.942 & 0.966 & 0.942 & 0.981 & 0.903 & 0.972 & 0.466  & 0.789\\
\bottomrule
\end{tabular}
\end{table*}

\begin{table*}[ht]
\centering
\caption{\textbf{Hyperparameter sensitivity of {\normalfont\pj}} (w.r.t. learning rate $\eta$). We vary the learning rate ($\eta$) and set $T$=1,000 and $m$=10,000. The performance of \pj~is stable across different learning rates.}
\vspace{0.05in}
\label{table:table-sensitivity-lr-appendix}
\begin{tabular}{@{}ccccccccccc@{}}
\toprule
\multirow{2}{*}{Dataset} 
& \multicolumn{2}{c}{$\eta$=\texttt{1e-3}}
& \multicolumn{2}{c}{$\eta$=\texttt{5e-4}} 
& \multicolumn{2}{c}{$\eta$=\texttt{1e-4}} 
\\ \cmidrule(l){2-7}
& $R^2$ & $\rho$  & $R^2$ & $\rho$ & $R^2$ & $\rho$ 
\\ \midrule
\multirow{1}{*}{CIFAR10} 
 & 0.962  & \textbf{0.992}   & 0.984 & 0.991 & \textbf{0.986} & 0.988 \\
\midrule
\multirow{1}{*}{CIFAR100} 
 & 0.978 & \textbf{0.989}  & \textbf{0.982} & \textbf{0.989} & 0.969 & 0.984 \\
\bottomrule
\end{tabular}
\end{table*}

\begin{figure*}[ht]
    \vspace{-0.1in}
    \centering
    \subfigure[ResNet18.]{\includegraphics[width=.33\textwidth]{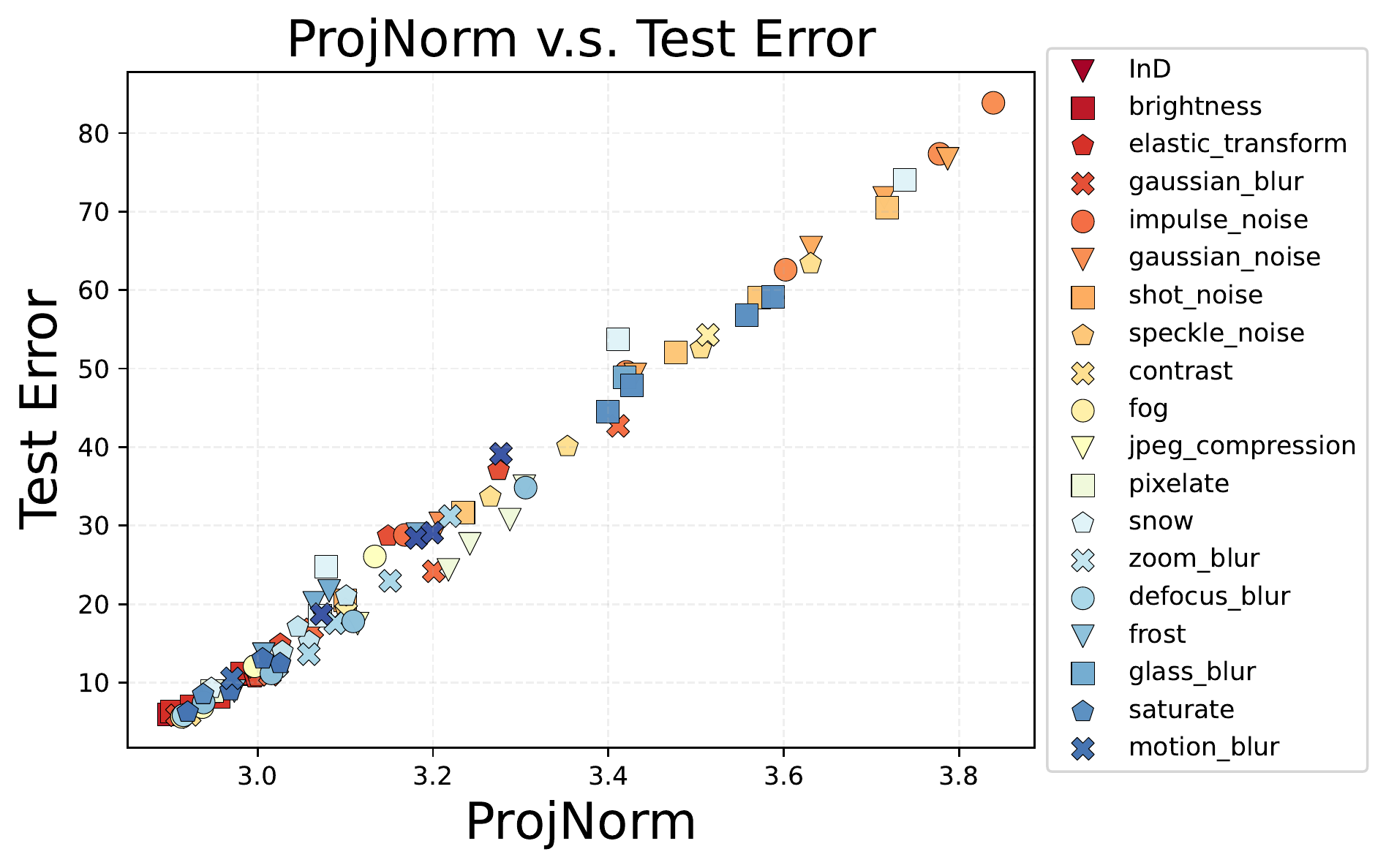}}
    \subfigure[ResNet50.]{\includegraphics[width=.33\textwidth]{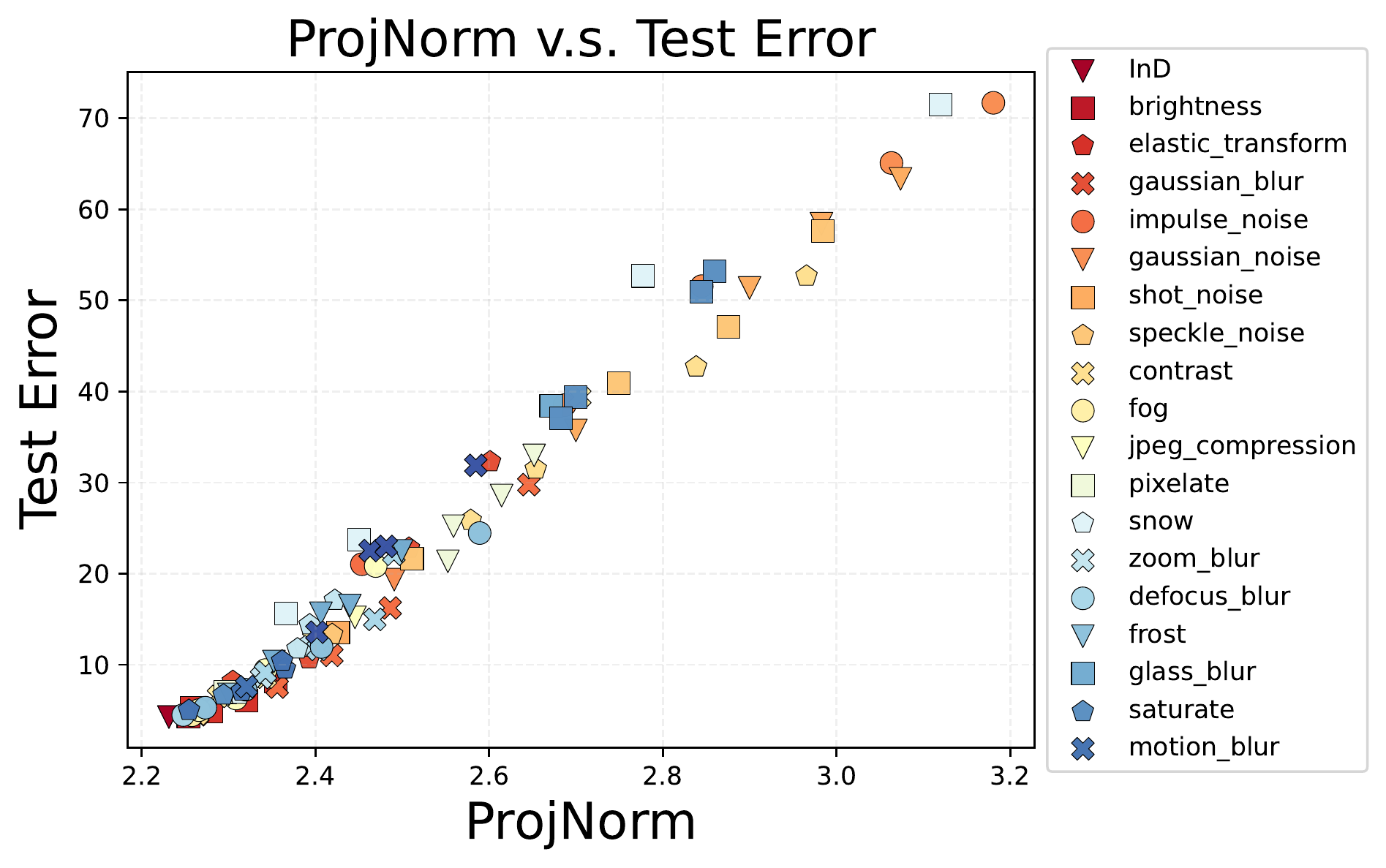}}
    \subfigure[VGG11.]{\includegraphics[width=.33\textwidth]{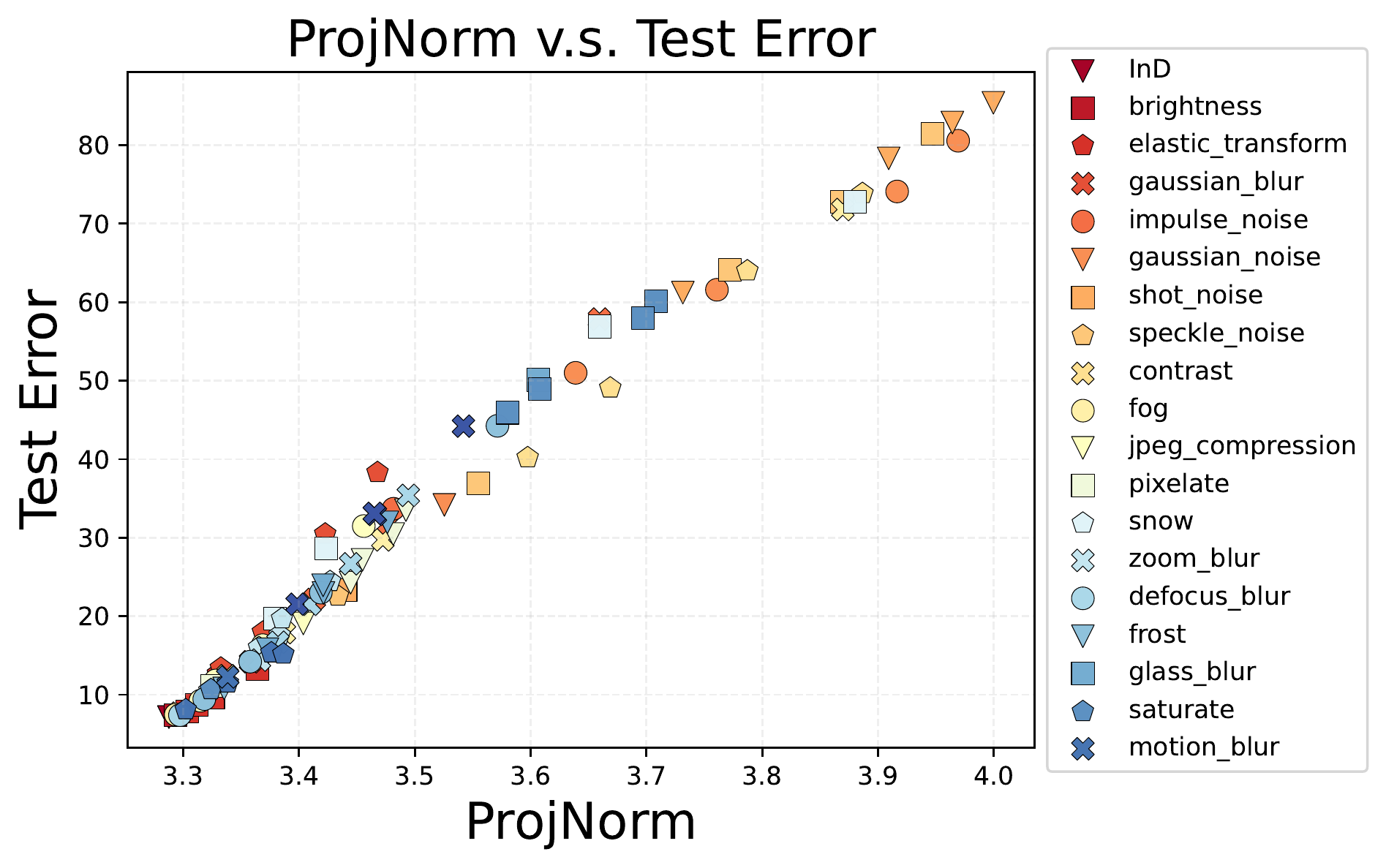}}
    \vspace{-0.1in}
    \caption{\textbf{Generalization prediction versus test error on CIFAR10 with ResNet18/ResNet50/VGG11.} We plot the actual test error and the prediction of \pj~(when $\bthetah_{\text{ref}}=\bthetah$) on each OOD dataset.
    Each point represents one InD/OOD dataset, and points with the same color and marker shape are the same corruption but with different severity levels.
    }
    \label{fig:compare-appendix-cifar10-thetahat}
    \vspace{-0.1in}
\end{figure*}

\clearpage
\paragraph{Comparing $\|\bthetat - \bthetah\|_{2}$ and $\|\bthetat - \bthetah_{\text{\normalfont ref}}\|_{2}$.} We study the performance of \pj~when we use $\bthetah$ as $\bthetah_{\text{ref}}$ on CIFAR10. We do not train a new reference model on the training dataset and use the fine-tuned model $\bthetah$ to measure the difference $\|\bthetat - \bthetah\|_{2}$. As shown in Figure~\ref{fig:compare-appendix-cifar10-thetahat}, applying $\bthetah_{\text{ref}}=\bthetah$ does not degrade the performance of \pj.

\begin{table*}[ht]
\vspace{-0.1in}
\centering
\caption{\textbf{Comparing {\normalfont\pj{}} with different reference models on CIFAR10}. We study the performance of \pj{} when using $\bthetah_{\text{ref}}=\bthetah$ and compare it with the default version of \pj. \pj{} with $\bthetah_{\text{ref}}=\bthetah$ achieves similar or even better performance compared to the default version. We set $T=500$ and $\eta=0.001$.}
\vspace{0.1in}
\label{table:table-sensitivity-thetahat-appendix}
\begin{tabular}{@{}ccccccccccc@{}}
\toprule
\multirow{2}{*}{Dataset} 
& \multicolumn{2}{c}{ResNet18}
& \multicolumn{2}{c}{ResNet50} 
& \multicolumn{2}{c}{VGG11} 
\\ \cmidrule(l){2-7}
& $R^2$ & $\rho$  & $R^2$ & $\rho$ & $R^2$ & $\rho$ 
\\ \midrule
\multirow{1}{*}{Default} 
 & 0.980   &  0.989   & 0.972  & 0.986  &  \textbf{0.982} & 0.993  \\
\midrule
\multirow{1}{*}{$\bthetah_{\text{ref}}=\bthetah$} 
 & \textbf{0.989}  &  \textbf{0.991}  & \textbf{0.980} & \textbf{0.987}  &  \textbf{0.982} & \textbf{0.994} \\
\bottomrule
\end{tabular}
\end{table*}

\begin{figure}[ht]
    \centering
    \subfigure[ResNet18.]{\includegraphics[width=.3\textwidth]{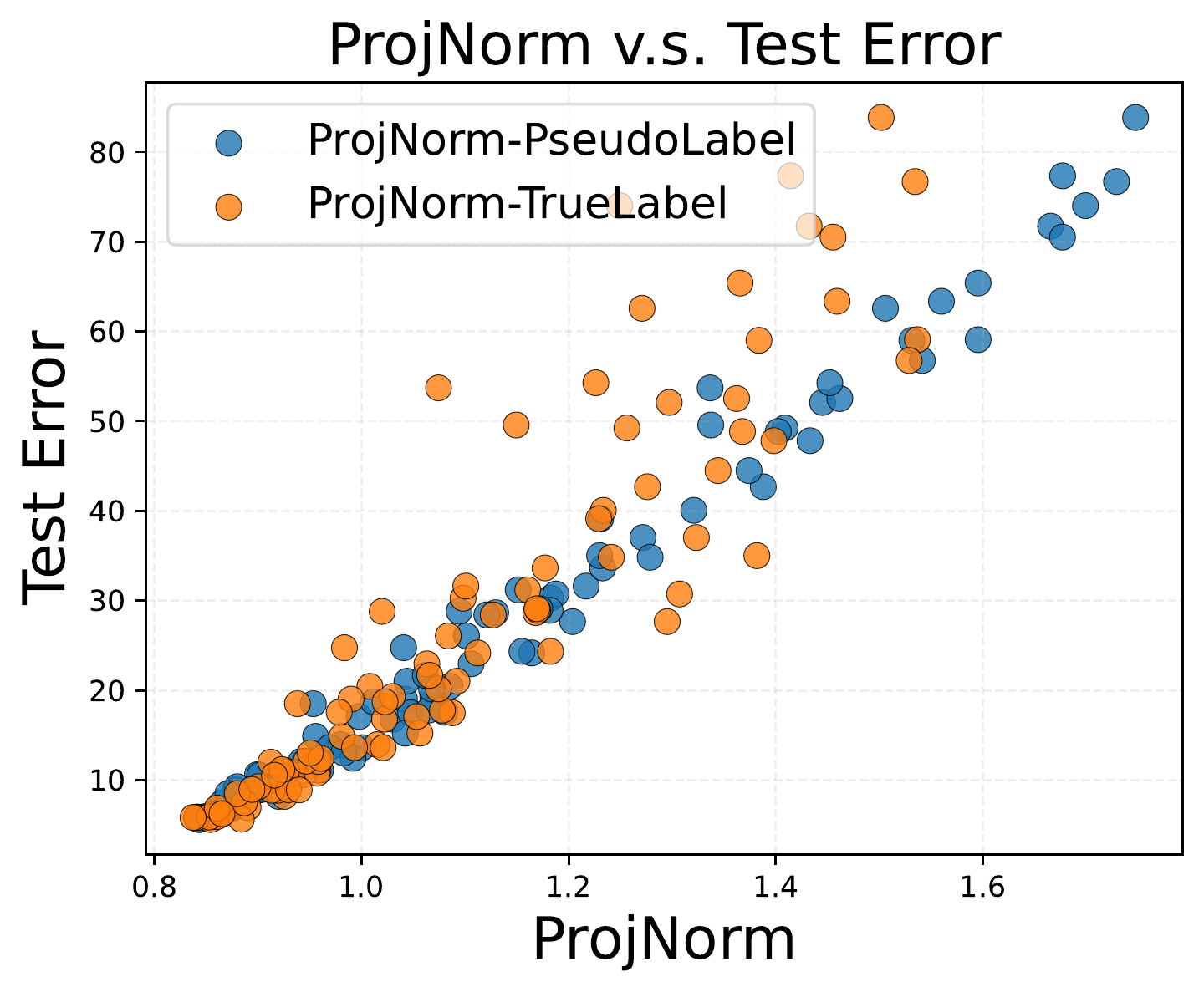}}
    \subfigure[ResNet50.]{\includegraphics[width=.3\textwidth]{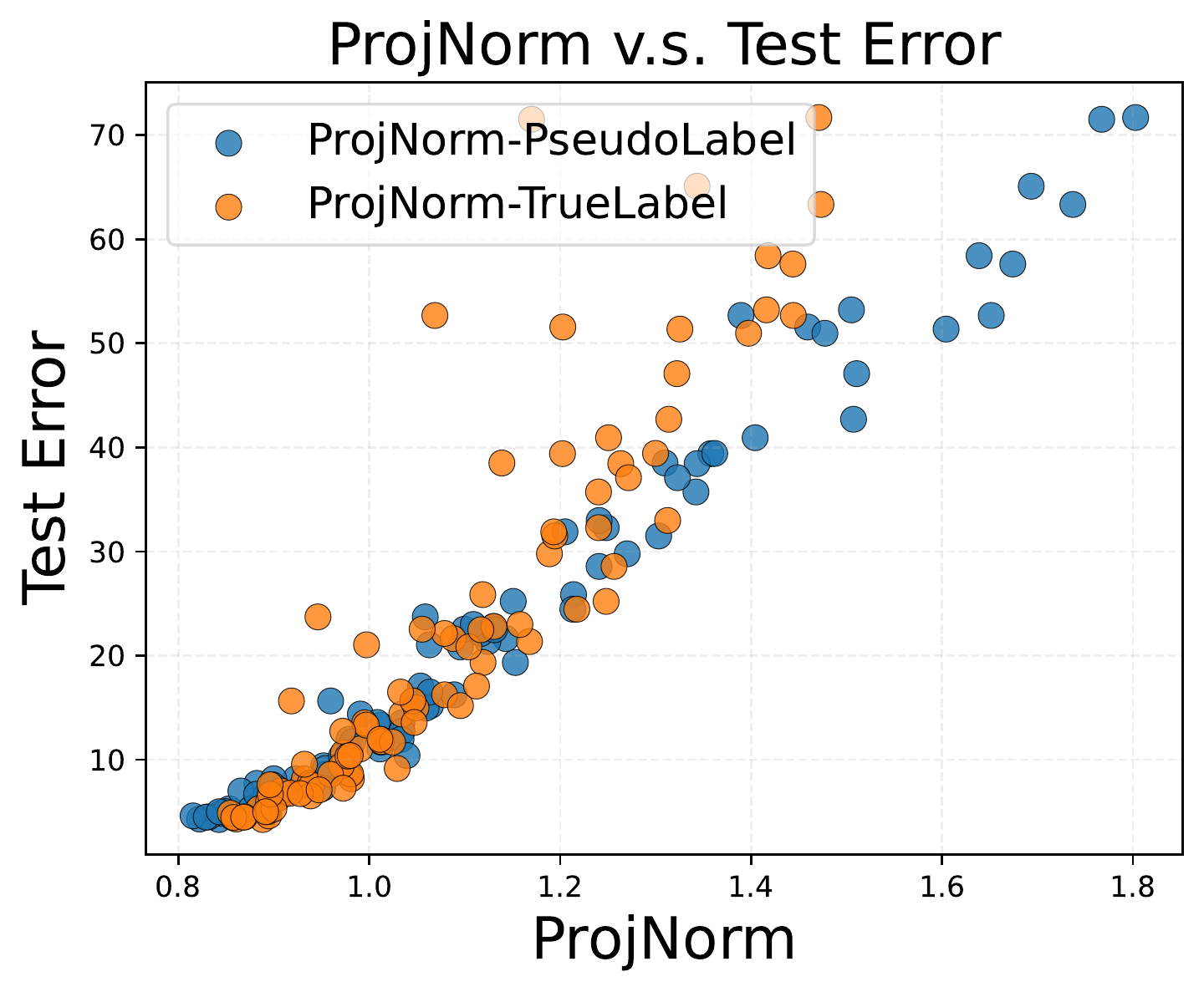}}
    \subfigure[VGG11.]{\includegraphics[width=.3\textwidth]{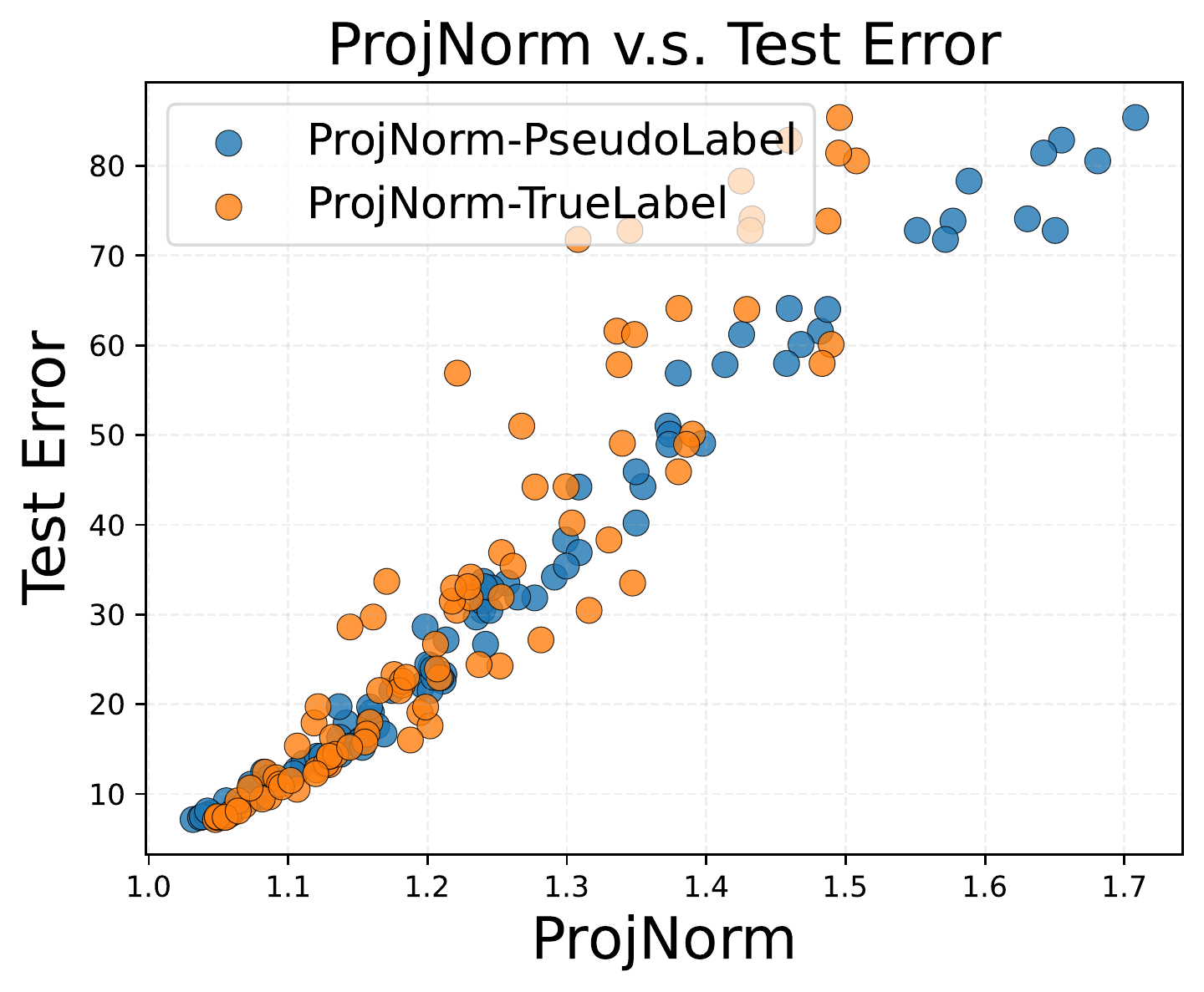}}
    \vspace{-0.15in}
    \caption{\textbf{Comparing the performance of {\normalfont \pj}~when using pseudo-labels and ground truth labels on CIFAR10.} We plot the actual test error and the prediction of \pj~on each OOD dataset. Blue circles are results when using pseudo-labels, and orange circles are results when using ground truth labels.
    }
    \label{fig:compare-appendix-cifar10-truelabel}
    \vspace{-0.15in}
\end{figure}

\paragraph{Role of pseudo-labels.} We investigate the role of pseudo-labels in \pj. Specifically, we modify \textbf{Step 2} of \pj{} by training $\bthetat$ using the ground truth labels of the OOD data. We compare the performance of \pj~when using pseudo-labels and ground truth labels. As shown in Figure~\ref{fig:compare-appendix-cifar10-truelabel}, we find that \pj~with pseudo-label performs much better than \pj~with ground truth label, which suggests that pseudo-labeling is an essential component of \pj.

\begin{table*}[ht]
\vspace{-0.1in}
\centering
\caption{\textbf{Comparing {\normalfont\pj{}} with pseudo-labels and ground truth labels on CIFAR10}. We study the performance of \pj{} when using ground truth labels of OOD data (in \textbf{Step 2}) and compare it with the default version of \pj. \pj{} with ground truth labels achieves worse performance compared to the default version. We set $T=500$ and $\eta=0.001$.}
\vspace{0.1in}
\label{table:table-sensitivity-truelabel-appendix}
\begin{tabular}{@{}ccccccccccc@{}}
\toprule
\multirow{2}{*}{Dataset} 
& \multicolumn{2}{c}{ResNet18}
& \multicolumn{2}{c}{ResNet50} 
& \multicolumn{2}{c}{VGG11} 
\\ \cmidrule(l){2-7}
& $R^2$ & $\rho$  & $R^2$ & $\rho$ & $R^2$ & $\rho$ 
\\ \midrule
\multirow{1}{*}{Default} 
 & \textbf{0.980}   &  \textbf{0.989}   & \textbf{0.972}  & \textbf{0.986}  &  \textbf{0.982} & \textbf{0.993}  \\
\midrule
\multirow{1}{*}{Ground truth labels} 
 & 0.833  &  0.952  & 0.813 &  0.946  & 0.870   & 0.961 \\
\bottomrule
\end{tabular}
\vspace{-0.2in}
\end{table*}

\begin{figure*}[ht]
    \vspace{-0.1in}
    \centering
    \subfigure[ConfScore.]{\includegraphics[width=.3\textwidth]{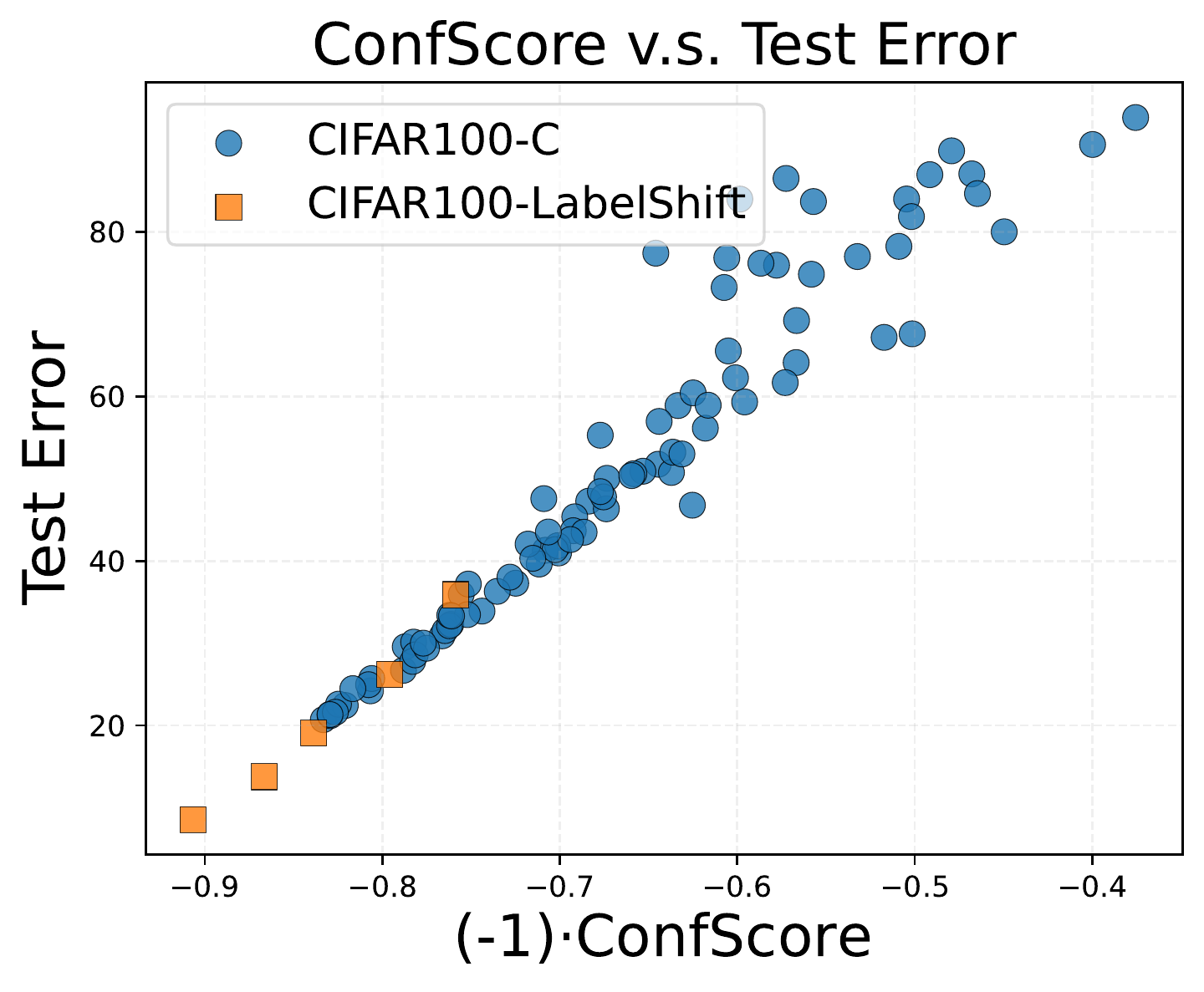}}
    \subfigure[Entropy.]{\includegraphics[width=.3\textwidth]{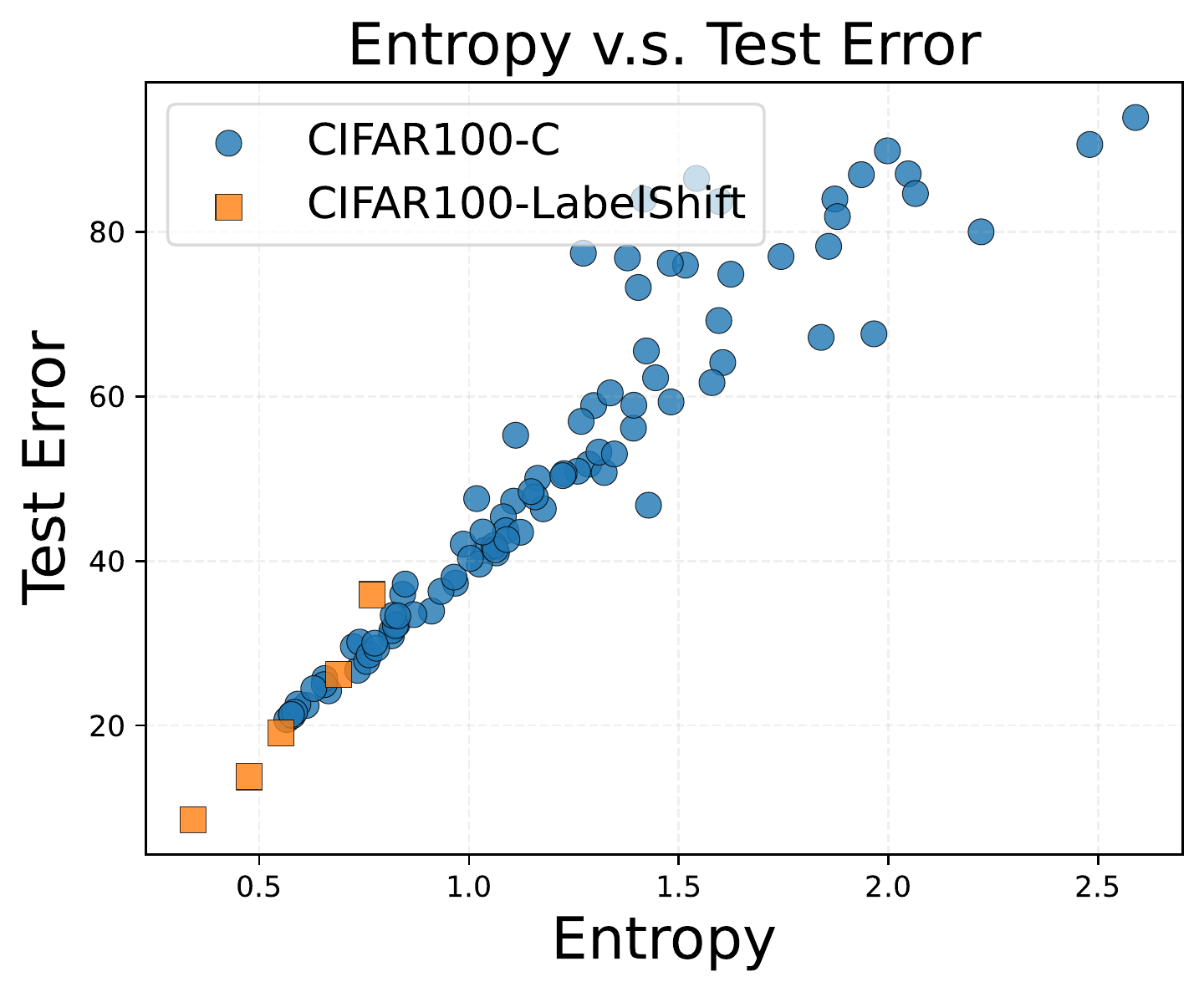}}
    \subfigure[AgreeScore.]{\includegraphics[width=.3\textwidth]{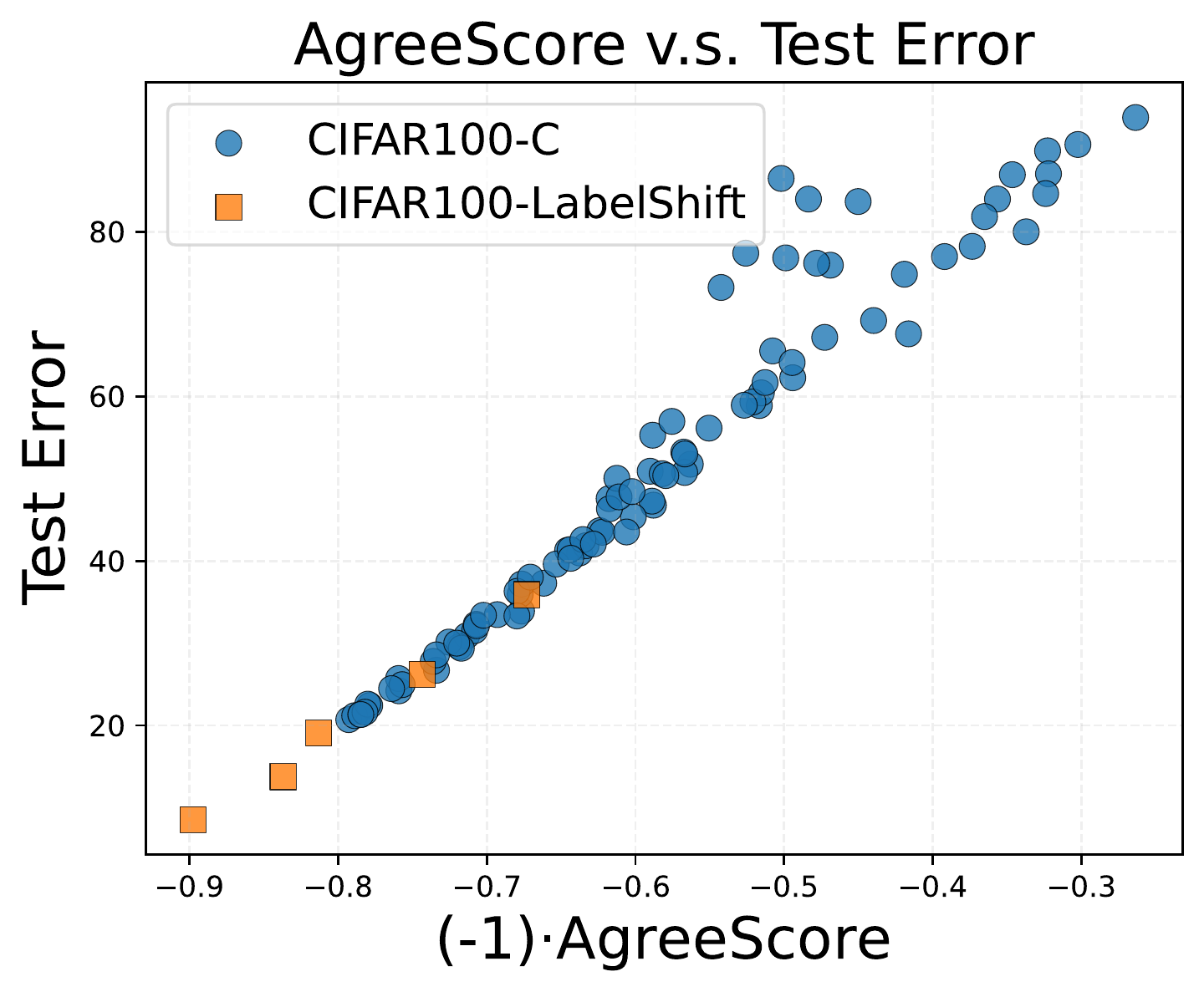}}
    \subfigure[ATC.]{\includegraphics[width=.3\textwidth]{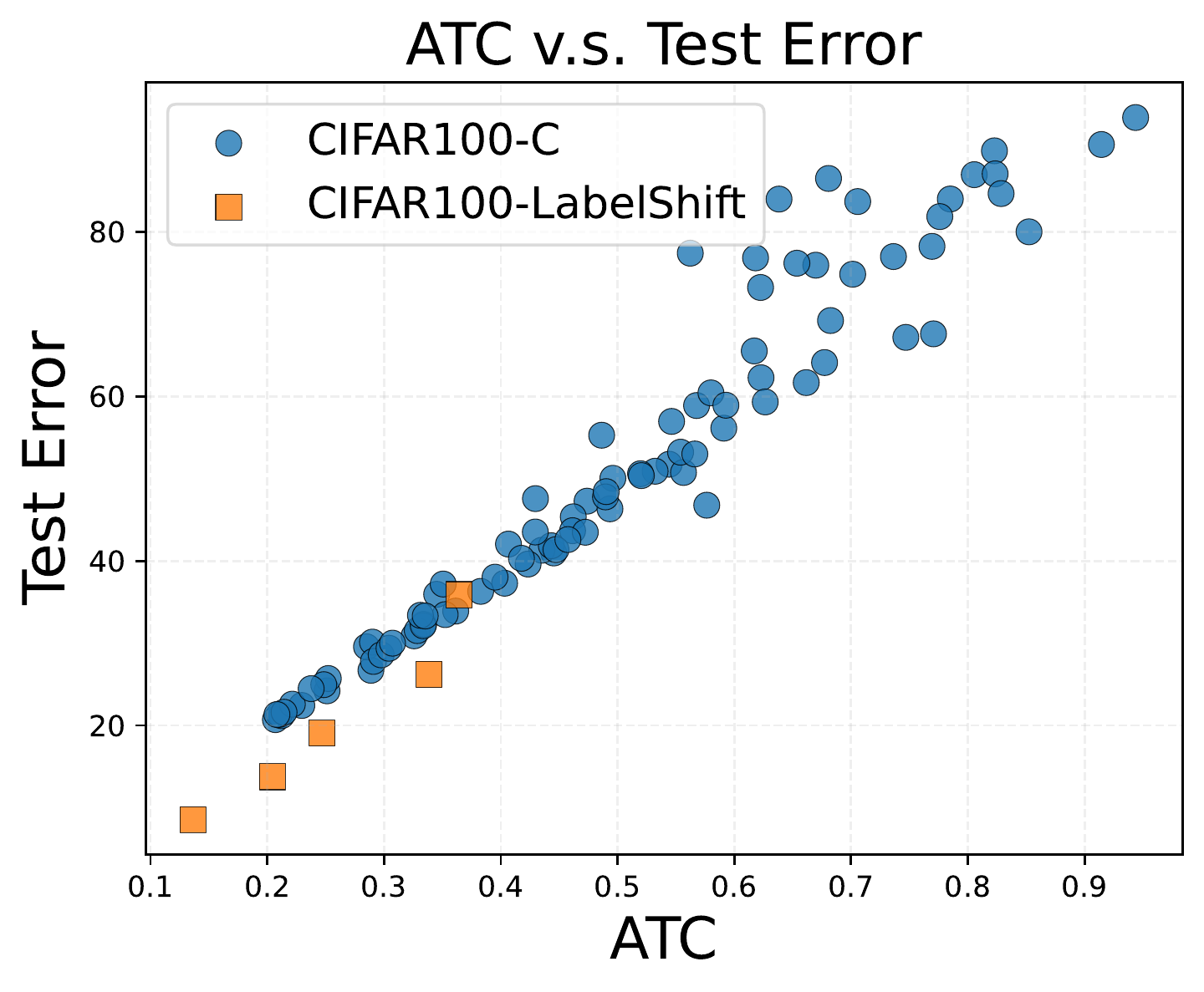}}
    \subfigure[ProjNorm.]{\includegraphics[width=.3\textwidth]{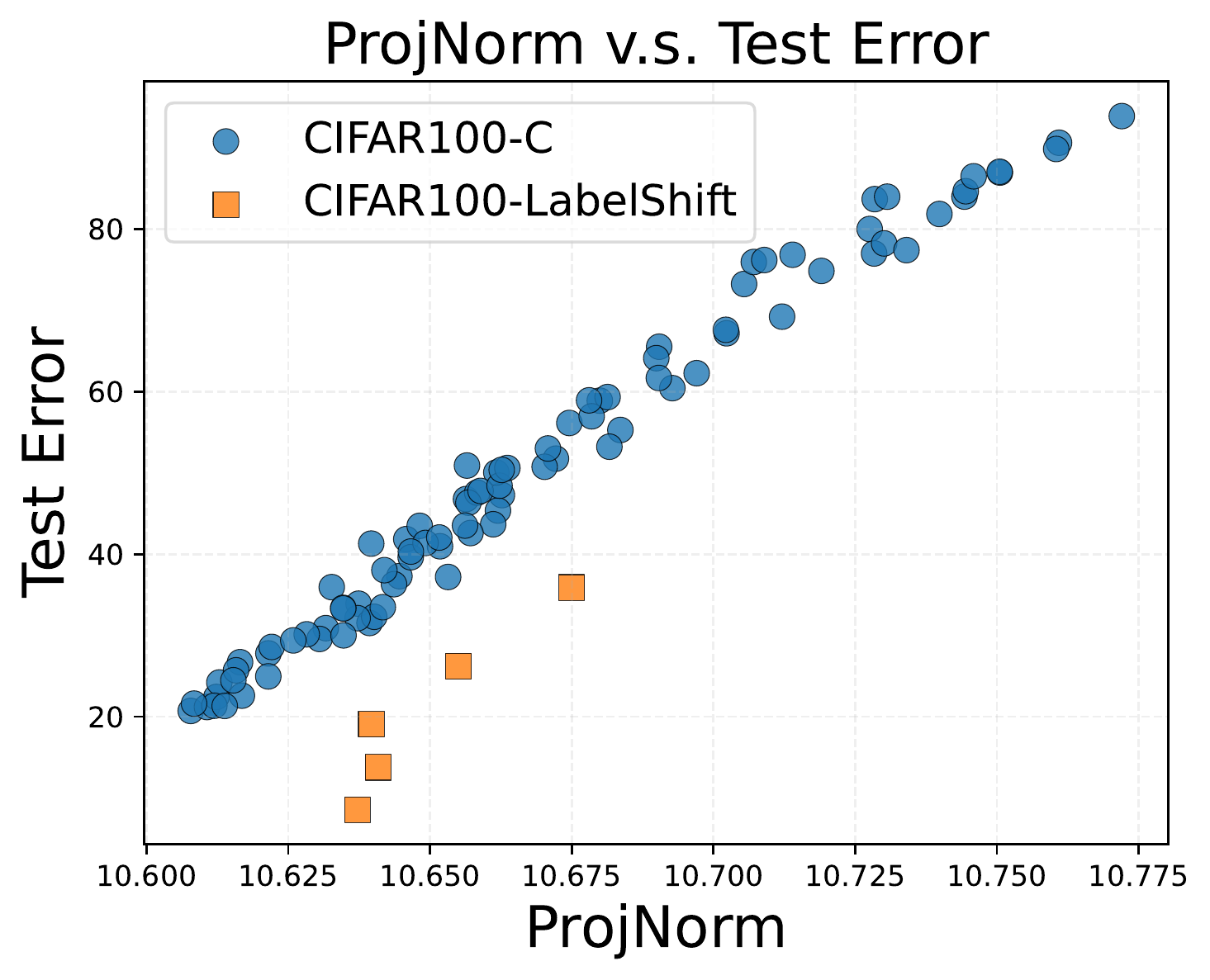}}
    \vspace{-0.1in}
    \caption{\textbf{Generalization prediction versus test error on CIFAR100 with ResNet18.} Compare out-of-distribution prediction performance of all methods. We plot the actual test error and the method prediction on each InD/OOD dataset. Blue circles are results evaluated on CIFAR100-C, and orange squares are results evaluated on 5 test datasets under label shift.
    }
    \label{fig:compare-appendix-labelshift}
    \vspace{-0.15in}
\end{figure*}

\newpage
\paragraph{Evaluation on label shift.} We evaluate our method and existing methods on CIFAR100 under label shift. Specifically, we measure the test error of each class from the in-distribution test dataset. Then, we rank the classes by the test error of each class (in descending order), i.e., $(c_1^{r}, c_2^{r}, \dots, c_{100}^{r})$. Finally, we partition the test dataset into five datasets $(D_1, \dots, D_5)$, where $D_j$ contains classes $((j-1)\cdot 20 + 1, \dots, j\cdot 20))$. The results are summarized in Figure~\ref{fig:compare-appendix-labelshift}. We find that \pj~performs worse than existing methods.

\newpage
\section{Details for the Toy Experiments}\label{sec:appendix-toy}
We construct a synthetic classification task with $\bx\in\R^d$ with
\[
\text{Training coviarte distribution:~} \cN\left(\bzero, \left[ \begin{array}{cc} \bI_{d_1} & \bzero \\ \bzero & \bzero \end{array} \right] \right).
\]
\[
\text{Test coviarte distribution:~} \cN\left(\bzero, \left[ \begin{array}{cc} \bI_{d_1} & \bzero \\ \bzero & \sigma^2 \bI_{d_2} \end{array} \right]\right).
\]
We set $d_1=1000$ and $d_2 = 500$. For both the training and test distributions, we assume class membership is given by
\[
 y|\bx = \text{sign}(\bx[1] + \bx[1500]).
\]
Given the definition of the training and test distributions, we sample $n=500$ training samples and $m=500$ test samples. Then, we perform the two-class linear regression to obtain Figure~\ref{fig:toy}.

\section{Details for NTK Experiments}\label{sec:appendix-ntk}
\begin{figure*}[ht!]
    \centering
    \subfigure{\includegraphics[width=.4\textwidth]{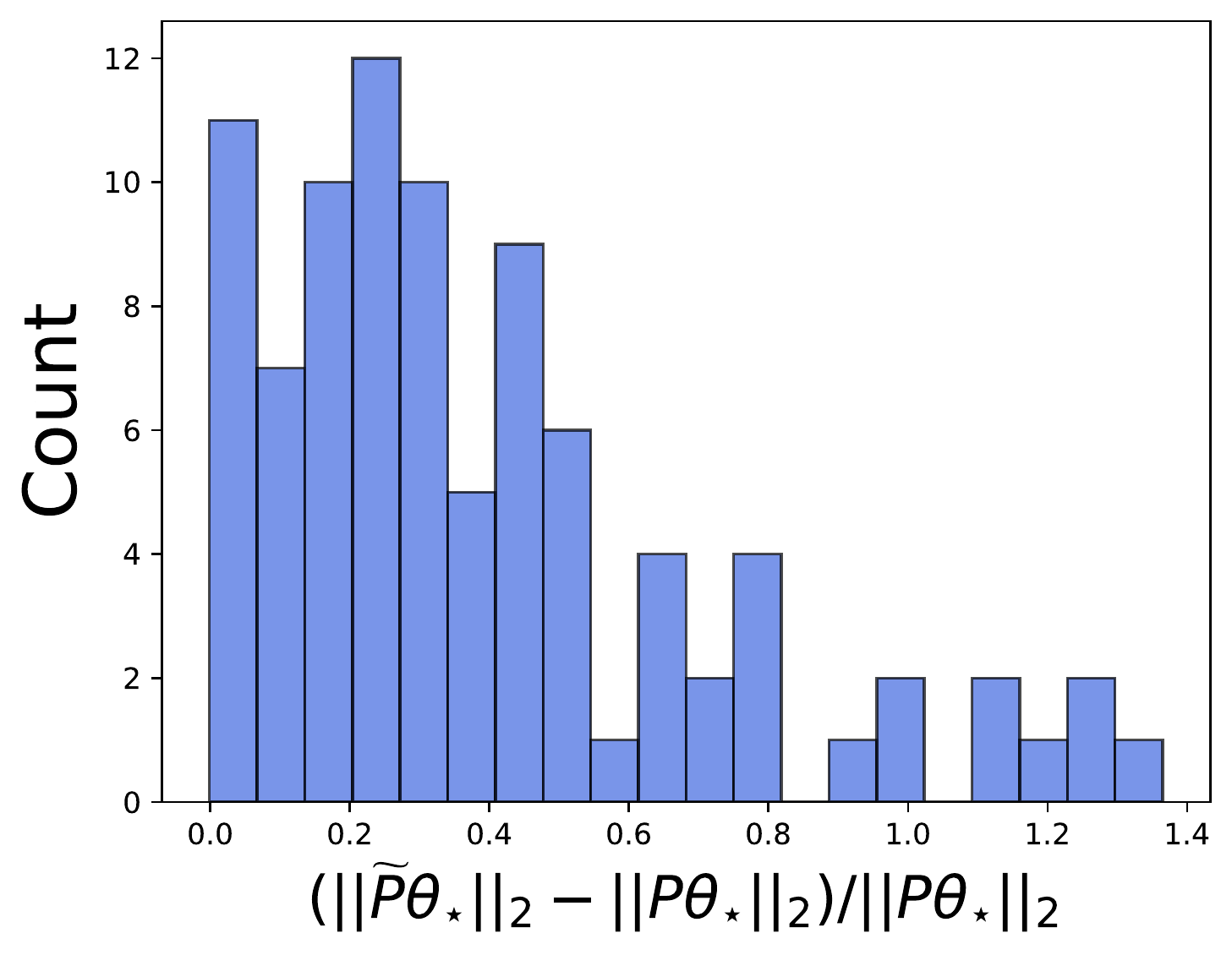}}
    \vspace{-0.1in}
    \caption{\textbf{Evaluation of $(\|\bPt\btheta_\star\| - \|\bP\btheta_\star\|)/\|\bP\btheta_\star\|$ on all OOD datasets from CIFAR10-C.} We empirically study Assumption~\ref{ass:norm} on CIFAR10-C. For each dataset in CIFAR10-C, we first randomly subsample 5,000 data points, $(\bXt_{\text{input}}, \byt)$. Then we use the ImageNet pre-trained ResNet18 to obtain NTK representations of the OOD data, i.e., $\bXt$. Then we set $\bPt\btheta_\star=\text{argmin}_{\btheta}\|\bXt\btheta-\byt\|_2^2$, and measure $\|\bPt\btheta_\star\|$.}
    \label{fig:thetastar-appendix}
    \vspace{-0.2in}
\end{figure*}

\begin{figure*}[h]
    \centering
    \subfigure{\includegraphics[width=.24\textwidth]{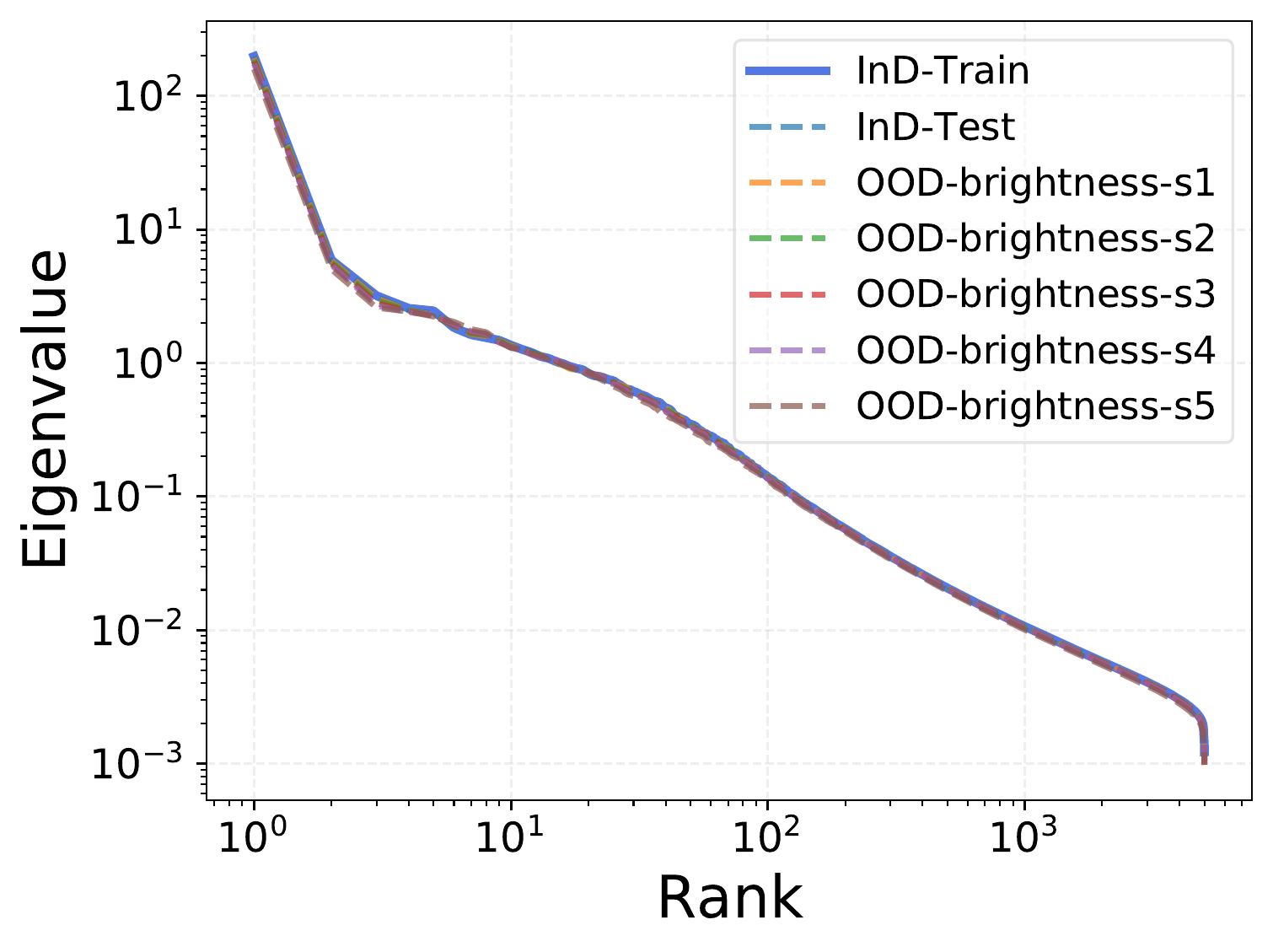}}
    \subfigure{\includegraphics[width=.24\textwidth]{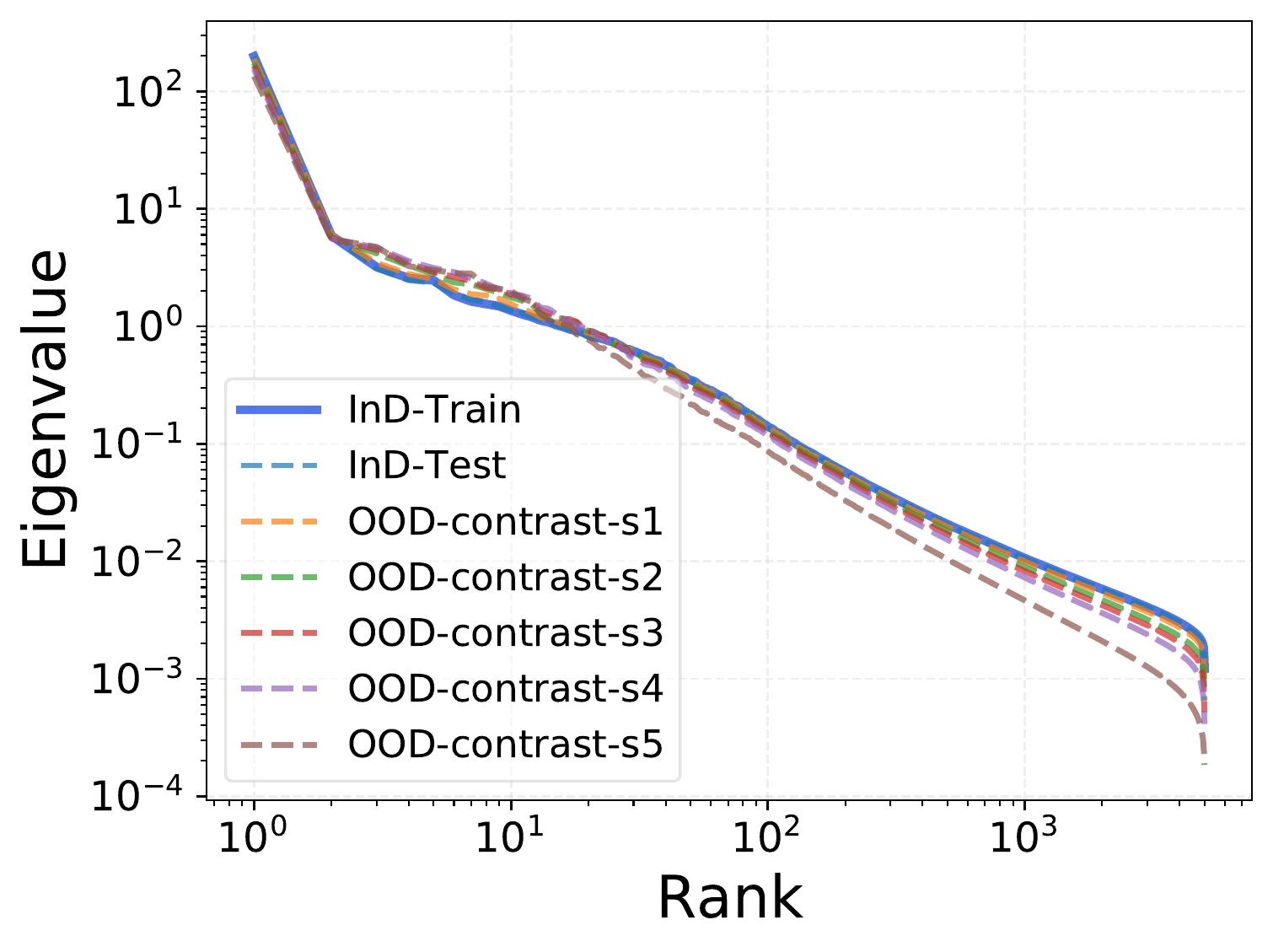}}
    \subfigure{\includegraphics[width=.24\textwidth]{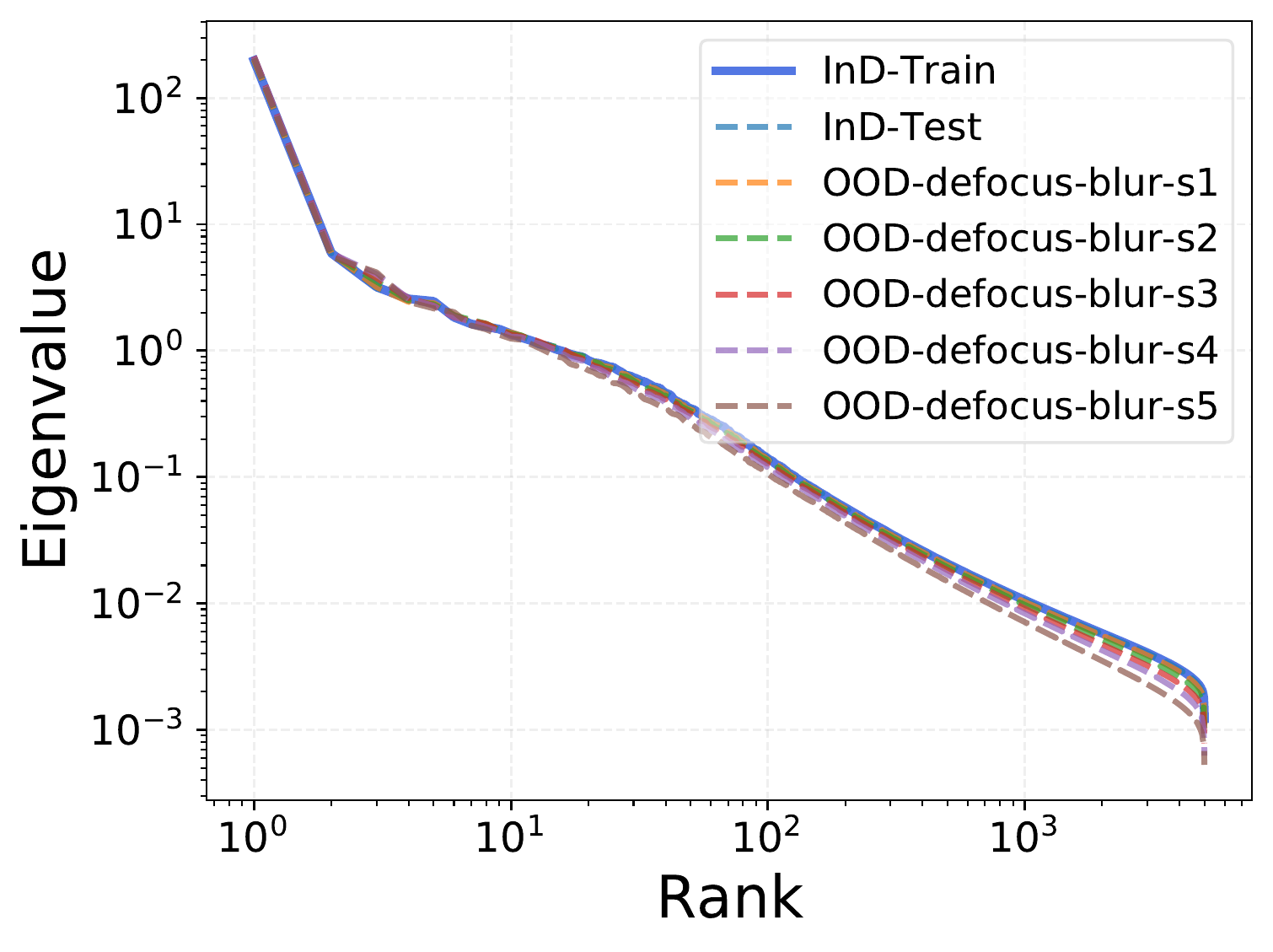}}
    \subfigure{\includegraphics[width=.24\textwidth]{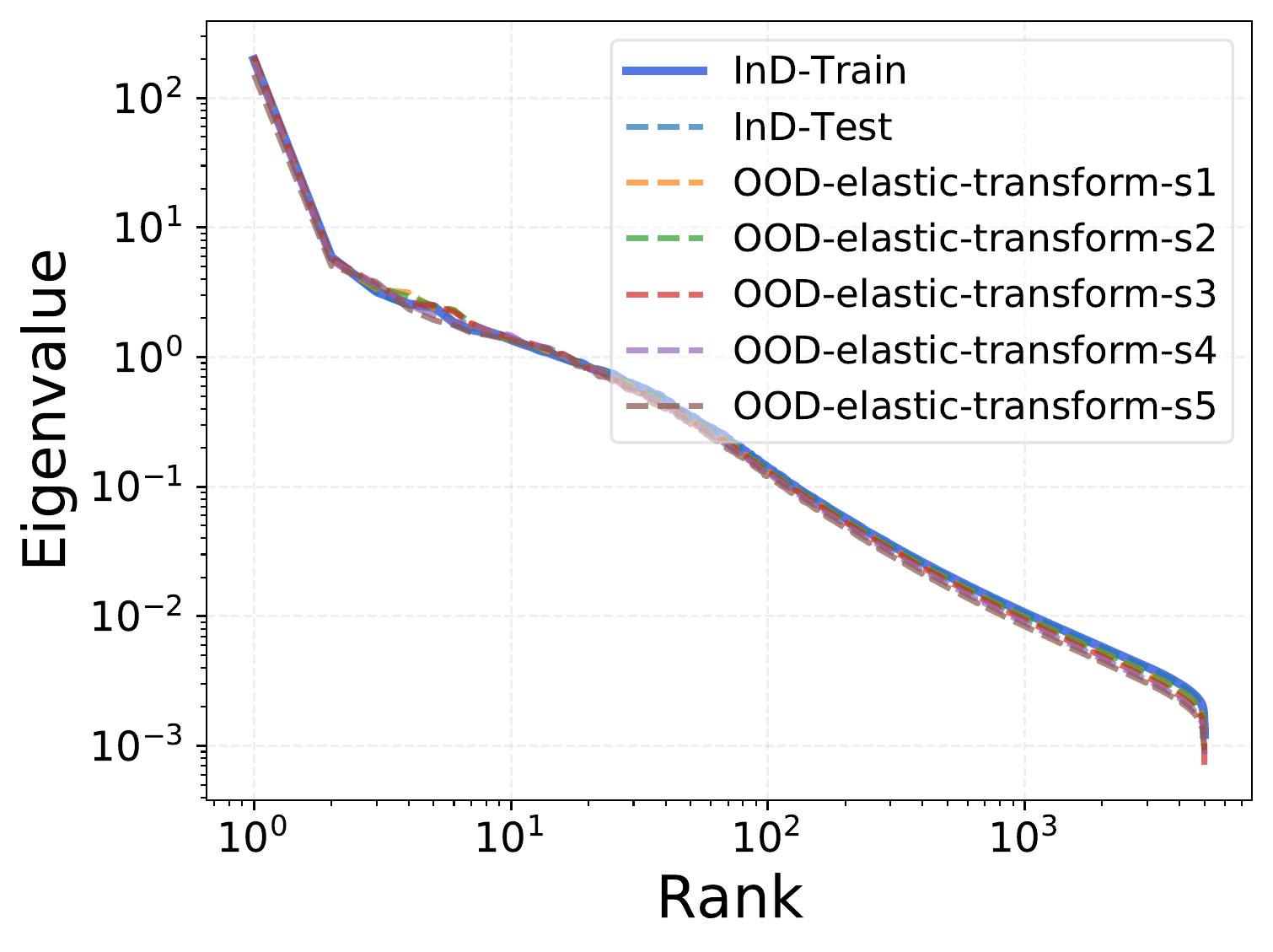}}
    \subfigure{\includegraphics[width=.24\textwidth]{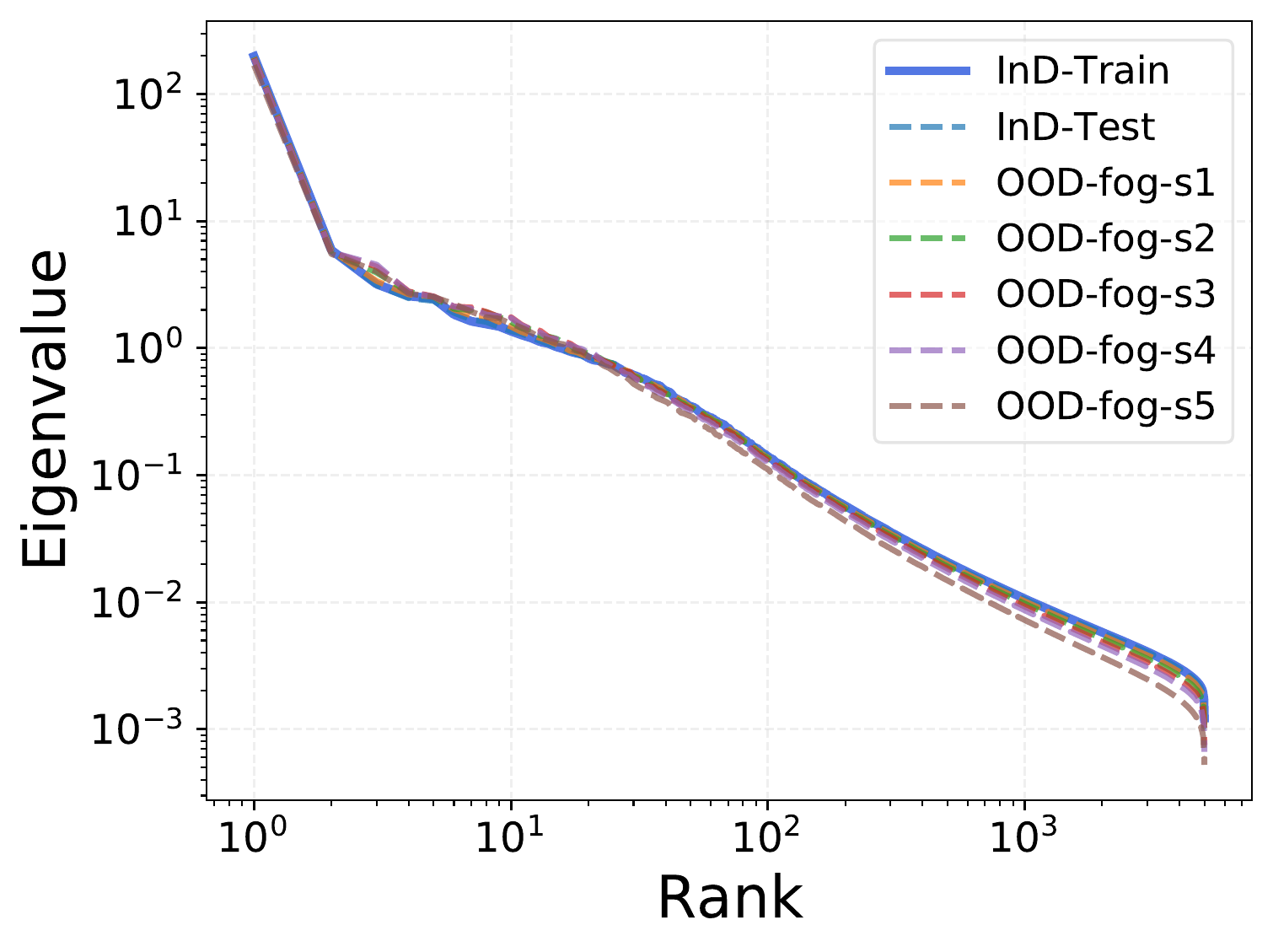}}
    \subfigure{\includegraphics[width=.24\textwidth]{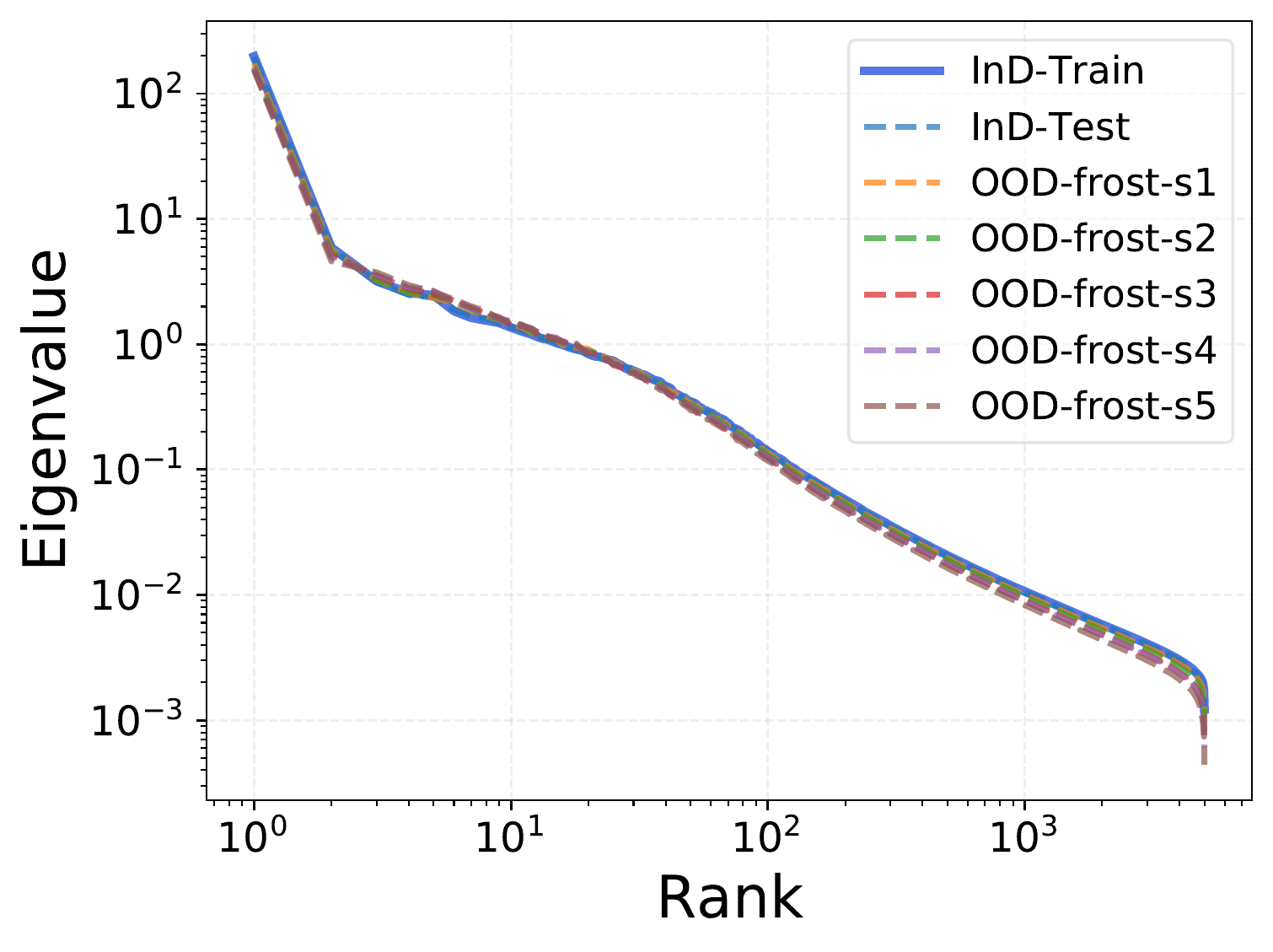}}
    \subfigure{\includegraphics[width=.24\textwidth]{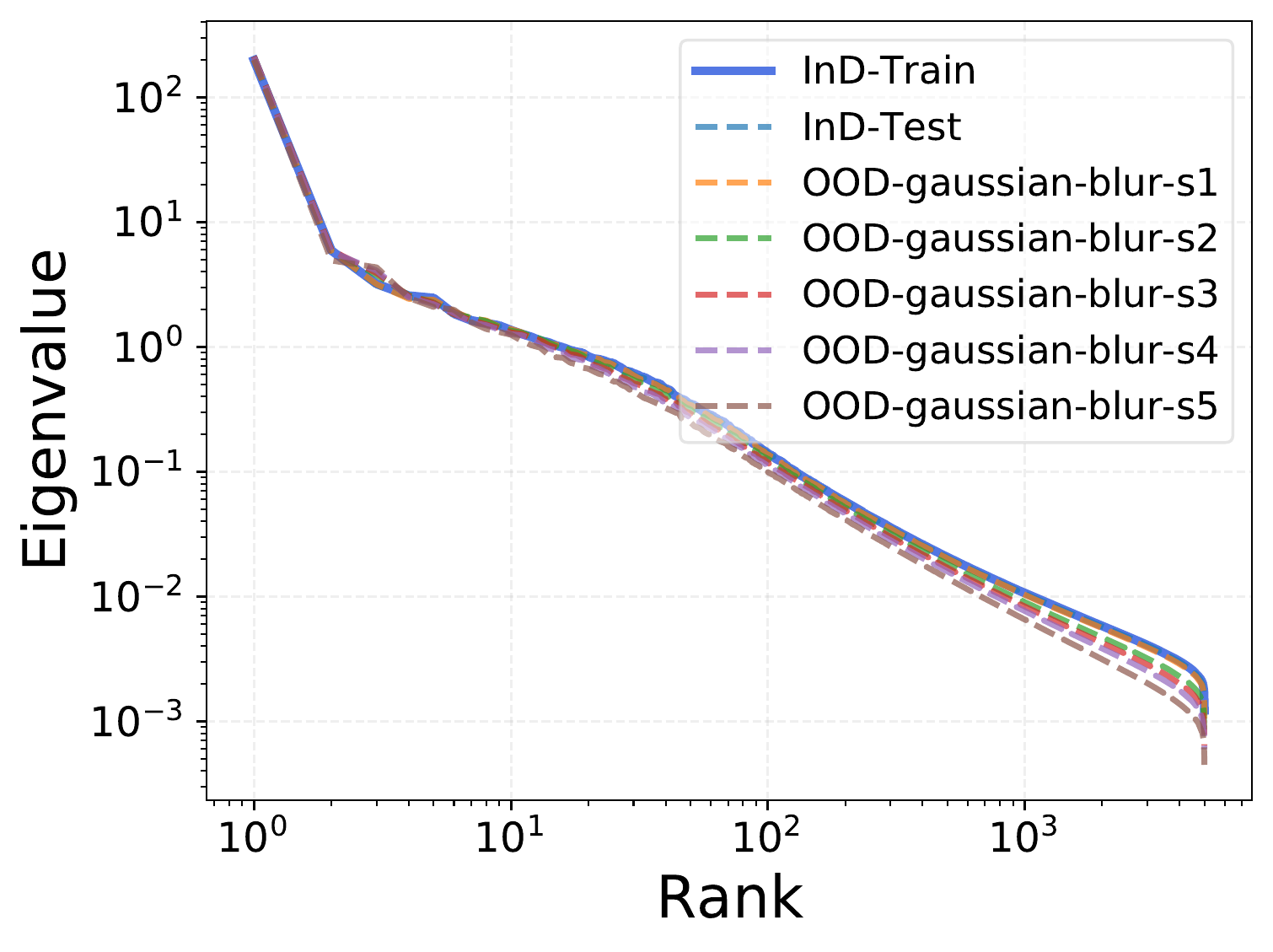}}
    \subfigure{\includegraphics[width=.24\textwidth]{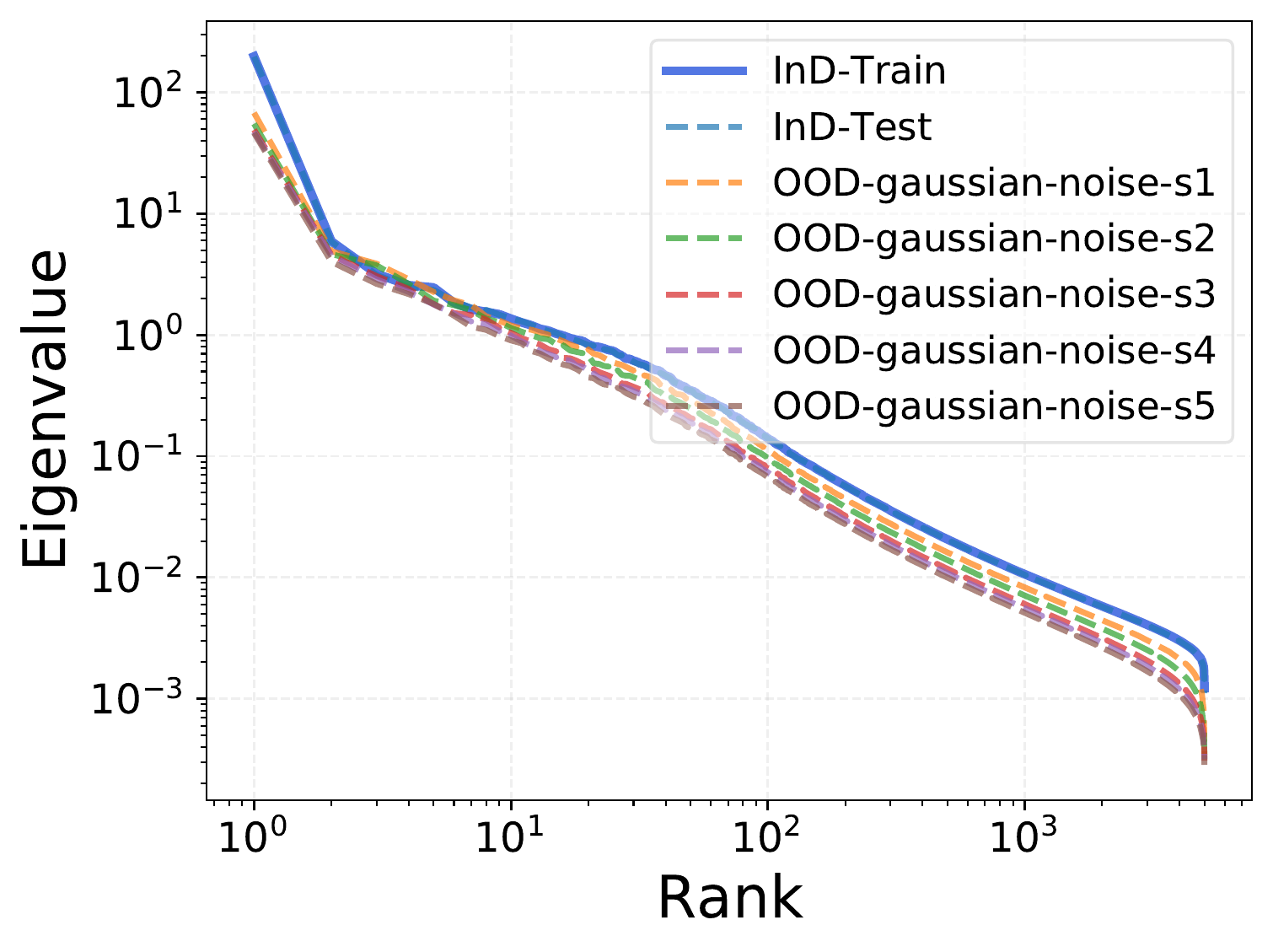}}
    \subfigure{\includegraphics[width=.24\textwidth]{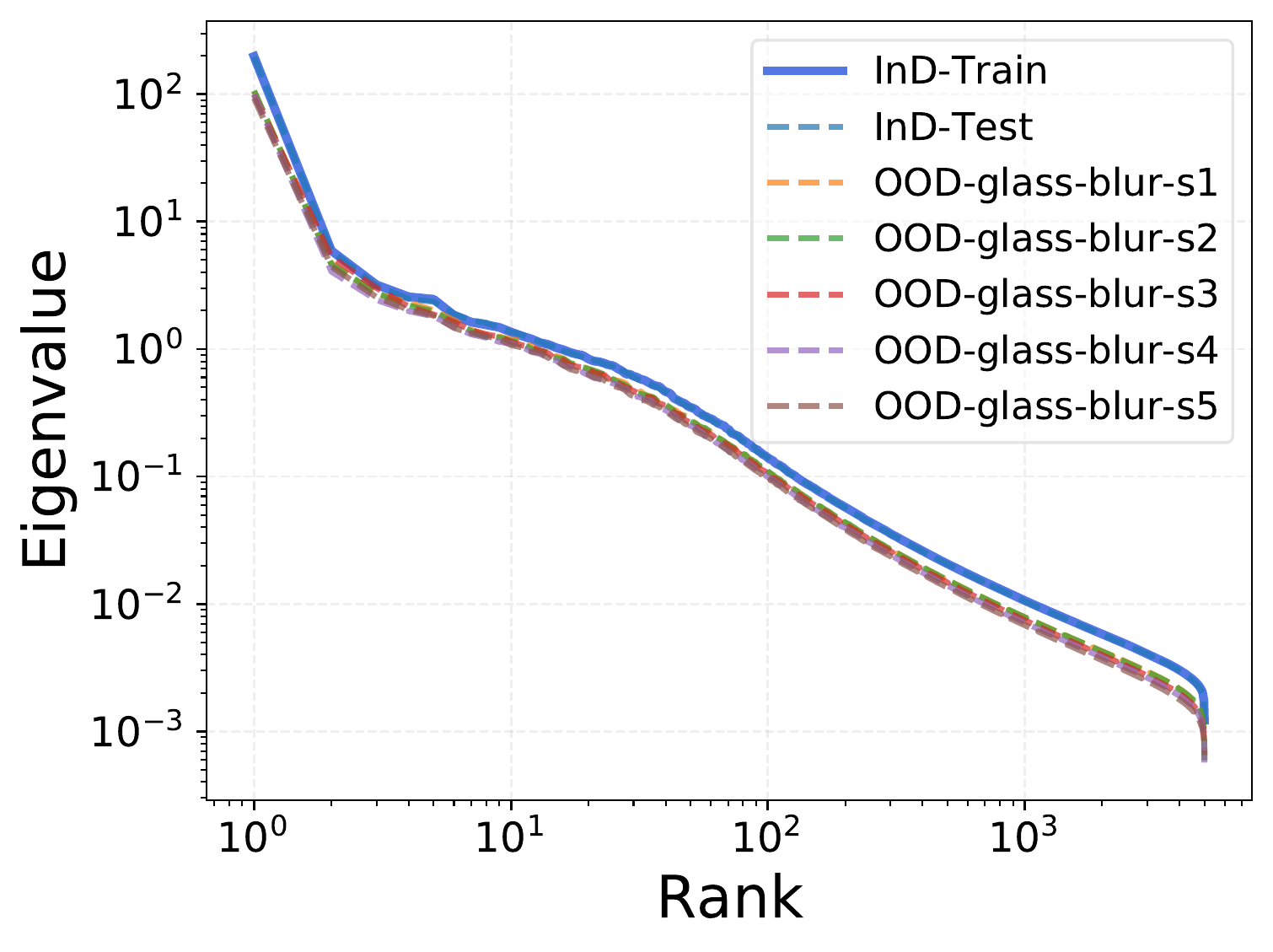}}
    \subfigure{\includegraphics[width=.24\textwidth]{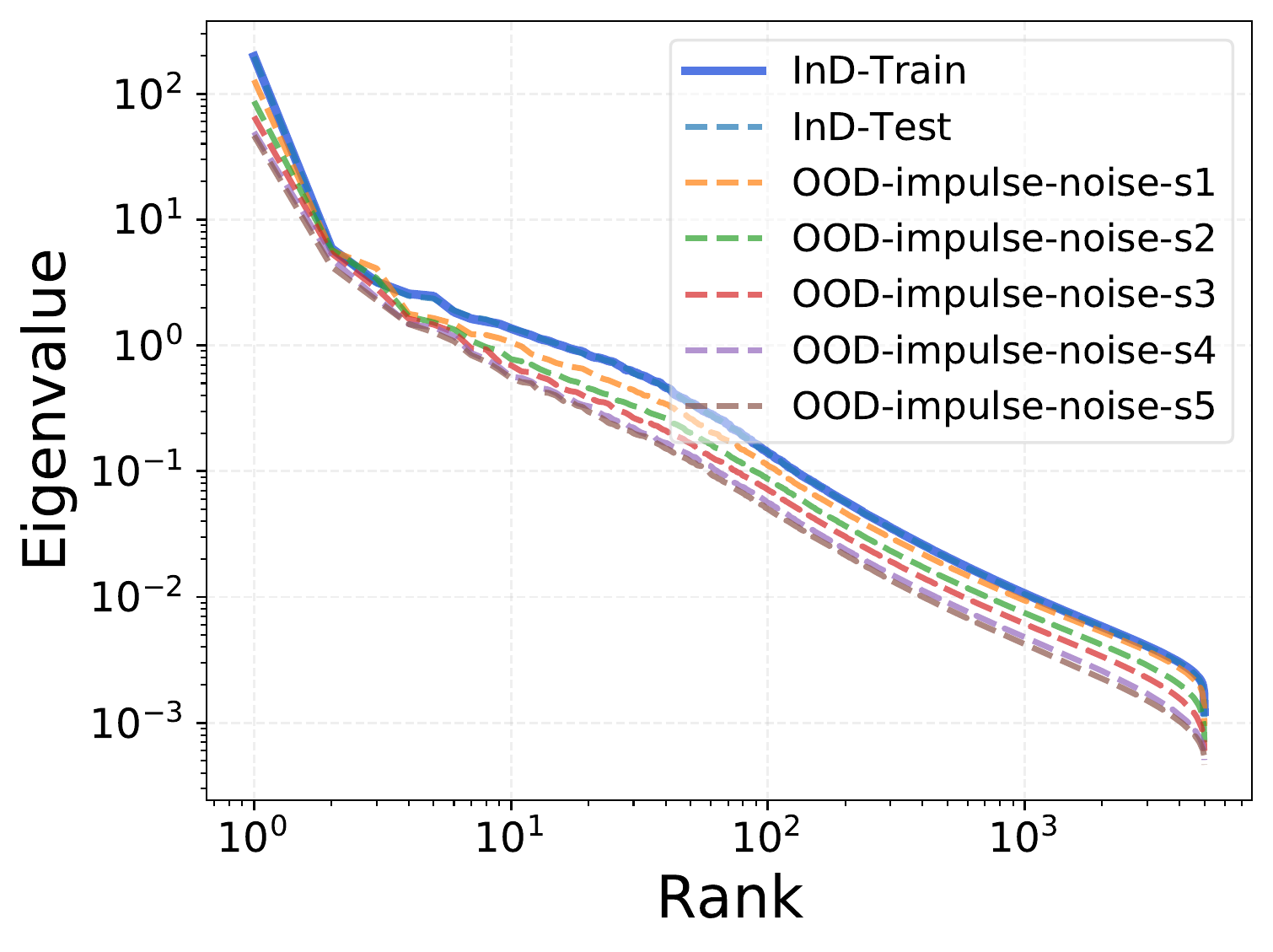}}
    \subfigure{\includegraphics[width=.24\textwidth]{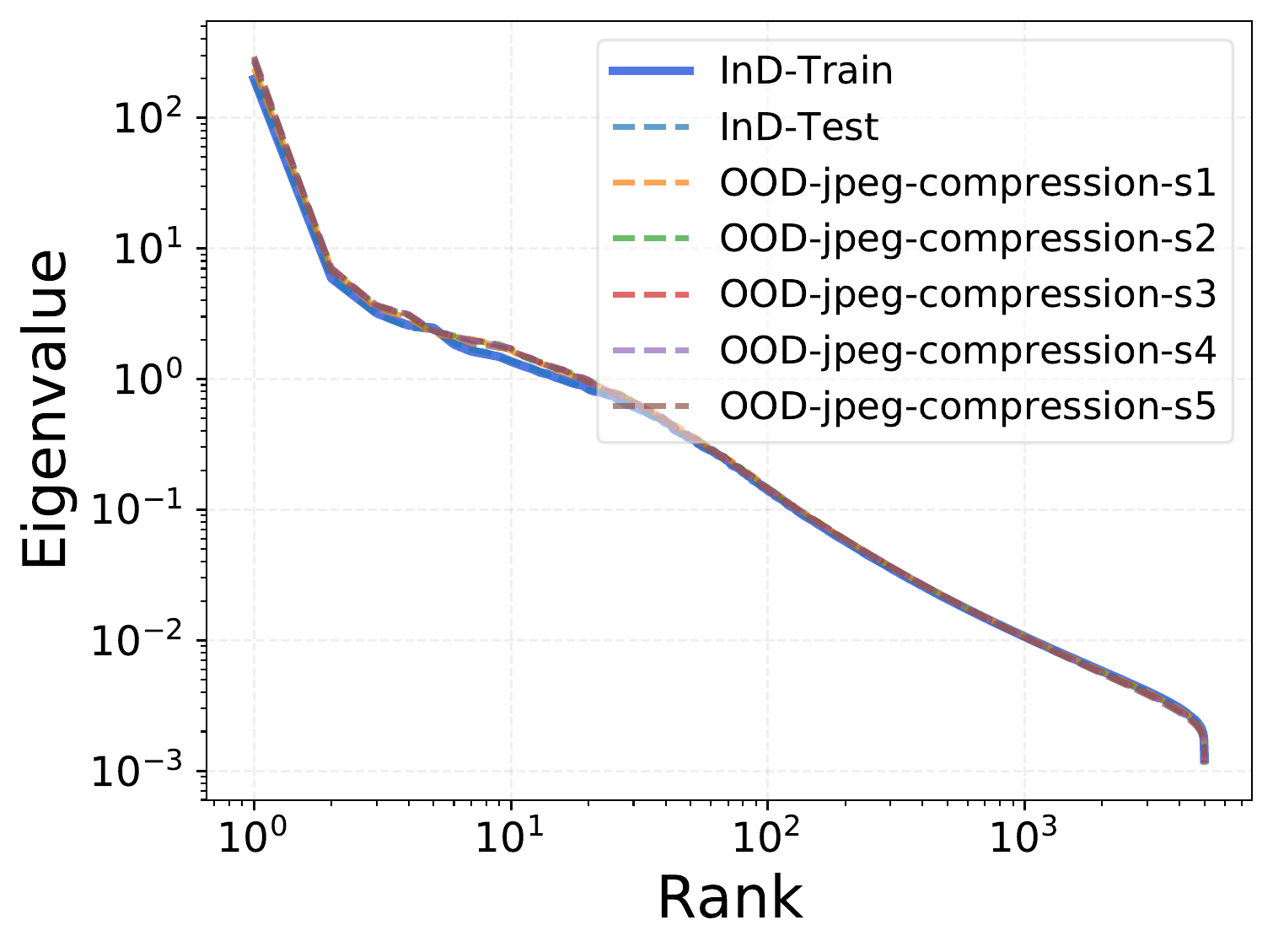}}
    \subfigure{\includegraphics[width=.24\textwidth]{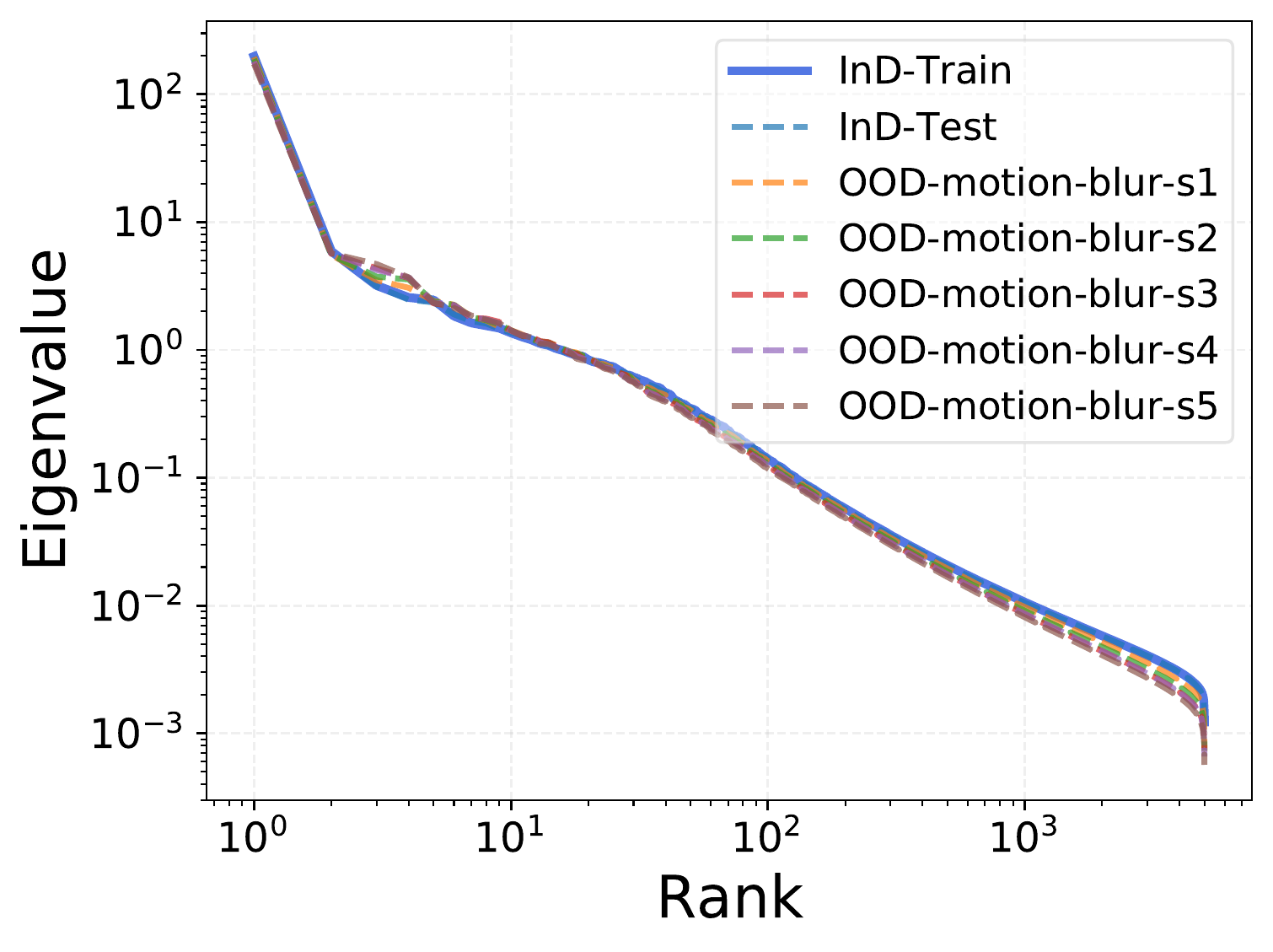}}
    \subfigure{\includegraphics[width=.24\textwidth]{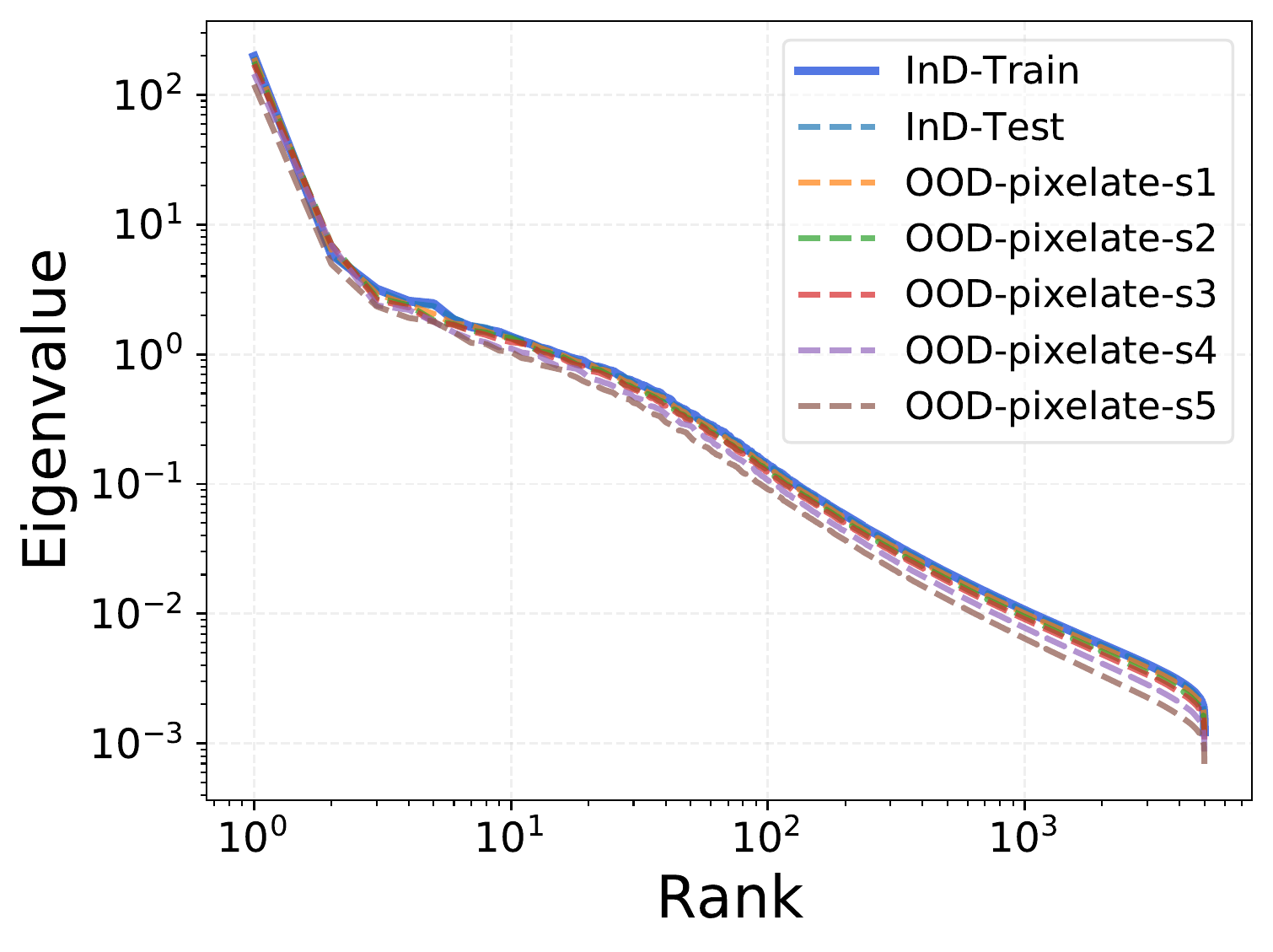}}
    \subfigure{\includegraphics[width=.24\textwidth]{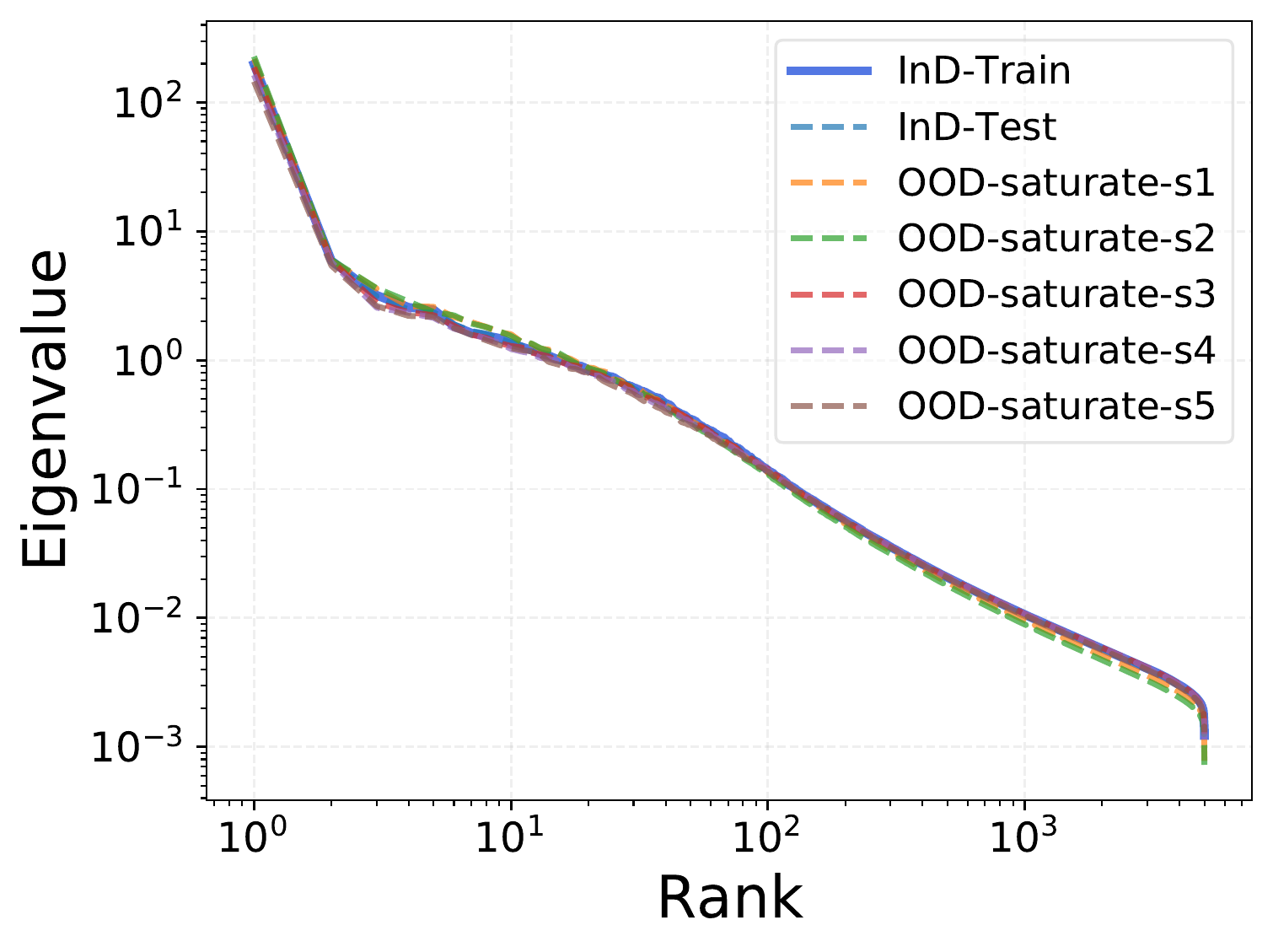}}
    \subfigure{\includegraphics[width=.24\textwidth]{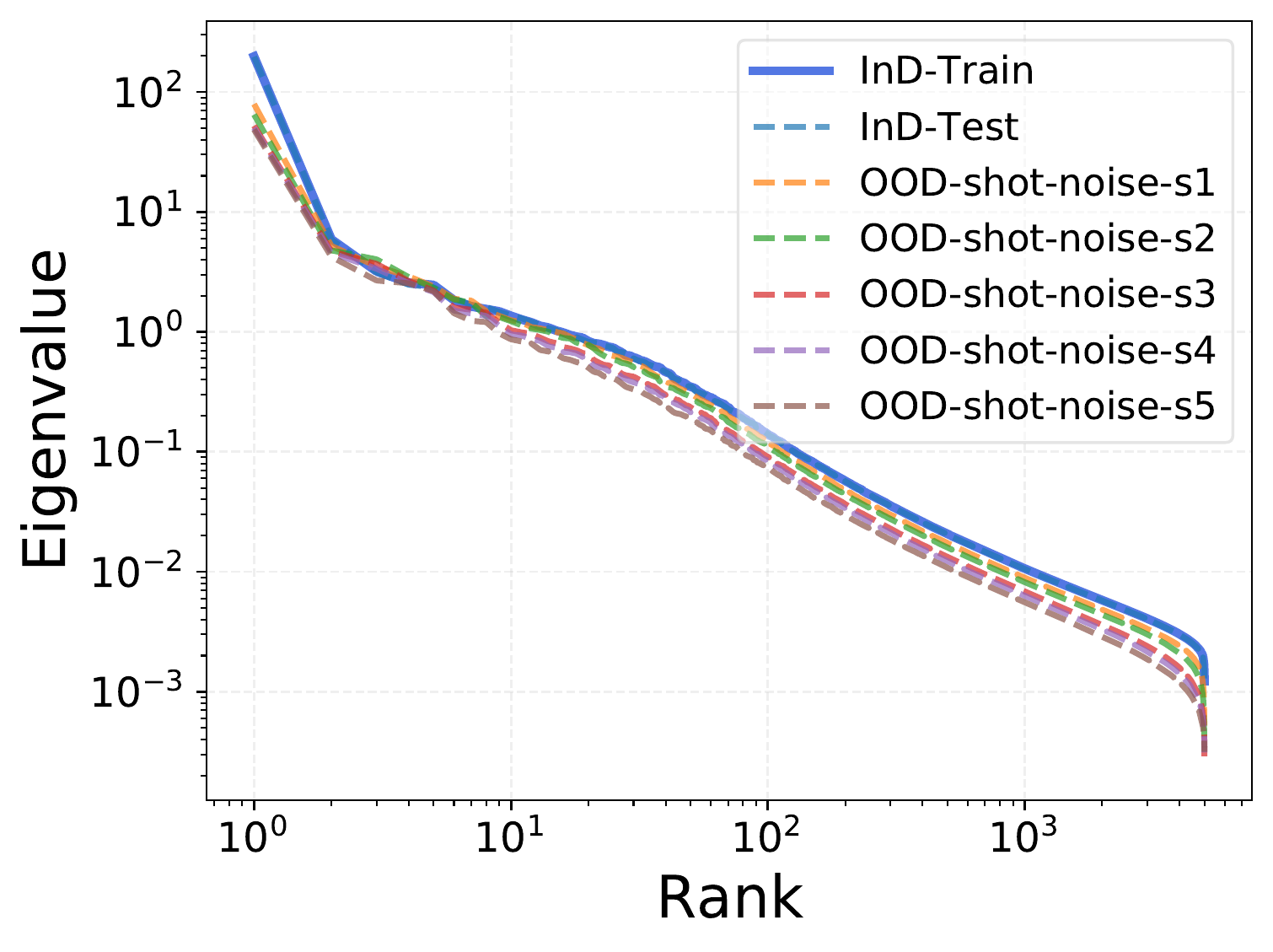}}
    \subfigure{\includegraphics[width=.24\textwidth]{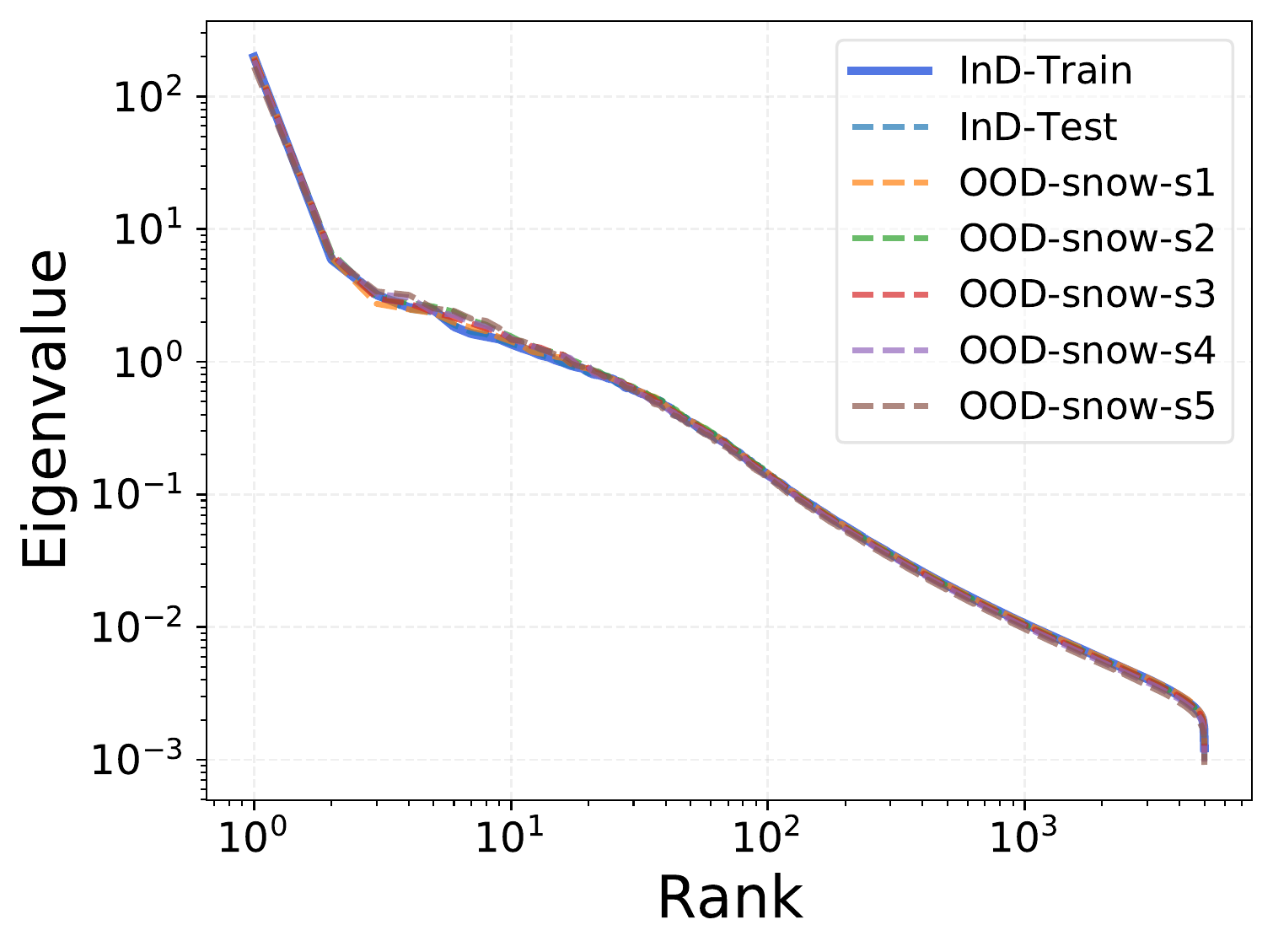}}
    \subfigure{\includegraphics[width=.24\textwidth]{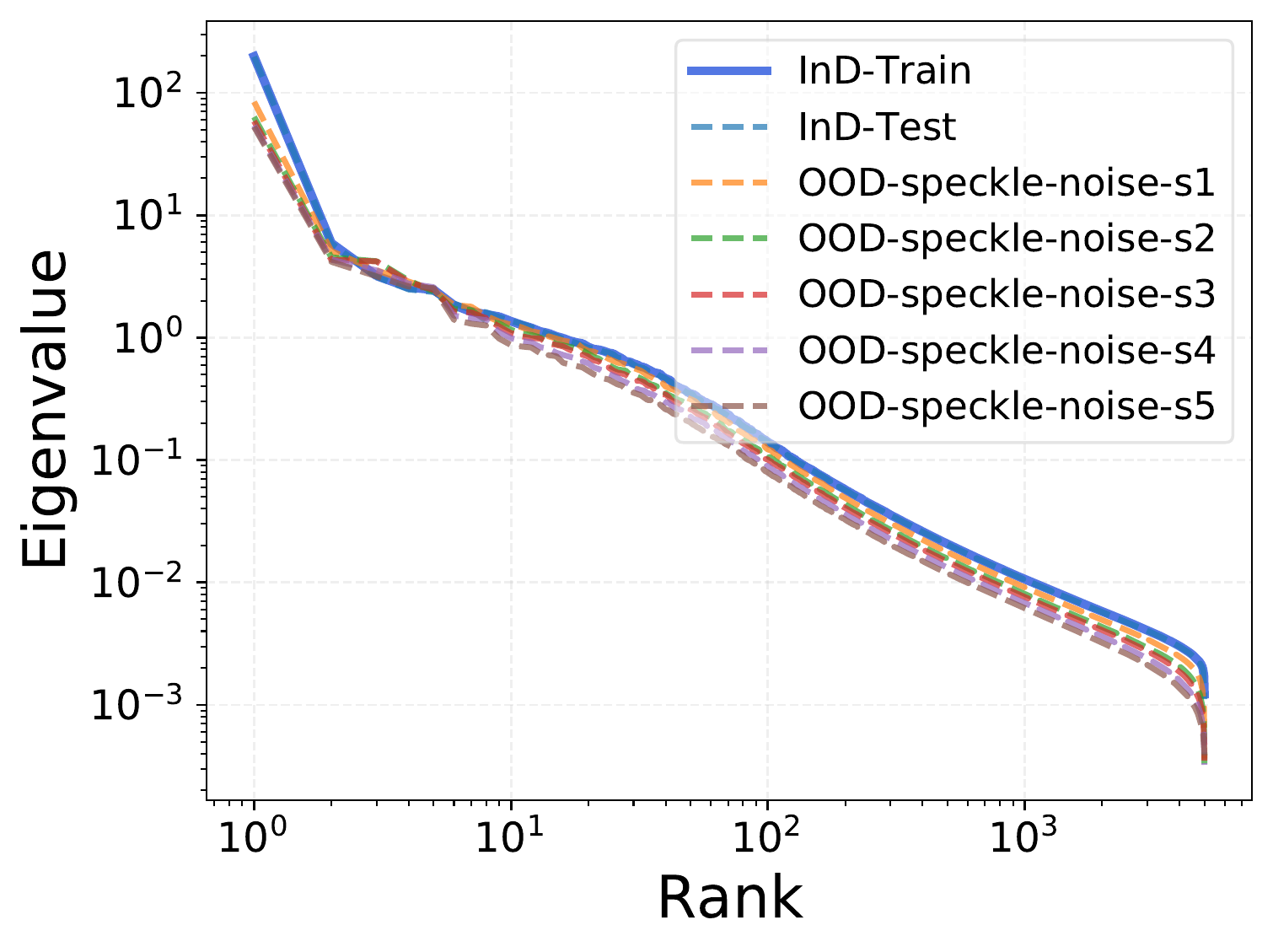}}
    \subfigure{\includegraphics[width=.24\textwidth]{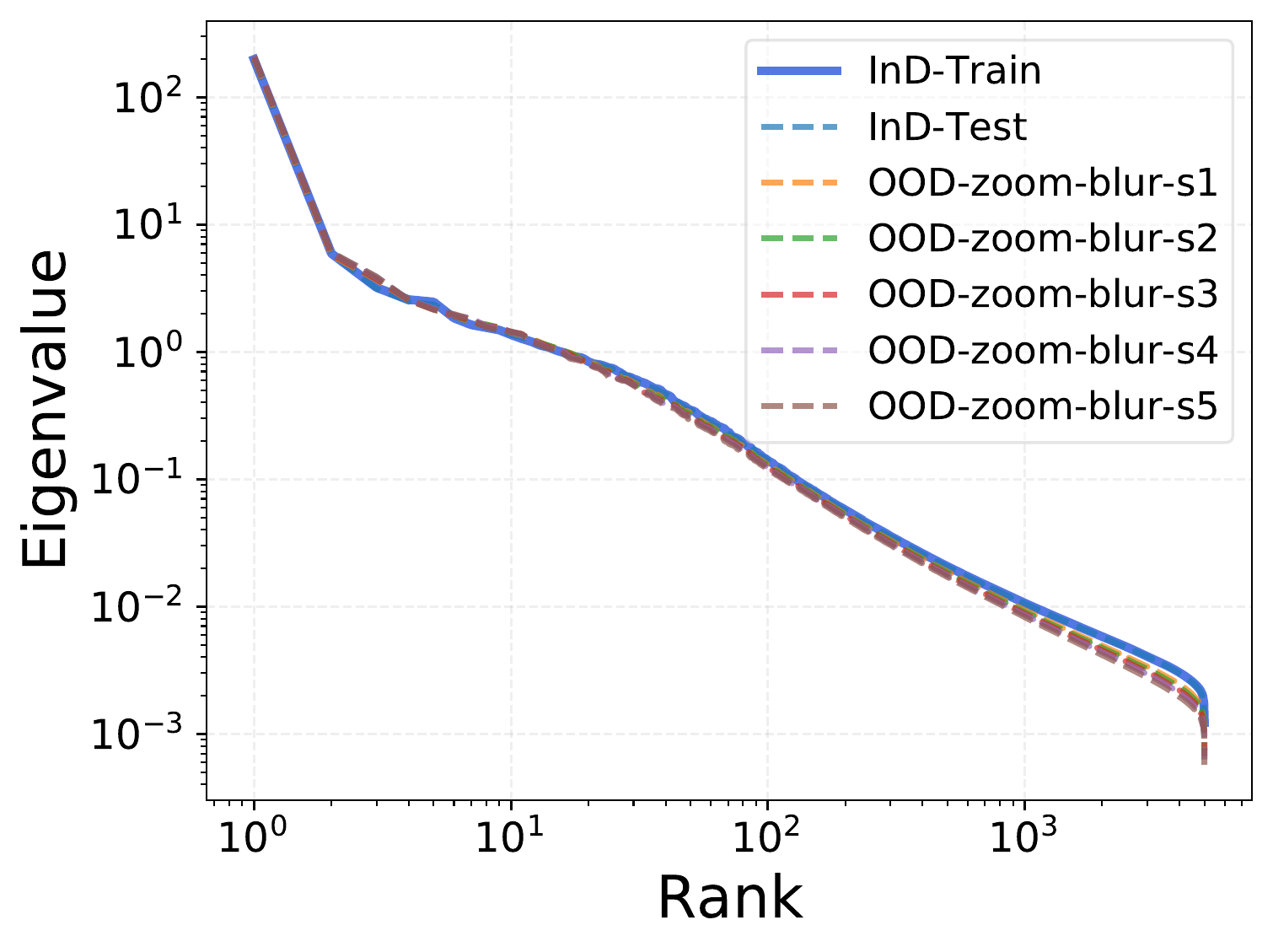}}
    \vspace{-0.1in}
    \caption{Results on eigenvalue decay in $\log$-$\log$ scale, including InD train, InD test, and all corruptions in CIFAR10-C.}
    \label{fig:ntk-eigen-appendix}
    \vspace{-0.1in}
\end{figure*}

As shown in Figure~\ref{fig:thetastar-appendix}, we visualize the evaluations of $(\|\bPt\btheta_\star\| - \|\bP\btheta_\star\|)/\|\bP\btheta_\star\|$ for all corruptions in CIFAR10-C. We present the eigenvalue decay results in Figure~\ref{fig:ntk-eigen-appendix}, which include all corruptions in CIFAR10-C.

\clearpage
\section{More Experimental Results on Adversarial Examples}\label{sec:appendix-adv}
\begin{figure*}[ht]
    \centering
    \subfigure[ConfScore.]{\includegraphics[width=.24\textwidth]{figs/adv/dnn_cifar10_ConfScore.pdf}}
    \subfigure[AgreeScore.]{\includegraphics[width=.24\textwidth]{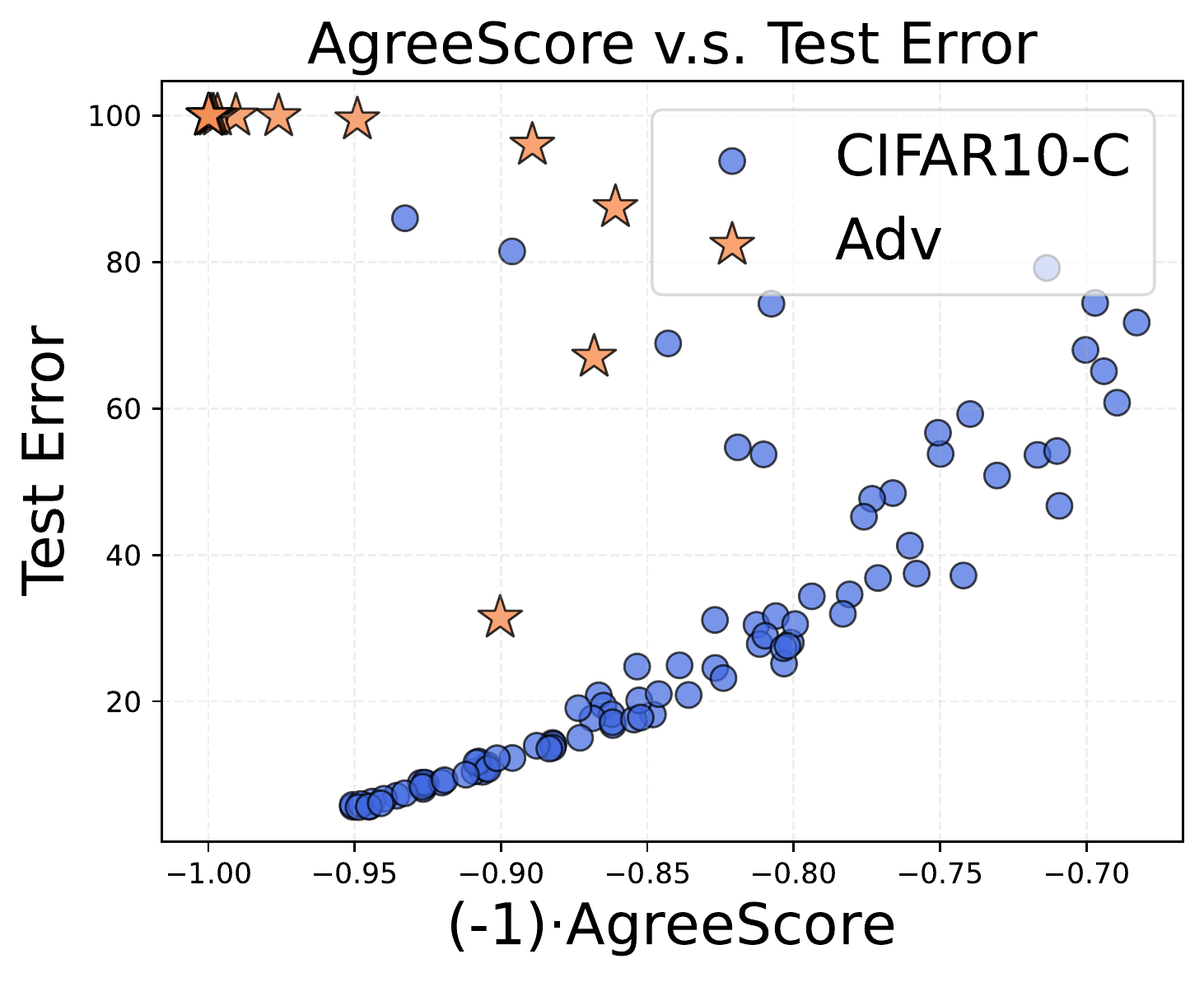}}
    \subfigure[Entropy.]{\includegraphics[width=.24\textwidth]{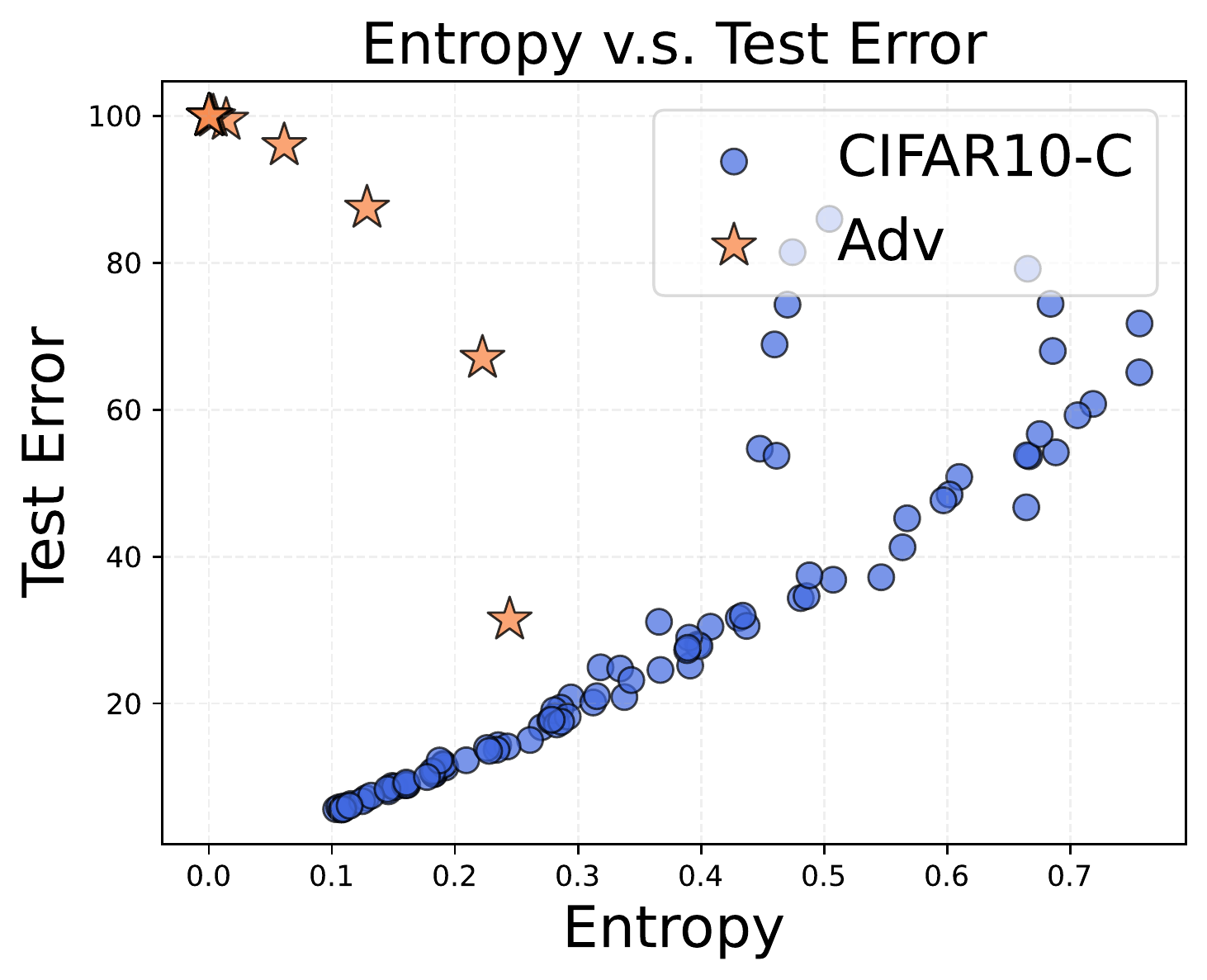}}
    \subfigure[ATC.]{\includegraphics[width=.24\textwidth]{figs/adv/dnn_cifar10_ATC.pdf}}
    \vspace{-0.1in}
    \caption{Evaluation of existing methods on predicting OOD error under adversarial attack. Blue circles are results evaluated on CIFAR10-C~(each point corresponds to one corrupted test dataset), and orange stars are results evaluated on adversarial examples~(each point corresponds to one perturbation radius $\varepsilon$). }
    \label{fig:adv-appendix}
    \vspace{-0.2in}
\end{figure*}

We provide additional experimental results for Section~\ref{sec:stress}. The prediction performance results of existing methods and \pj~are summarized in Table~\ref{table:table-adv-appendix}~(measured in MSE) and Table~\ref{table:table-detail-adv-appendix}. We also present the scatter plots of prediction on adversarial examples versus test error for existing methods in Figure~\ref{fig:adv-appendix}.

\begin{table*}[ht]
\centering
\caption{\textbf{Prediction performance under adversarial attack of different methods measure in MSE.} We first fir a linear regression model on CIFAR10-C results, ($\textsf{Prediction}(\Dtest)$, $\err(\Dtest, \bthetah)$), for each method. Then we use the learned linear model to predict the OOD error of adversarial examples with perturbation size varying from $0.25$ to $8.0$. The prediction performance is measured by MSE. Lower is better.}
\vspace{0.05in}
\label{table:table-adv-appendix}
\begin{tabular}{@{}ccccccc@{}}
\toprule
 & ConfScore  & Entropy  & AgreeScore & ATC & ProjNorm\\
\midrule
\multirow{1}{*}{CIFAR10} & 0.875 & 0.895 &
0.796 & 0.823 & \textbf{0.432}\\
\bottomrule
\end{tabular}
\vspace{-0.15in}
\end{table*}

\begin{table*}[ht]
\centering
\caption{\textbf{Prediction performance under adversarial attack of different methods measure in MSE.} We first fir a linear regression model on CIFAR10-C results, ($\textsf{Prediction}(\Dtest)$, $\err(\Dtest, \bthetah)$), for each method. Then we use the learned linear model to predict the OOD error of adversarial examples with perturbation size varying from $0.25$ to $8.0$. For each perturbation $\varepsilon$, we present the actual test error (``Test Error'' in the table) and the predictions by \pj\, and other methods.}
\vspace{0.1in}
\label{table:table-detail-adv-appendix}
\begin{tabular}{@{}cccccccccc@{}}
\toprule
 & $\varepsilon=0.0$  & $\varepsilon=0.25$  & $\varepsilon=0.5$ & $\varepsilon=0.75$ & $\varepsilon=1.0$ & $\varepsilon=1.5$ & $\varepsilon=2.0$ & $\varepsilon=2.5$ \\
\midrule
\multirow{1}{*}{Test Error} & 5.6 & 31.4 &
67.0 & 87.4 & 96.0 & 99.4 & 99.9 & 99.9\\
\multirow{1}{*}{ConfScore} & 3.5 & 19.3 & 17.4 & 7.0 & -0.3 & -5.2 & -6.2 & -6.4 &  \\
\multirow{1}{*}{Entropy} & 3.0 & 17.4 & 15.1 & 5.5 & -1.3 & -6.1 & -7.2 & -7.4 &  \\
\multirow{1}{*}{AgreeScore} & 5.2 & 16.0 & 23.0 & 24.6 & 18.4 & 5.5 & -0.3 & -3.4 &  \\
\multirow{1}{*}{ATC} & 4.5 & 14.6 & 12.0 & 5.8 & 1.3 & -1.4 & -2.2 & -2.4 &  \\
\multirow{1}{*}{\pj} & 5.2 & 7.2 & 22.5 & 28.7 & 31.8 & 29.1 & 26.4 & 25.1 &  \\
\midrule
& $\varepsilon=3.0$ & $\varepsilon=3.5$ & $\varepsilon=4.0$  & $\varepsilon=5.0$  & $\varepsilon=6.0$ & $\varepsilon=7.5$ & $\varepsilon=8.0$\\
\midrule
\multirow{1}{*}{Test Error} & 100.0 & 100.0 & 100.0 & 100.0 & 100.0 & 100.0 & 100.0\\
\multirow{1}{*}{ConfScore} & -6.4 & -6.5 & -6.5 &-6.5 &-6.5 &-6.5 &-6.5  \\
\multirow{1}{*}{Entropy} & -7.5 & -7.5 & -7.5 & -7.5 & -7.5 & -7.5 & -7.5 \\
\multirow{1}{*}{ATC} & -2.4 & -2.4 & -2.4 & -2.4 & -2.4 & -2.4 & -2.4   \\
\multirow{1}{*}{\pj} & 24.4 & 24.5 & 25.0 & 25.2 & 25.8 & 27.0 & 28.1   \\
\bottomrule
\end{tabular}
\end{table*}

\newpage
\clearpage
\section{Proof of Proposition \ref{proposition:ntk-eigen}}
\begin{proof}
Recall that we decompose the empirical covariance of training and test set as
\begin{equation*}
    \bSigma = \frac{1}{n}\bX^\sT\bX = \frac{1}{n}\sum_{i=1}^n \mu_i \bu_i\bu_i^\sT,
\end{equation*}
\begin{equation*}
    \bSigmat = \frac{1}{m}\bXt^\sT\bXt = \frac{1}{m}\sum_{j=1}^m \lambda_j \bv_j\bv_j^\sT.
\end{equation*}
Then given $k$ from Assumption \ref{ass:spectral}, we define the projection matrices
\[
\begin{aligned}
\bP_0 &= \sum_{i=1}^k\bu_i\bu_i^\sT = \sum_{j=1}^k\bv_j\bv_j^\sT, \\
\bP_\perp &= \bP-\bP_0 = \sum_{i=k+1}^n \bu_i\bu_i^\sT, \\
\bPt_\perp &= \bPt-\bPt_0 = \sum_{j=k+1}^m \bv_j\bv_j^\sT.
\end{aligned}
\]
The test loss can be written as
\[\tl = \frac{1}{m}\|\bXt(\bI-\bP)\btheta_\star\|_2^2 = \frac{1}{m}\|\bXt\bPt(\bI-\bP)\btheta_\star\|_2^2.\]
Under Assumption~$\ref{ass:spectral}$,
\[\bPt(\bI-\bP) = (\bP_0 + \bPt_\perp)(\bI - \bP_0 - \bP_\perp) = \bPt_\perp.\]
This allows us to simply write the test loss as
\[\tl = \frac{1}{m}\|\bXt\bPt_\perp\btheta_\star\|_2^2 
= \frac{1}{m}\sum_{j=k+1}^m \lambda_j \<\bv_j, \btheta_\star\>^2.\]
Since $\lambda_j$ is a the decreasing sequence of eigenvalues
\[
\frac{\lambda_m}{m} \sum_{j=k+1}^m \<\bv_j, \btheta_\star\>^2 \leq \tl \leq \frac{\lambda_{k+1}}{m}\sum_{j=k+1}^m \<\bv_j, \btheta_\star\>^2.
\]
Note that with Assumption \ref{ass:norm}
\[
\sum_{j=k+1}^m \<\bv_j, \btheta_\star\>^2 = \|\bPt_\perp\btheta_\star\|_2^2 = \|\bP_\perp\btheta_\star\|_2^2 = \|\bP\btheta_\star\|_2^2 - \|\bP_0\btheta_\star\|_2^2 =\|\bP\btheta_\star\|_2^2 - \|\bPt\bP\btheta_\star\|_2^2 = \pjl^2.
\]
This completes the proof of Proposition \ref{proposition:ntk-eigen}. 
\end{proof}